\documentclass[10pt,twocolumn,letterpaper]{article}

\usepackage[pagenumbers]{cvpr} 
\usepackage[dvipsnames]{xcolor}

\usepackage{graphicx}
\usepackage{amsmath}
\usepackage{amssymb}
\usepackage{booktabs}
\usepackage{multirow}
\usepackage{array}
\usepackage{appendix}
\usepackage{tabularx}

\newcommand \blfootnote[1]{
    \begingroup
        \renewcommand
        \thefootnote{}\footnote{#1}
        \addtocounter{footnote}{-1}
        \vspace{-1ex}
    \endgroup
}

\definecolor{cvprblue}{rgb}{0.21,0.49,0.74}
\usepackage[pagebackref,breaklinks,colorlinks,citecolor=cvprblue]{hyperref}

\begin{document}

\title{Cross Initialization for Personalized Text-to-Image Generation}

\author{Lianyu Pang$^{1}$, Jian Yin$^{1,2}$, Haoran Xie$^{3}$, Qiping Wang$^{4}$, Qing Li$^{5}$, Xudong Mao$^{1,2}$\footnotemark[1]\>\,\\
{$^1$Sun Yat-sen University\ \ $^2$Guangdong Key Laboratory of Big Data Analysis and Processing}\\
{$^3$Linnan University\ \ $^4$East China Normal University\ \ $^5$The Hong Kong Polytechnic University }\\[-20pt]
}

\twocolumn[{%
\vspace{-1em}
\maketitle
\renewcommand\twocolumn[1][]{#1}%
\vspace{-0.1in}
\begin{center}
    \centering
    \includegraphics[width=0.98\textwidth]{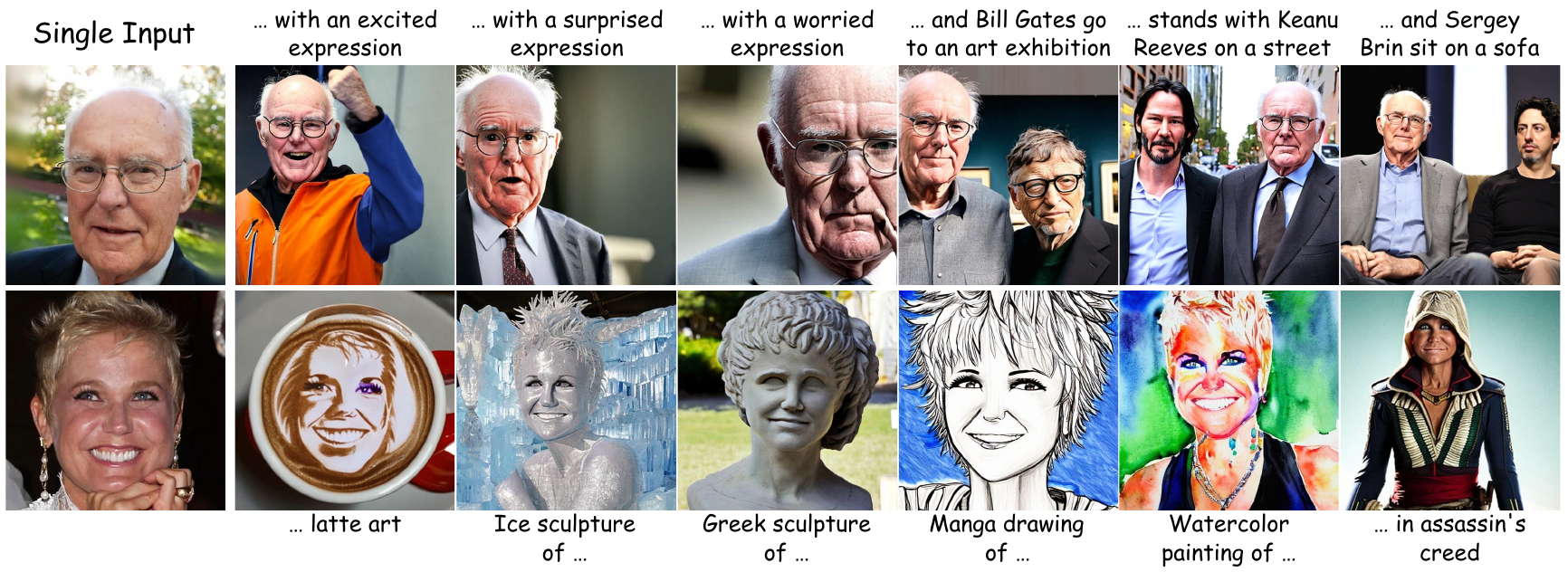}
    \vspace{-0.2cm}
    \captionof{figure}{
    Personalization results of our method using a single input image. Our method enables a variety of novel personalized face generations with high visual fidelity, such as facial expression editing, interaction with other individuals, and stylization. Moreover, it significantly speeds up the personalization process by reducing the optimization steps from 5,000 to 320.
    }
    \label{fig:teaser}
\end{center}%
}]

\blfootnote{$^*$Corresponding author (xudong.xdmao@gmail.com).}

\begin{abstract}
Recently, there has been a surge in face personalization techniques, benefiting from the advanced capabilities of pretrained text-to-image diffusion models. Among these, a notable method is Textual Inversion, which generates personalized images by inverting given images into textual embeddings. However, methods based on Textual Inversion still struggle with balancing the trade-off between reconstruction quality and editability. In this study, we examine this issue through the lens of initialization. Upon closely examining traditional initialization methods, we identified a significant disparity between the initial and learned embeddings in terms of both scale and orientation. The scale of the learned embedding can be up to 100 times greater than that of the initial embedding. Such a significant change in the embedding could increase the risk of overfitting, thereby compromising the editability. Driven by this observation, we introduce a novel initialization method, termed Cross Initialization, that significantly narrows the gap between the initial and learned embeddings. This method not only improves both reconstruction and editability but also reduces the optimization steps from 5,000 to 320. Furthermore, we apply a regularization term to keep the learned embedding close to the initial embedding. We show that when combined with Cross Initialization, this regularization term can effectively improve editability. We provide comprehensive empirical evidence to demonstrate the superior performance of our method compared to the baseline methods. Notably, in our experiments, Cross Initialization is the only method that successfully edits an individual's facial expression. Additionally, a fast version of our method allows for capturing an input image in roughly 26 seconds, while surpassing the baseline methods in terms of both reconstruction and editability. Code will be available at https://github.com/lyuPang/CrossInitialization.
\end{abstract}

\section{Introduction}
\label{sec:intro}
Recent advancements in large-scale diffusion models~\cite{stable-diffusion,imagen,dalle2} have significantly advanced the field of text-to-image generation, paving the way for a variety of generative tasks~\cite{textual-inversion,p2p,balaji2023ediffi}. Text-to-image personalization~\cite{textual-inversion}, when provided with several images of a target concept, enables users to produce personalized images in novel contexts or styles. This personalization is achieved either by inverting the target concept into the textual embedding space~\cite{textual-inversion,voynov2023p,alaluf2023neural} or by fine-tuning the pretrained diffusion model~\cite{dreambooth,kumari2022customdiffusion}. Among these, Textual Inversion~\cite{textual-inversion} is one notable method that learns the target concept by inverting given images into textual embeddings.

\begin{figure}[t]
  \centering
  \begin{subfigure}{0.49\linewidth}
    \includegraphics[width=\linewidth]{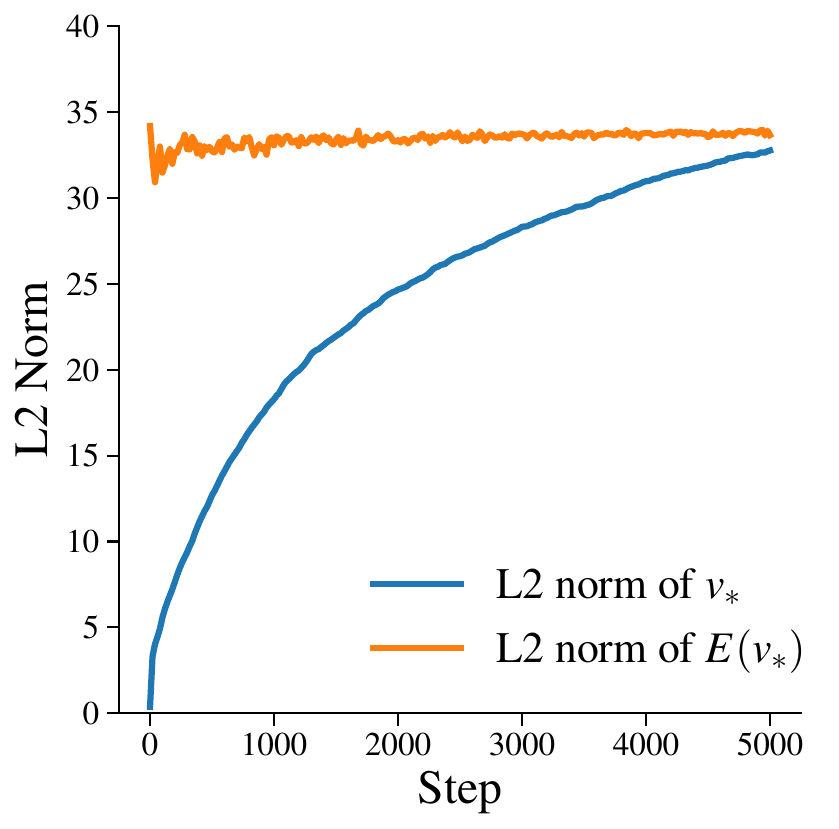}
    \label{fig:ti_comparison_a}
  \end{subfigure}
  \hfill
  \begin{subfigure}{0.49\linewidth}
    \includegraphics[width=\linewidth]{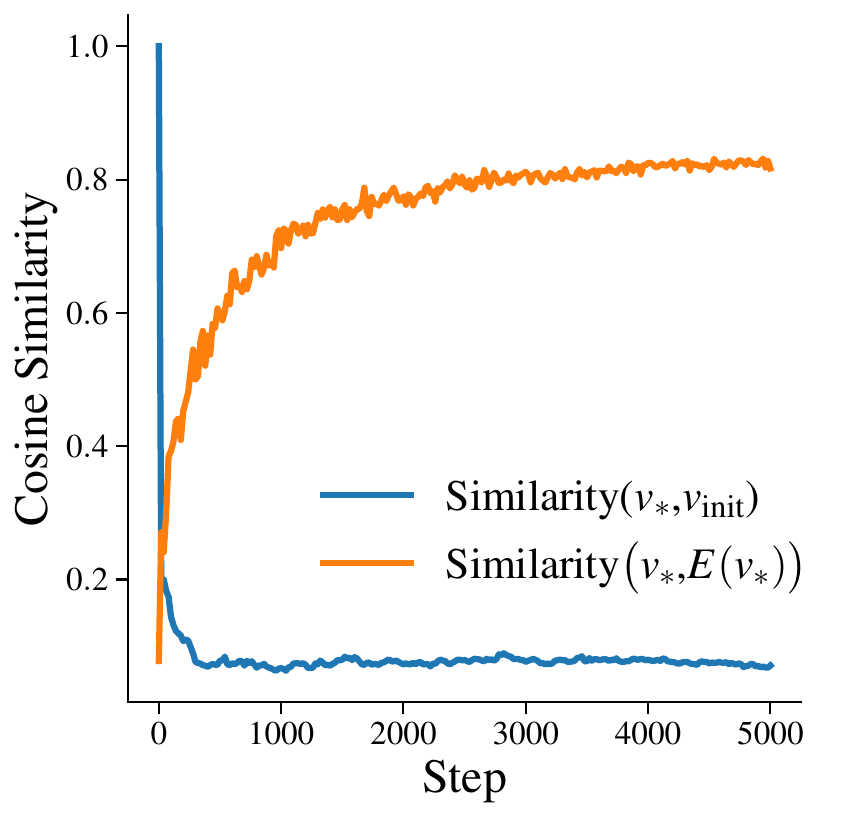}
    \label{fig:ti_comparison_b}
  \end{subfigure}
  \vspace{-1.6em}
  \caption{
Scale (left) and orientation (right) of the textual embedding $v_*$, as initialized by the traditional method. The term $E(v_*)$ represents the output vector of the text encoder, and $v_\text{init}$ represents the initial state of the embedding. After optimization, both the scale and orientation of $v_*$ undergo substantial alterations, aligning more closely with $E(v_*)$.
  }
  \vspace{-0.8em}
  \label{fig:ti_comparison}
\end{figure}

Face personalization~\cite{zhou2023enhancing,gal2023encoderbased,yuan2023inserting} focuses on the personalized generation of a particular individual. An effective face personalization model should be able to synthesize the individual in novel scenes or styles based on text prompts while preserving the individual's unique identity. However, many existing methods are prone to overfitting and often struggle to generate images that align with the prompt while accurately capturing the individual's identity.

In this work, we investigate the overfitting problem in Textual Inversion~\cite{textual-inversion} through the lens of initialization. Traditional methods typically initialize the textual embedding with a super-category token (e.g., ``face'' or ``person'')~\cite{textual-inversion,voynov2023p,alaluf2023neural}. However, after optimization, this approach often leads to significant deviations from the initial embedding in both scale and orientation, as depicted in \cref{fig:ti_comparison}. Such drastic changes may increase the risk of overfitting and compromise the editability of the embedding.

\begin{figure}[t]
    \centering
    \renewcommand{\arraystretch}{0.3}
    \setlength{\tabcolsep}{0.5pt}
    {\small
    \begin{tabular}{m{1.67cm}<{\centering} m{1.6cm}<{\centering} m{1.6cm}<{\centering} m{1.6cm}<{\centering} m{1.6cm}<{\centering}}
        
        Conditioning & Apple & House & Giraffe & Face \\
        \\
        $c(v)=E(v)$ &
        \includegraphics[width=0.092\textwidth]{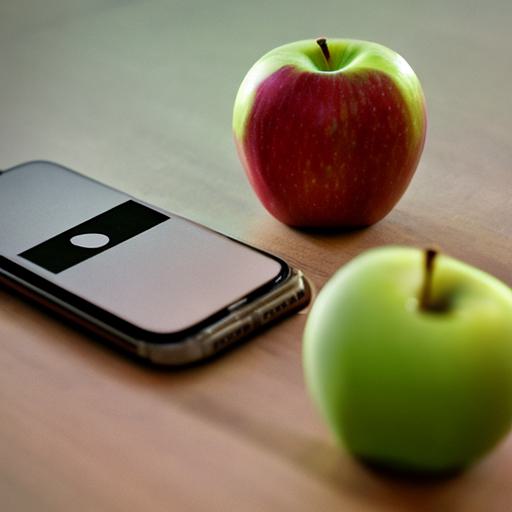} &
        \includegraphics[width=0.092\textwidth]{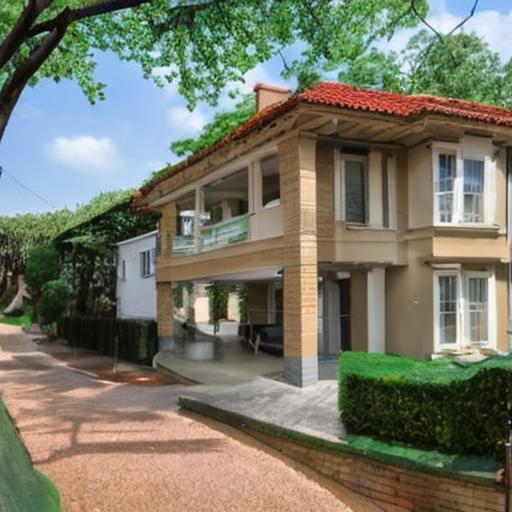}&
        \includegraphics[width=0.092\textwidth]{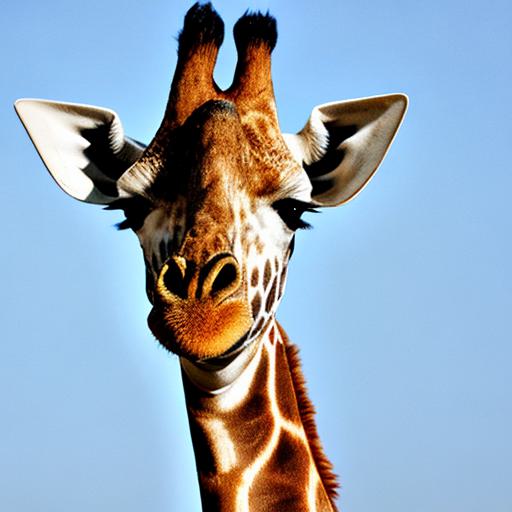}&
        \includegraphics[width=0.092\textwidth]{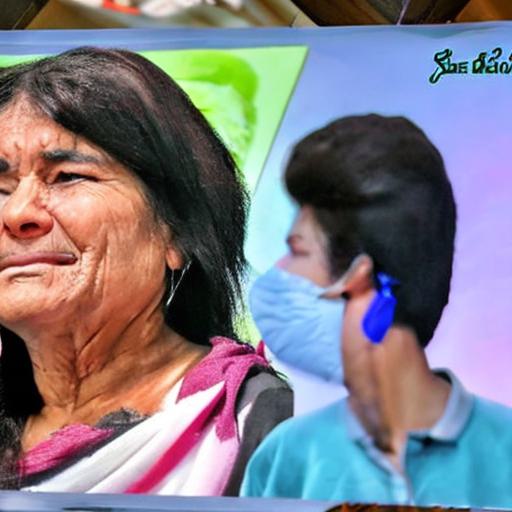}
        \\
        $c(v)=E(E(v))$ &
        \includegraphics[width=0.092\textwidth]{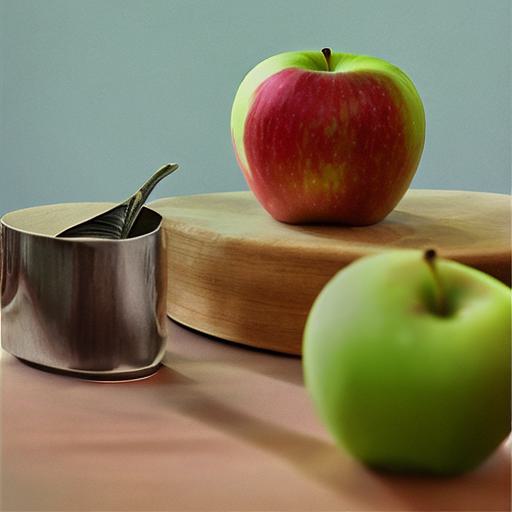} &
        \includegraphics[width=0.092\textwidth]{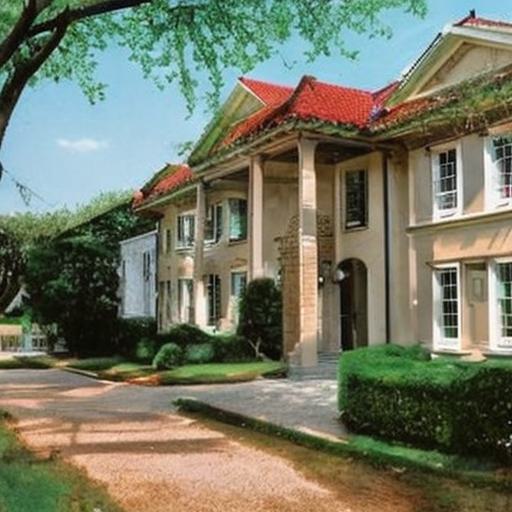}&
        \includegraphics[width=0.092\textwidth]{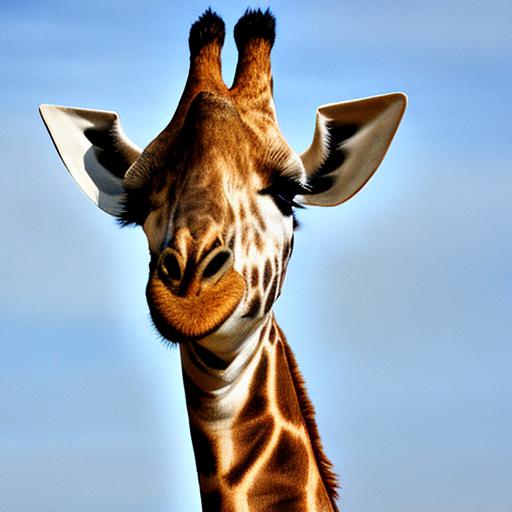}&
        \includegraphics[width=0.092\textwidth]{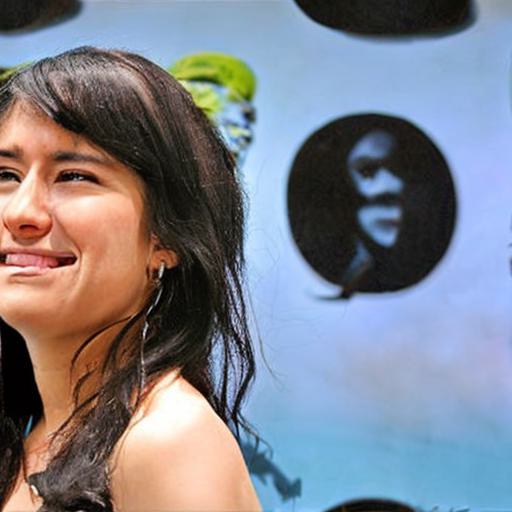}
        \\
        
    \end{tabular}
    }
    \caption{
    Top row: Images generated using standard textual embeddings as input for the text encoder, for instance, $v_\text{apple}$. Bottom row: Images generated using the output of the text encoder as its input, for instance, $E(v_\text{apple})$. Here, $c(v)$ denotes the conditioning vector in diffusion models. The images produced by $v$ and $E(v)$ are remarkably similar.
    }
    \vspace{-0.1cm}
    \label{fig:v_and_E_v_images}
\end{figure}

\begin{figure*}
    \centering
    \hspace*{0.95cm} 
    \includegraphics[width=0.86\linewidth]{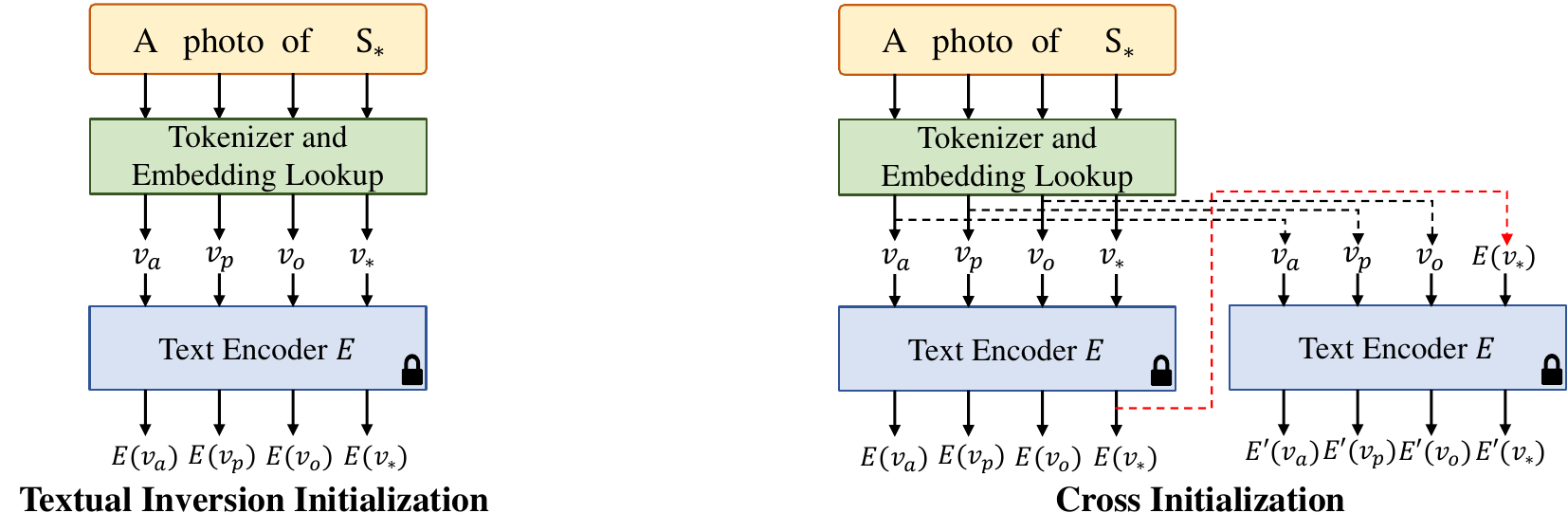}
    \caption{
    Comparison of Textual Inversion Initialization and Cross Initialization techniques. Textual Inversion~\cite{textual-inversion} (left) initializes the textual embedding $v_*$ with a super-category token (e.g., ``face''). Cross Initialization (right) begins by obtaining the output vector from the text encoder $E(v_*)$, which is subsequently used to initialize the embedding. This approach reduces the disparity between the initial and learned embeddings.
    }
    \vspace{-1em}
    \label{fig:cross_initialization}
\end{figure*}

To address this issue, our approach aims to minimize the disparity between the initial and learned embeddings. Our method is inspired by two main observations. Firstly, after optimization, the learned embedding tends to align with the output of the CLIP~\cite{clip} text encoder in terms of both scale and orientation, as illustrated in \cref{fig:ti_comparison}. Secondly, using the text encoder's output as its input typically produces an image nearly identical to the original, as shown in \cref{fig:v_and_E_v_images}. Drawing from these insights, we introduce \textit{Cross Initialization}, a method where the textual embedding is initialized with the text encoder's output, as depicted in \cref{fig:cross_initialization}. This approach effectively narrows the gap between the initial and learned embeddings, facilitating more effective optimizations compared to traditional methods. Our results demonstrate that Cross Initialization not only enhances reconstruction quality and editability but also significantly speeds up the personalization process.

To further improve editability, we incorporate a regularization term designed to keep the learned embedding close to its initial state throughout the optimization process. In Textual Inversion, the effectiveness of this regularization is often limited due to the substantial disparity between the initial and learned embeddings. In contrast, when used in conjunction with Cross Initialization, this regularization strategy becomes significantly more effective. This improvement is primarily attributed to the reduced gap between the initial and learned embeddings facilitated by Cross Initialization.

We demonstrate the superior performance of Cross Initialization compared to the baseline methods through both qualitative and quantitative evaluations. Our method enables a variety of novel personalized face generations with high visual fidelity. Notably, in our experiments, Cross Initialization is the only method capable of editing an individual's facial expression. Furthermore, a fast version of our method allows for capturing an input image in roughly 26 seconds, while surpassing the baseline methods in terms of both reconstruction and editability.

\section{Related Works}
\paragraph{Text-to-Image Synthesis.} 
Text-to-image synthesis is the task of generating realistic and diverse images from natural language descriptions. Various deep generative models have been widely explored for this task, such as GANs~\cite{Reed2016,sauer2023stylegant}, VAEs~\cite{Ramesh2021,ding2021cogview}, and Autoregressive Models~\cite{yu2022scaling,dalle2}. Recently, diffusion models~\cite{stable-diffusion,ddim,ddpm} have demonstrated remarkable capabilities in generating high-fidelity images aligned with textual prompts~\cite{dalle2,nichol2021glide,stable-diffusion,imagen,balaji2023ediffi}.

\paragraph{Inversion.} 
Image inversion involves reconstructing an image by mapping it into the latent space of a pretrained generator. This process can be accomplished either through direct optimization of the latent code~\cite{abdal2019image2stylegan,gu2020image,zhu2020improved} or by employing an encoder network to map the image into a latent space~\cite{Pidhorskyi_2020_CVPR, richardson2020encoding, bai2022high, parmar2022spatially, tov2021designing, wang2021HFGI, zhu2020domain}. Image inversion has been applied to various image manipulation tasks~\cite{shamsian2021personalized,patashnik2021styleclip,gu2020image}. In the context of diffusion models, image inversion aims to identify an initial noise latent code that can be denoised back to the input image~\cite{dalle2,dhariwal2021diffusion,null}.  This inverted noise latent code is then leveraged for text-guided image manipulation, as explored in recent studies~\cite{p2p, couairon2022diffedit, kawar2023imagic, liew2022magicmix, pnpDiffusion2022}.

\paragraph{Personalization.} Personalization adapts pretrained generative models to capture new concepts depicted in several given images. In the realm of text-to-image diffusion models, this allows for the creation of personalized images guided by text prompts. Techniques for this task include optimizing textual embeddings to learn new concepts~\cite{cohen2022my, textual-inversion, voynov2023p, alaluf2023neural, dreamartist,vinker2023concept}, fine-tuning diffusion models for concept acquisition~\cite{he2023data,ruiz2023hyperdreambooth,choi2023customedit,hao2023vico,cohen2022my, tewel2023keylocked, dreambooth, kumari2022customdiffusion, continualdiffusion,avrahami2023breakascene}, and training encoders for mapping new concepts to textual representations~\cite{arar2023domainagnostic,gal2023encoderbased, shi2023instantbooth, zhou2023enhancing, umm, suti, jia2023taming}. These methods facilitate applications like image editing~\cite{kawar2023imagic, valevski2022unitune} and personalized 3D generation~\cite{lin2023magic3d, metzer2023latent, raj2023dreambooth3d, richardson2023texture}. Particularly, some studies~\cite{zhou2023enhancing, yuan2023inserting, gal2023encoderbased,wu2023singleinsert,chen2023photoverse,gu2023mixofshow,hyung2023magicapture} focus on the personalized generation of individual human images. However, existing methods often face the overfitting problem, hindering the creation of text-aligned personalized images. Our work addresses this challenge by examining the overfitting problem through the lens of initialization. Our approach enables more efficient learning of new concepts, leading to faster personalized face generation with improved identity preservation and enhanced editability.

\section{Preliminaries}
\paragraph{Latent Diffusion Models.}
We implement our method on the publicly available Stable Diffusion (SD) model, a Latent Diffusion Model (LDM)~\cite{stable-diffusion} for text-to-image synthesis. This model is composed of an encoder, $\mathcal{E}$, which maps an image $x$ to a latent code $z=\mathcal{E}(x)$, and a decoder, $\mathcal{D}$, which reconstructs the image from this code $\mathcal{D}(\mathcal{E}(x))\approx x$. A Denoising Diffusion Probabilistic Model (DDPM)~\cite{ddpm} is trained to generate latent codes within the latent space of a pretrained autoencoder. For text-to-image generation, the model is conditioned on a vector $c(y)$ derived from a text prompt $y$. The training objective of LDM is defined by:
\begin{equation}
  \mathcal{L}_{\text{diffusion}}=\mathbb{E}_{z \sim \mathcal{E}(x), y, \varepsilon \sim \mathcal{N}(0,1), t}\left[\left\|\varepsilon-\varepsilon_{\theta}\left(z_{t}, t, c(y)\right)\right\|_{2}^{2}\right].
  \label{eq:ldm}
\end{equation}
Given the timestep $t$, the noised latent $z_{t}$, and the conditioning vector $c(y)$, the denoising network $\varepsilon_{\theta}$ aims to remove the noise that was added to the original latent code $z_0$.

\paragraph{Text Embeddings.}
Given a text prompt $y$, the sentence is first tokenized into several tokens. Each token is then mapped to a textual embedding $v_i$ using a predefined embedding lookup. Subsequently, these textual embeddings are passed through a pretrained CLIP text encoder $E$, which outputs a series of vectors that constitute the conditioning vector $c(y) = [E(v_1), \dots, E(v_n)]$. For a textual embedding $v_i \in \mathbb{R}^{1024}$, its corresponding output of the text encoder is denoted by $E(v_i) \in \mathbb{R}^{1024}$. Note that in the SD v2.1 model, the dimensionality of both $v_i$ and $E(v_i)$ is $1024$.

\paragraph{Textual Inversion.}
Textual Inversion~\cite{textual-inversion} is a technique that captures novel concepts from a few example images. It is achieved by injecting new concepts into the pretrained diffusion models. Specifically, Textual Inversion introduces a new token $S_*$ and its corresponding textual embedding $v_*$, representing the new concept. To learn the new concept, Textual Inversion fixes the LDM and optimizes only $v_*$, minimizing the objective of LDM given in \cref{eq:ldm}. The optimization objective is defined by:
\begin{equation}
  v_{*}=\arg\min _{v} \mathbb{E}_{z, y, \varepsilon, t}\left[\left\|\varepsilon-\varepsilon_{\theta}\left(z_{t}, t, c(y, v)\right)\right\|_{2}^{2}\right],
  \label{eq:ti}
\end{equation}
where $c(y, v)$ is the conditioning vector obtained from the prompt $y$ and the textual embedding $v$.

\section{Method}
Our method is based on the Textual Inversion technique, in which the textual embedding is typically initialized with a super-category token (e.g., ``face''). In this section, we analyze how Textual Inversion suffers from a severe overfitting problem through the lens of initialization, as detailed in \cref{sec:analysis}. To address this issue, we propose a novel initialization method, named Cross Initialization, as described in \cref{sec:CI}. This method facilitates more efficient optimizations, enhancing both reconstruction and editability. To further improve editability, we introduce a regularization term in \cref{sec:regularization}.

\begin{figure}[t]
    \centering
    \renewcommand{\arraystretch}{0.3}
    \setlength{\tabcolsep}{0.5pt}
    {\small
    \begin{tabular}{c@{\hspace{0.1cm}} c c@{\hspace{0.1cm}} c c@{\hspace{0.25cm}}}
        Input & 
        \multicolumn{2}{c}{\begin{tabular}{c}A sand sculpture of $S_*$\end{tabular}} & 
        \multicolumn{2}{c}{$S_*$ Funko Pop} \\
        \\
        \includegraphics[width=0.087\textwidth]{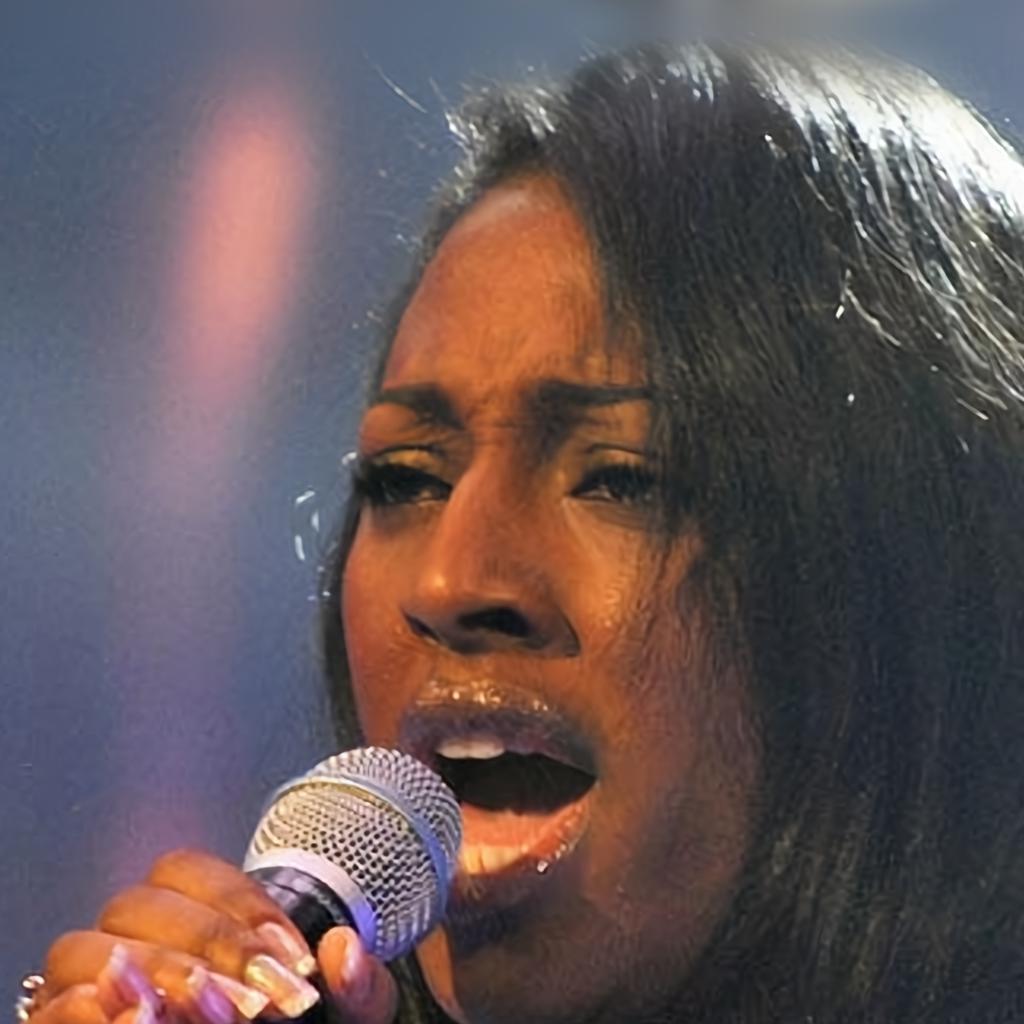} &
        \includegraphics[width=0.087\textwidth]{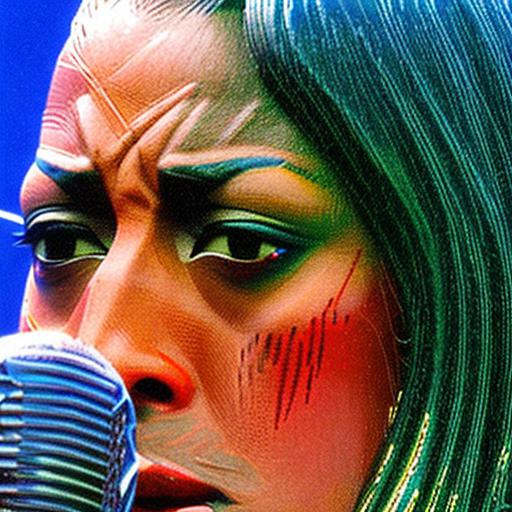} &
        \includegraphics[width=0.087\textwidth]{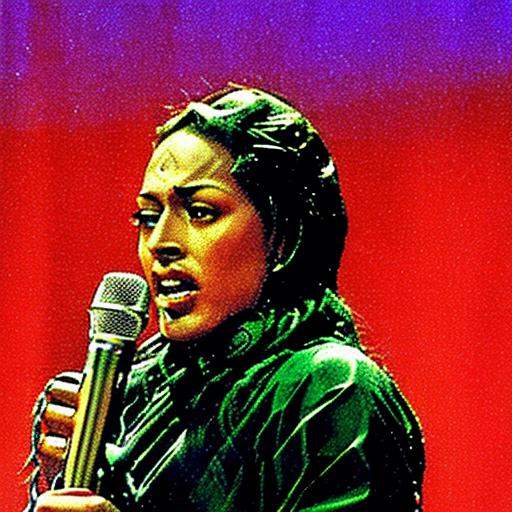}&
        \includegraphics[width=0.087\textwidth]{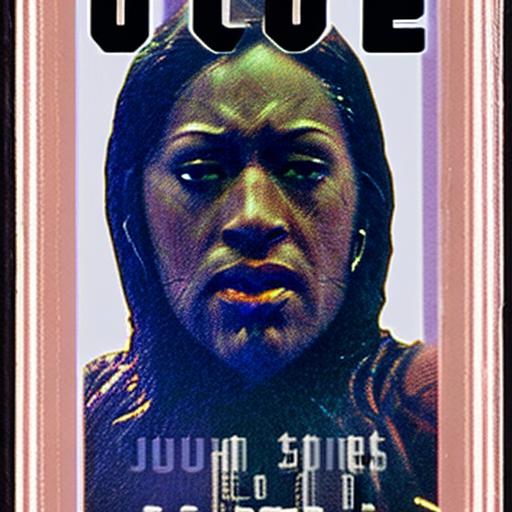}&
        \includegraphics[width=0.087\textwidth]{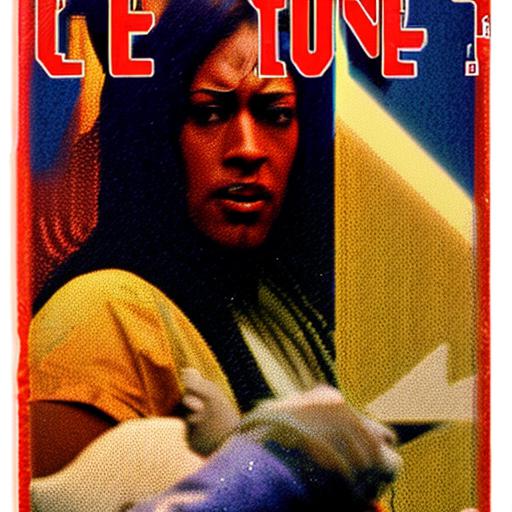}
        \\
        \includegraphics[width=0.087\textwidth]{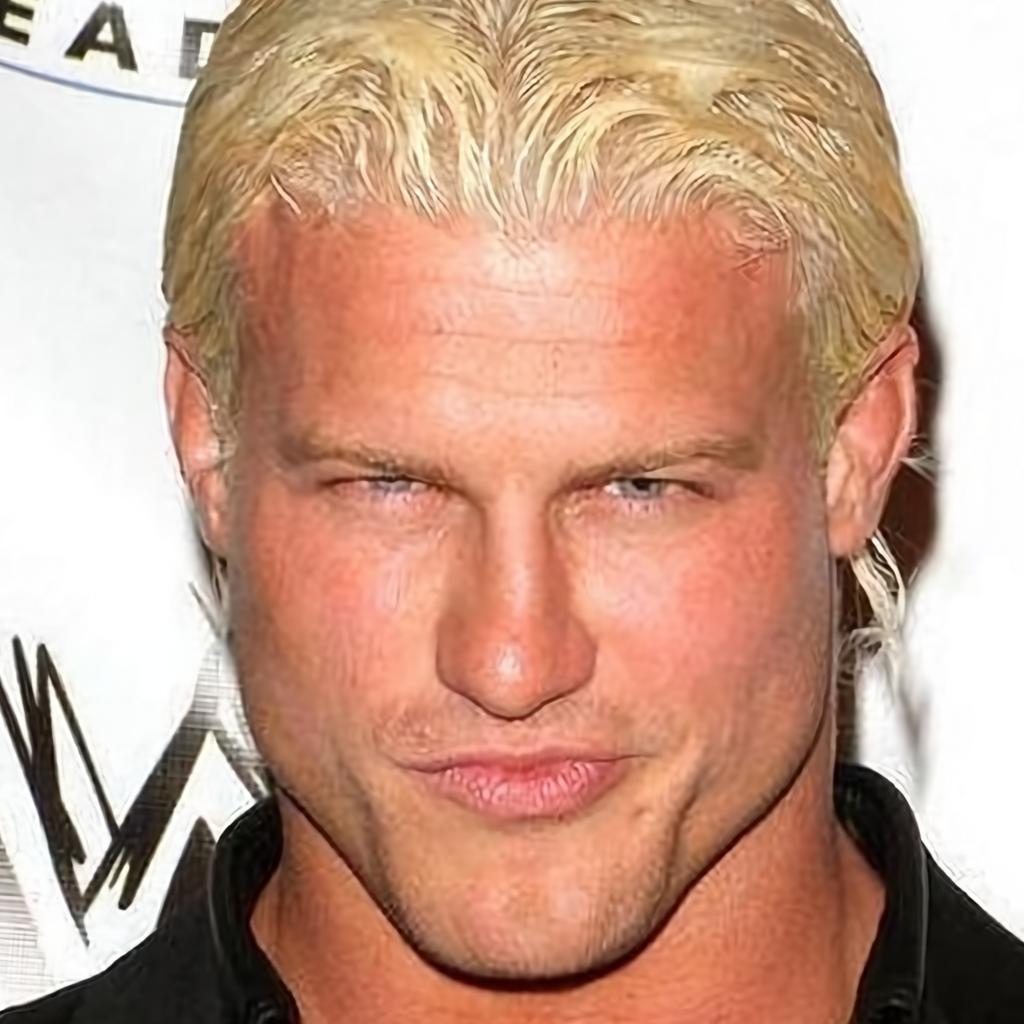} &
        \includegraphics[width=0.087\textwidth]{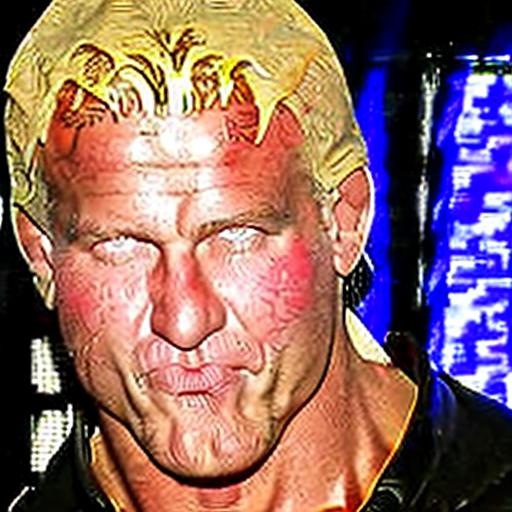} &
        \includegraphics[width=0.087\textwidth]{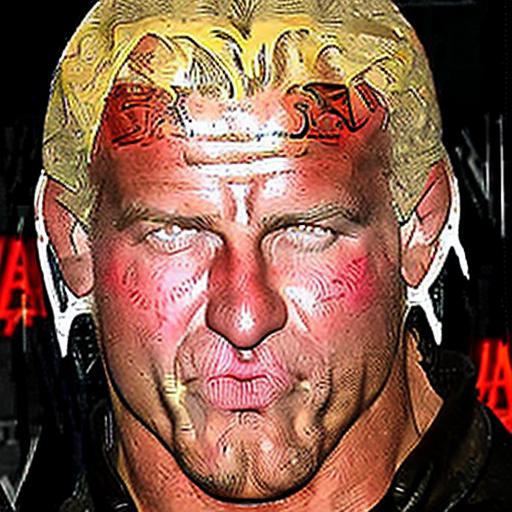}&
        \includegraphics[width=0.087\textwidth]{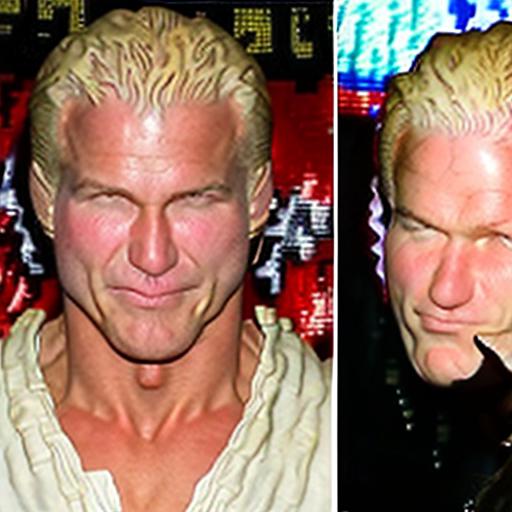}&
        \includegraphics[width=0.087\textwidth]{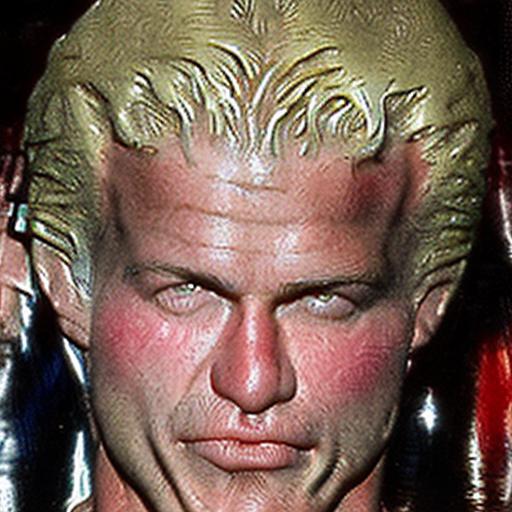}
        \\
    \end{tabular}
    }
    \caption{
    Images generated by Textual Inversion. This method fails to place the given individual in new styles, primarily due to its tendency to overfit the input image.
    }

    \vspace{-0.1cm}
    \label{fig:ti_overfitting}
\end{figure}

\subsection{Analysis}
\label{sec:analysis}
\begin{figure}[t]
  \centering
  \begin{subfigure}{0.49\linewidth}
    \includegraphics[width=\linewidth]{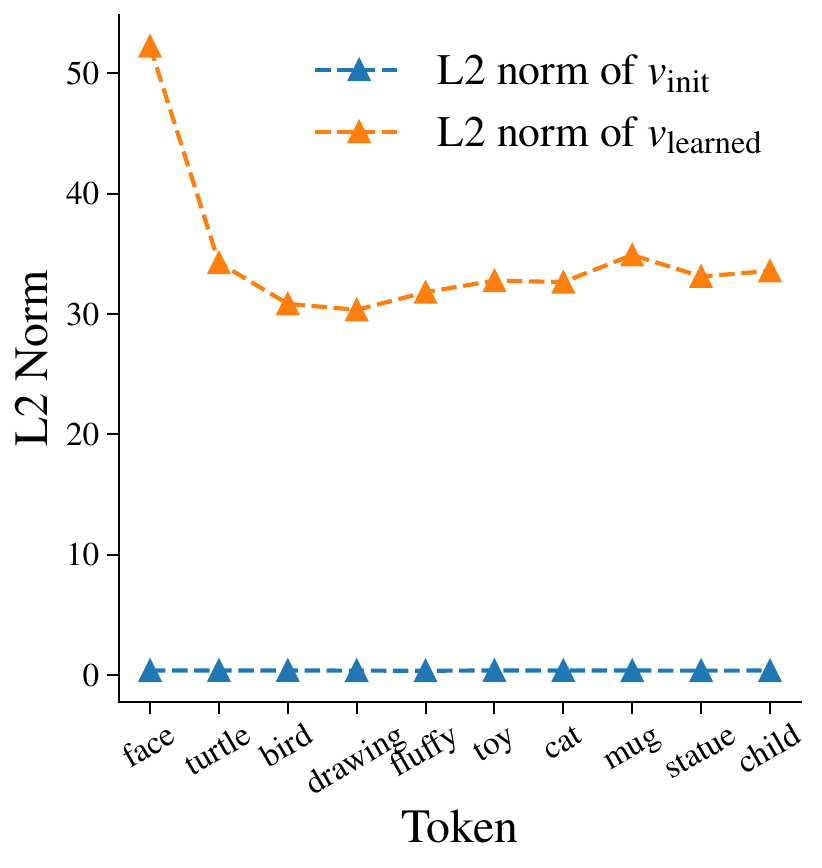}
    \label{fig:learned_embed_for_different_object_a}
  \end{subfigure}
  \hfill
  \begin{subfigure}{0.49\linewidth}
    \includegraphics[width=\linewidth]{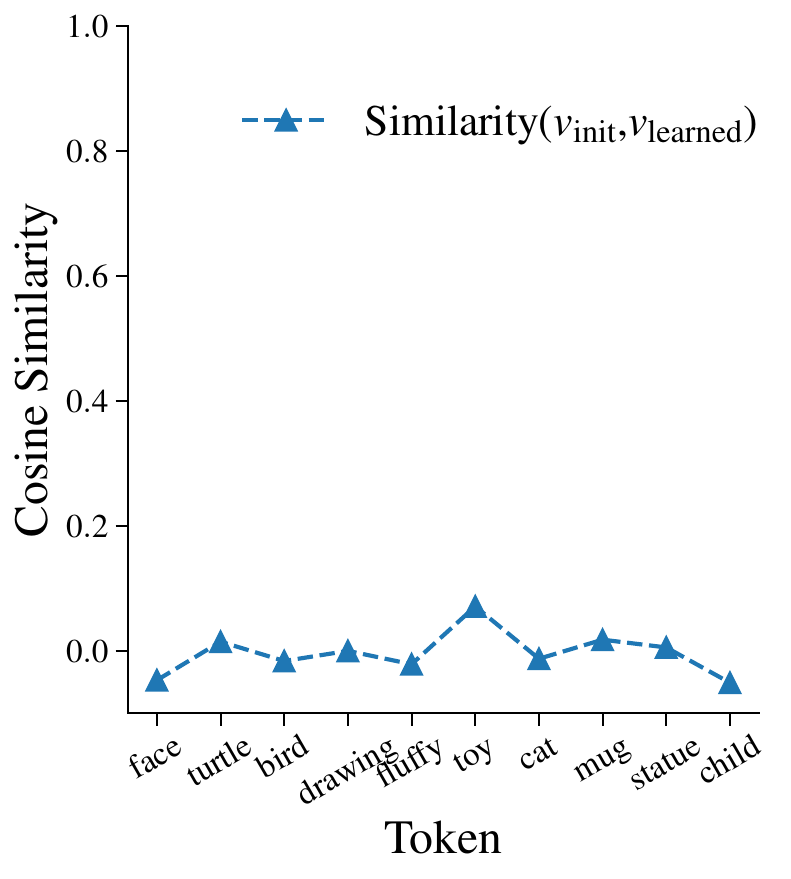}
    \label{fig:learned_embed_for_different_object_b}
  \end{subfigure}
  \vspace{-1.6em}
  \caption{
  More examples illustrating that, after optimization, the textual embedding $v_*$ experiences significant changes in both scale (left) and orientation (right). Here, $v_{\text{init}}$ denotes the embedding's initial state, and $v_{\text{learned}}$ denotes the embedding's final state.
  }
  \label{fig:learned_embed_for_different_object}
\end{figure}

In \cref{fig:ti_overfitting}, we show several examples generated by Textual Inversion. This method fails to place the person in new styles and generates images similar to the input image, indicating a severe overfitting problem. In this section, we delve into this overfitting problem in Textual Inversion from the perspective of initialization. Existing methods based on Textual Inversion typically initialize the textual embedding with a super-category token~\cite{textual-inversion,voynov2023p,alaluf2023neural}. However, our experiments consistently show that, after optimization, the learned embedding becomes significantly different from its initial state, both in scale and orientation. \cref{fig:ti_comparison,fig:learned_embed_for_different_object} show several examples where the scale of the learned embedding can be up to 100 times greater than that of the initial embedding. Such drastic changes in the embedding may increase the risk of overfitting and degrade the editability of the embedding.

\begin{figure}[t]
  \centering
  \begin{subfigure}{0.49\linewidth}
    \includegraphics[width=\linewidth]{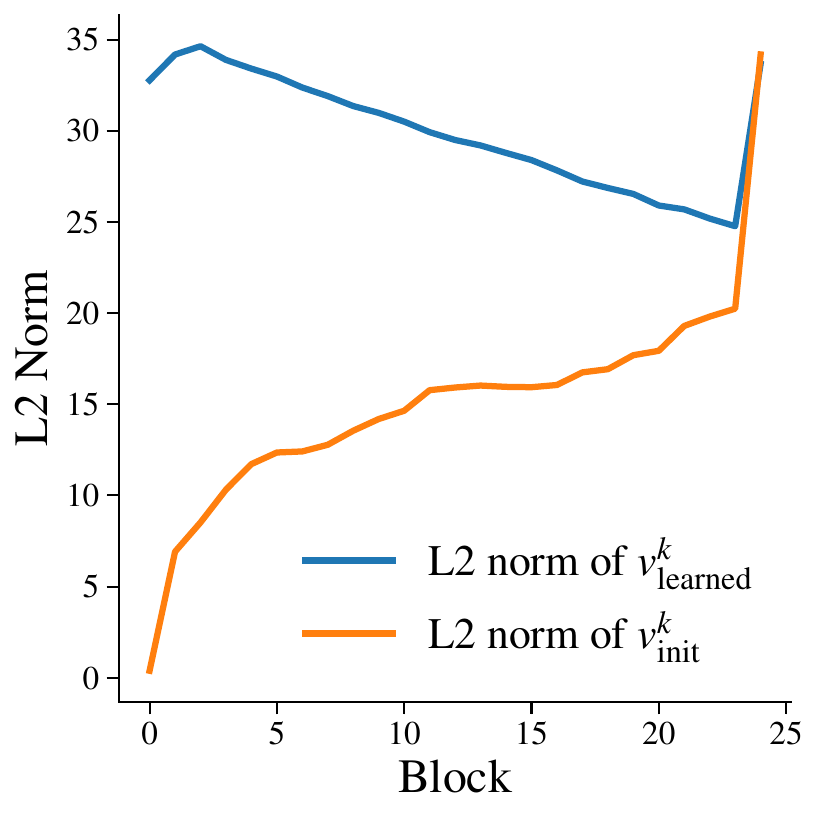}
    \label{fig:embedding_in_encoder_a}
  \end{subfigure}
  \hfill
  \begin{subfigure}{0.49\linewidth}
    \includegraphics[width=\linewidth]{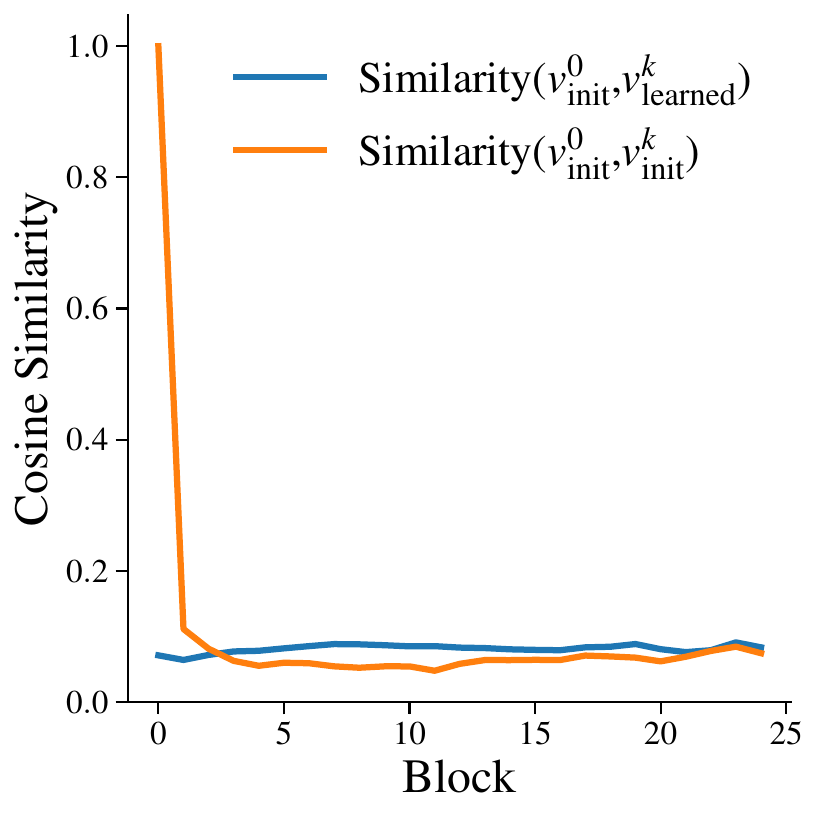}
    \label{fig:embedding_in_encoder_b}
  \end{subfigure}
  \vspace{-1.6em}
  \caption{
  Scale (left) and orientation (right) of the textual embedding processed by the $k$-th self-attention block of the text encoder. The symbols $v_{\text{init}}^k$ and $v_{\text{learned}}^k$ denote the outputs of the $k$-th self-attention block using the initial and learned embeddings as inputs, respectively. Note that an additional LayerNorm layer is present after the final block. In each block, the LayerNorm layer and the self-attention layer gradually modify the scale and orientation of the embedding. After optimization, the output vectors derived from the initial and learned embeddings exhibit a similarity in both scale and orientation.
  }

  \label{fig:embedding_in_encoder}
\end{figure}

Given that the learned embedding significantly differs from the initial embedding of a coarse descriptor, a question arises: How does the learned embedding manage to produce images that accurately represent the given concept? To investigate this, we examine the outputs of the intermediate layers in the text encoder. The text encoder comprises several self-attention blocks~\cite{shaw2018self}, with a LayerNorm layer~\cite{ba2016layer} preceding the input of each sub-block. We observe that the LayerNorm layer normalizes the scale of the embedding, while the self-attention layer modifies its orientation. \cref{fig:embedding_in_encoder} illustrates this process: each sub-block progressively alters the scale and orientation of the embedding, and ultimately the output vectors of the initial and learned embeddings exhibit a similarity in both scale and orientation. 

To mitigate the overfitting issue in Textual Inversion, this analysis motivates us to seek an initial embedding that can be close to the learned embedding.

\subsection{Cross Initialization}
\label{sec:CI}
Based on the analysis in \cref{sec:analysis}, our goal is to design an initial embedding that meets two criteria: 1) it is close to the learned embedding, and 2) it roughly captures the target concept. Our method is inspired by two key observations. First, as shown in \cref{fig:ti_comparison}, the learned embedding becomes similar to the output of the text encoder after optimization. Second, when we use the text encoder's output as its input, the diffusion model produces an image nearly identical to the original, as shown in \cref{fig:v_and_E_v_images}. The reason for these two phenomena is that the LayerNorm and self-attention layers in the text encoder gradually alter the scale and orientation of the embedding, making it converge to a specific vector, as discussed in \cref{sec:analysis}. Based on these insights, we propose initializing the textual embedding with the output of the text encoder, a method we term Cross Initialization, as depicted in \cref{fig:cross_initialization}.

Formally, given a single face image, we first set the textual embedding to the mean of 691 well-known names' embeddings, denoted as $\bar{v}_{691}$. The computation of $\bar{v}_{691}$ is elaborated in the following subsection. Subsequently, we feed $\bar{v}_{691}$ into the text encoder $E$, obtaining the output vector $E(\bar{v}_{691})$. We then initialize the textual embedding $v_{\text{init}}$ with this output vector:
\begin{equation}
    v_{\text{init}}=E(\bar{v}_{691}).
\label{eq:cross_ti}
\end{equation}
Finally, we optimize the textual embedding by minimizing the LDM loss given in \cref{eq:ti}.

\begin{figure}[t]
  \centering
  \begin{subfigure}{0.49\linewidth}
    \includegraphics[width=\linewidth]{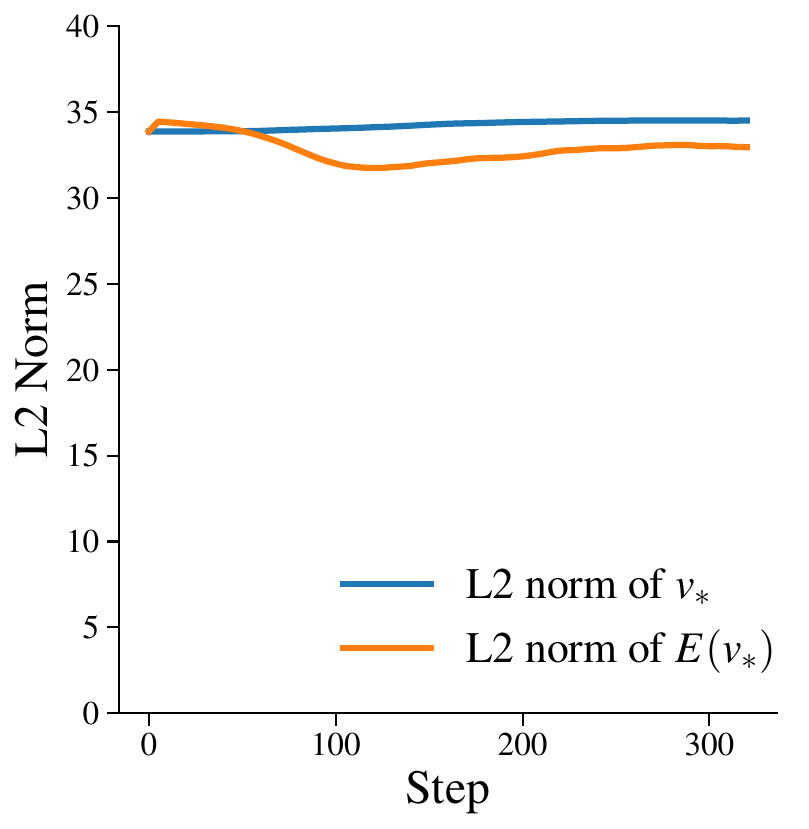}
    \label{fig:v_cross_comparison_a}
  \end{subfigure}
  \hfill
  \begin{subfigure}{0.49\linewidth}
    \includegraphics[width=\linewidth]{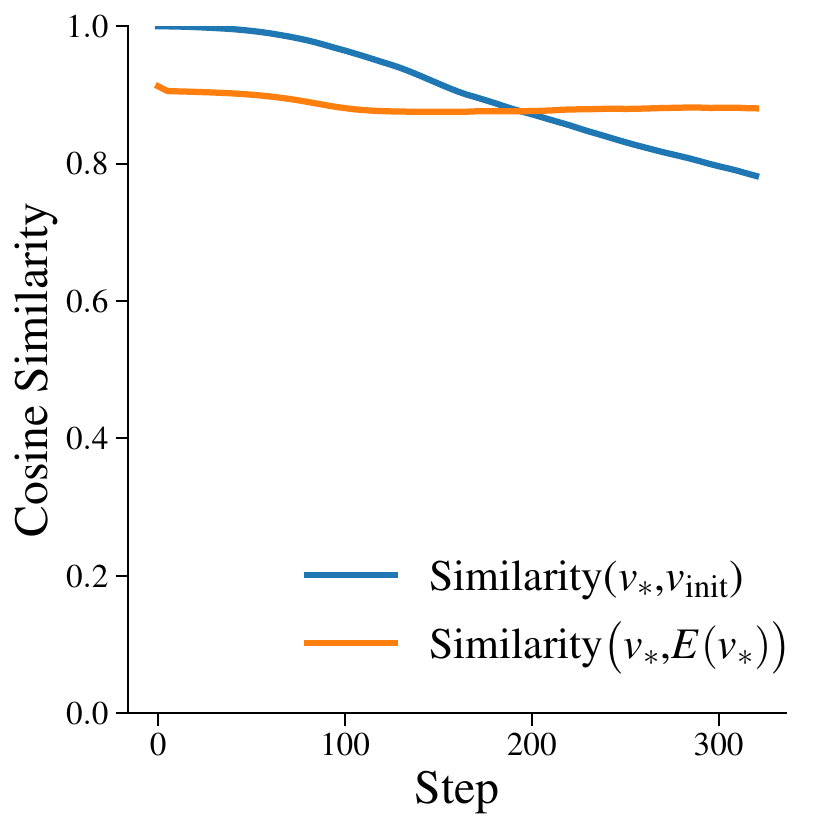}
    \label{fig:v_cross_comparison_b}
  \end{subfigure}
  \vspace{-1.6em}
  \caption{
  Scale (left) and orientation (right) of the textual embedding $v_*$, as initialized by Cross Initialization. Here, $E(v_*)$ represents the output vector of the text encoder, and $v_\text{init}$ represents the initial state of the embedding. In contrast to the examples in \cref{fig:ti_comparison}, Cross Initialization maintains the learned embedding close to the initial state in terms of both scale and orientation.
  }
  \label{fig:v_cross_comparison}
\end{figure}

The aforementioned two observations ensure that the initial embedding $E(\bar{v}_{691})$ is close to the learned embedding, while also roughly representing the target concept. As shown in \cref{fig:v_cross_comparison}, using Cross Initialization, the learned embedding retains proximity to its initial state throughout the optimization process. This facilitates more efficient optimizations, leading to more identity-preserved, prompt-aligned, and faster face personalization.

\paragraph{Mean Textual Embedding.}
We follow~\cite{yuan2023inserting} to construct the mean textual embedding $\bar{v}_{691}$. A total of 691 well-known names are used to form an embedding set $C=\{v_1,\dots,v_m\}$, where $m=691$ and each textual embedding $v_i$ is obtained from the pre-defined embedding lookup. 
The mean textual embedding is calculated as $\bar{v}_{691} = \frac{1}{m}\sum^m_{i=1}v_i$. Moreover, we represent each name with two tokens (i.e., the first and last names), resulting in the final mean textual embedding as $\bar{v}_{691}=[\bar{v}_{691}^{f},\bar{v}_{691}^{l}]$, where $\bar{v}_{691}^{f}$ and $\bar{v}_{691}^{l}$ are calculated using the embedding sets of the first and last names, respectively.

\paragraph{Comparison with Directly Optimizing $E(v)$.}
In Cross Initialization, we set the text encoder's output as its input, i.e. $v_{\text{init}}=E(\bar{v})$, and optimize the input vector $v_{\text{init}}$. An alternative method is to directly optimize the output vector $E(\bar{v})$. However, this approach eliminates the interaction between the new concept and other prompt tokens, as the new concept is not passed through the text encoder along with the other prompt tokens, leading to poor editability. This issue is also indicated in~\cite{alaluf2023neural}. In contrast, Cross Initialization optimizes the input vector, thereby preserving the ability to create new compositions for the new concept.

\subsection{Regularization}
\label{sec:regularization}
As illustrated in \cref{sec:CI}, the initial embedding is constructed using the mean center of embeddings from 691 well-known names. We assume that the region around this central embedding represents the subspace corresponding to the concept of the individual. High editability is expected when the learned embedding lies close to this subspace. Therefore, we introduce a regularization term to keep the learned embedding close to the central embedding throughout the optimization process. Specifically, we minimize the L2 distance between them, defined as: 
\begin{equation}
  \mathcal{L}_{\text{reg}}=||v-v_{\text{init}}||_2^2.
  \label{eq:reg}
\end{equation}
Overall, our final optimization objective is defined as:
\begin{equation}
  v_{*}=\arg\min _{v} \mathcal{L}_{\text{diffusion}} + \lambda \mathcal{L}_{\text{reg}}.
  \label{eq:total}
\end{equation}
Note that this regularization approach, also investigated in~\cite{textual-inversion}, faces challenges when applied in Textual Inversion. This is primarily due to the significant disparity between the initial and learned embeddings, as well as the coarseness of the super-category token. These factors limit the effectiveness of this regularization approach.

\begin{figure*}
    \centering
    \renewcommand{\arraystretch}{0.3}
    \setlength{\tabcolsep}{0.5pt}

    {\footnotesize
    \begin{tabular}{c@{\hspace{0.15cm}} c c @{\hspace{0.15cm}} c c @{\hspace{0.15cm}} c c @{\hspace{0.15cm}} c c @{\hspace{0.15cm}} c c }

        \normalsize Real Sample &
        \multicolumn{2}{c}{\normalsize Textual Inversion} &
        \multicolumn{2}{c}{\normalsize DreamBooth} &
        \multicolumn{2}{c}{\normalsize NeTI} &
        \multicolumn{2}{c}{\normalsize Celeb Basis} &
        \multicolumn{2}{c}{\normalsize Ours} \\
        \includegraphics[width=0.08\textwidth]{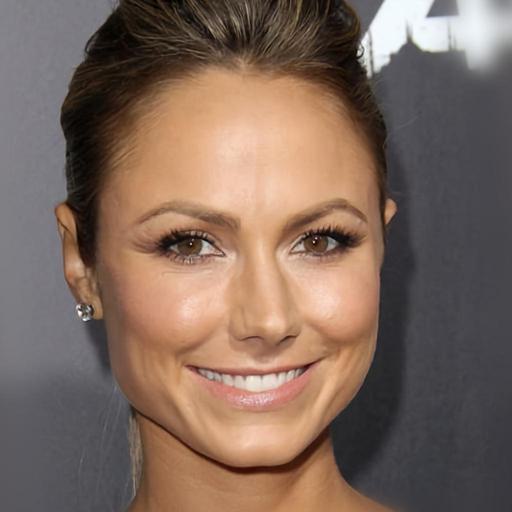} &
        \includegraphics[width=0.08\textwidth]{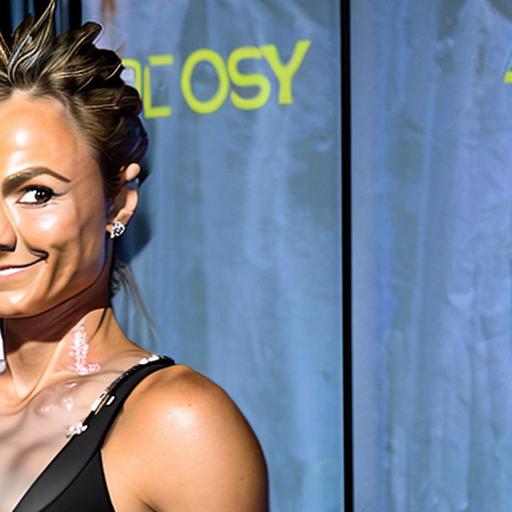} &
        \includegraphics[width=0.08\textwidth]{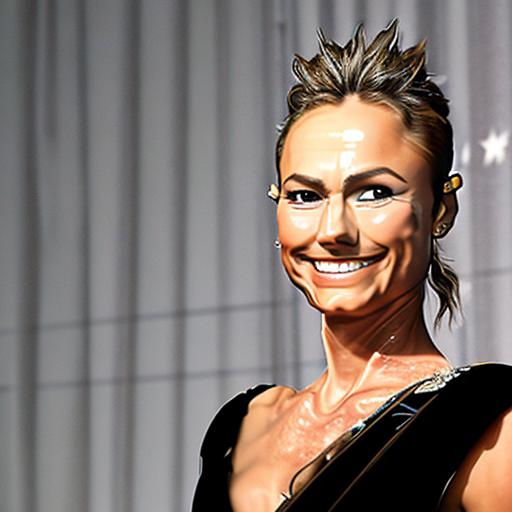} &
        \hspace{0.05cm}
        \includegraphics[width=0.08\textwidth]{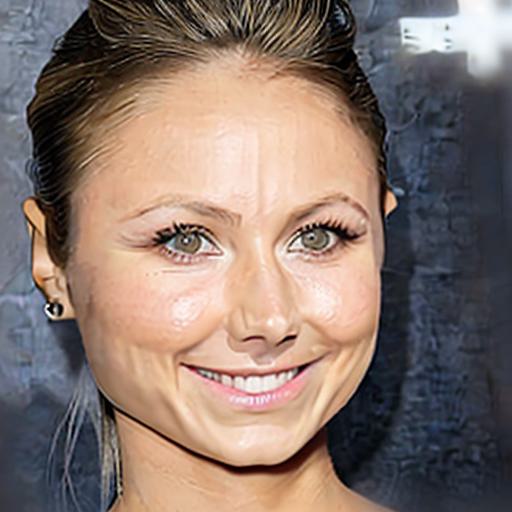} &
        \includegraphics[width=0.08\textwidth]{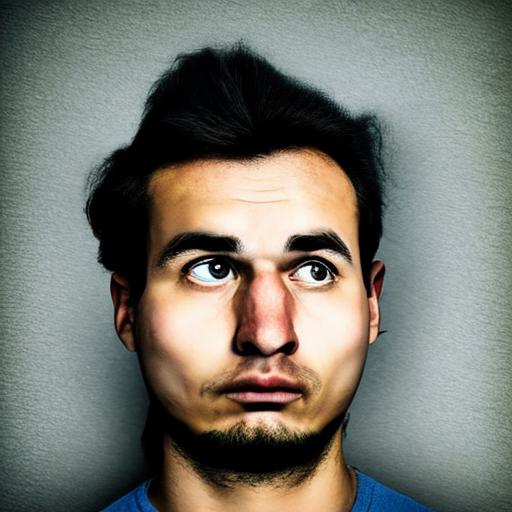} &
        \hspace{0.05cm}
        \includegraphics[width=0.08\textwidth]{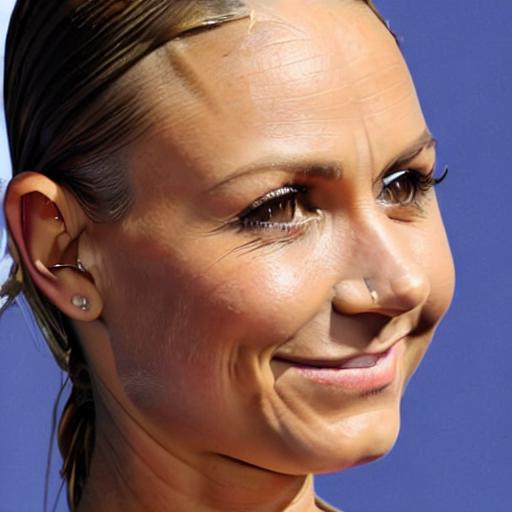} &
        \includegraphics[width=0.08\textwidth]{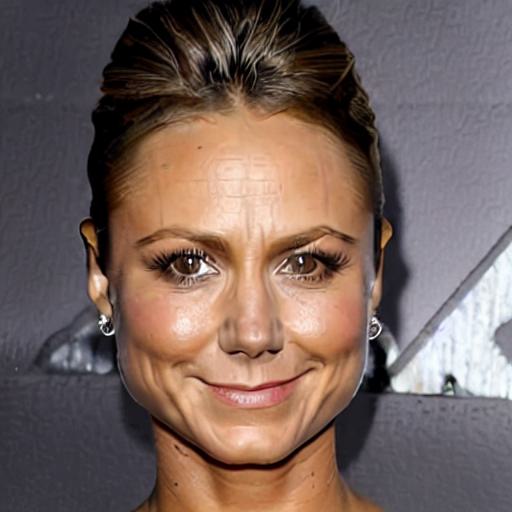} &
        \hspace{0.05cm}
        \includegraphics[width=0.08\textwidth]{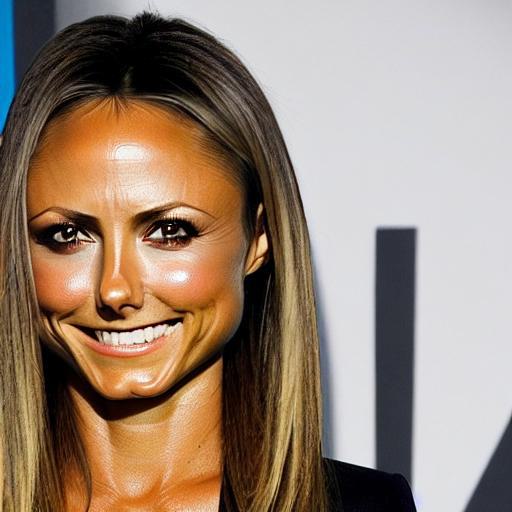} &
        \includegraphics[width=0.08\textwidth]{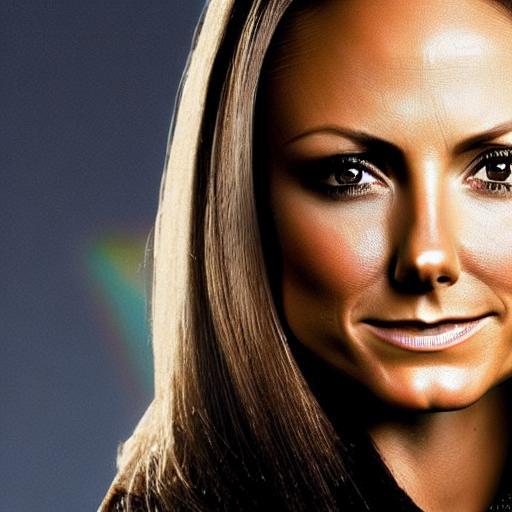} &
        \hspace{0.05cm}
        \includegraphics[width=0.08\textwidth]{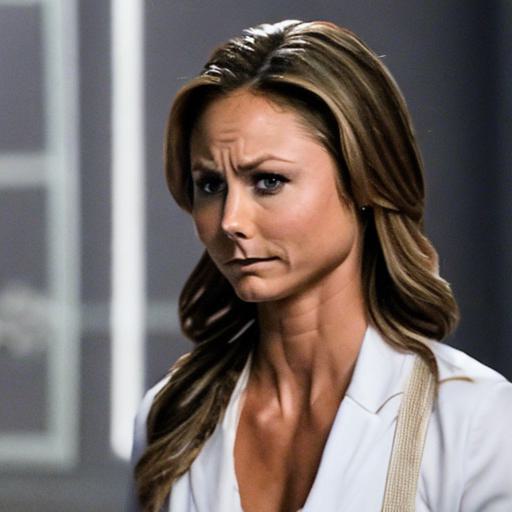} &
        \includegraphics[width=0.08\textwidth]{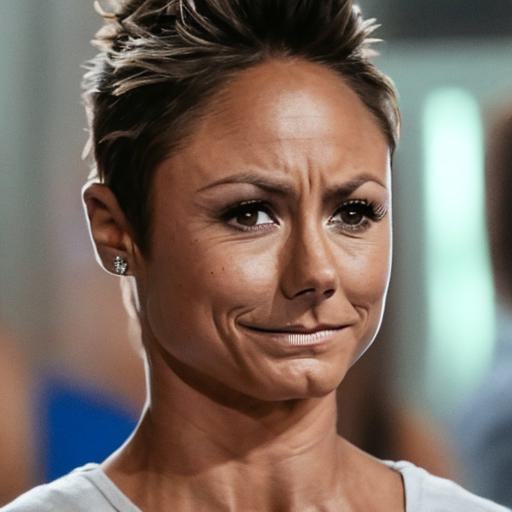} \\
        
        \raisebox{0.325in}{\begin{tabular}{c} ``$S_*$ with \\ \\[-0.05cm] a puzzled \\ \\[-0.05cm]expression''\end{tabular}} &
        \includegraphics[width=0.08\textwidth]{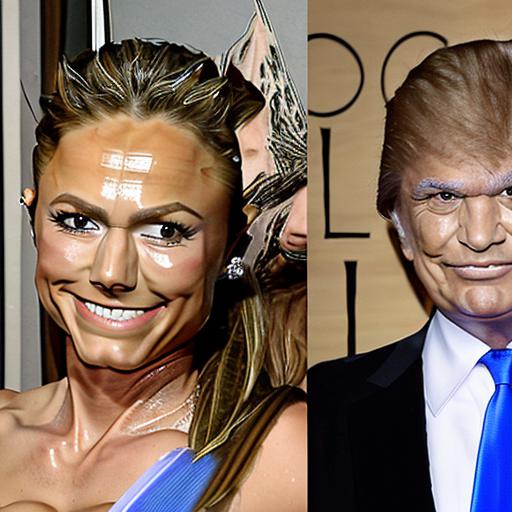} &
        \includegraphics[width=0.08\textwidth]{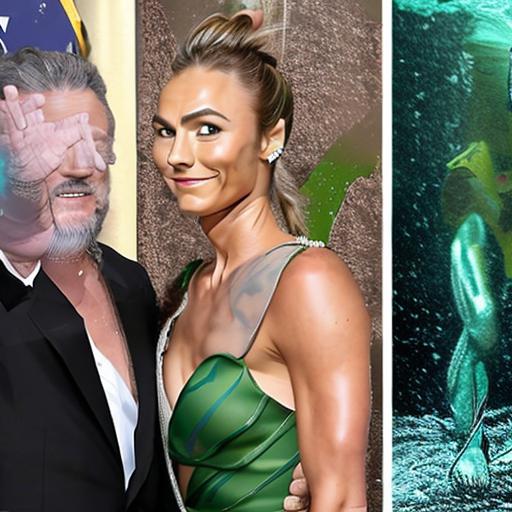} &
        \hspace{0.05cm}
        \includegraphics[width=0.08\textwidth]{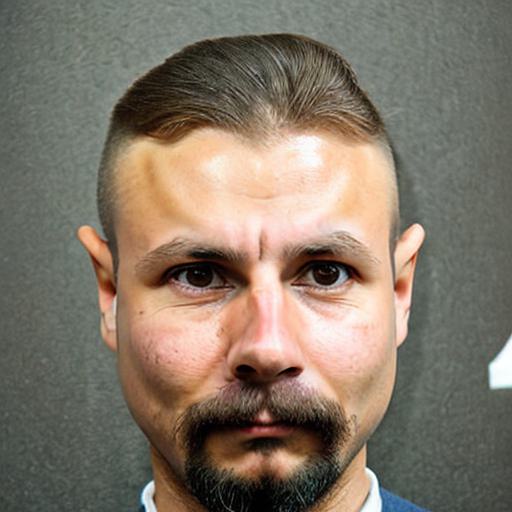} &
        \includegraphics[width=0.08\textwidth]{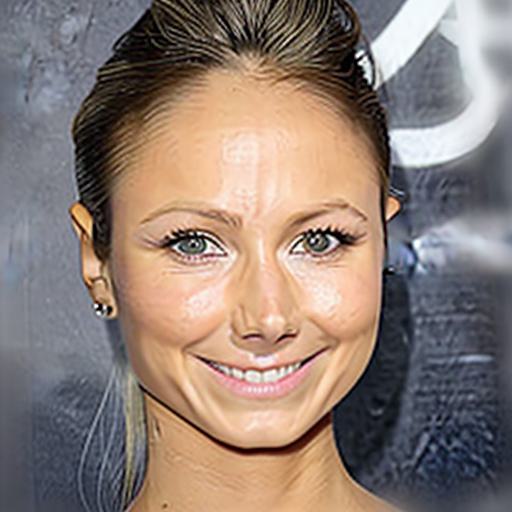} &
        \hspace{0.05cm}
        \includegraphics[width=0.08\textwidth]{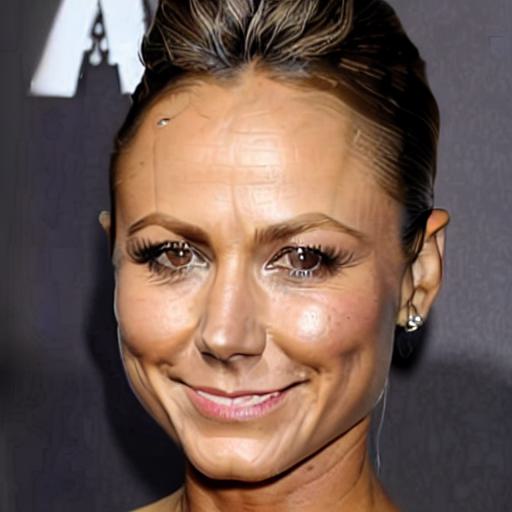} &
        \includegraphics[width=0.08\textwidth]{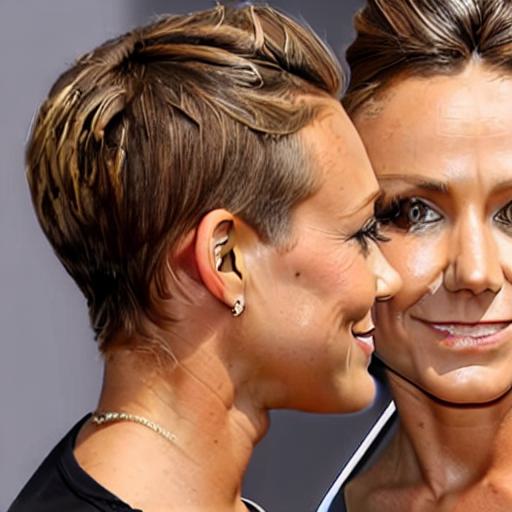} &
        \hspace{0.05cm}
        \includegraphics[width=0.08\textwidth]{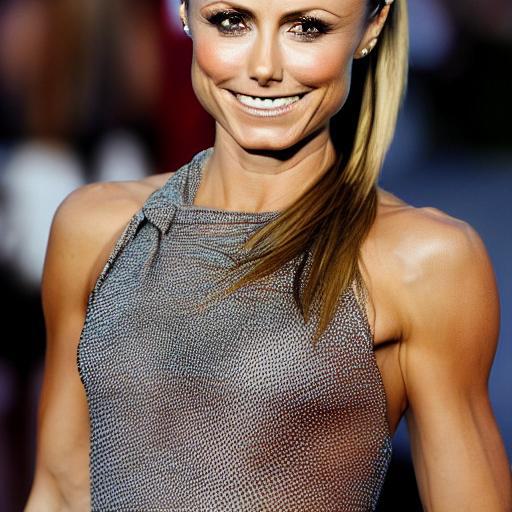} &
        \includegraphics[width=0.08\textwidth]{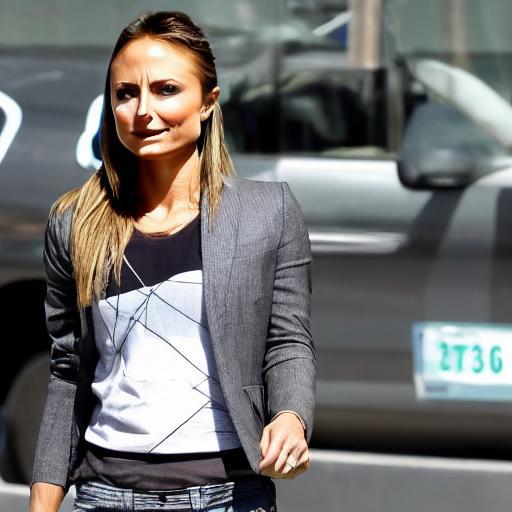} &
        \hspace{0.05cm}
        \includegraphics[width=0.08\textwidth]{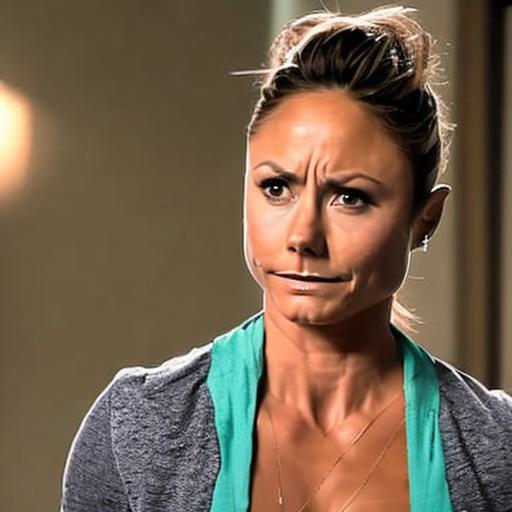} &
        \includegraphics[width=0.08\textwidth]{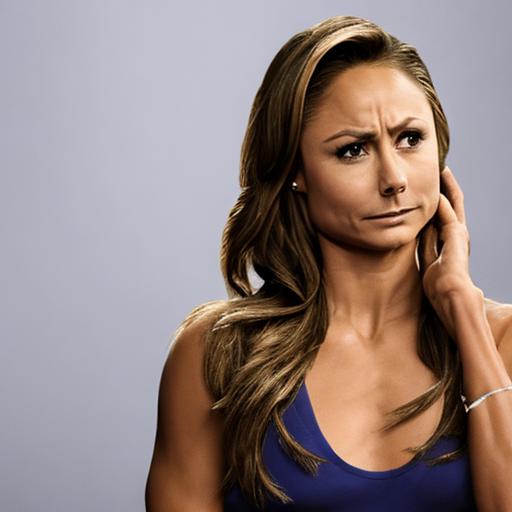} \\ \\
        
        \includegraphics[width=0.08\textwidth]{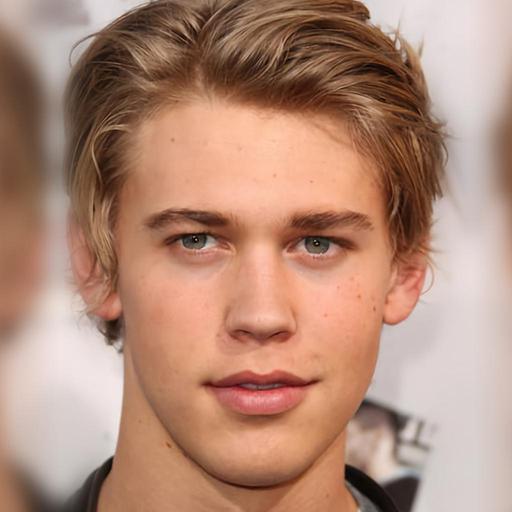} &
        \includegraphics[width=0.08\textwidth]{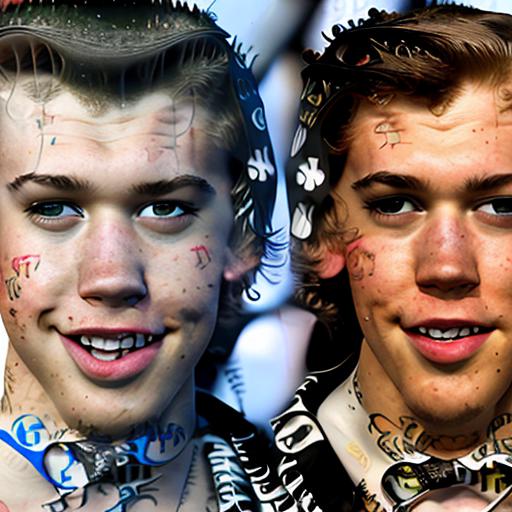} &
        \includegraphics[width=0.08\textwidth]{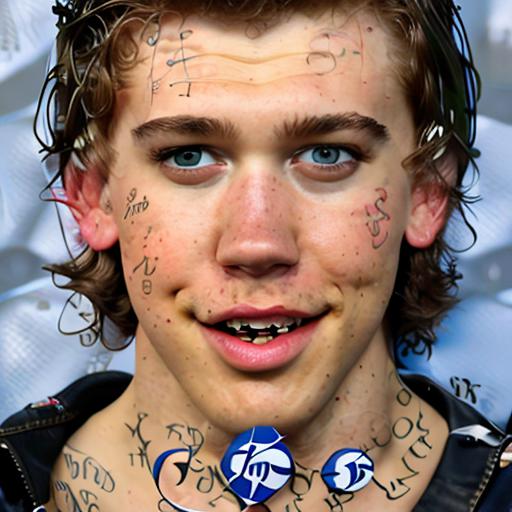} &
        \hspace{0.05cm}
        \includegraphics[width=0.08\textwidth]{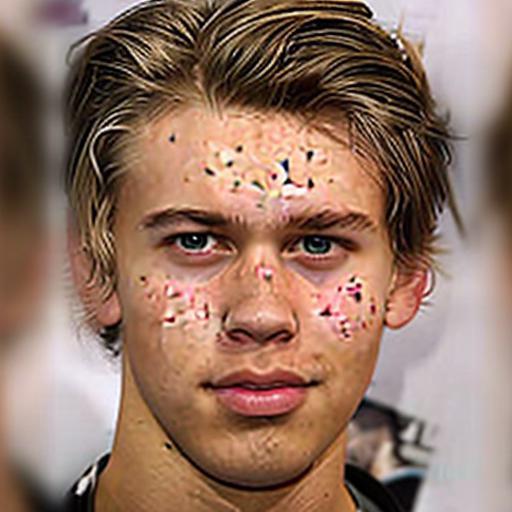} &
        \includegraphics[width=0.08\textwidth]{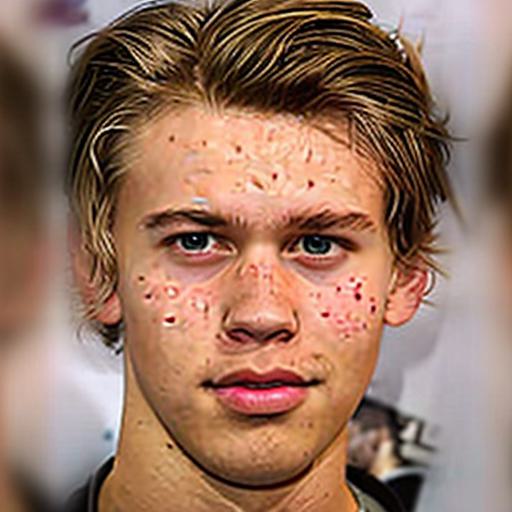} &
        \hspace{0.05cm}
        \includegraphics[width=0.08\textwidth]{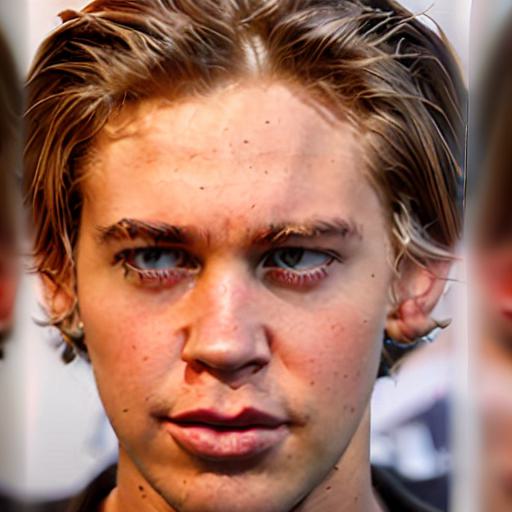} &
        \includegraphics[width=0.08\textwidth]{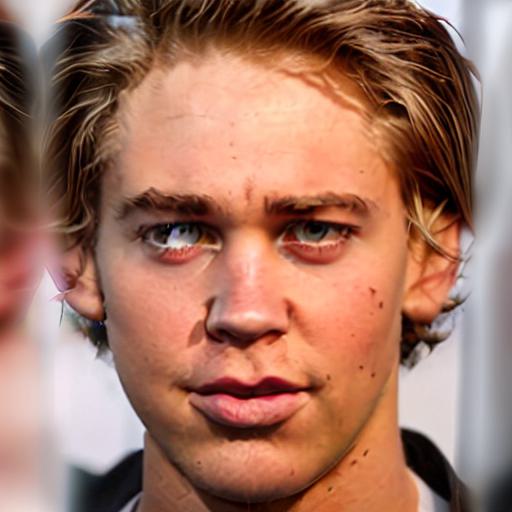} &
        \hspace{0.05cm}
        \includegraphics[width=0.08\textwidth]{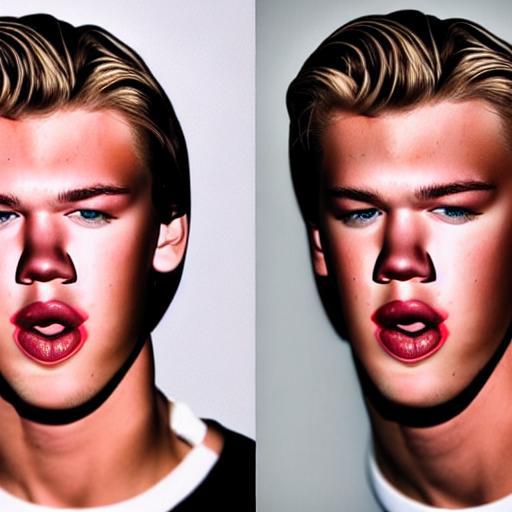} &
        \includegraphics[width=0.08\textwidth]{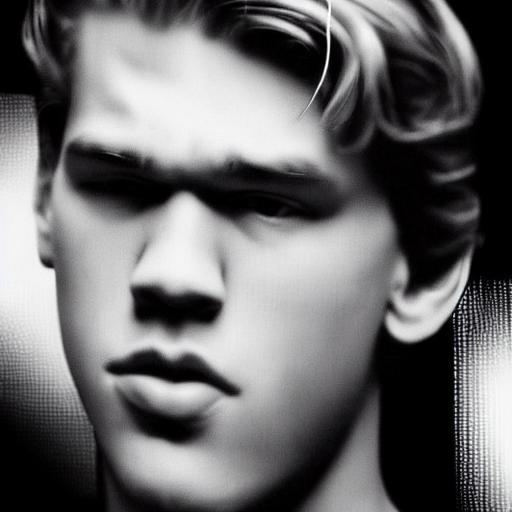} &
        \hspace{0.05cm}
        \includegraphics[width=0.08\textwidth]{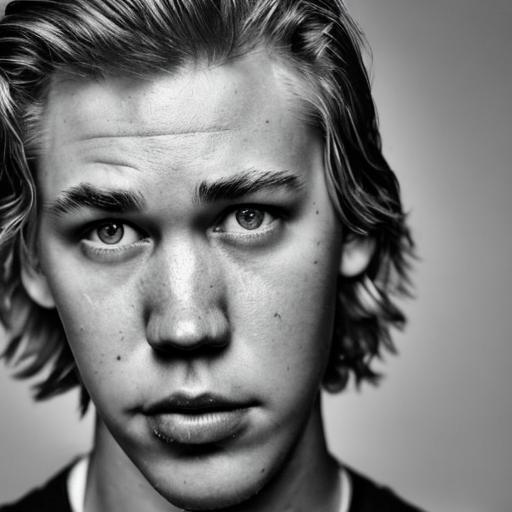} &
        \includegraphics[width=0.08\textwidth]{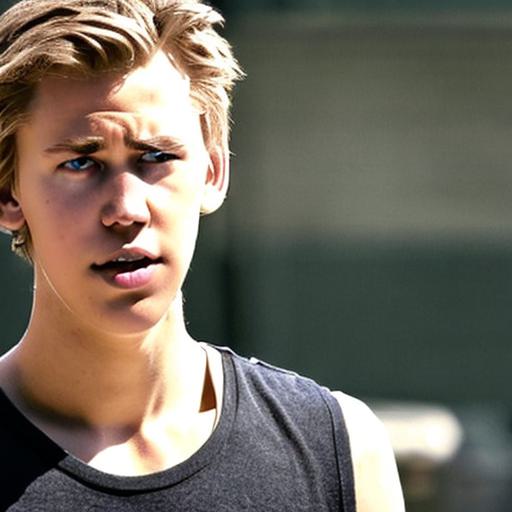} \\
        
        \raisebox{0.325in}{\begin{tabular}{c} ``$S_*$ with \\ \\[-0.05cm] an angry \\ \\[-0.05cm]expression''\end{tabular}} &
        \includegraphics[width=0.08\textwidth]{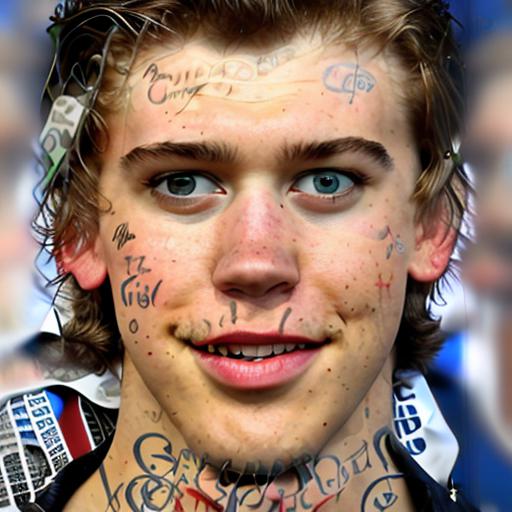} &
        \includegraphics[width=0.08\textwidth]{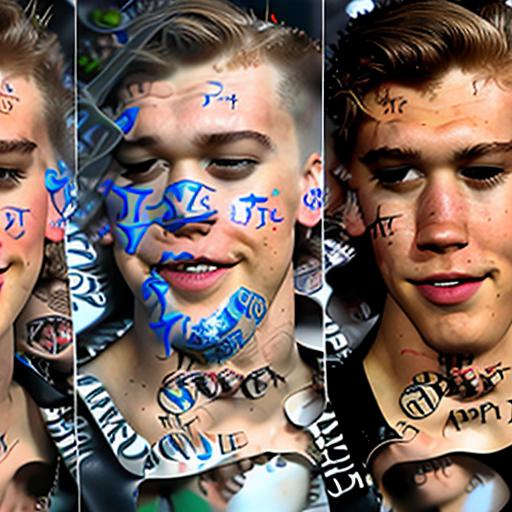} &
        \hspace{0.05cm}
        \includegraphics[width=0.08\textwidth]{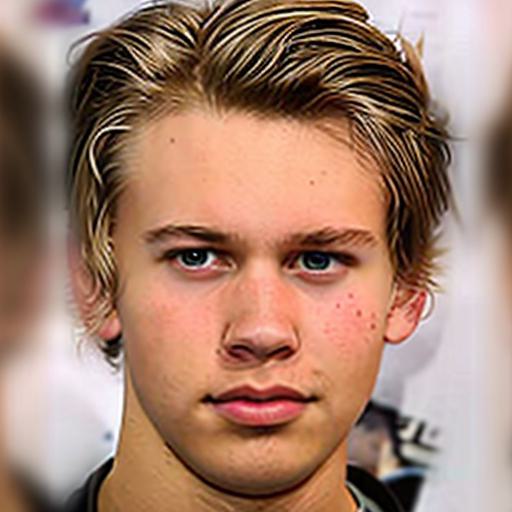} &
        \includegraphics[width=0.08\textwidth]{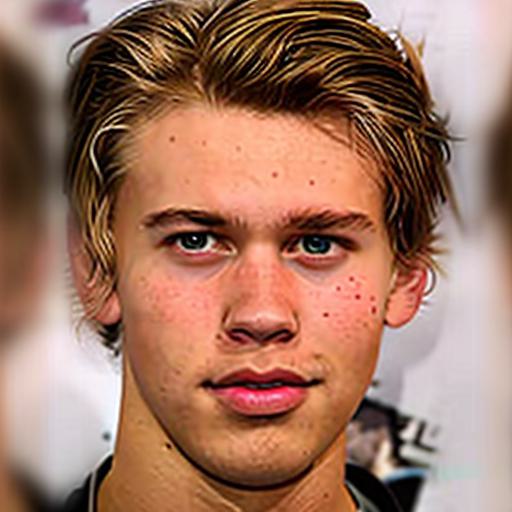} &
        \hspace{0.05cm}
        \includegraphics[width=0.08\textwidth]{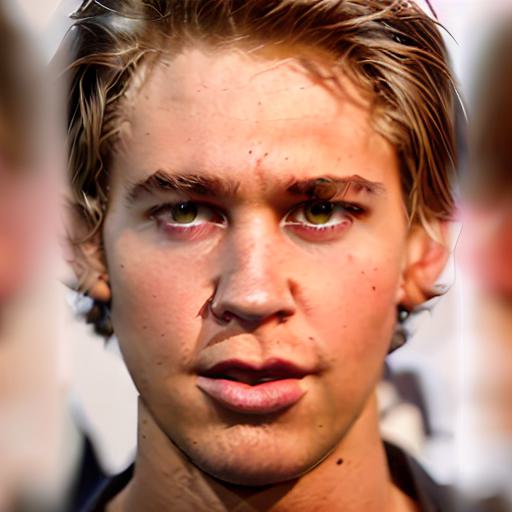} &
        \includegraphics[width=0.08\textwidth]{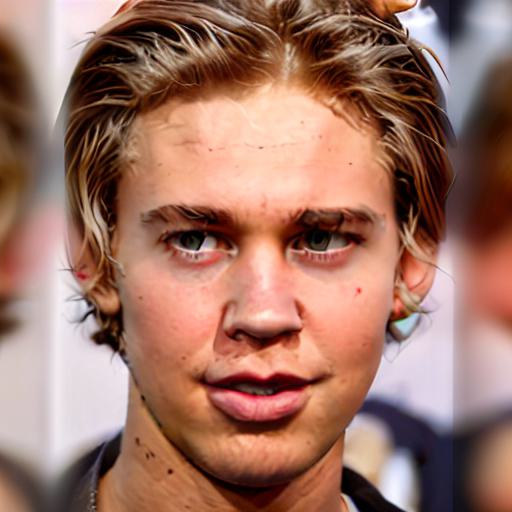} &
        \hspace{0.05cm}
        \includegraphics[width=0.08\textwidth]{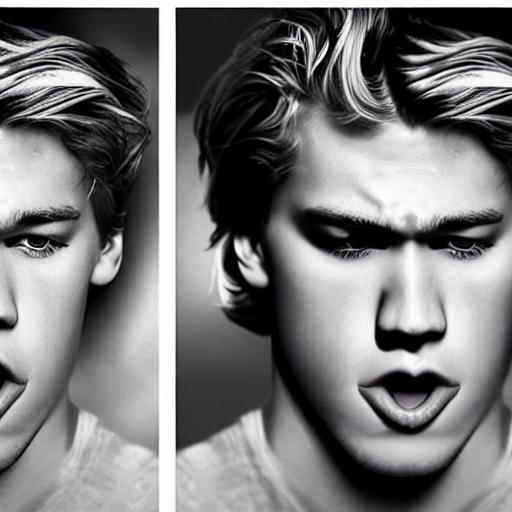} &
        \includegraphics[width=0.08\textwidth]{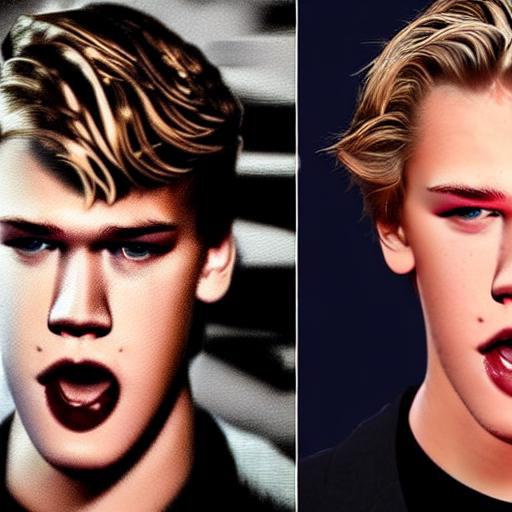} &
        \hspace{0.05cm}
        \includegraphics[width=0.08\textwidth]{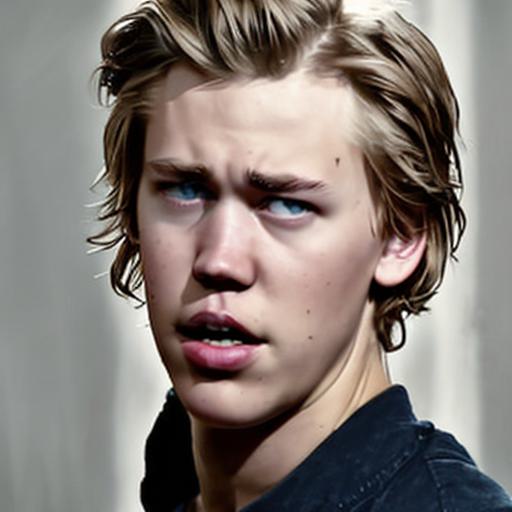} &
        \includegraphics[width=0.08\textwidth]{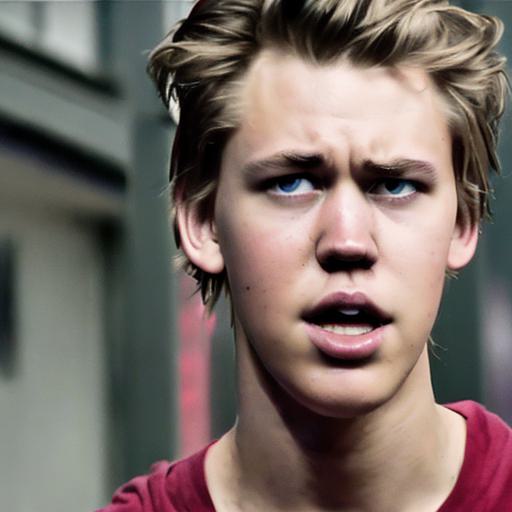} \\ \\

        \includegraphics[width=0.08\textwidth]{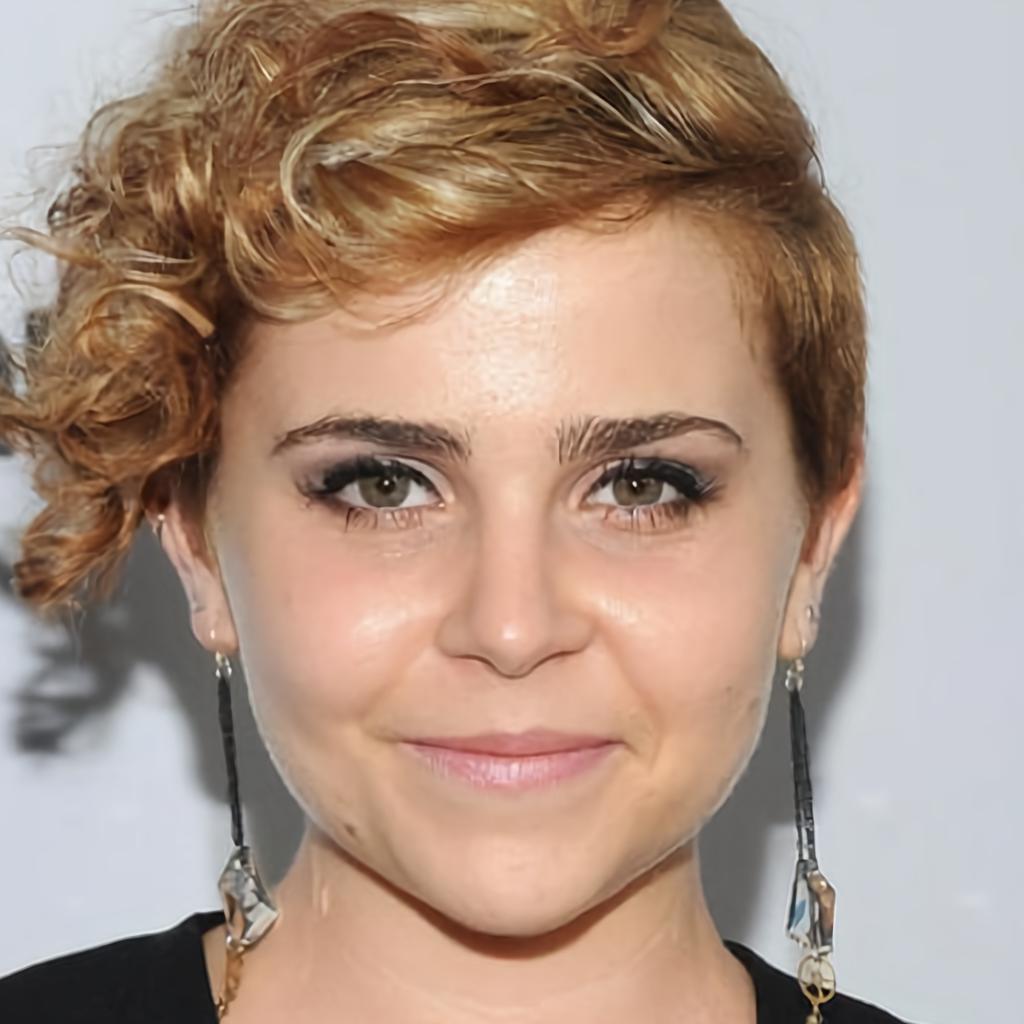} &
        \includegraphics[width=0.08\textwidth]{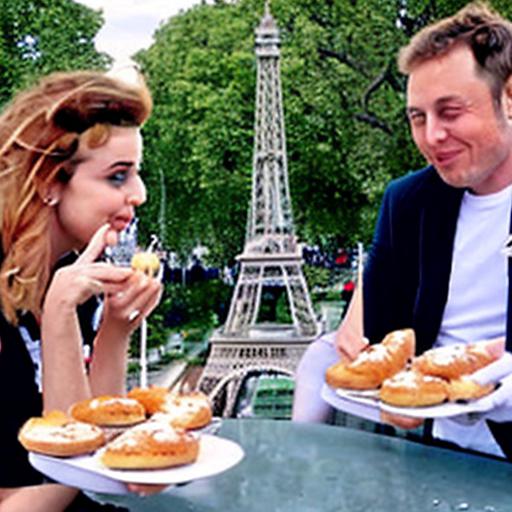} &
        \includegraphics[width=0.08\textwidth]{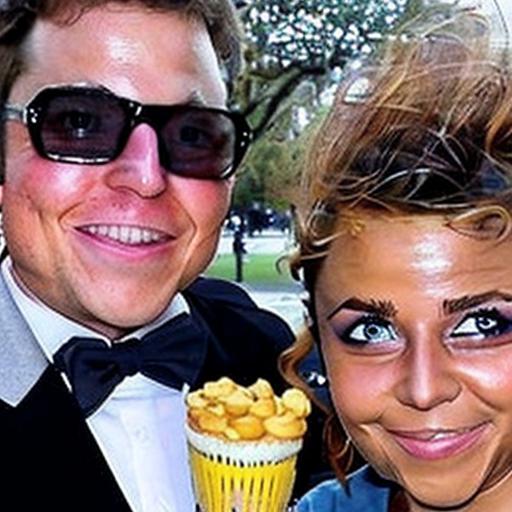} &
        \hspace{0.05cm}
        \includegraphics[width=0.08\textwidth]{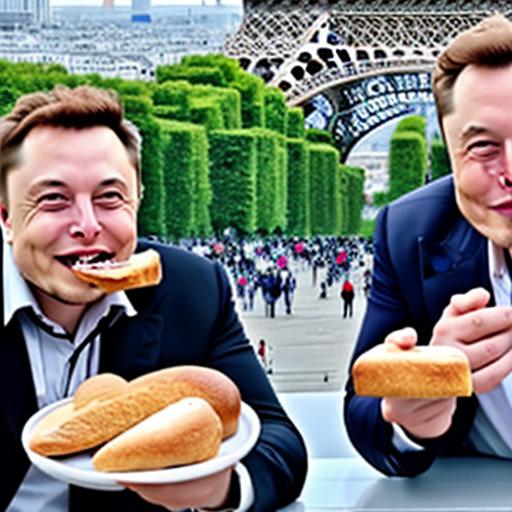} &
        \includegraphics[width=0.08\textwidth]{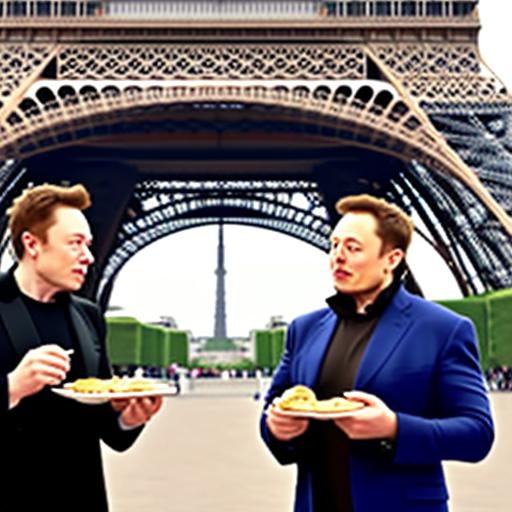} &
        \hspace{0.05cm}
        \includegraphics[width=0.08\textwidth]{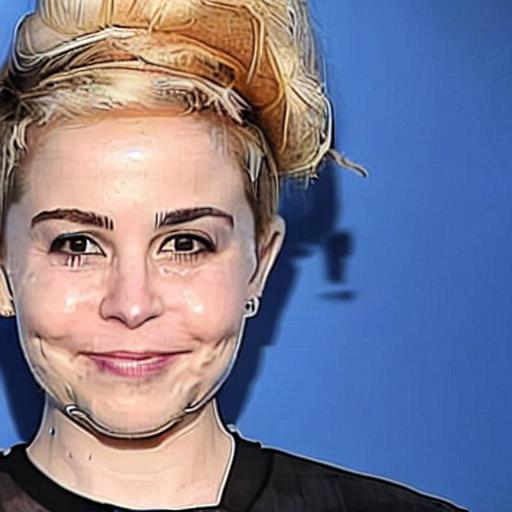} &
        \includegraphics[width=0.08\textwidth]{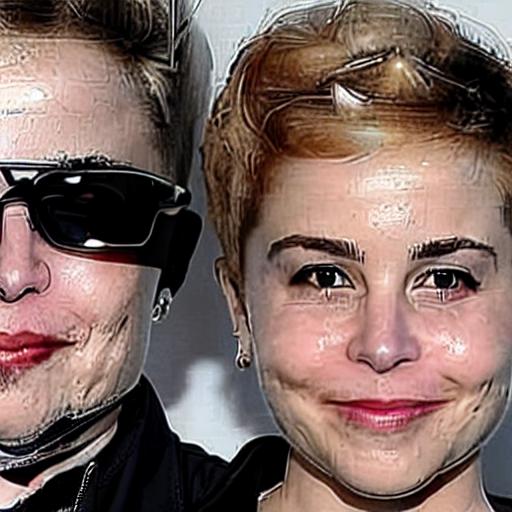} &
        \hspace{0.05cm}
        \includegraphics[width=0.08\textwidth]{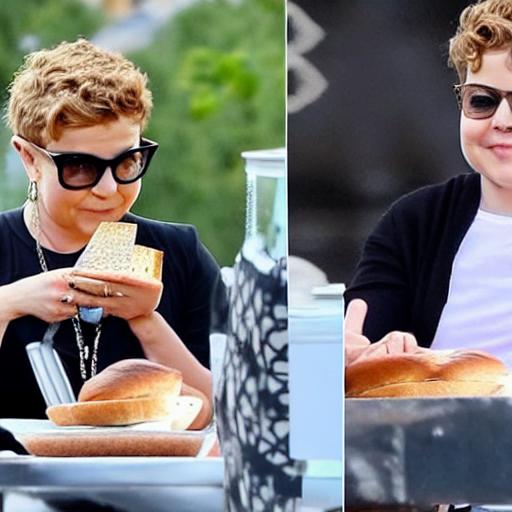} &
        \includegraphics[width=0.08\textwidth]{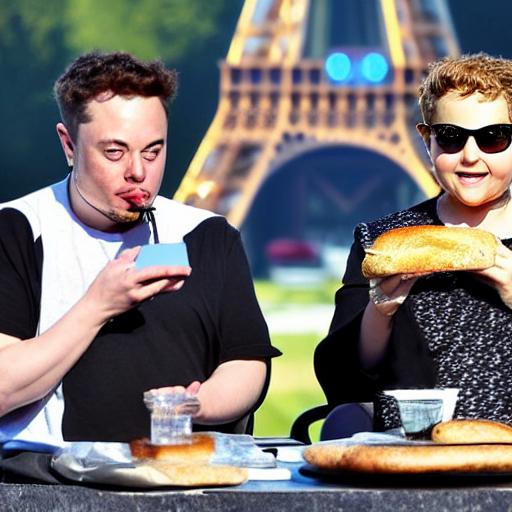} &
        \hspace{0.05cm}
        \includegraphics[width=0.08\textwidth]{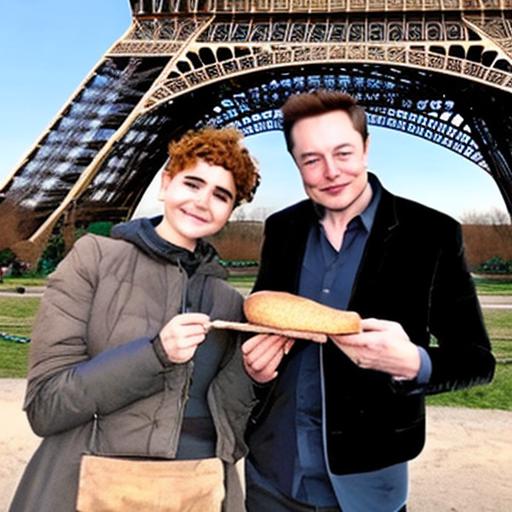} &
        \includegraphics[width=0.08\textwidth]{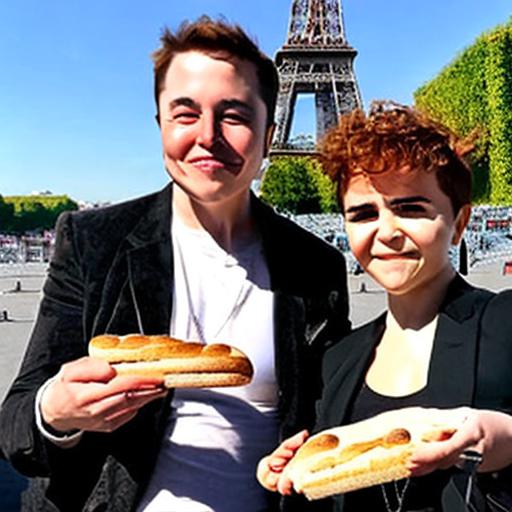} \\
        
        \raisebox{0.325in}{
        \begin{tabular}{c} ``$S_*$ and Elon \\ \\
        [-0.05cm]  Musk  are eating \\ \\
        [-0.05cm] bread in front of  \\ \\
        [-0.05cm] the Eiffel Tower''\end{tabular}} &
        \includegraphics[width=0.08\textwidth]{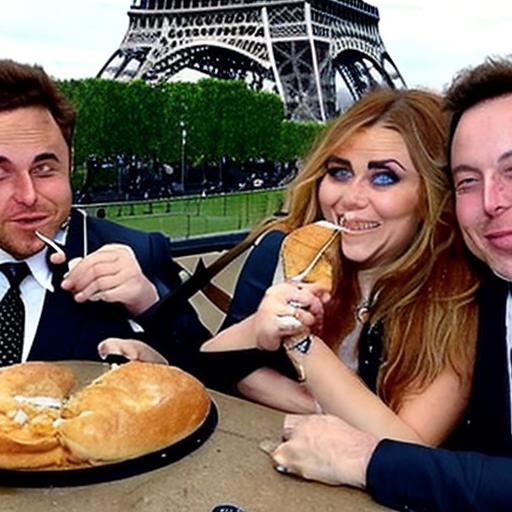} &
        \includegraphics[width=0.08\textwidth]{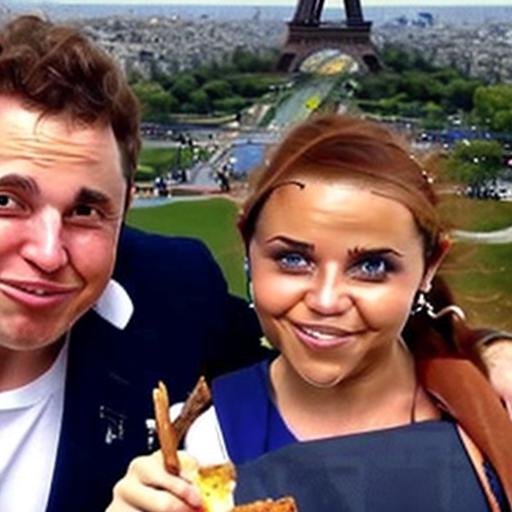} &
        \hspace{0.05cm}
        \includegraphics[width=0.08\textwidth]{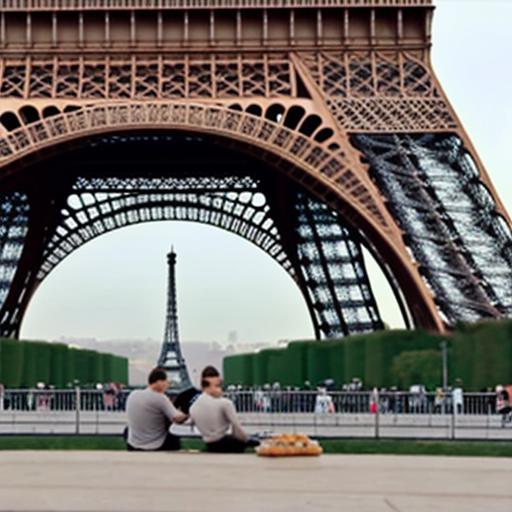} &
        \includegraphics[width=0.08\textwidth]{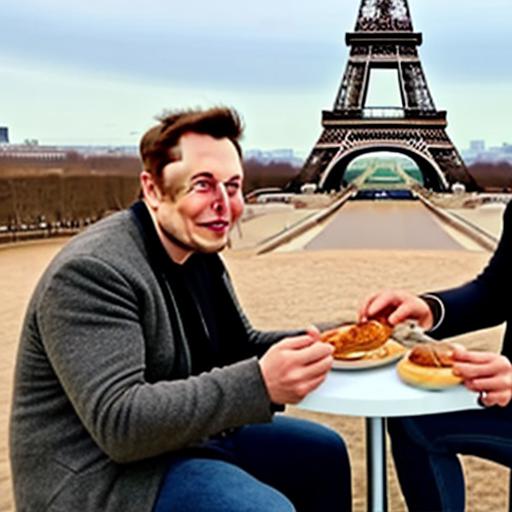} &
        \hspace{0.05cm}
        \includegraphics[width=0.08\textwidth]{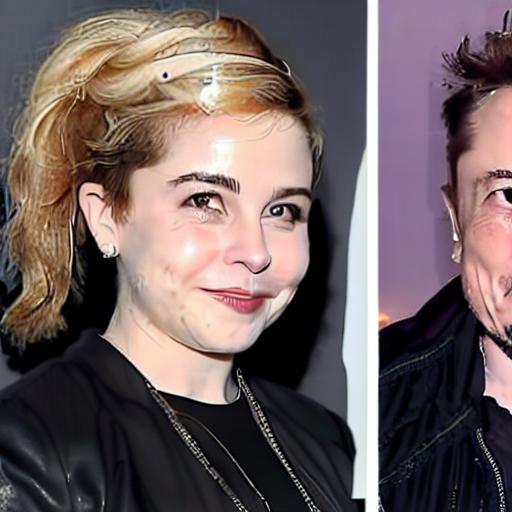} &
        \includegraphics[width=0.08\textwidth]{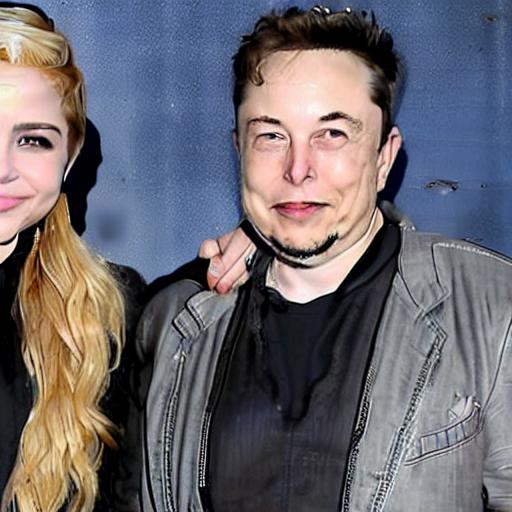} &
        \hspace{0.05cm}
        \includegraphics[width=0.08\textwidth]{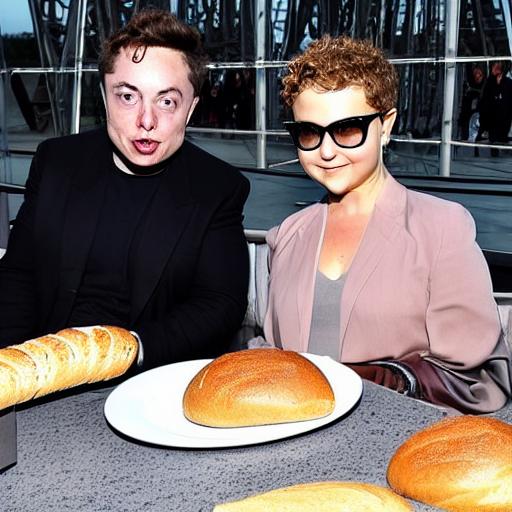} &
        \includegraphics[width=0.08\textwidth]{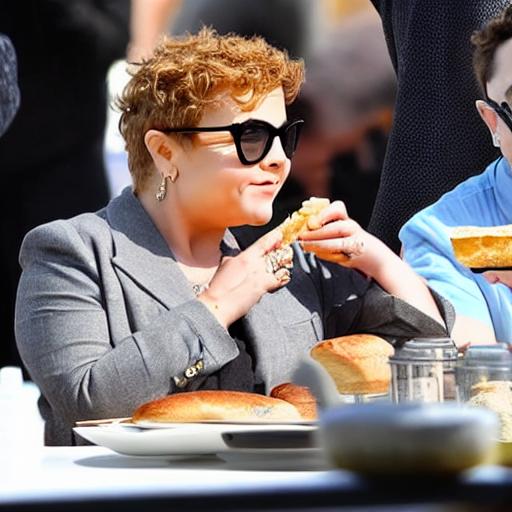} &
        \hspace{0.05cm}
        \includegraphics[width=0.08\textwidth]{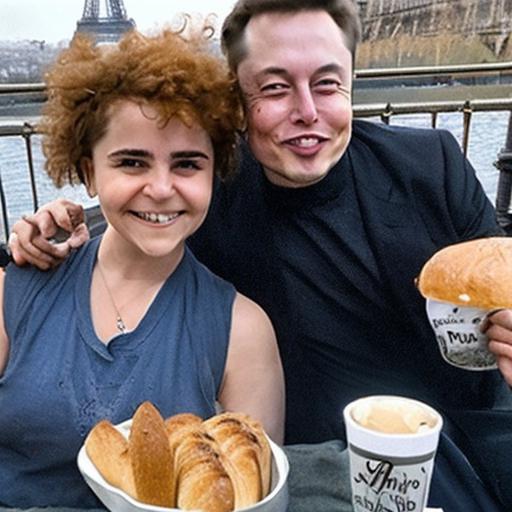} &
        \includegraphics[width=0.08\textwidth]{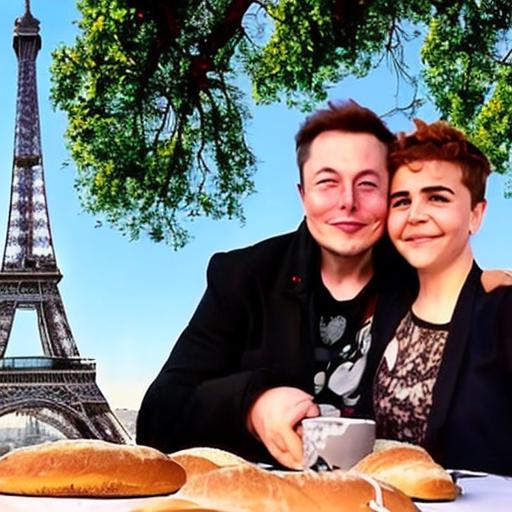} \\ \\
        
        \includegraphics[width=0.08\textwidth]{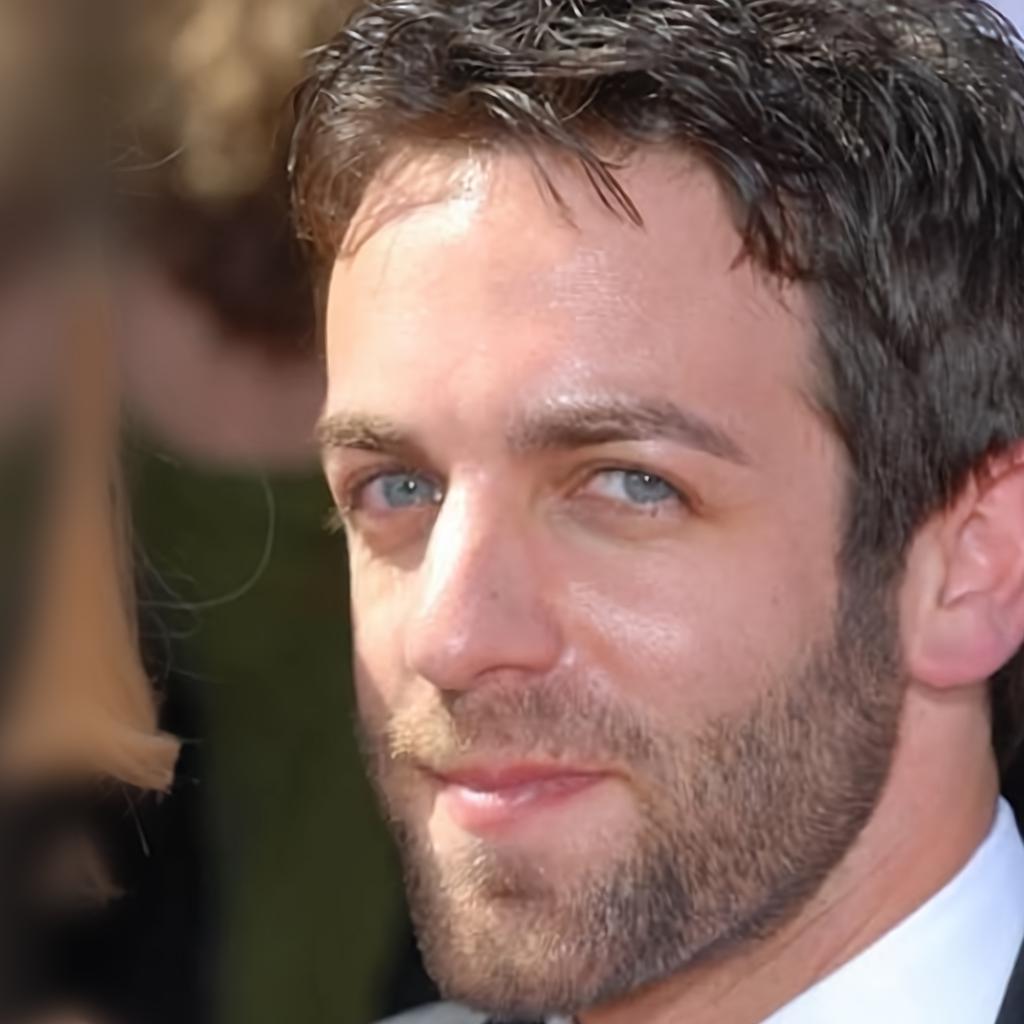} &
        \includegraphics[width=0.08\textwidth]{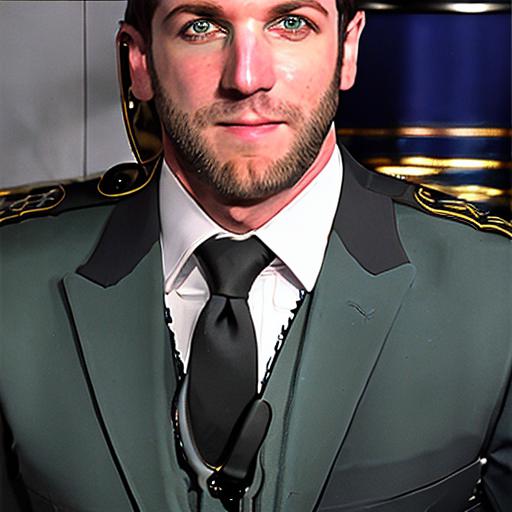} &
        \includegraphics[width=0.08\textwidth]{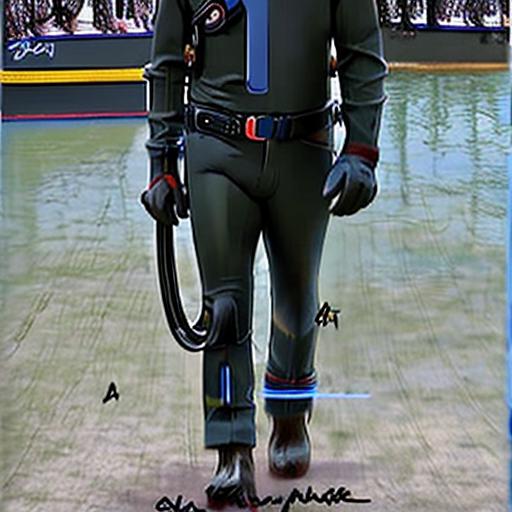} &
        \hspace{0.05cm}
        \includegraphics[width=0.08\textwidth]{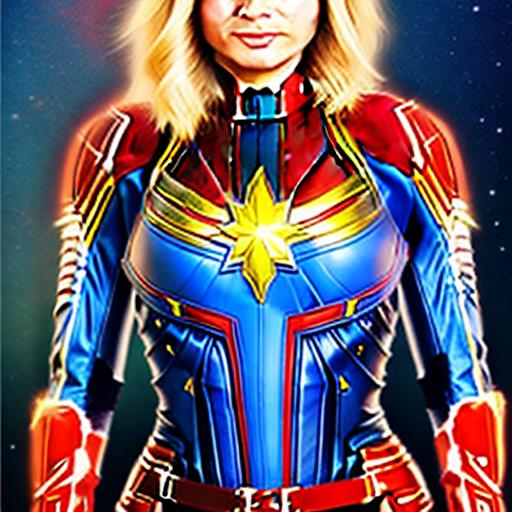} &
        \includegraphics[width=0.08\textwidth]{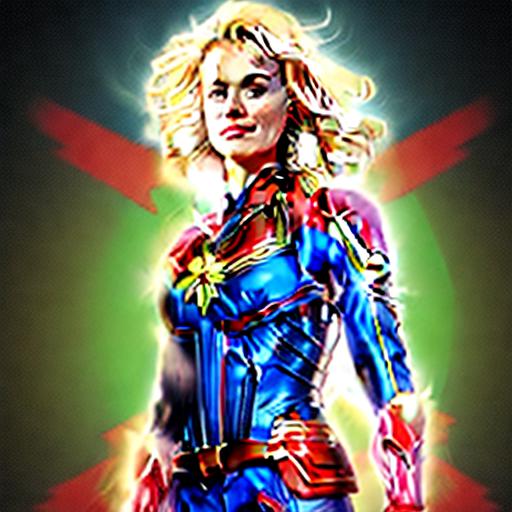} &
        \hspace{0.05cm}
        \includegraphics[width=0.08\textwidth]{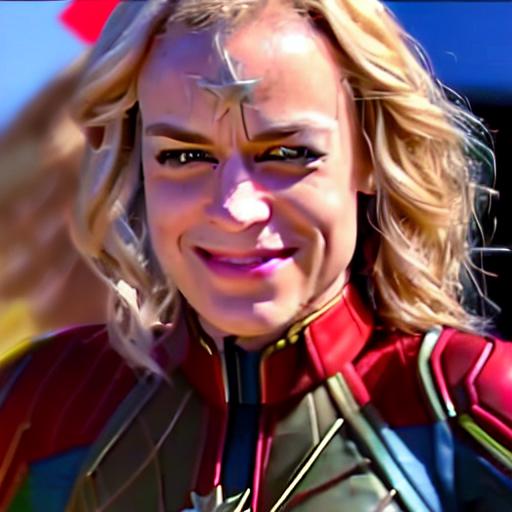} &
        \includegraphics[width=0.08\textwidth]{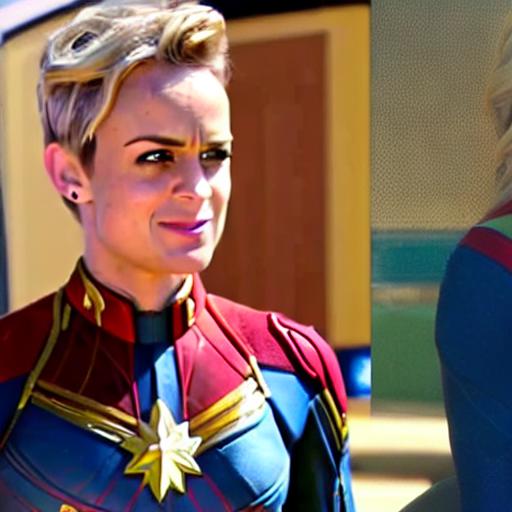} &
        \hspace{0.05cm}
        \includegraphics[width=0.08\textwidth]{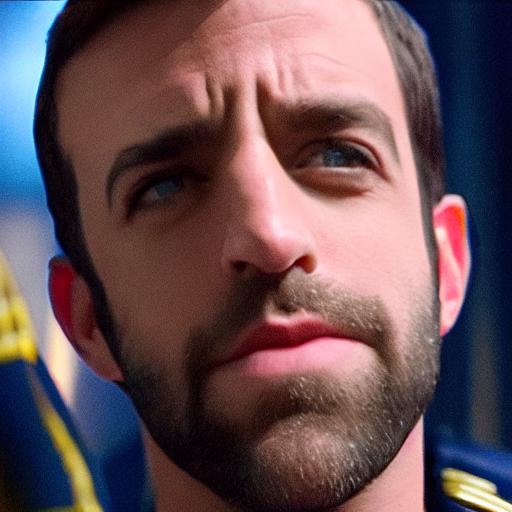} &
        \includegraphics[width=0.08\textwidth]{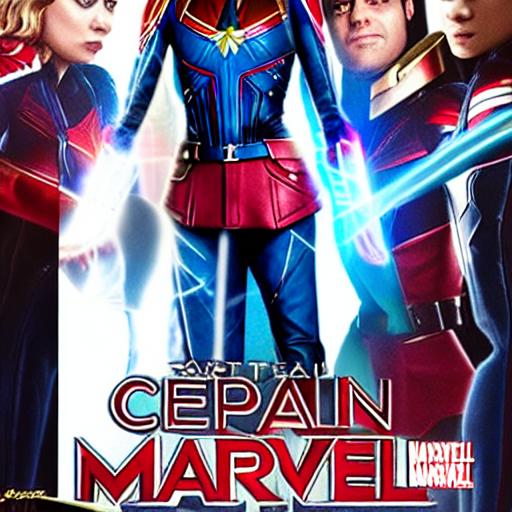} &
        \hspace{0.05cm}
        \includegraphics[width=0.08\textwidth]{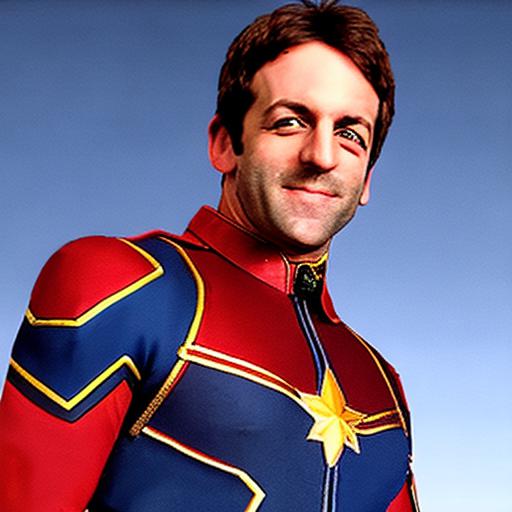} &
        \includegraphics[width=0.08\textwidth]{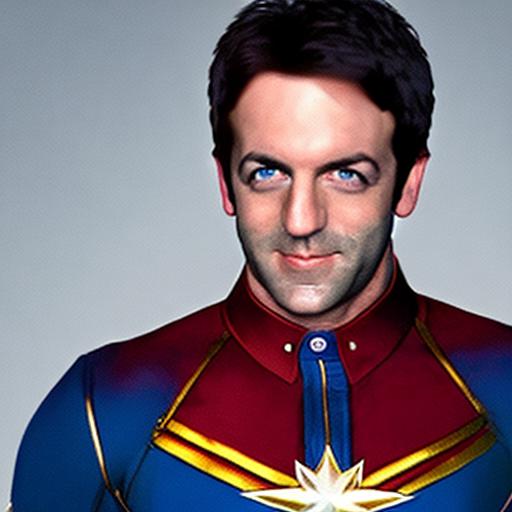} \\
        
        \raisebox{0.325in}{\begin{tabular}{c} ``$S_*$ as \\ Captain \\ Marvel''\end{tabular}} &
        \includegraphics[width=0.08\textwidth]{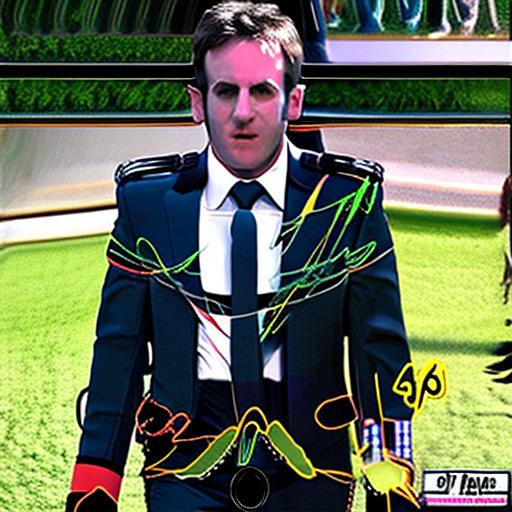} &
        \includegraphics[width=0.08\textwidth]{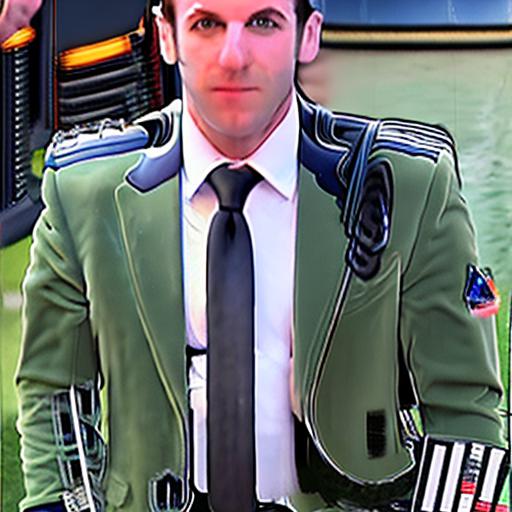} &
        \hspace{0.05cm}
        \includegraphics[width=0.08\textwidth]{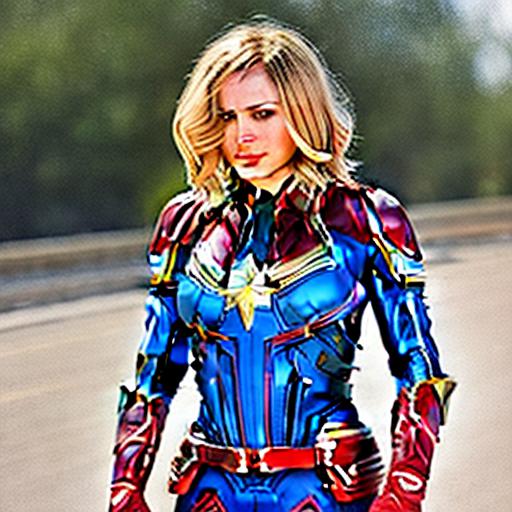} &
        \includegraphics[width=0.08\textwidth]{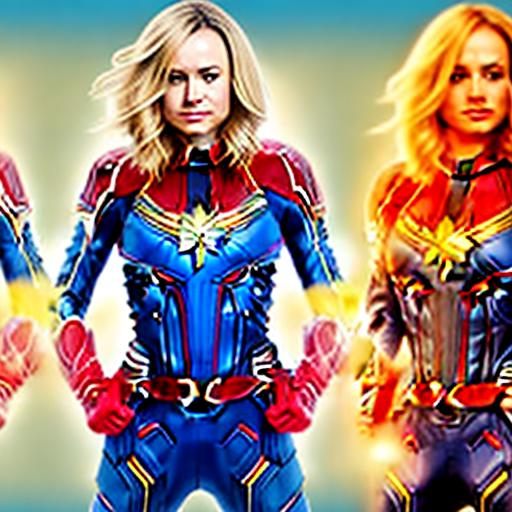} &
        \hspace{0.05cm}
        \includegraphics[width=0.08\textwidth]{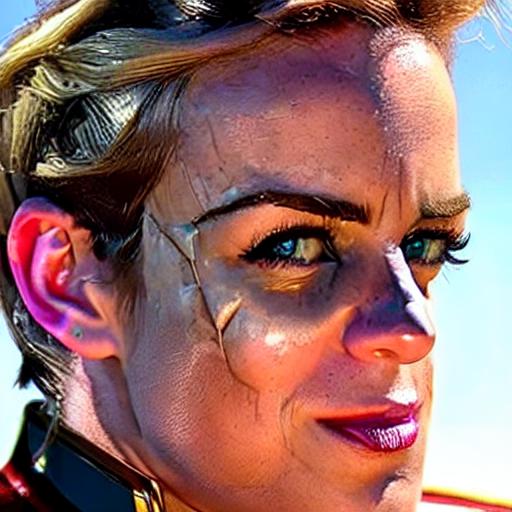} &
        \includegraphics[width=0.08\textwidth]{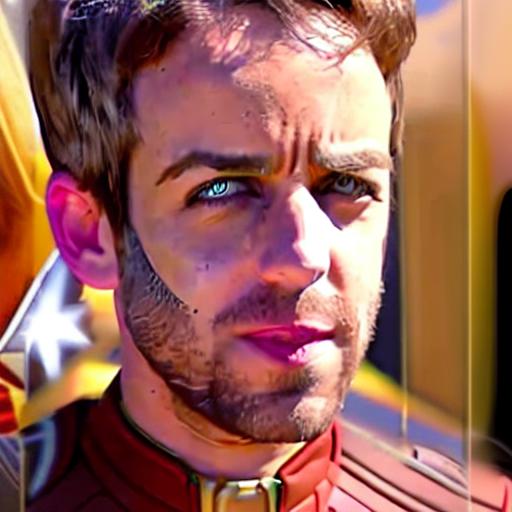} &
        \hspace{0.05cm}
        \includegraphics[width=0.08\textwidth]{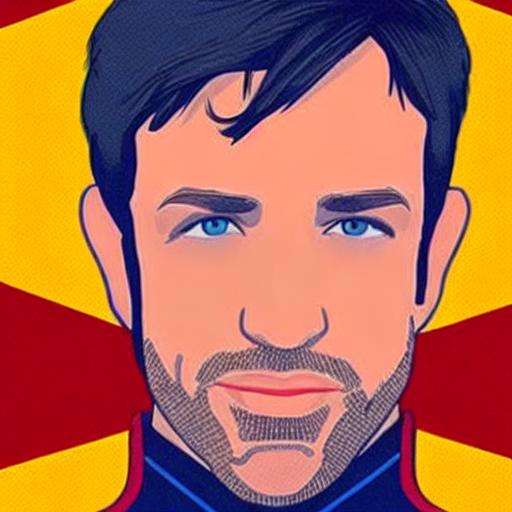} &
        \includegraphics[width=0.08\textwidth]{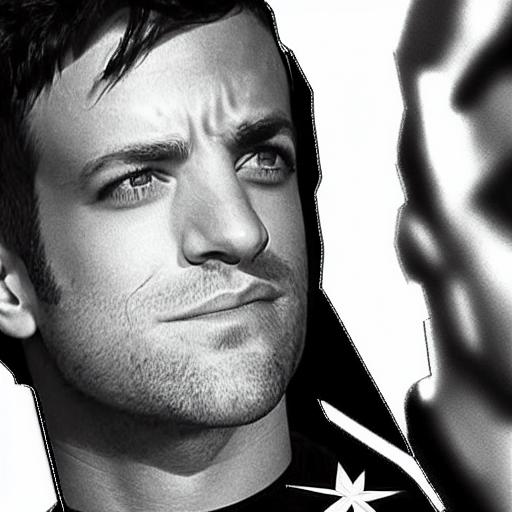} &
        \hspace{0.05cm}
        \includegraphics[width=0.08\textwidth]{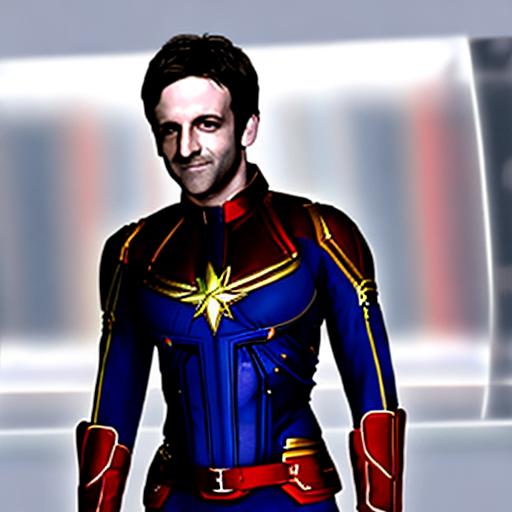} &
        \includegraphics[width=0.08\textwidth]{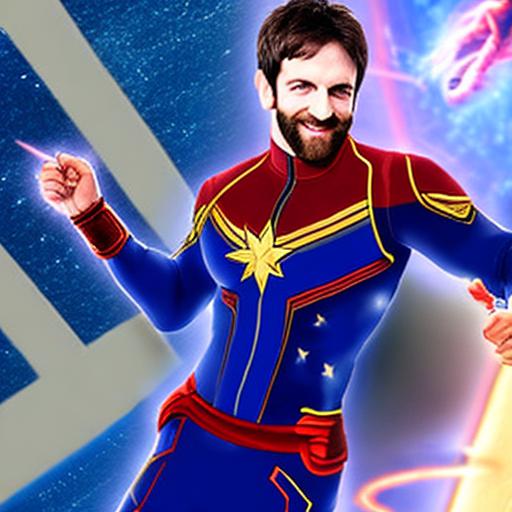} \\ \\
        
        \includegraphics[width=0.08\textwidth]{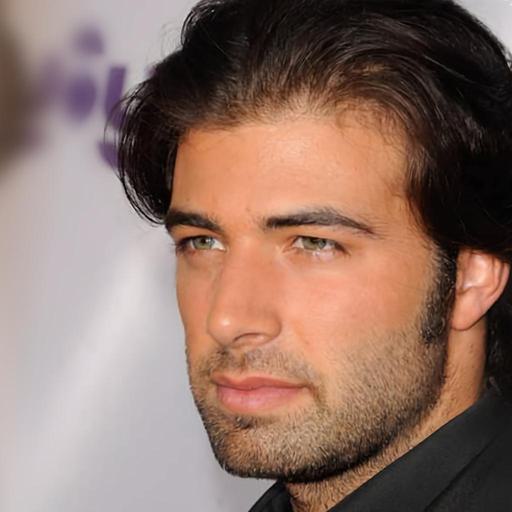} &
        \includegraphics[width=0.08\textwidth]{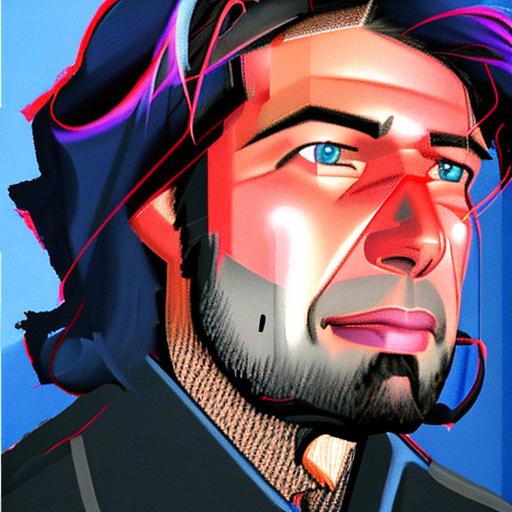} &
        \includegraphics[width=0.08\textwidth]{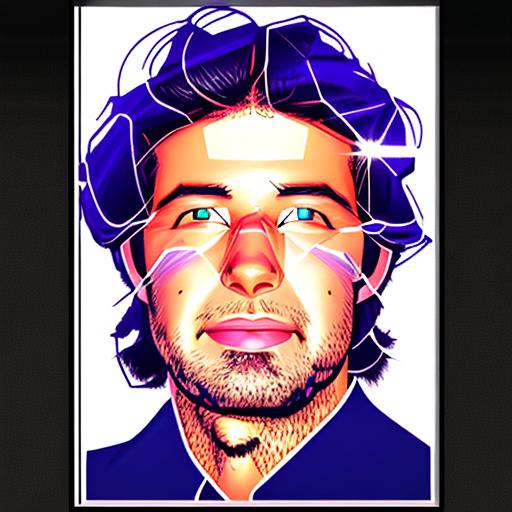} &
        \hspace{0.05cm}
        \includegraphics[width=0.08\textwidth]{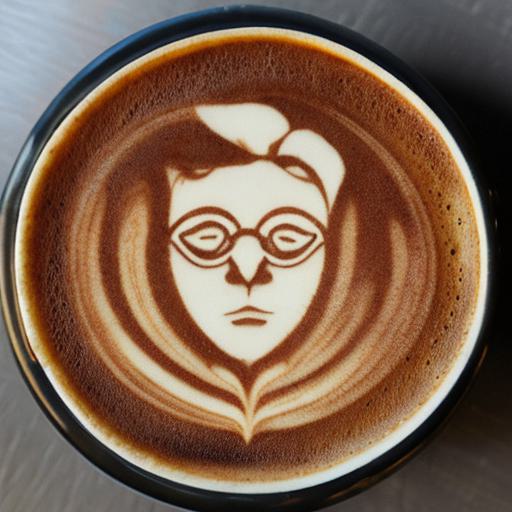} &
        \includegraphics[width=0.08\textwidth]{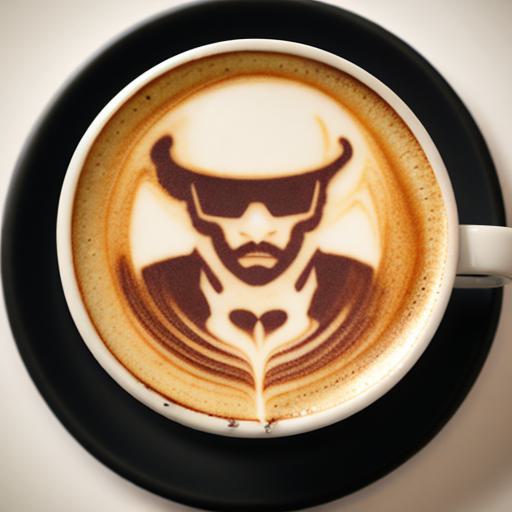} &
        \hspace{0.05cm}
        \includegraphics[width=0.08\textwidth]{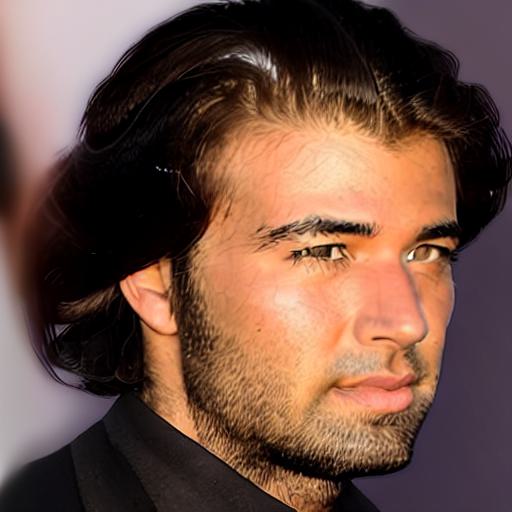} &
        \includegraphics[width=0.08\textwidth]{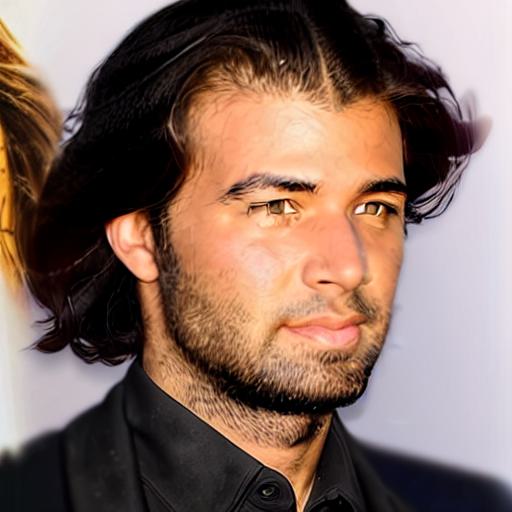} &
        \hspace{0.05cm}
        \includegraphics[width=0.08\textwidth]{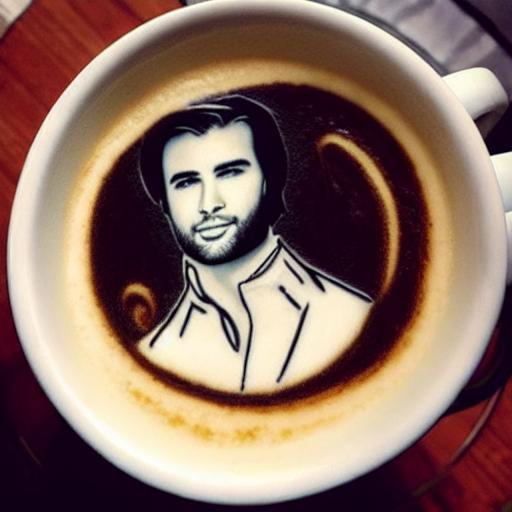} &
        \includegraphics[width=0.08\textwidth]{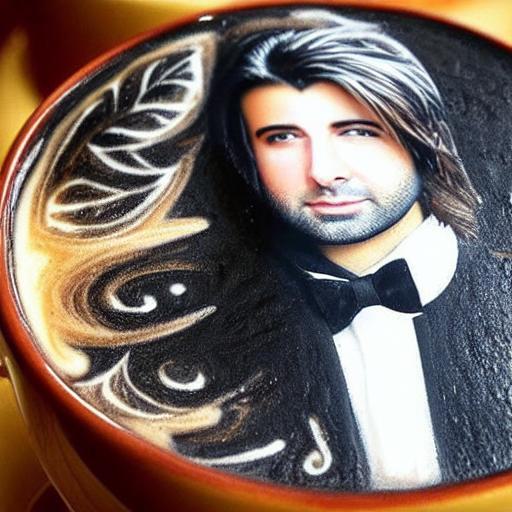} &
        \hspace{0.05cm}
        \includegraphics[width=0.08\textwidth]{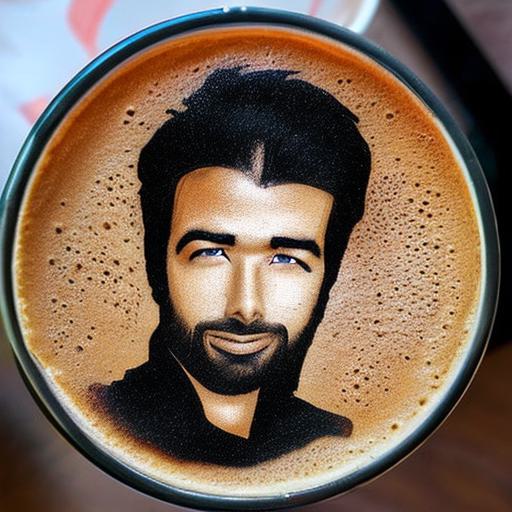} &
        \includegraphics[width=0.08\textwidth]{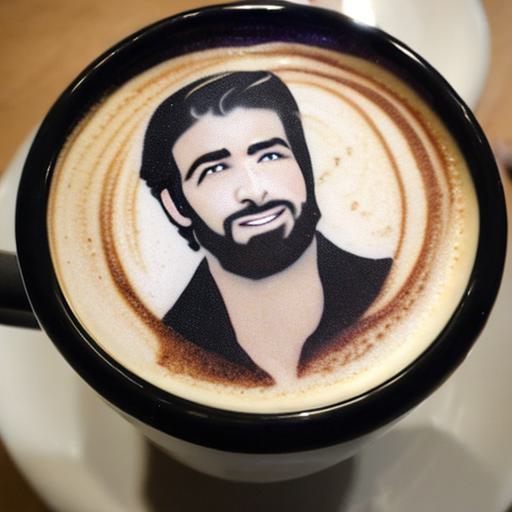} \\
        
        \raisebox{0.325in}{\begin{tabular}{c} ``$S_*$ latte \\ \\[-0.05cm] art''\end{tabular}} &
        \includegraphics[width=0.08\textwidth]{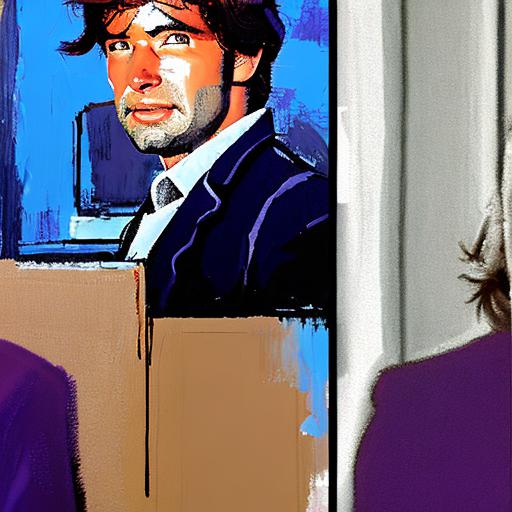} &
        \includegraphics[width=0.08\textwidth]{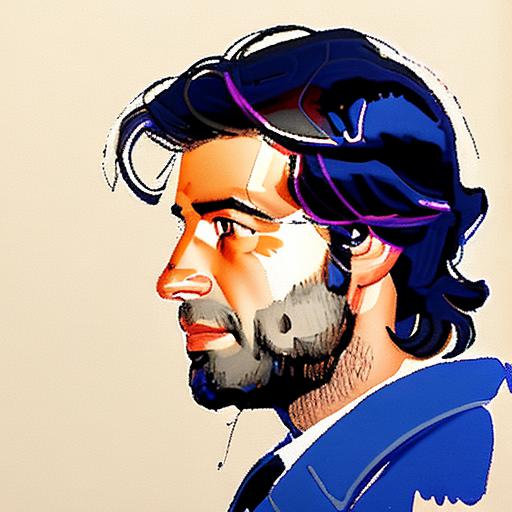} &
        \hspace{0.05cm}
        \includegraphics[width=0.08\textwidth]{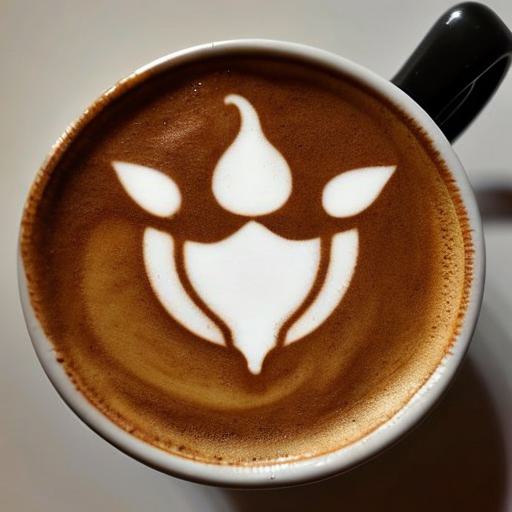} &
        \includegraphics[width=0.08\textwidth]{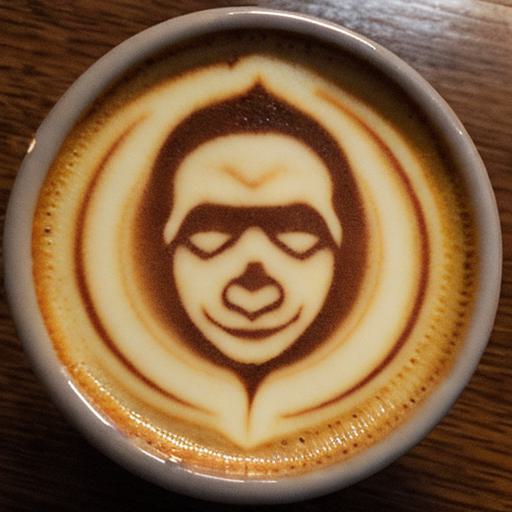} &
        \hspace{0.05cm}
        \includegraphics[width=0.08\textwidth]{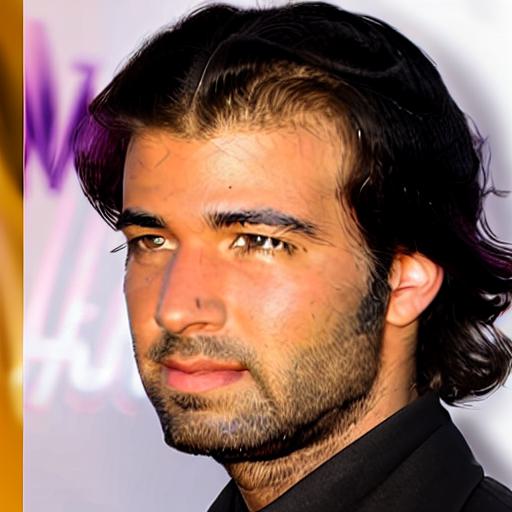} &
        \includegraphics[width=0.08\textwidth]{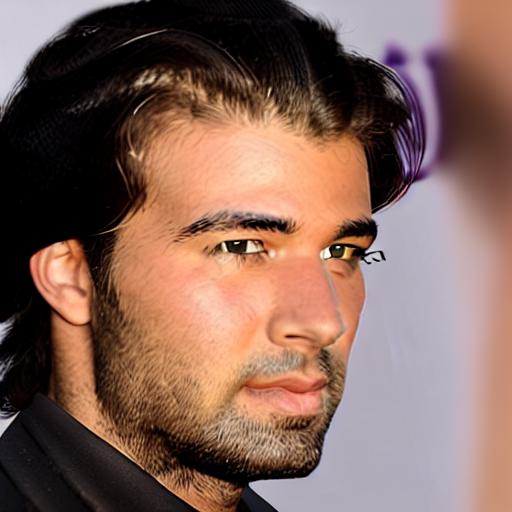} &
        \hspace{0.05cm}
        \includegraphics[width=0.08\textwidth]{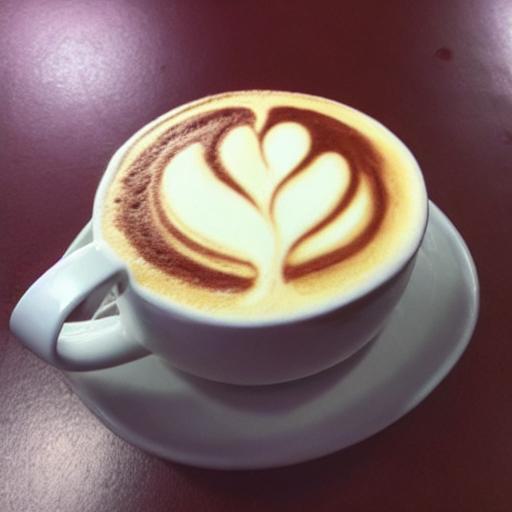} &
        \includegraphics[width=0.08\textwidth]{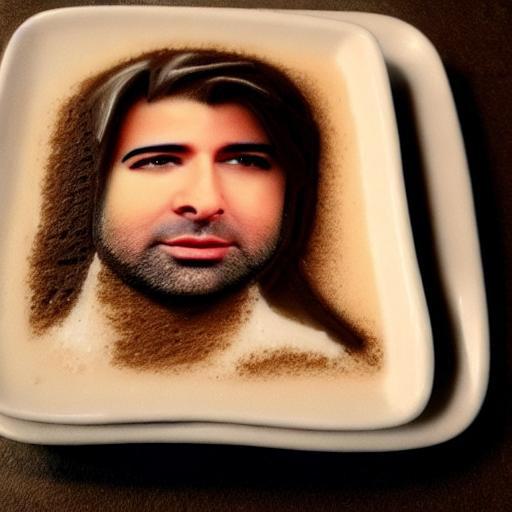} &
        \hspace{0.05cm}
        \includegraphics[width=0.08\textwidth]{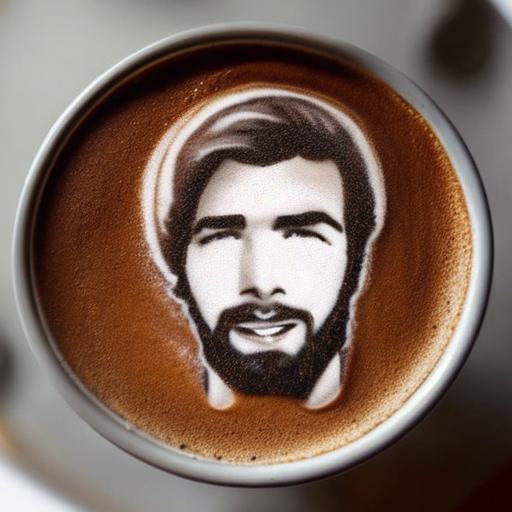} &
        \includegraphics[width=0.08\textwidth]{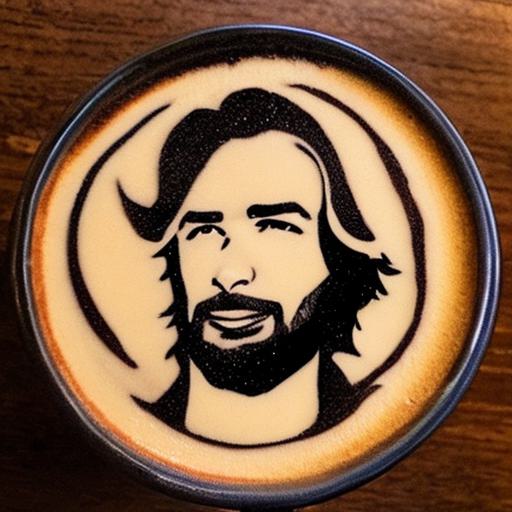} \\ \\

    \end{tabular}
    \\[-0.1cm]
    }
    \caption{
    Qualitative comparisons. Given a single input image, we present four images generated by each method using identical random seeds. Our approach demonstrates superior performance in identity preservation and editability. Notably, Cross Initialization is the only method that successfully edits an individual's facial expression.
    }
    \label{fig:qualitative_comparison}
\end{figure*}

\begin{figure*}
    \centering
    \setlength{\tabcolsep}{0.1pt}
    {\footnotesize
    \begin{tabular}{c@{\hspace{0.25cm}} c@{\hspace{0.25cm}} c@{\hspace{0.25cm}} c@{\hspace{0.25cm}} c@{\hspace{0.25cm}} c@{\hspace{0.25cm}} c@{\hspace{0.25cm}}}

        \includegraphics[width=0.15\textwidth]{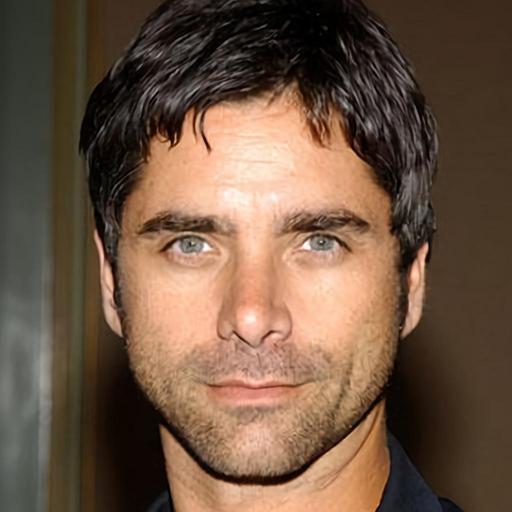} &
        \includegraphics[width=0.15\textwidth]{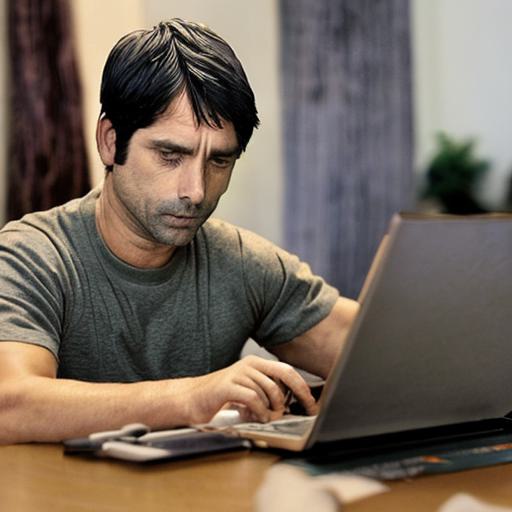} &
        \includegraphics[width=0.15\textwidth]{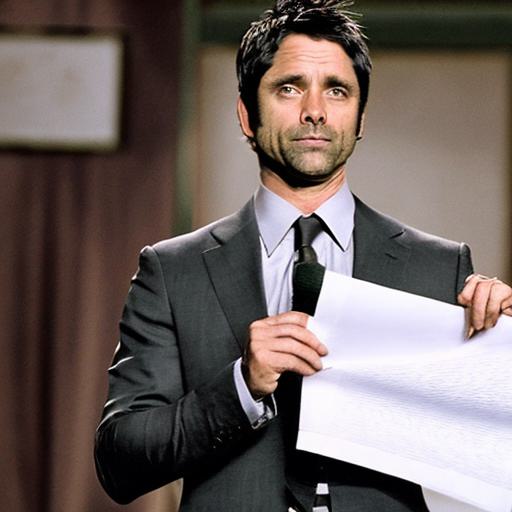} &
        \includegraphics[width=0.15\textwidth]{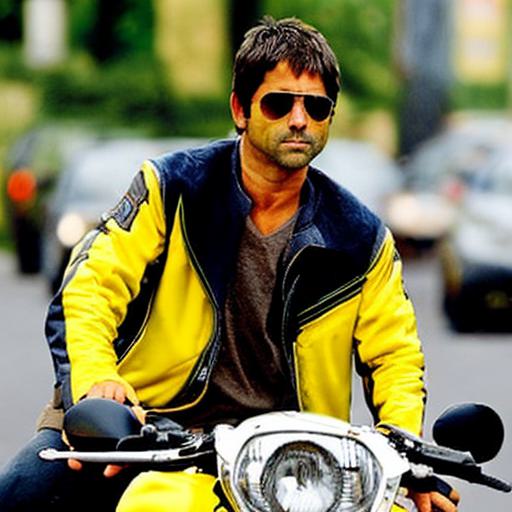} &
        \includegraphics[width=0.15\textwidth]{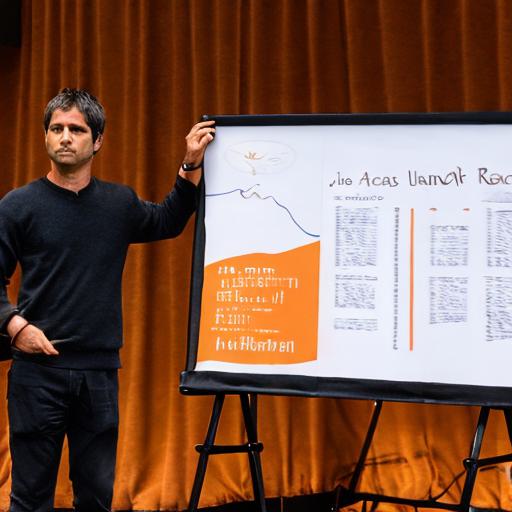} &
        \includegraphics[width=0.15\textwidth]{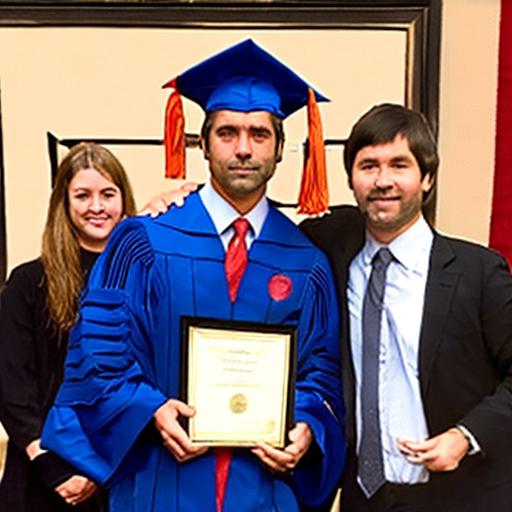} \\

        Real Sample &
        \begin{tabular}{c} A photo of $S_*$ \\ typing a paper \\ on a laptop \end{tabular} &
        \begin{tabular}{c} $S_*$ holding up \\ his accepted paper \end{tabular} &
        \begin{tabular}{c} $S_*$ wearing  \\  yellow jacket and \\ driving a motorbike \end{tabular} &
        \begin{tabular}{c} $S_*$ presenting a poster \\ at a conference \end{tabular} &
        \begin{tabular}{c} A photo of $S_*$ \\ graduating after \\ finishing his PhD \end{tabular} \\ \\[-0.185cm]

        \includegraphics[width=0.15\textwidth]{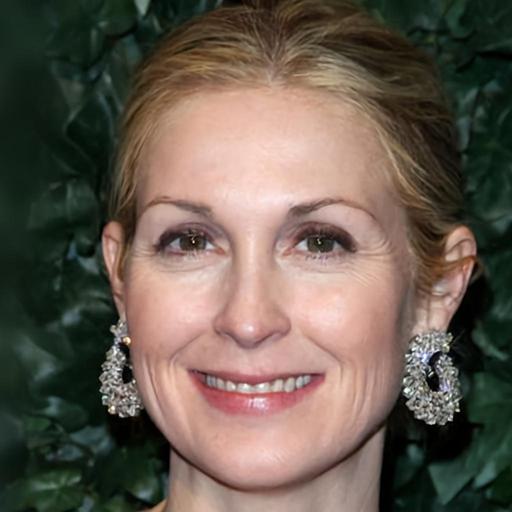} &
        \includegraphics[width=0.15\textwidth]{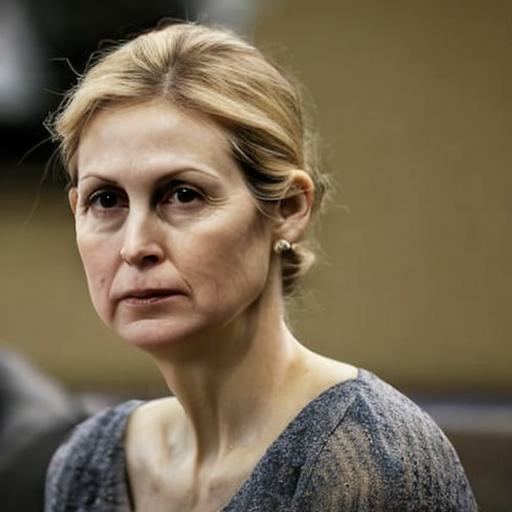} &
        \includegraphics[width=0.15\textwidth]{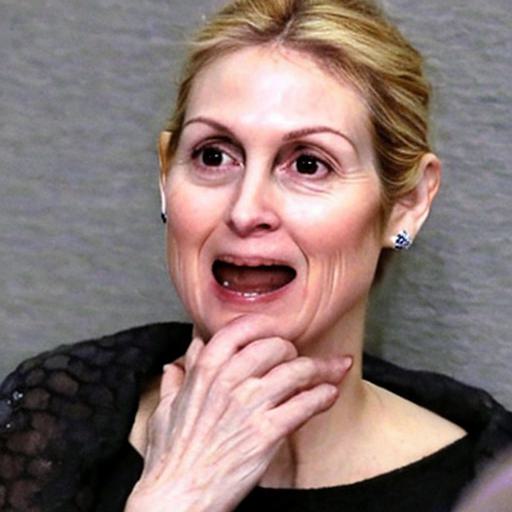} &
        \includegraphics[width=0.15\textwidth]{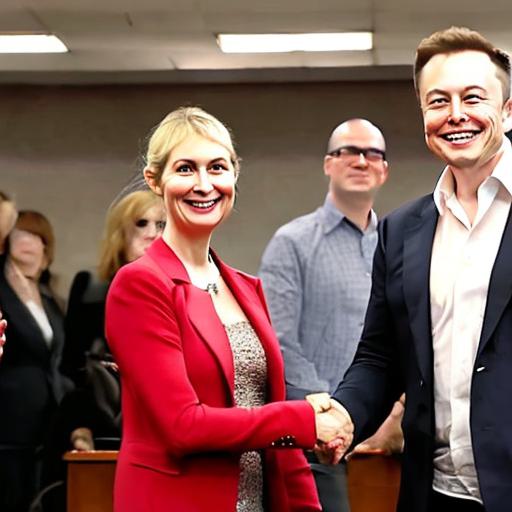} &
        \includegraphics[width=0.15\textwidth]{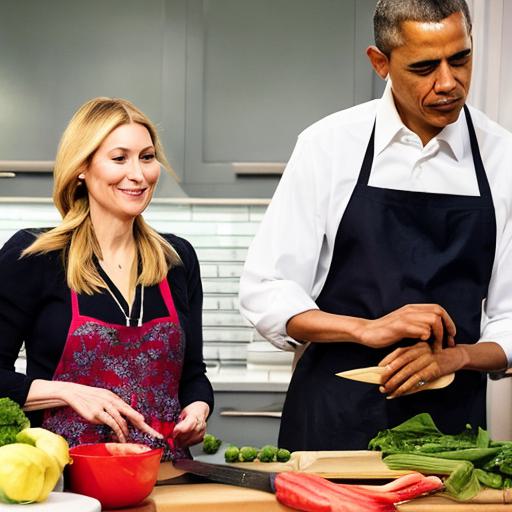} &
        \includegraphics[width=0.15\textwidth]{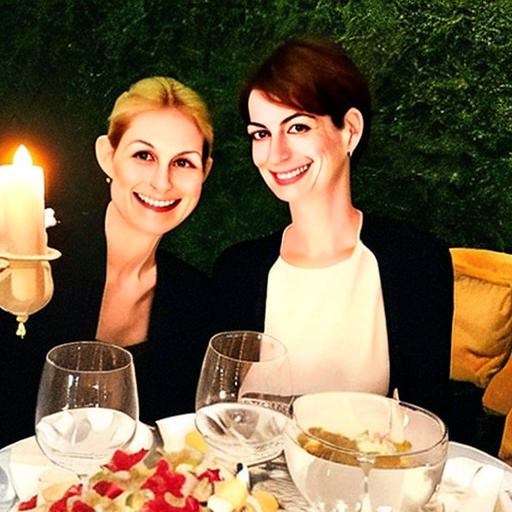} \\

        Real Sample &
        \begin{tabular}{c} $S_*$ with \\ a sad expression \end{tabular} &
        \begin{tabular}{c} $S_*$ with \\ a terrified expression \end{tabular} &
        \begin{tabular}{c} $S_*$ shakes hands \\ with Elon Musk \\ in news conference \end{tabular} &
        \begin{tabular}{c} $S_*$ and Barack Obama \\ cooking together \\ in a kitchen \end{tabular} &
        \begin{tabular}{c} $S_*$ and Anne Hathaway \\ enjoy a delicate \\ candlelight dinner \end{tabular} \\ \\[-0.185cm]

        \includegraphics[width=0.15\textwidth]{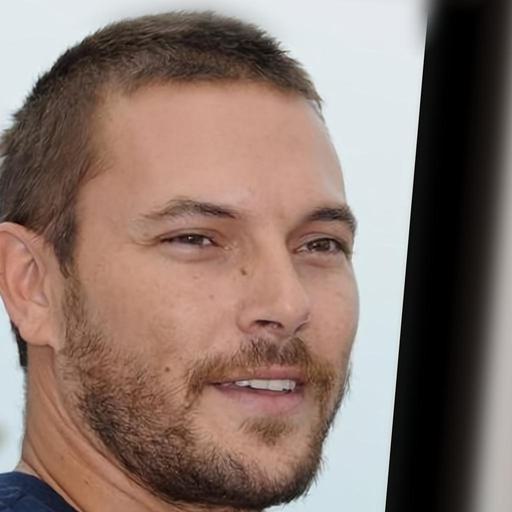} &
        \includegraphics[width=0.15\textwidth]{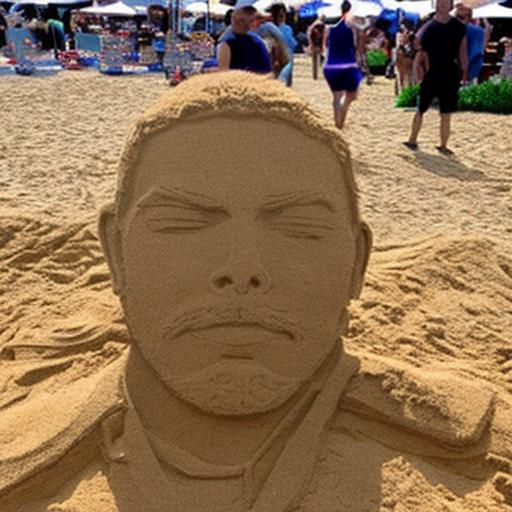} &
        \includegraphics[width=0.15\textwidth]{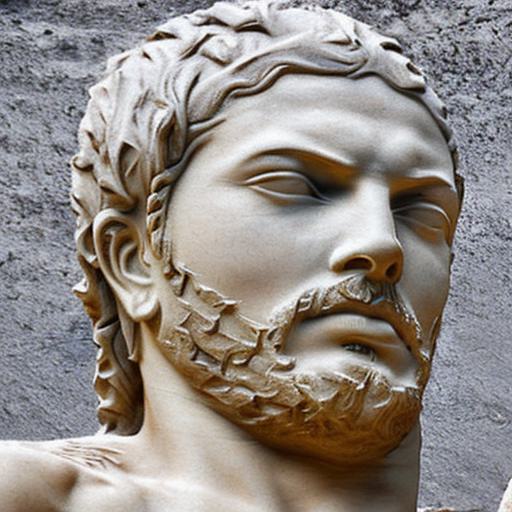} &
        \includegraphics[width=0.15\textwidth]{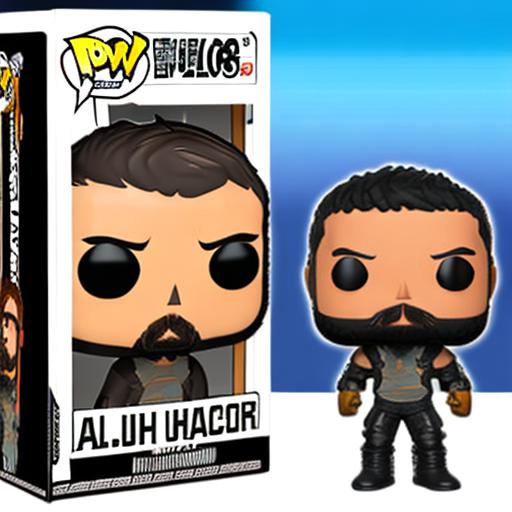} &
        \includegraphics[width=0.15\textwidth]{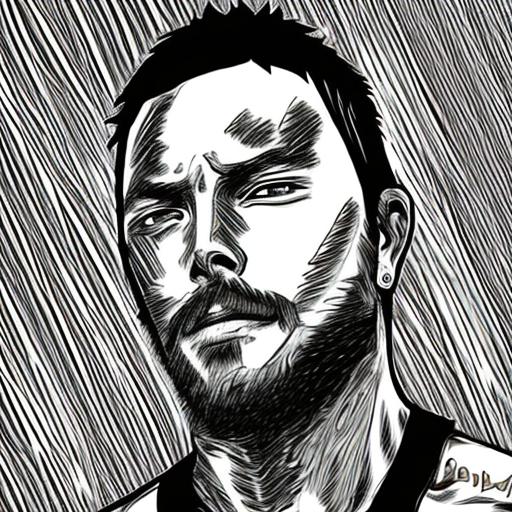} &
        \includegraphics[width=0.15\textwidth]{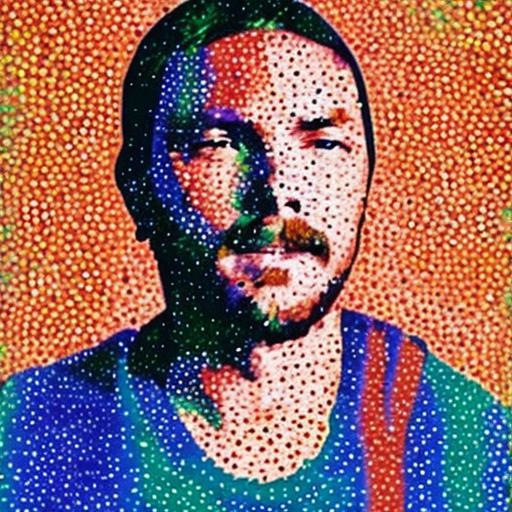} \\

        Real Sample &
        \begin{tabular}{c} A sand sculpture \\ of $S_*$ \end{tabular} &
        \begin{tabular}{c} Greek sculpture \\ of $S_*$ \end{tabular} &
        \begin{tabular}{c} $S_*$ Funko Pop \end{tabular} &
        \begin{tabular}{c} Manga drawing \\ of $S_*$ \end{tabular} &
        \begin{tabular}{c} Pointillism painting \\ of $S_*$ \end{tabular} \\

    \\[-0.4cm]        
    \end{tabular}
    }
    \caption{
    Examples of personalized text-to-image generation obtained with Cross Initialization.
    }
    \label{fig:our_results}
\end{figure*}

\section{Experiments}
In this section, we first present the implementation details of our method. Subsequently, we demonstrate its effectiveness by conducting a comparative analysis with four state-of-the-art personalization methods, focusing on aspects such as identity preservation, editability, and optimization time.

\subsection{Implementation and Evaluation Setup}
\paragraph{Implementation.}
We utilize the publicly available Stable Diffusion v2.1~\cite{stable-diffusion} as our base model. Images are generated at a resolution of \(512 \times 512\). The hyper-parameter \(\lambda\) is set to \(10^{-5}\) for all experiments. Given a single image as input, our experiments are conducted on a single A800 GPU, using a batch size of 8 and a learning rate of 0.005. All results are obtained using 320 optimization steps.

\paragraph{Evaluation Setup.}
We evaluate each method using the images from CelebA-HQ test set~\cite{liu2015faceattributes,karras2018progressive}. The prompts used are primarily sourced from~\cite{yuan2023inserting} and~\cite{gal2023encoderbased}. We compare our method with four state-of-the-art personalization methods: Textual Inversion~\cite{textual-inversion}, DreamBooth~\cite{dreambooth}, NeTI~\cite{alaluf2023neural}, and Celeb Basis~\cite{yuan2023inserting}. The implementation details of baselines are presented in \cref{sec:baseline_implementation}. All methods are implemented for one-shot personalization. For quantitative evaluation, each method is evaluated on the first 200 images from CelebA-HQ test set using two metrics, including identity similarity and prompt similarity. For identity similarity, ArcFace~\cite{arcface}, a pretrained face recognition model, is used to measure the identity preservation in generated images. Prompt similarity is measured by computing the CLIP score between generated images and text prompts. We exclude the prompts for stylization in the identity similarity assessment, as ArcFace is trained on real images.

\subsection{Results}
\label{sec:results}
\paragraph{Qualitative Evaluation.}
In \cref{fig:qualitative_comparison}, we present a visual comparison of personalized generation using four types of prompts: expression editing, background modification, individual interaction, and artistic style. Textual Inversion exhibits an overfitting problem, failing to compose the given individual in novel scenes. DreamBooth struggles to reconstruct the individual for complex editing prompts such as background modification and artistic style. It tends to disregard the new concept and generate images based solely on the remaining prompt tokens. In contrast, NeTI generates images based solely on the new concept without incorporating the other prompt tokens, indicating a severe overfitting problem. Both Celeb Basis and our method are capable of generating novel compositions of personalized concepts. Compared to Celeb Basis, our method shows superior identity preservation and excels in editing the individual's expression. For all prompts, Cross Initialization achieves high-fidelity reconstruction of the individual's identity while providing superior editability. Notably, it is the only method that successfully edits an individual's facial expression. \cref{fig:our_results} shows more results with different prompts from our method. Additional qualitative results can be found in \cref{sec:additional_comparison,sec:additional_qualitative}. We also provide results on synthetic facial images in \cref{sec:synthetic}.

\paragraph{Quantitative Evaluation.}
We quantitatively evaluate our approach in two aspects: 1) identity similarity between the generated and input images, and 2) prompt similarity between the generated image and the given text prompt. All methods are evaluated over 20 text prompts, see \cref{sec:prompts} for a full list. These prompts cover expression editing (e.g., ``$S_*$ with a sad expression''), background modification (e.g., ``$S_*$ on the beach''), individual interaction (e.g., ``$S_*$ shakes hands with Anne Hathaway in news conference''), and artistic style (e.g., ``$S_*$ latte art''). For each prompt, we generate 32 images using the same random seed for all methods. 

The results are shown in \cref{tab:quantitative_comparisons}. DreamBooth excels in prompt similarity but ranks lowest in identity similarity. This is consistent with the qualitative observations, where DreamBooth often overlooks the new concept, focusing solely on the other prompt tokens. In contrast, NeTI achieves the highest identity similarity scores but ranks lowest in prompt similarity, as NeTI tends to overfit the input image. Besides these two extreme cases, our method demonstrates superior performance in both identity and prompt similarity metrics.

\begin{table}
  \centering
  \caption{Quantitative comparisons. ``Identity'' denotes the identity similarity between the generated and input images. ``Prompt'' denotes the prompt similarity between the generated image and the given text prompt. ``Time'' denotes the average personalization time in seconds.
  }
  \vspace{-0.6em}
  \begin{tabular}{l@{\hspace{0.5cm}}ccc}
    \toprule
    \multirow{1}{*}{Methods}  & \multirow{1}{*}{Identity$\uparrow$}& \multirow{1}{*}{Prompt$\uparrow$} & \multirow{1}{*}{Time$\downarrow$}
    \\
    \midrule
    Textual Inversion~\cite{textual-inversion}    & 0.2115            & 0.2498            &6331           \\
    DreamBooth~\cite{raj2023dreambooth3d}         & 0.2053            & \textbf{0.3015}   &623           \\
    NeTI~\cite{alaluf2023neural}                  & \textbf{0.3789}   & 0.2325            &1527           \\
    Celeb Basis~\cite{yuan2023inserting}          & 0.2070            & 0.2683            &\underline{140} \\
    \midrule
    Ours-fast                                     &0.2225             &0.2800             &\textbf{26}    \\
    Ours                                          & \underline{0.2517}& \underline{0.2859} & 346         \\
    \bottomrule
  \end{tabular}

  \label{tab:quantitative_comparisons}
\end{table}
\begin{table}
  \centering
  \caption{User study results. We asked the participants to select the image that better preserves the identity and matches the prompt.
  }
  \vspace{-0.6em}
  \begin{tabular}{l@{\hspace{0.7cm}} c@{\hspace{0.25cm}} c@{\hspace{0.25cm}} }
    \toprule
    Baselines & Prefer Baseline    & Prefer Ours  \\
    \midrule
    Textual Inversion~\cite{textual-inversion}        & 22.0\%  & \textbf{78.0\%}           \\
    DreamBooth~\cite{raj2023dreambooth3d}               & 9.3\%  &  \textbf{90.7\%}           \\
    NeTI~\cite{alaluf2023neural}                     & 24.7\%  & \textbf{75.3\%}           \\
    Celeb Basis~\cite{yuan2023inserting}              & 26.7\%  & \textbf{73.3\%}           \\
    \bottomrule
  \end{tabular}
  \label{tab:user_study}
  \vspace{-0.2cm}
\end{table}

\vspace{-0.1cm}
\paragraph{Personalization Time.}
The average time for personalization using each method is reported in \cref{tab:quantitative_comparisons}. Compared to Textual Inversion, our method significantly reduces the optimization time from 106 minutes to 6 minutes. Additionally, We develop a fast version of our method, denoted as ``Ours-fast'', with a learning rate of 0.08. This fast version allows for learning the new concept in merely 25 optimization steps, taking only 26 seconds. As demonstrated in \cref{tab:quantitative_comparisons}, this fast version achieves the quickest personalization while surpassing Celeb Basis and Textual Inversion in both identity similarity and prompt similarity. The visual results of this fast version are presented in \cref{sec:fast_version}.

\vspace{-0.1cm}
\paragraph{User Study.}
We also evaluate our method from a human perspective by conducting a user study. We randomly selected one prompt from the prompt set and one image from the CelebA-HQ test set. These were used to generate personalized images for each method. In each question of the study, participants were presented with the input image and text prompt, as well as two generated images: one from our method and another from the baseline method. Participants were asked to select the image that better preserves the identity and matches the prompt. In total, we collected 600 responses from 30 participants, as shown in \cref{tab:user_study}. The results show a clear preference for our method.

\subsection{Ablation Study}
\label{sec:ablation}
\begin{figure}[t]
    \centering
    \renewcommand{\arraystretch}{0.3}
    \setlength{\tabcolsep}{0.5pt}
    \vspace{-0.2cm}
    {

    \begin{tabular}{c c c c c c }

        \begin{tabular}{c} Input \end{tabular} &
        \begin{tabular}{c}  w/o CI \end{tabular} &
        \begin{tabular}{c} w/o Mean \end{tabular} &
        \begin{tabular}{c} w/o Reg \end{tabular} &
        \begin{tabular}{c} Full \end{tabular} \\

        \includegraphics[width=0.09\textwidth]{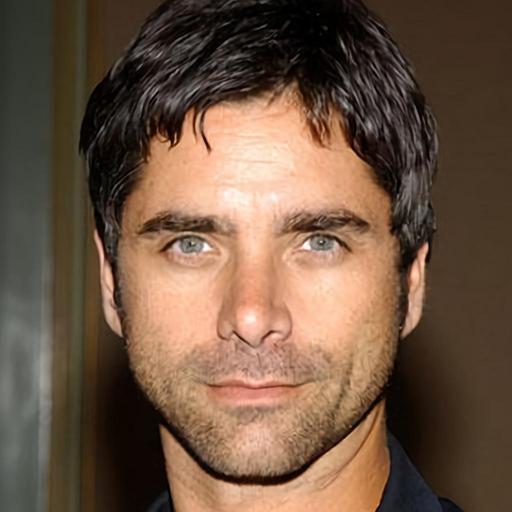} &
        \includegraphics[width=0.09\textwidth]{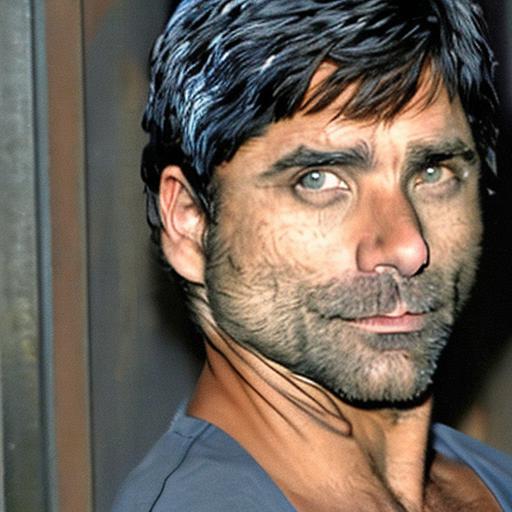} &
        \includegraphics[width=0.09\textwidth]{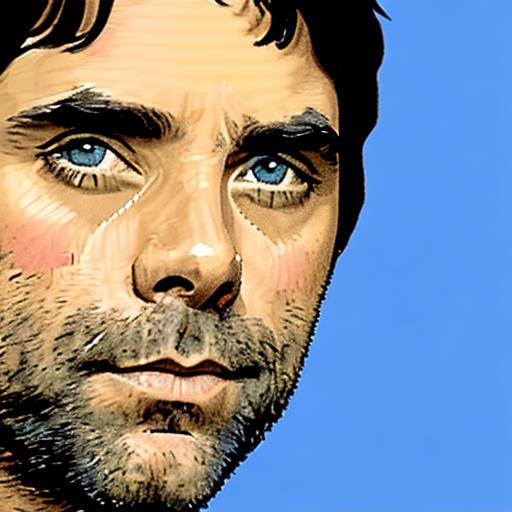} &
        \includegraphics[width=0.09\textwidth]{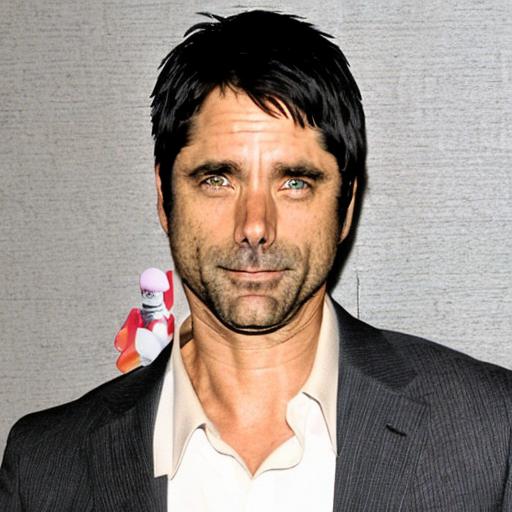}&
        \includegraphics[width=0.09\textwidth]{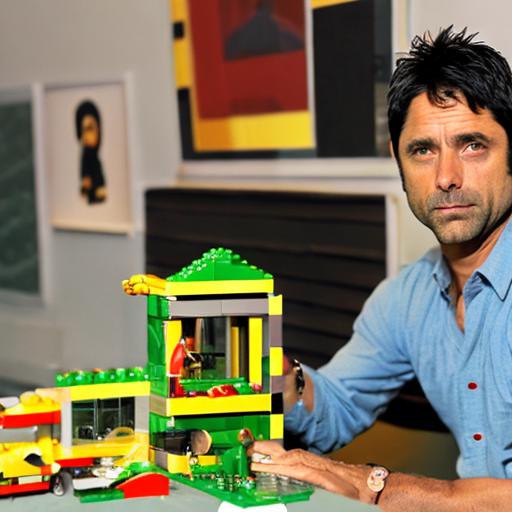} \\

        \includegraphics[width=0.09\textwidth]{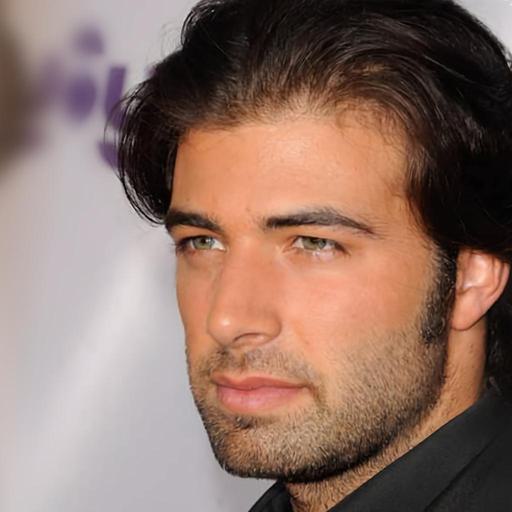} &
        \includegraphics[width=0.09\textwidth]{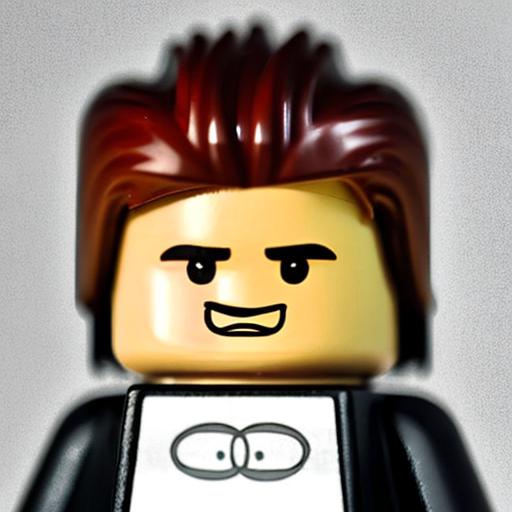} &
        \includegraphics[width=0.09\textwidth]{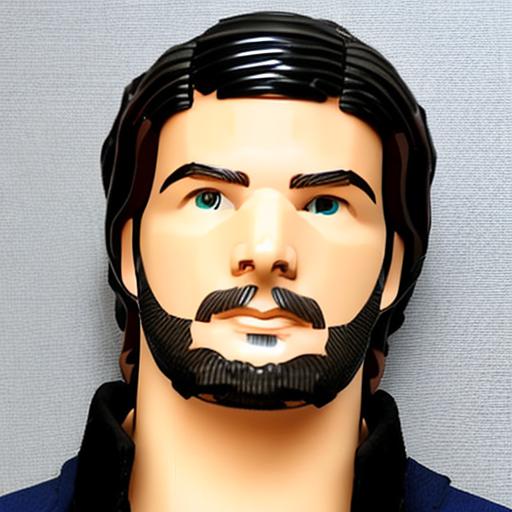} &
        \includegraphics[width=0.09\textwidth]{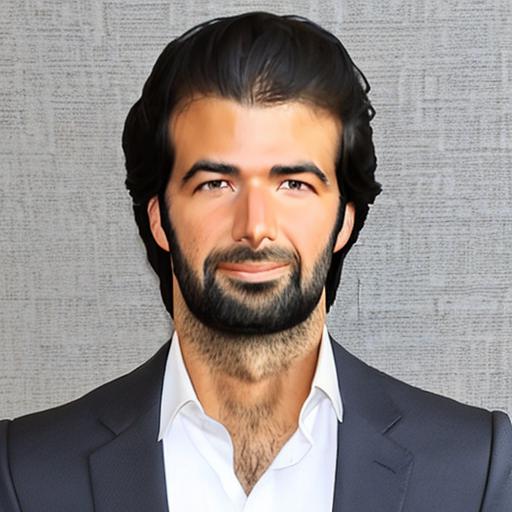} &
        \includegraphics[width=0.09\textwidth]{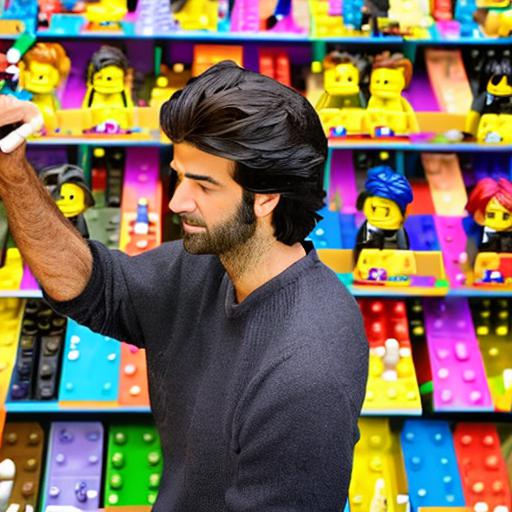} \\

    \end{tabular}
    \\[-0.2cm]
    }
    \caption{
    Ablation study. The prompt is ``$S_*$ plays the LEGO toys''. We compare the models trained without Cross Initialization (w/o CI), without mean textual embedding (w/o Mean), and without regularization (w/o Reg). As can be seen, all sub-modules are essential for achieving identity-preserved and prompt-aligned personalized face generation.
    }
    \label{fig:ablation_study}
    \vspace{-0.4cm}
\end{figure}

We conduct an ablation study by separately removing each sub-module from our method. Specifically, we sequentially remove the following sub-modules: 1) Cross Initialization, 2) mean textual embedding, and 3) the regularization term. In \cref{fig:ablation_study}, we present a visual comparison of the personalized images generated by each variant. The results indicate that all sub-modules are crucial for achieving identity-preserved and prompt-aligned personalized face generation. Specifically, the model without Cross Initialization produces results similar to those by Textual Inversion. This variant tends to generate images focusing either solely on the given concept or exclusively on the other prompt tokens. The models without mean textual embedding or the regularization term lead to degradation in editability, struggling to create consistent scenes as described in the prompt. More ablation study results are provided in \cref{sec:more_ablation}.

\section{Conclusions and Future Work}
We introduced a new initialization method for personalized text-to-image generation. We identified a significant disparity between the initial and learned embeddings in Textual Inversion, which often leads to an overfitting problem. Our approach, ``Cross Initilization'', addresses this issue by initializing the textual embedding with the output of the text encoder. Cross Initialization enables more identity-preserved, prompt-aligned, and faster face personalization. In this work, we mainly examined the performance of Cross Initialization on the human being concept. For general concepts, we found that Cross Initialization is not as effective as it is for the human being concept. In future work, we plan to further investigate the applicability of Cross Initialization to a broader range of concepts.

{\small
\bibliographystyle{ieeenat_fullname}
 \bibliography{egbib}
}

\clearpage
\appendix
\appendixpage
\section{Implementation Details of Baselines}
\label{sec:baseline_implementation}
We compare our method with four baseline methods: Textual Inversion~\cite{textual-inversion}, DreamBooth~\cite{dreambooth}, NeTI~\cite{alaluf2023neural}, and Celeb Basis~\cite{yuan2023inserting}. For Textual Inversion, we use the diffusers implementation~\cite{diffusers} with Stable Diffusion v2.1 as the base model. The textual embeddings are initialized with the embeddings of ``human face''. We perform 5,000 optimization steps using a learning rate of 5e-3 and a batch size of 8. For DreamBooth, we also use the diffusers implementation and tune the U-Net with prior preservation loss. We perform 800 fine-tuning steps using a learning rate of 2e-6 and a batch size of 1. For NeTI and Celeb Basis, we use their official implementations and follow the official hyperparameters described in their papers. Moreover, we apply the textual bypass and Nested Dropout~\cite{rippel2014learning} techniques for NeTI. 

\begin{table}[b]
\small
\centering
\setlength{\tabcolsep}{3pt}   
\renewcommand{\arraystretch}{1.2}
\caption{The 20 prompts used in the quantitative evaluation.\\[-0.3cm]}
\begin{tabular}{@{\hspace{1cm}}c@{\hspace{1cm}}}
    \toprule
        a photo of a $S^*$ person \\
        \hline
        a $S^*$ person with a sad expression \\
        \hline
        a $S^*$ person with a happy expression \\
        \hline
        a $S^*$ person with a puzzled expression \\
        \hline
        a $S^*$ person with an angry expression \\
        \hline
        a $S^*$ person plays the LEGO toys \\
        \hline
        a $S^*$ person on the beach \\
        \hline
        a $S^*$ person piloting a fighter jet \\
        \hline
        a $S^*$ person wearing the sweater, a backpack and \\ camping stove, outdoors, RAW, ultra high res \\
        \hline
        a $S^*$ person wearing a scifi spacesuit in space \\
        \hline
        a $S^*$ person and Anne Hathaway \\ are baking a birthday cake \\
        \hline
        a $S^*$ person and Anne Hathaway \\ taking a relaxing hike in the mountains \\
        \hline
        a $S^*$ person and Anne Hathaway sit on a sofa \\
        \hline
        a $S^*$ person and Anne Hathaway \\ enjoying a day at an amusement park \\
        \hline
        a $S^*$ person shakes hands with \\ Anne Hathaway in news conference \\
        \hline
        cubism painting of a $S^*$ person \\
        \hline
        fauvism painting of a $S^*$ person \\
        \hline
        cave mural depicting a $S^*$ person \\
        \hline
        pointillism painting of a $S^*$ person \\
        \hline
        a $S^*$ person latte art \\
    \bottomrule
\end{tabular}
\label{tab:prompts}
\vspace{-0.3cm}
\end{table}

\section{Text Prompts}
\label{sec:prompts}
In \cref{tab:prompts}, we list all 20 text prompts used in the quantitative evaluation. These prompts cover a range of modifications, including expression editing, background modification, individual interaction, and artistic style.

\section{Results for Our Fast Version Method}
\label{sec:fast_version}
\begin{figure*}
    \centering
    \setlength{\tabcolsep}{0.1pt}
    {\footnotesize
    \begin{tabular}{c@{\hspace{0.25cm}} c@{\hspace{0.25cm}} c@{\hspace{0.25cm}} c@{\hspace{0.25cm}} c@{\hspace{0.25cm}} c@{\hspace{0.25cm}} c@{\hspace{0.25cm}}}

        \includegraphics[width=0.15\textwidth]{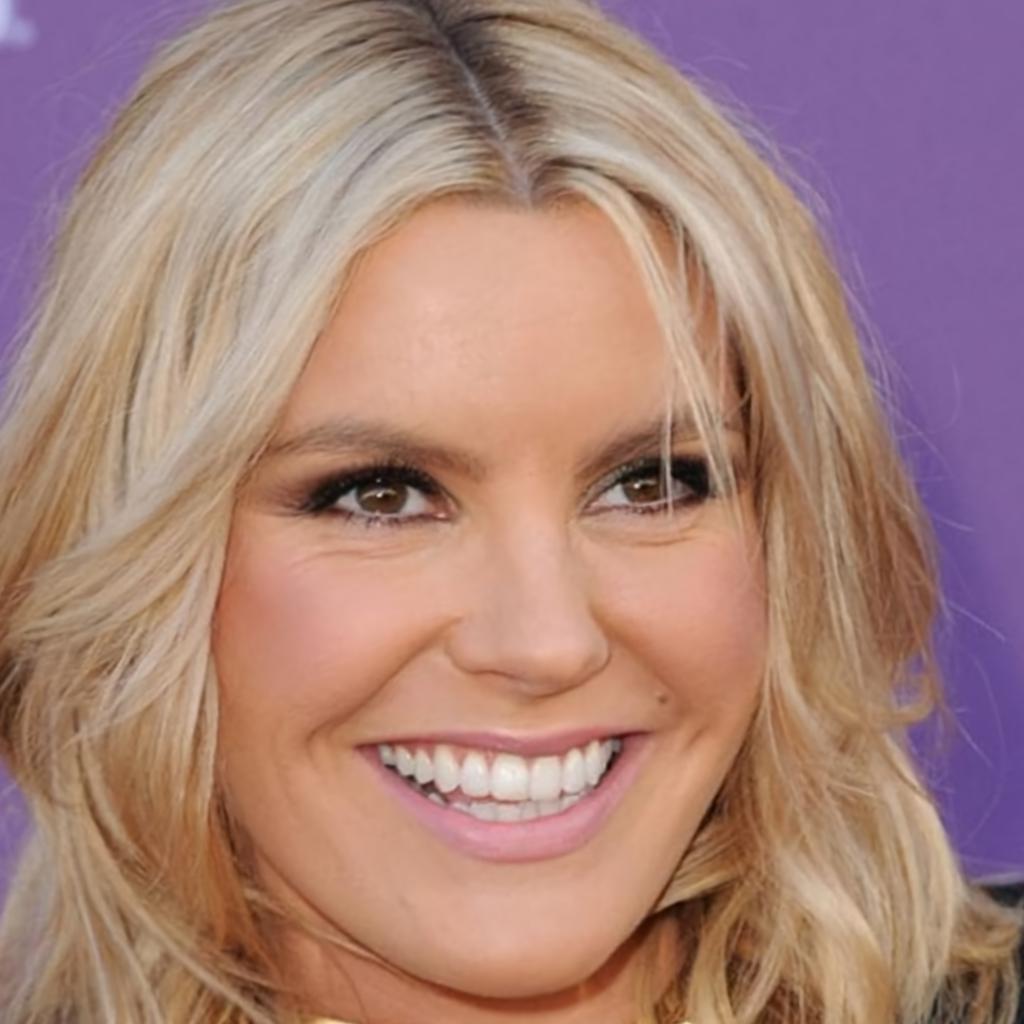} &
        \includegraphics[width=0.15\textwidth]{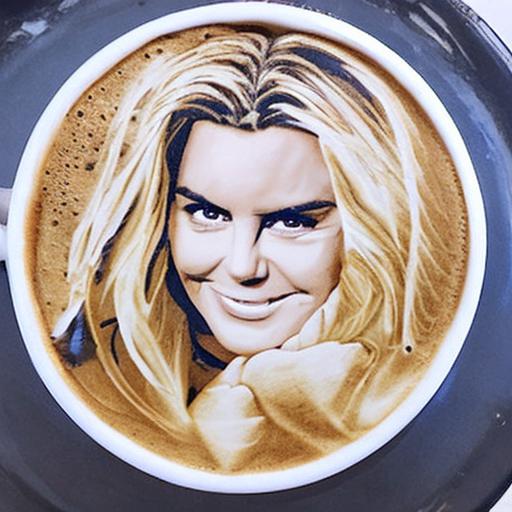} &
        \includegraphics[width=0.15\textwidth]{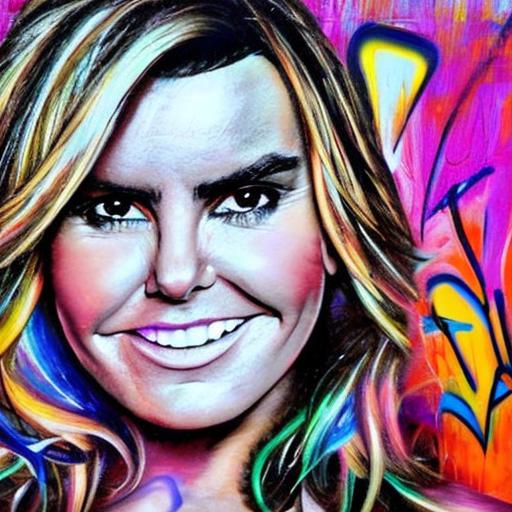} &
        \includegraphics[width=0.15\textwidth]{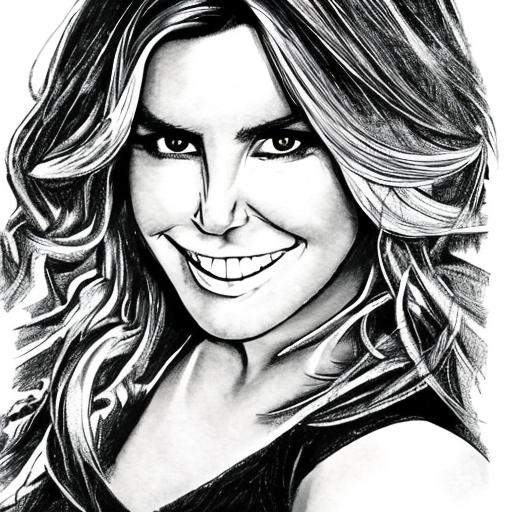} &
        \includegraphics[width=0.15\textwidth]{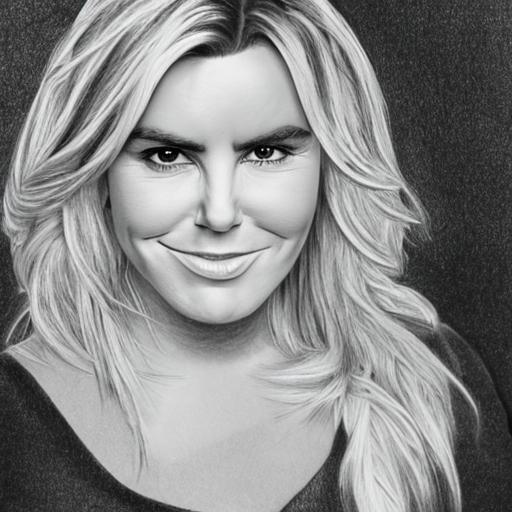} &
        \includegraphics[width=0.15\textwidth]{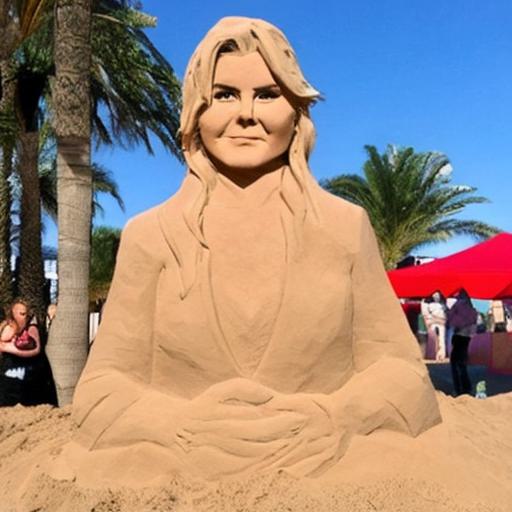} \\

        Real Sample &
        \begin{tabular}{c} ``$S_*$ latte \\ art'' \end{tabular} &
        \begin{tabular}{c} ``Colorful graffiti \\ of $S_*$'' \end{tabular} &
        \begin{tabular}{c} ``Manga drawing \\ of $S_*$'' \end{tabular} &
        \begin{tabular}{c} ``Pencil drawing \\ of $S_*$'' \end{tabular} &
        \begin{tabular}{c} ``A sand sculpture \\ of $S_*$'' \end{tabular} \\ \\[-0.185cm]

        \includegraphics[width=0.15\textwidth]{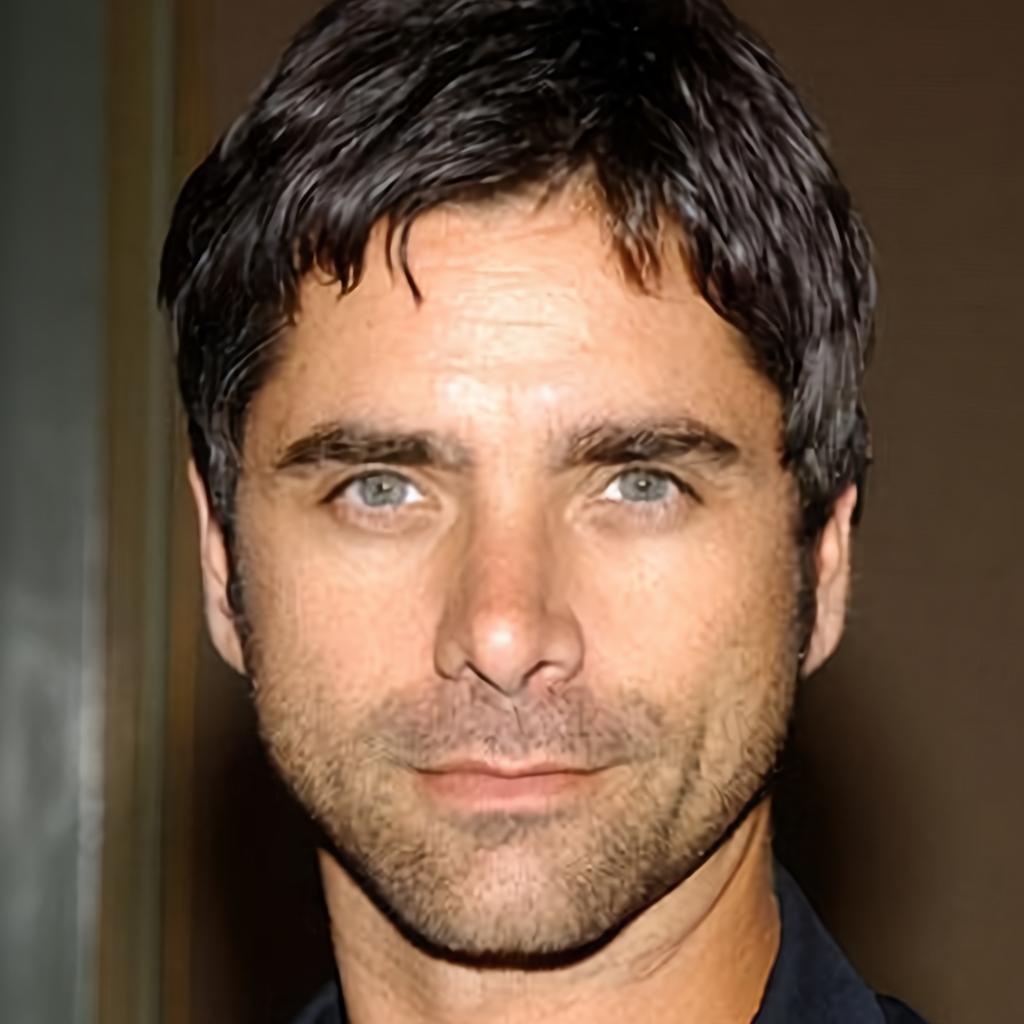} &
        \includegraphics[width=0.15\textwidth]{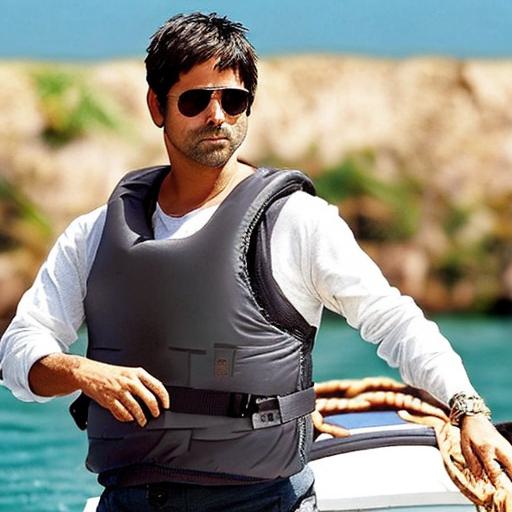} &
        \includegraphics[width=0.15\textwidth]{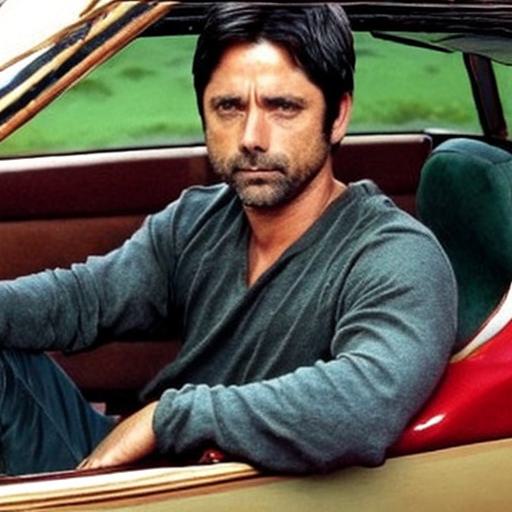} &
        \includegraphics[width=0.15\textwidth]{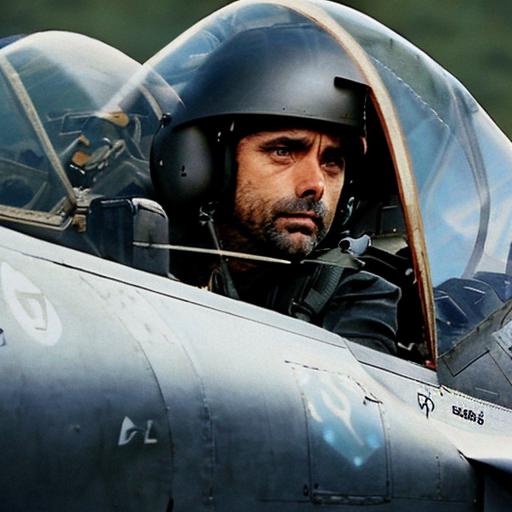} &
        \includegraphics[width=0.15\textwidth]{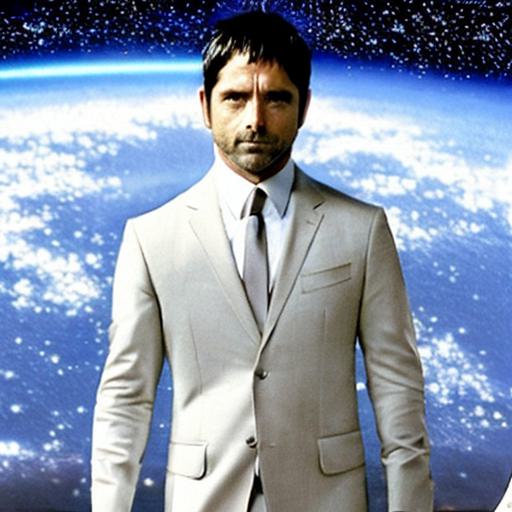} &
        \includegraphics[width=0.15\textwidth]{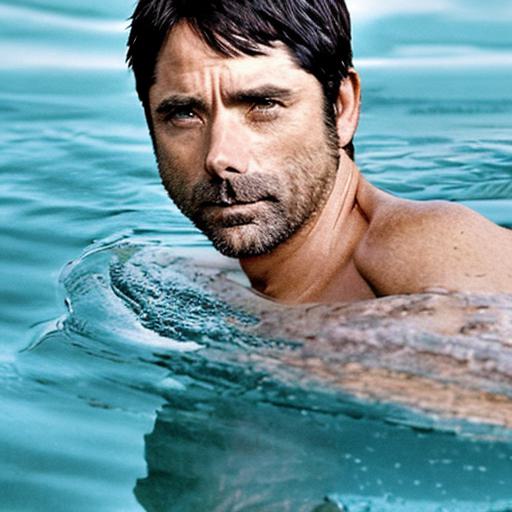} \\

        Real Sample &
        \begin{tabular}{c} ``$S_*$ wears a \\ sunglass and a life \\ jacket on a boat'' \end{tabular} &
        \begin{tabular}{c} ``$S_*$ is \\ driving a car'' \end{tabular} &
        \begin{tabular}{c} ``$S_*$ piloting \\ a fighter jet'' \end{tabular} &
        \begin{tabular}{c} ``$S_*$ wears a \\ suit in space'' \end{tabular} &
        \begin{tabular}{c} ``$S_*$ swims \\ in the ocean'' \end{tabular} \\ \\[-0.185cm]

        \includegraphics[width=0.15\textwidth]{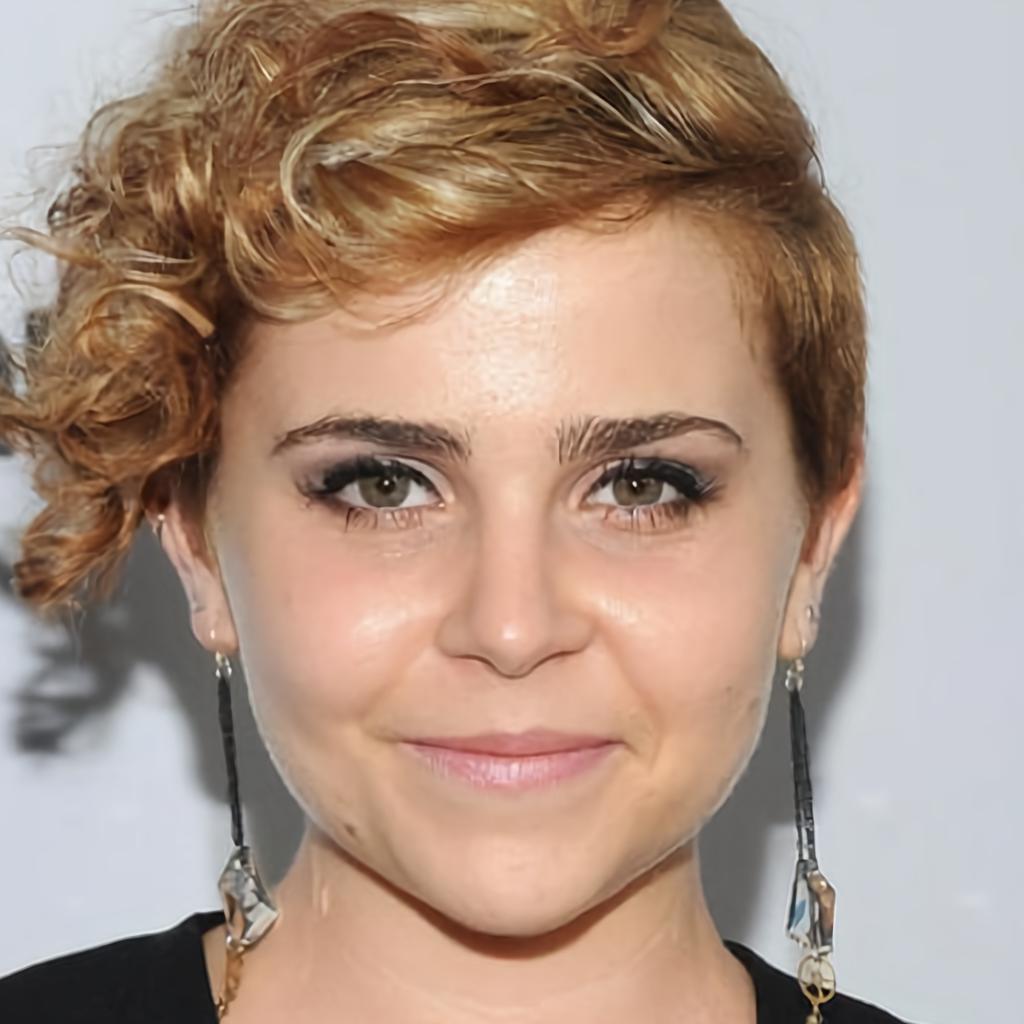} &
        \includegraphics[width=0.15\textwidth]{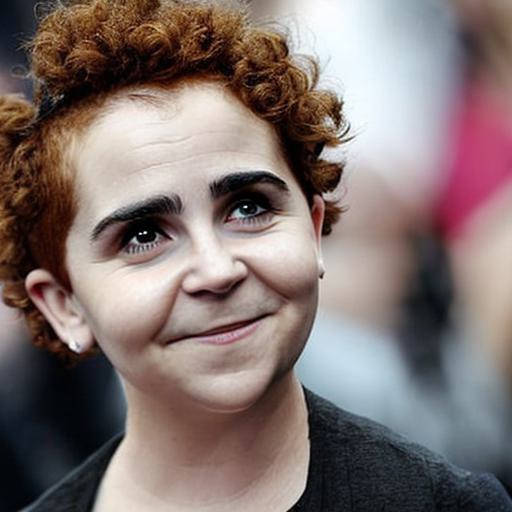} &
        \includegraphics[width=0.15\textwidth]{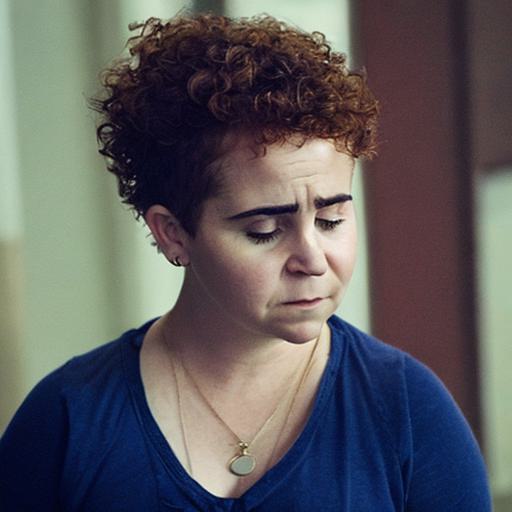} &
        \includegraphics[width=0.15\textwidth]{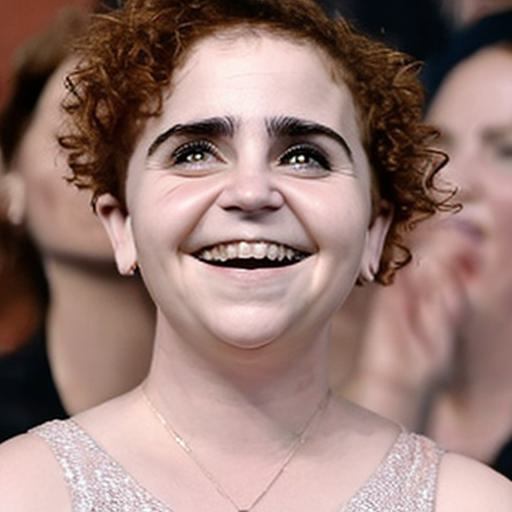} &
        \includegraphics[width=0.15\textwidth]{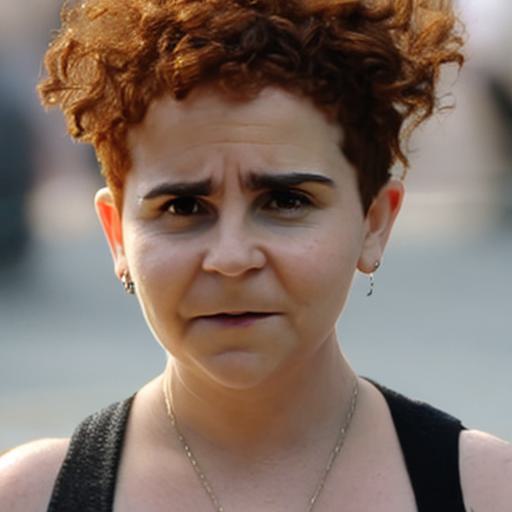} &
        \includegraphics[width=0.15\textwidth]{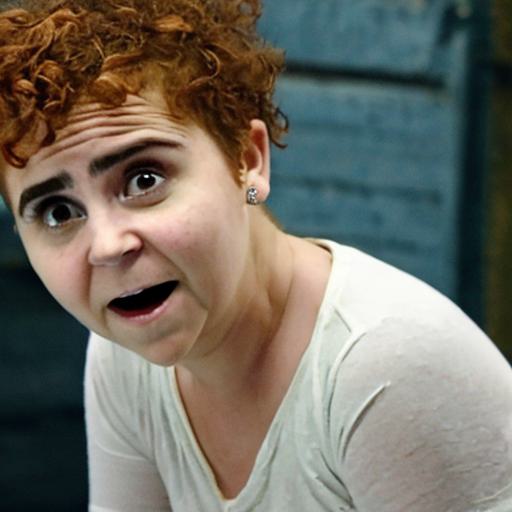} \\

        Real Sample &
        \begin{tabular}{c} ``$S_*$ with an \\ admiring expression'' \end{tabular} &
        \begin{tabular}{c} ``$S_*$ with a \\ depressed expression'' \end{tabular} &
        \begin{tabular}{c} ``$S_*$ with an \\ ecstatic expression'' \end{tabular} &
        \begin{tabular}{c} ``$S_*$ with a \\ puzzled expression'' \end{tabular} &
        \begin{tabular}{c} ``$S_*$ with a \\ terrified expression'' \end{tabular} \\ \\[-0.185cm]

        \includegraphics[width=0.15\textwidth]{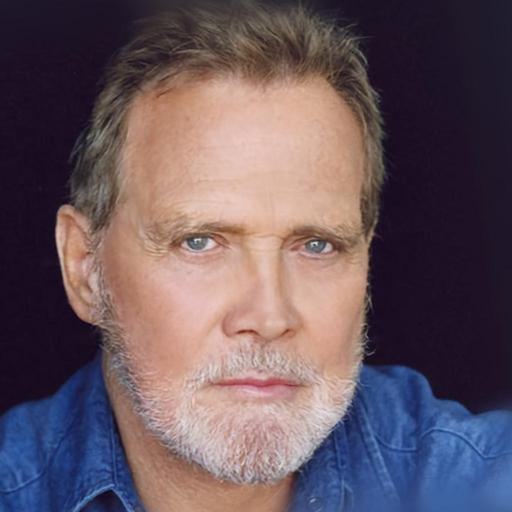} &
        \includegraphics[width=0.15\textwidth]{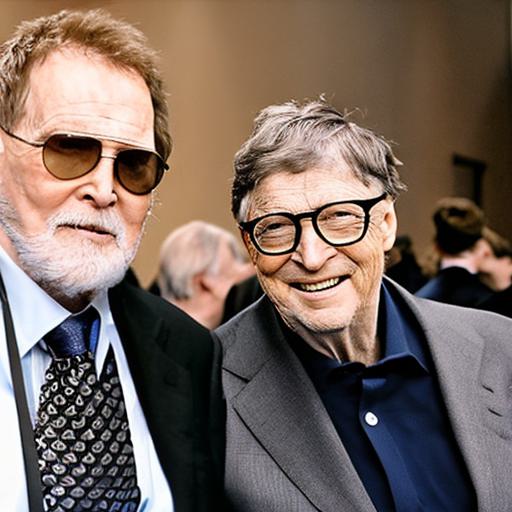} &
        \includegraphics[width=0.15\textwidth]{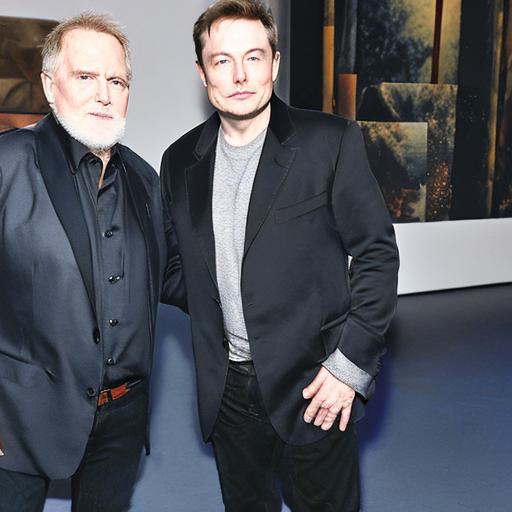} &
        \includegraphics[width=0.15\textwidth]{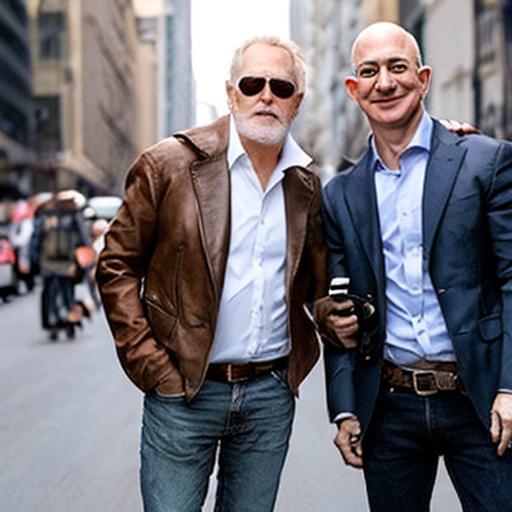} &
        \includegraphics[width=0.15\textwidth]{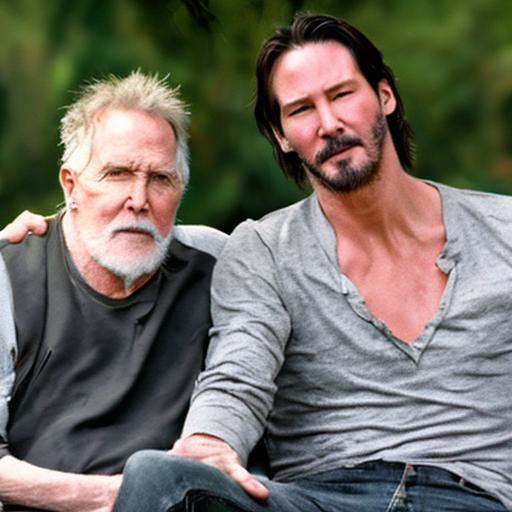} &
        \includegraphics[width=0.15\textwidth]{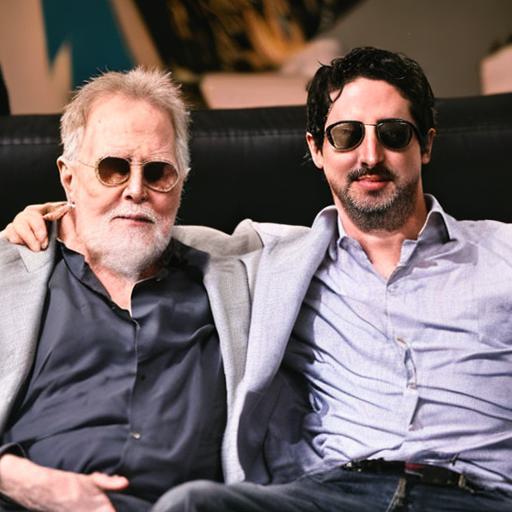} \\

        Real Sample &
        \begin{tabular}{c} ``$S_*$ and Bill \\ Gates go to a \\ technology \\ exhibition together'' \end{tabular} &
        \begin{tabular}{c} ``$S_*$ and Elon \\ Musk go to an art \\ exhibition together'' \end{tabular} &
        \begin{tabular}{c} ``$S_*$ is standing \\ with Jeff Bezos \\ on a street'' \end{tabular} &
        \begin{tabular}{c} ``$S_*$ and \\ Keanu Reeves \\ sit in the park'' \end{tabular} &
        \begin{tabular}{c} ``$S_*$ and \\ Sergey Brin \\ sit on a sofa'' \end{tabular} \\ \\[-0.185cm]

        \includegraphics[width=0.15\textwidth]{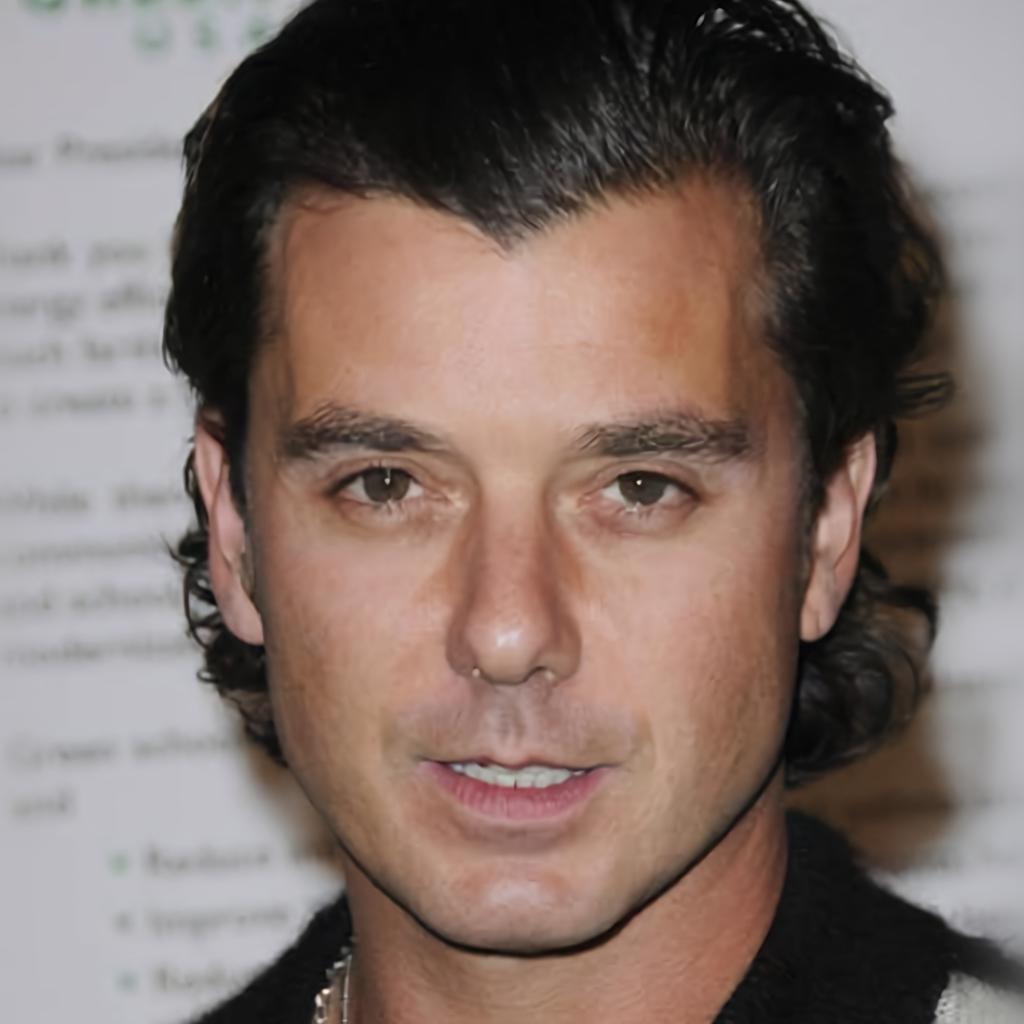} &
        \includegraphics[width=0.15\textwidth]{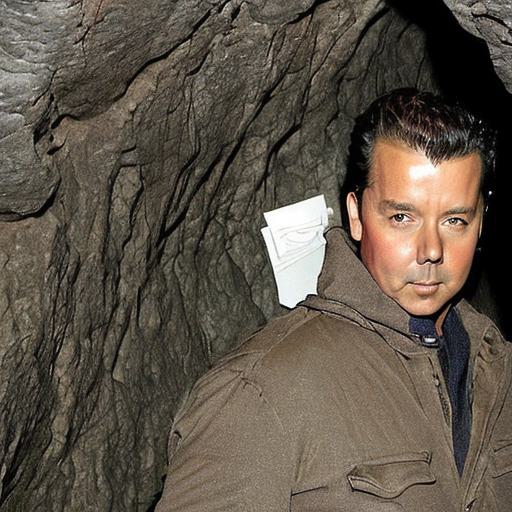} &
        \includegraphics[width=0.15\textwidth]{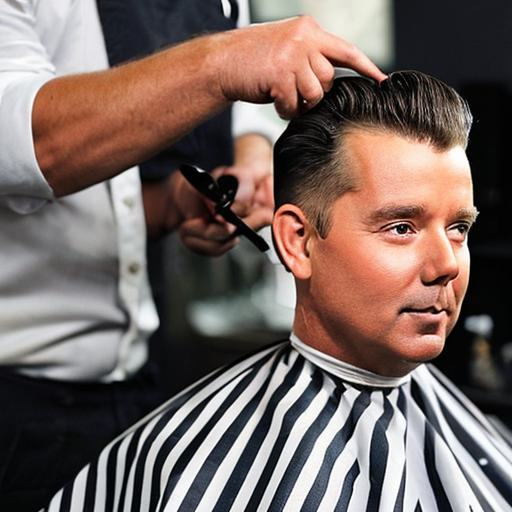} &
        \includegraphics[width=0.15\textwidth]{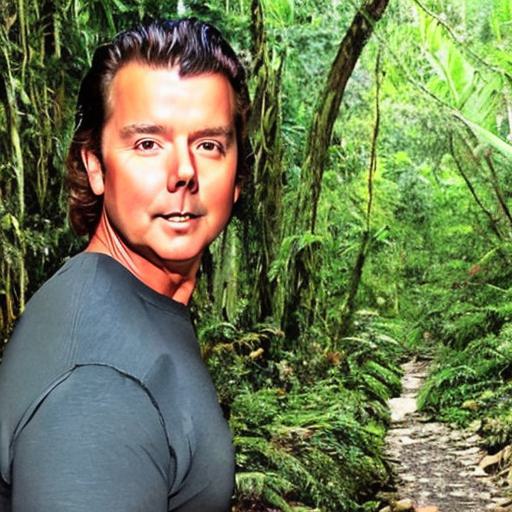} &
        \includegraphics[width=0.15\textwidth]{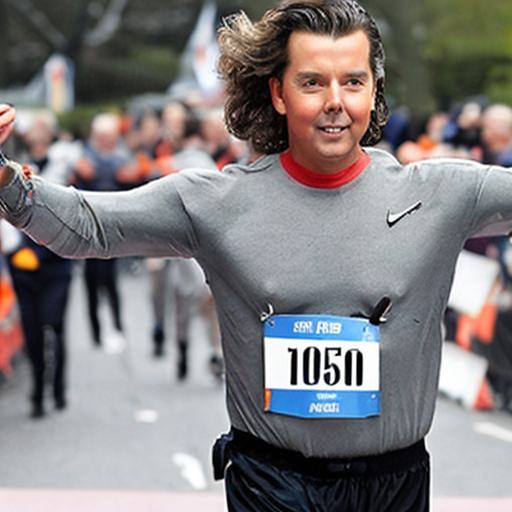} &
        \includegraphics[width=0.15\textwidth]{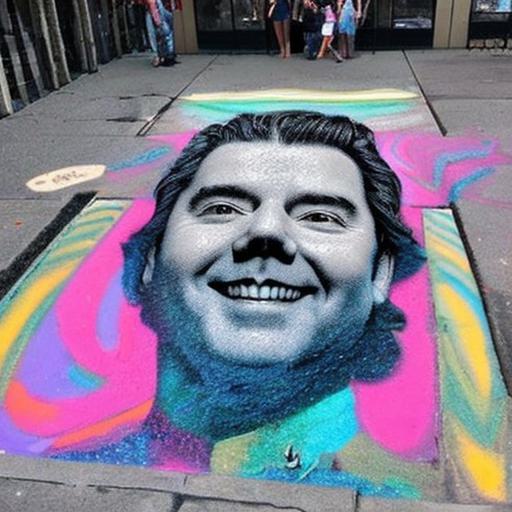} \\

        Real Sample &
        \begin{tabular}{c} ``$S_*$ is surveying \\ an underground cave'' \end{tabular} &
        \begin{tabular}{c} ``$S_*$ is having \\ a haircut in a \\ classic, retro-styled \\ barbershop'' \end{tabular} &
        \begin{tabular}{c} ``$S_*$ is hiking \\ in a dense, \\ lush rainforest'' \end{tabular} &
        \begin{tabular}{c} ``$S_*$ is crossing \\ the marathon \\ finish line'' \end{tabular} &
        \begin{tabular}{c} ``A vibrant, large-scale \\ chalk art of $S_*$ \\ on a sidewalk'' \end{tabular} \\ \\[-0.185cm]

    \\[-0.4cm]        
    \end{tabular}
    }
    \caption{Images generated by our fast version method with a learning rate of 0.08. Results are obtained after 25 optimization steps, taking only 26 seconds.}
    \label{fig:appendix_our_with_lr_008_1}
\end{figure*}

\begin{figure*}
    \centering
    \setlength{\tabcolsep}{0.1pt}
    {\footnotesize
    \begin{tabular}{c@{\hspace{0.25cm}} c@{\hspace{0.25cm}} c@{\hspace{0.25cm}} c@{\hspace{0.25cm}} c@{\hspace{0.25cm}} c@{\hspace{0.25cm}} c@{\hspace{0.25cm}}}

        \includegraphics[width=0.15\textwidth]{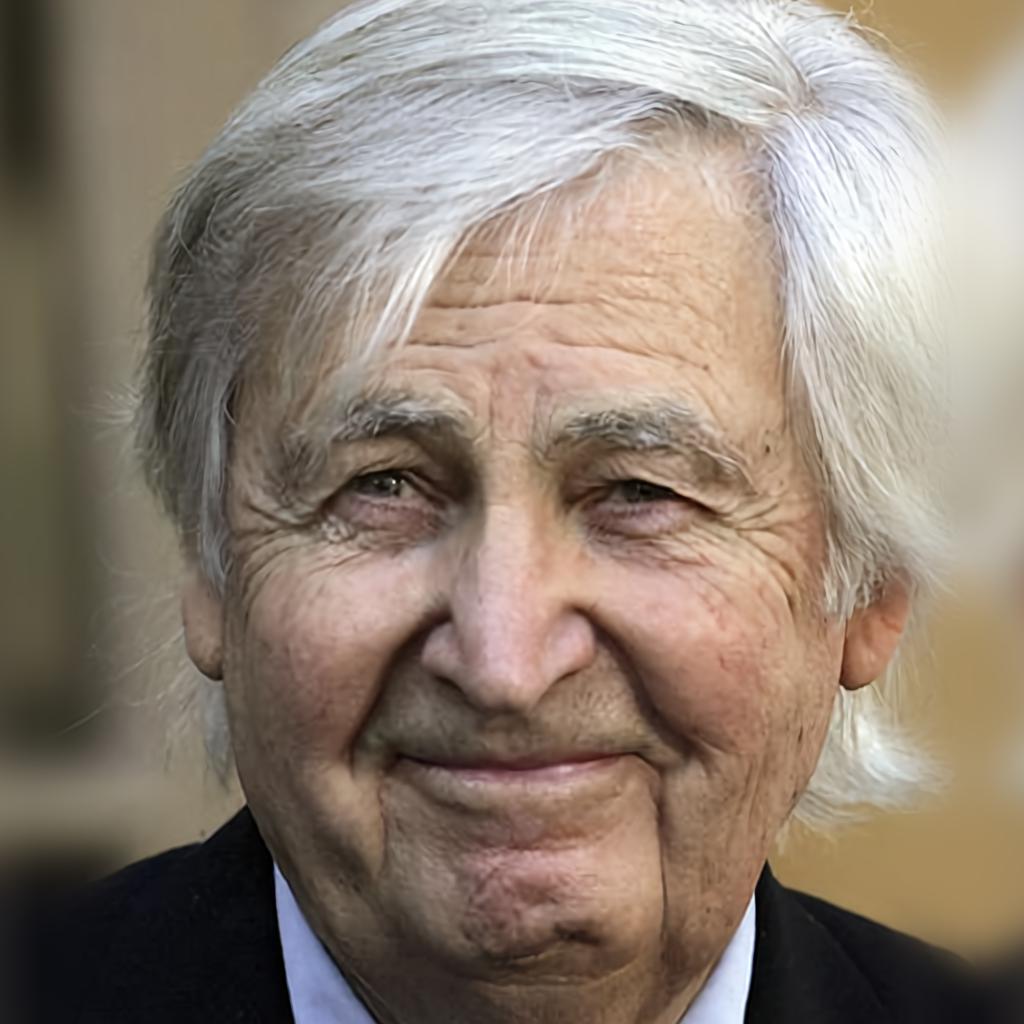} &
        \includegraphics[width=0.15\textwidth]{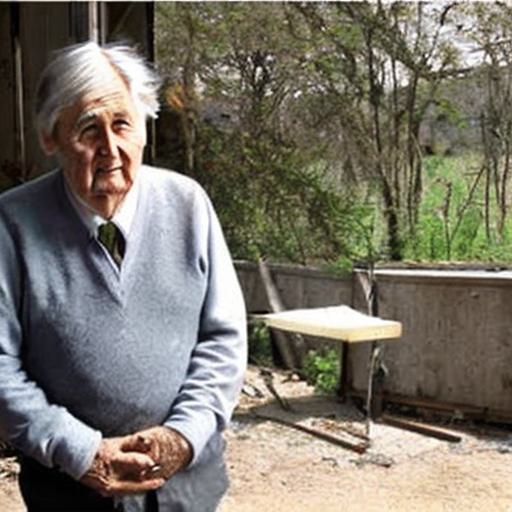} &
        \includegraphics[width=0.15\textwidth]{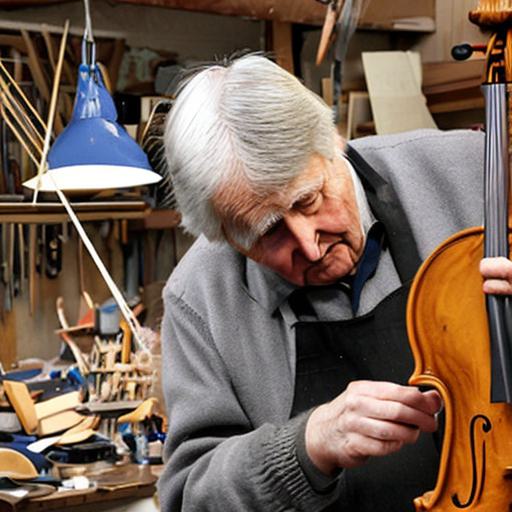} &
        \includegraphics[width=0.15\textwidth]{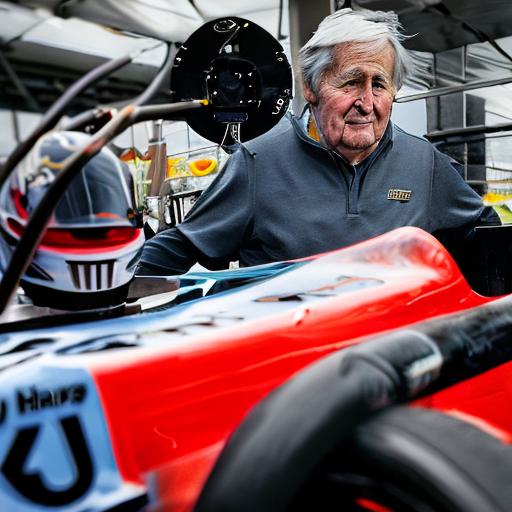} &
        \includegraphics[width=0.15\textwidth]{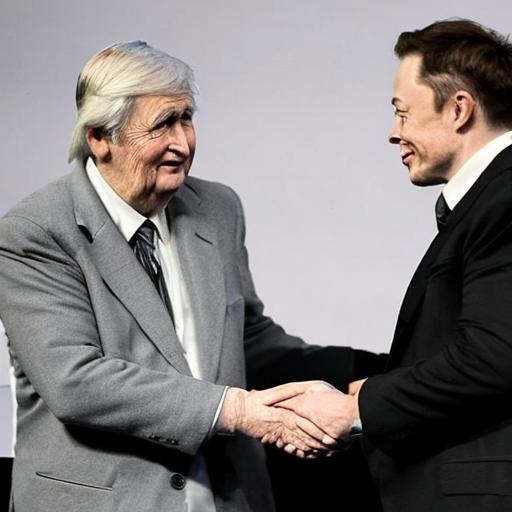} &
        \includegraphics[width=0.15\textwidth]{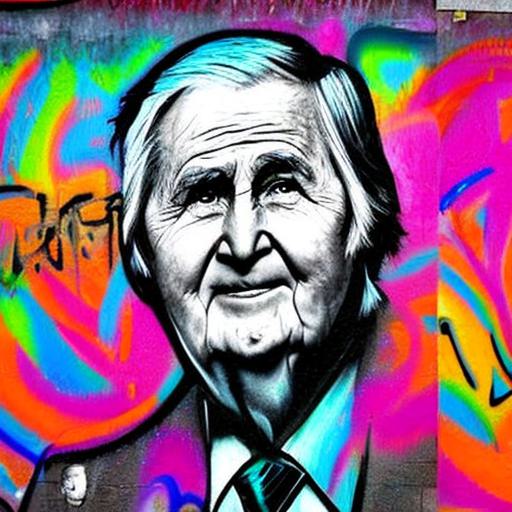} \\

        Real Sample &
        \begin{tabular}{c} ``$S_*$ is living in \\ an abandoned building
'' \end{tabular} &
        \begin{tabular}{c} ``$S_*$ is fine-tuning \\ a handmade violin \\ in a workshop'' \end{tabular} &
        \begin{tabular}{c} ``$S_*$ race car driver \\ is gearing up \\ in the pit lane \\ before a race'' \end{tabular} &
        \begin{tabular}{c} ``$S_*$ shakes hands \\ with Elon Musk \\ in a news conference'' \end{tabular} &
        \begin{tabular}{c} ``colorful graffiti \\ of $S_*$'' \end{tabular} \\ \\[-0.185cm]

        \includegraphics[width=0.15\textwidth]{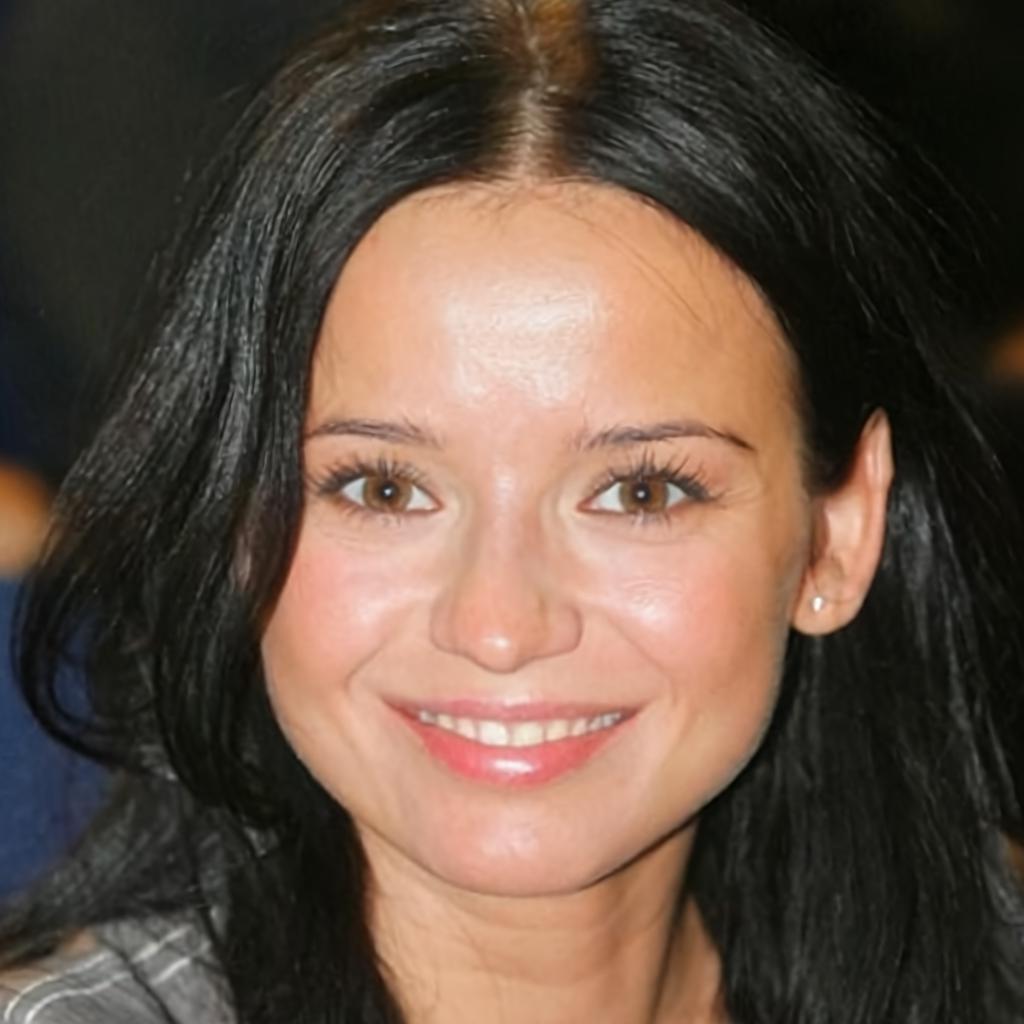} &
        \includegraphics[width=0.15\textwidth]{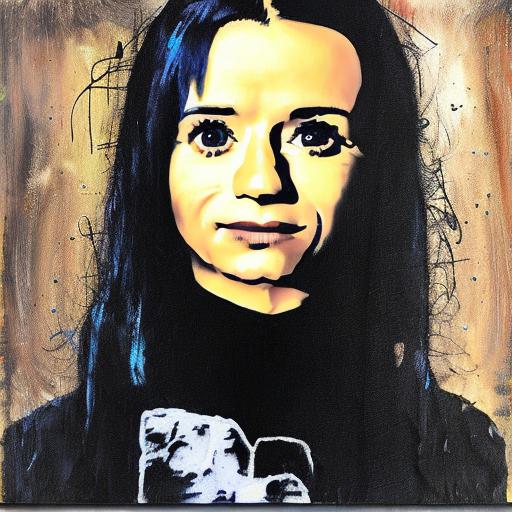} &
        \includegraphics[width=0.15\textwidth]{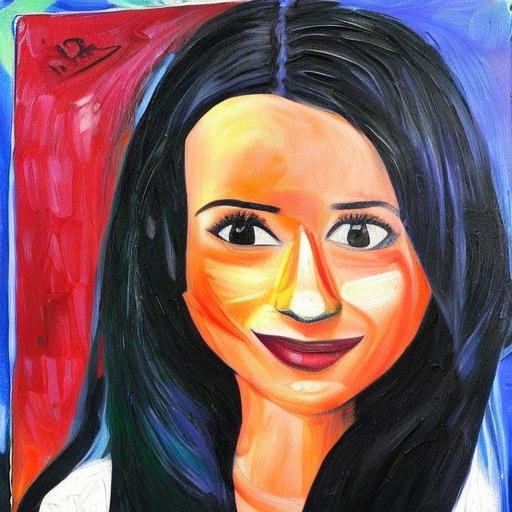} &
        \includegraphics[width=0.15\textwidth]{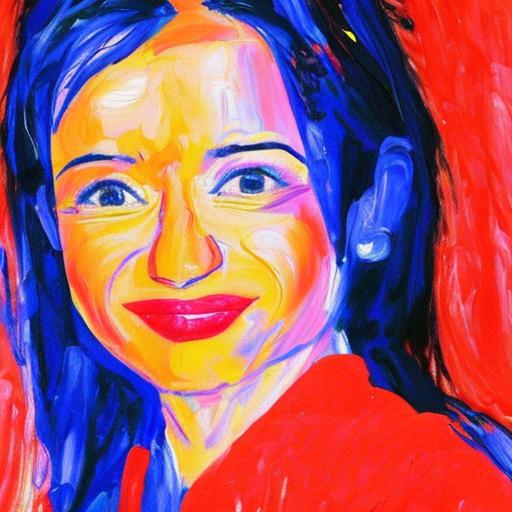} &
        \includegraphics[width=0.15\textwidth]{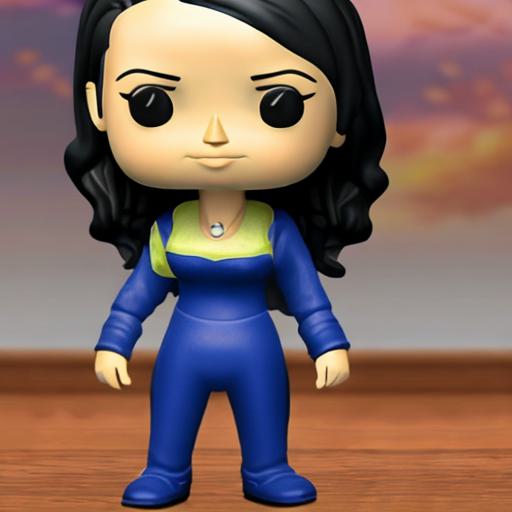} &
        \includegraphics[width=0.15\textwidth]{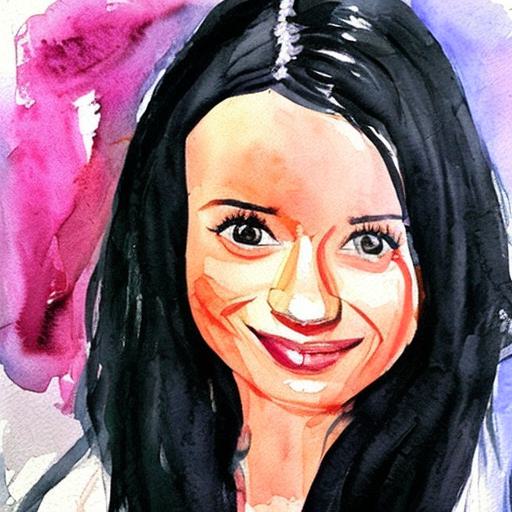} \\

        Real Sample &
        \begin{tabular}{c} ``Banksy art \\ of $S_*$'' \end{tabular} &
        \begin{tabular}{c} ``Cubism painting \\ of $S_*$'' \end{tabular} &
        \begin{tabular}{c} ``Fauvism painting \\ of $S_*$'' \end{tabular} &
        \begin{tabular}{c} ``$S_*$ Funko pop'' \end{tabular} &
        \begin{tabular}{c} ``Watercolor painting \\ of $S_*$'' \end{tabular} \\ \\[-0.185cm]

        \includegraphics[width=0.15\textwidth]{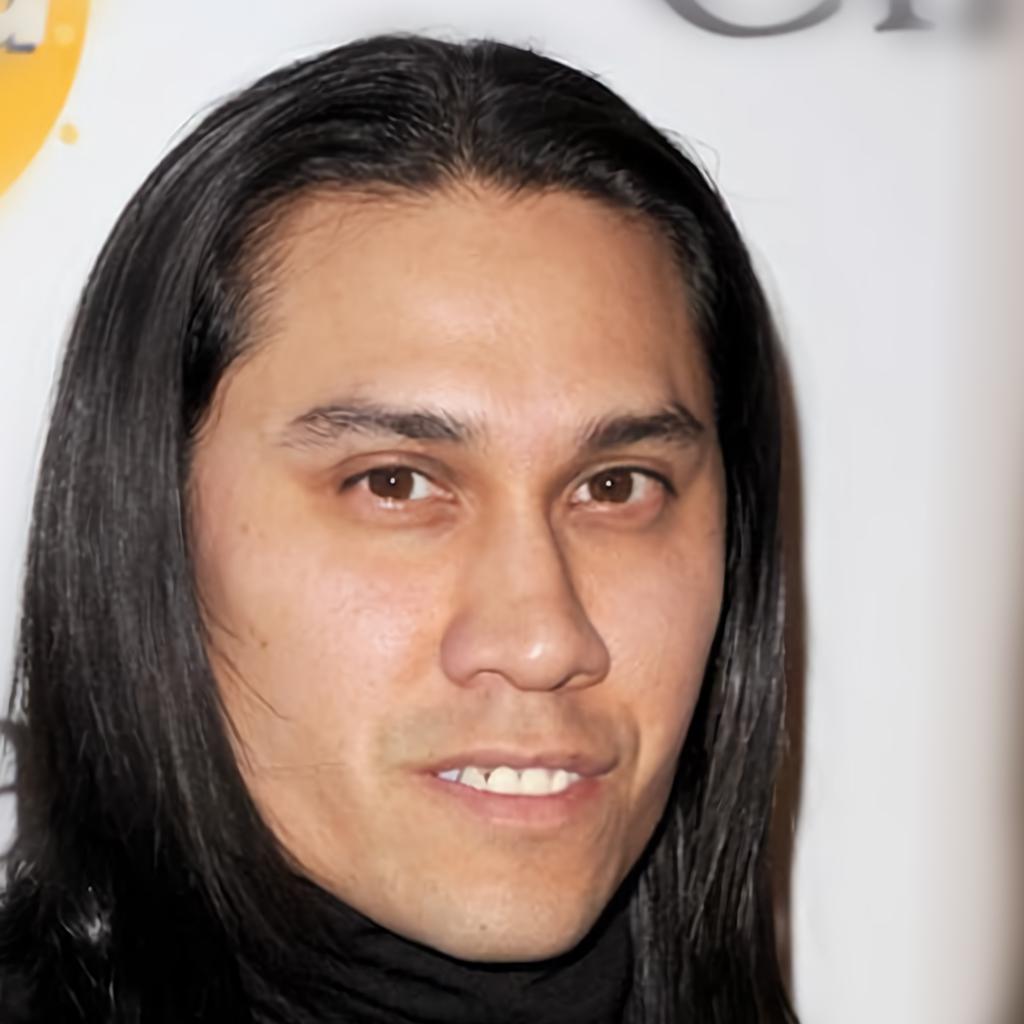} &
        \includegraphics[width=0.15\textwidth]{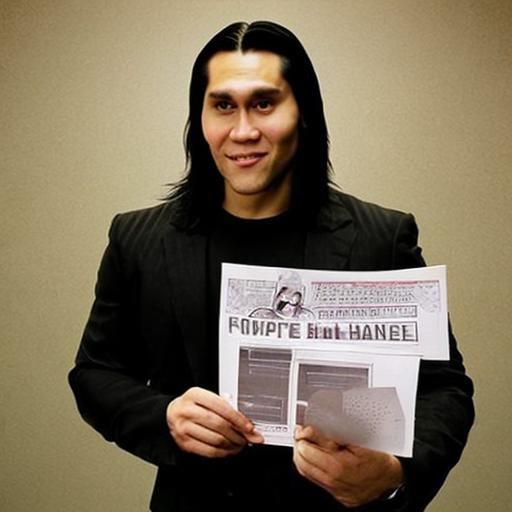} &
        \includegraphics[width=0.15\textwidth]{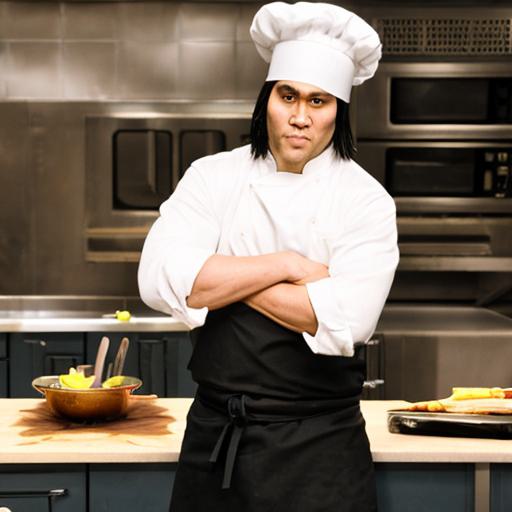} &
        \includegraphics[width=0.15\textwidth]{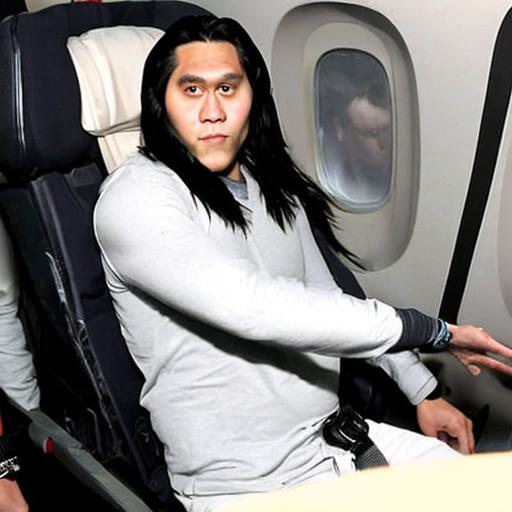} &
        \includegraphics[width=0.15\textwidth]{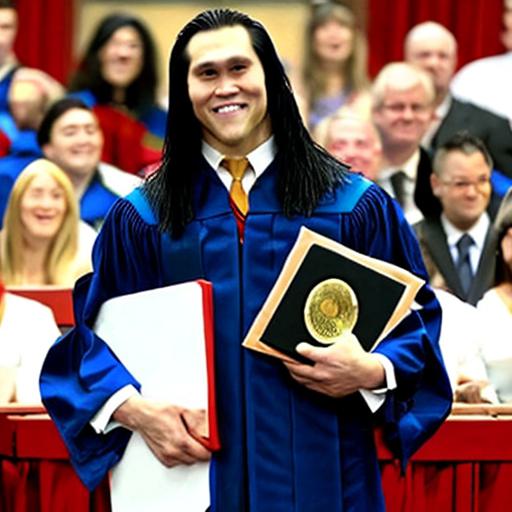} &
        \includegraphics[width=0.15\textwidth]{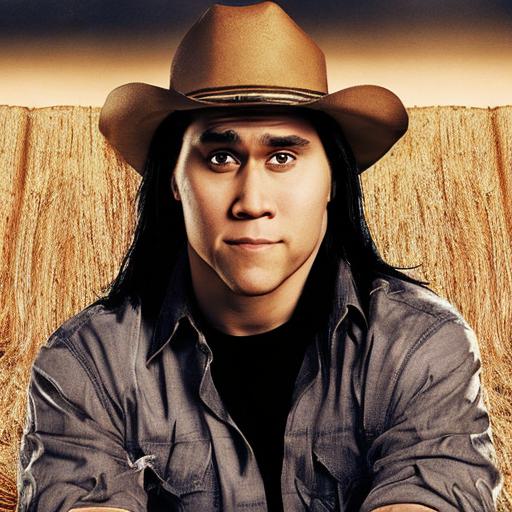} \\

        Real Sample &
        \begin{tabular}{c} ``$S_*$ holding up \\ his accepted paper'' \end{tabular} &
        \begin{tabular}{c} ``$S_*$ wears a \\ chefs hat \\ in the kitchen'' \end{tabular} &
        \begin{tabular}{c} ``$S_*$ buckled in \\ his seat \\ on a plane'' \end{tabular} &
        \begin{tabular}{c} ``A photo of $S_*$ \\ graduating after \\ finishing his PhD'' \end{tabular} &
        \begin{tabular}{c} ``$S_*$ as a cowboy \\ sitting on hay '' \end{tabular} \\ \\[-0.185cm]

        \includegraphics[width=0.15\textwidth]{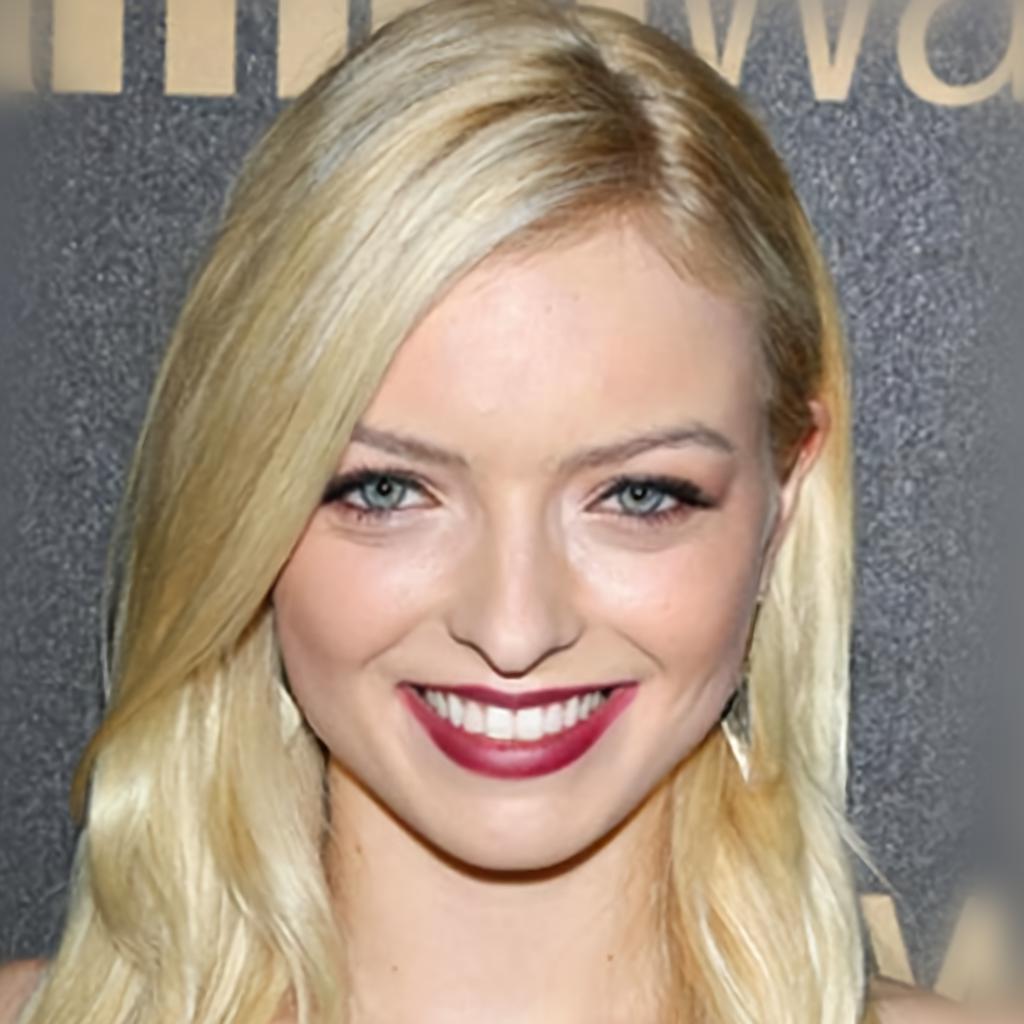} &
        \includegraphics[width=0.15\textwidth]{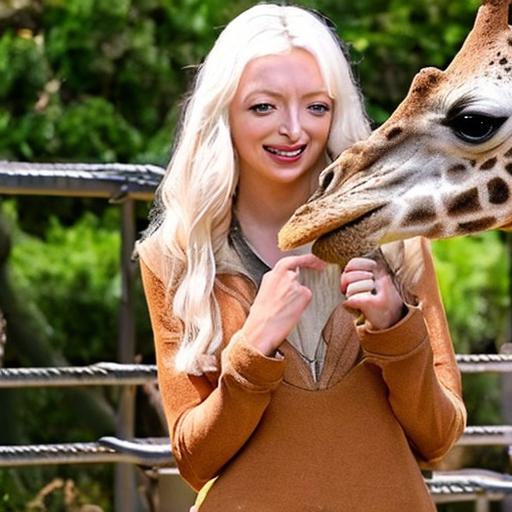} &
        \includegraphics[width=0.15\textwidth]{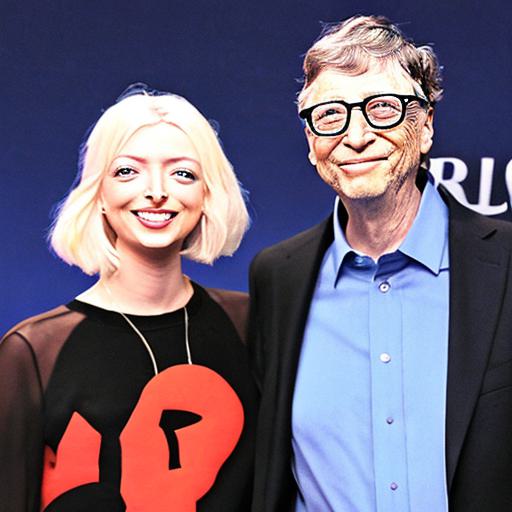} &
        \includegraphics[width=0.15\textwidth]{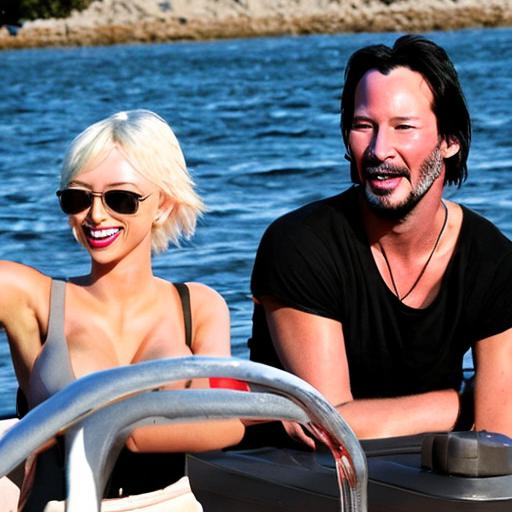} &
        \includegraphics[width=0.15\textwidth]{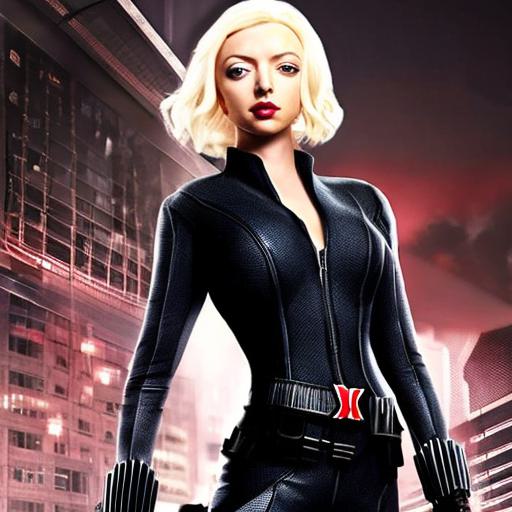} &
        \includegraphics[width=0.15\textwidth]{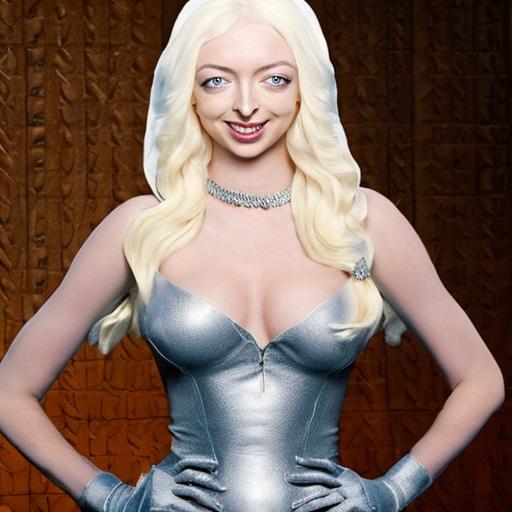} \\

        Real Sample &
        \begin{tabular}{c} ``$S_*$ is feeding \\ giraffes in a sunny \\ open zoo enclosure'' \end{tabular} &
        \begin{tabular}{c} ``$S_*$ and \\ Bill Gates go \\ to a technology \\ exhibition together'' \end{tabular} &
        \begin{tabular}{c} ``$S_*$ and \\ Keanu Reeves \\ on a boat'' \end{tabular} &
        \begin{tabular}{c} ``$S_*$ as \\ Black Widow'' \end{tabular} &
        \begin{tabular}{c} ``$S_*$ as \\ White Queen'' \end{tabular} \\ \\[-0.185cm]

        \includegraphics[width=0.15\textwidth]{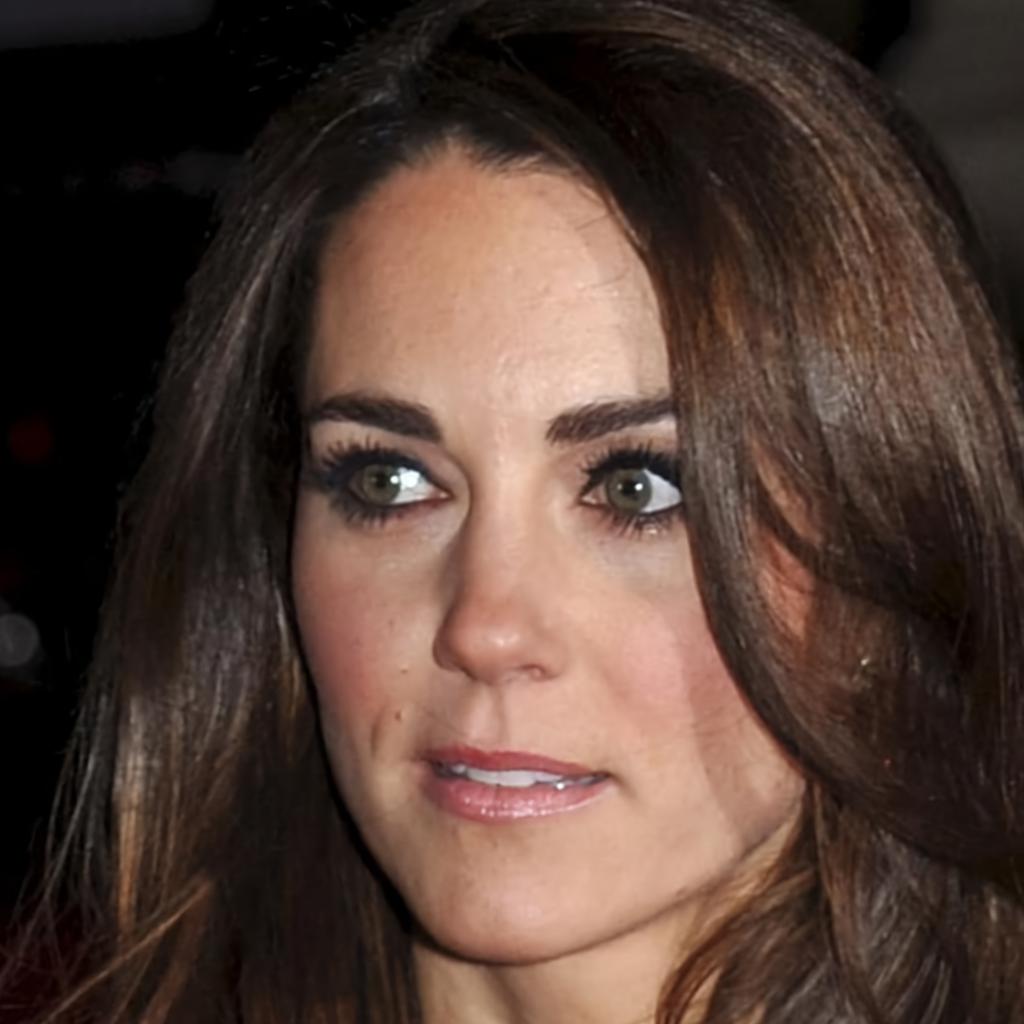} &
        \includegraphics[width=0.15\textwidth]{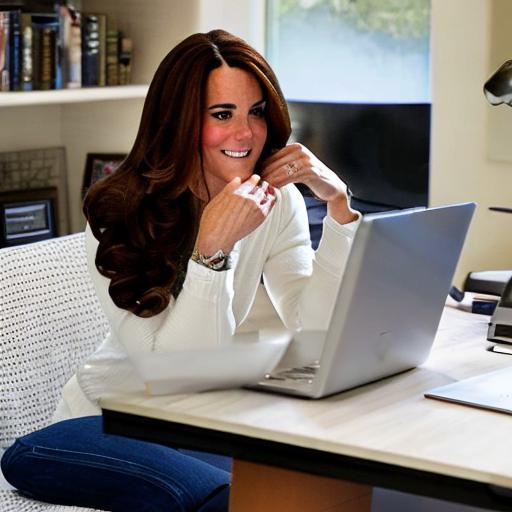} &
        \includegraphics[width=0.15\textwidth]{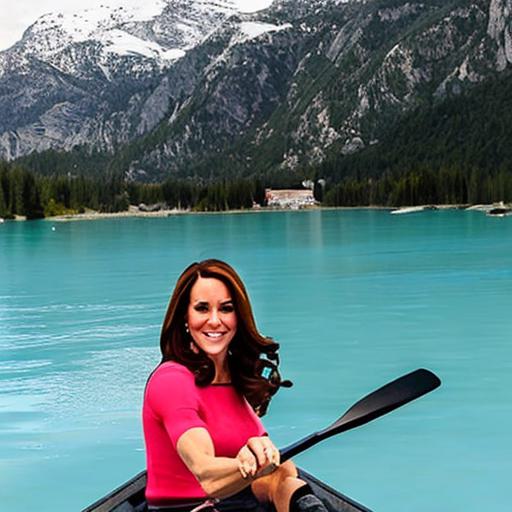} &
        \includegraphics[width=0.15\textwidth]{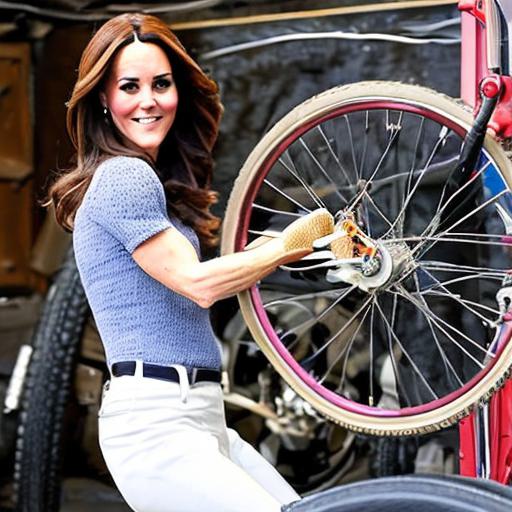} &
        \includegraphics[width=0.15\textwidth]{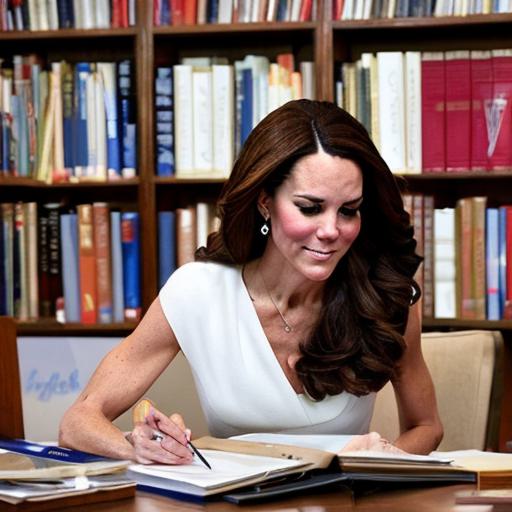} &
        \includegraphics[width=0.15\textwidth]{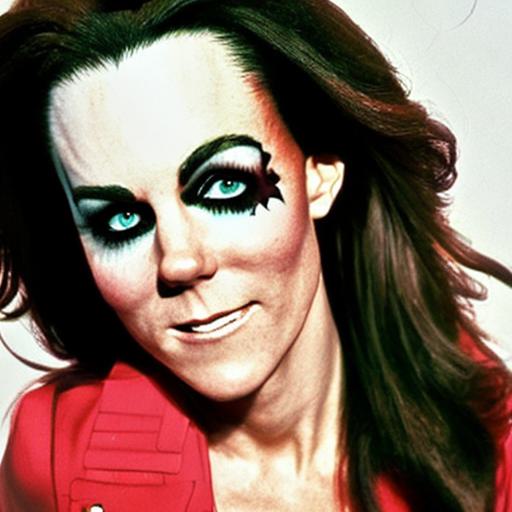} \\

        Real Sample &
        \begin{tabular}{c} ``$S_*$ is coding in \\ a cozy home office'' \end{tabular} &
        \begin{tabular}{c} ``$S_*$ is paddling \\ on a crystal-clear \\ alpine lake'' \end{tabular} &
        \begin{tabular}{c} ``$S_*$ is repairing \\ a vintage bike \\ in a garage'' \end{tabular} &
        \begin{tabular}{c} ``$S_*$ is writing \\ a novel in \\ a home library'' \end{tabular} &
        \begin{tabular}{c} ``$S_*$ as \\ Ziggy Stardust'' \end{tabular} \\ \\[-0.185cm]

    \\[-0.4cm]        
    \end{tabular}
    }
    \caption{Images generated by our fast version method with a learning rate of 0.08. Results are obtained after 25 optimization steps, taking only 26 seconds.}
    \label{fig:appendix_our_with_lr_008_2}
\end{figure*}

As illustrated in \cref{sec:results}, we developed a fast version of our method with a learning rate of 0.08. This fast version enables learning of the new concept in 25 optimization steps, taking only 26 seconds. In \cref{fig:appendix_our_with_lr_008_1,fig:appendix_our_with_lr_008_2}, we provide qualitative results of applying this fast version to a variety of prompts. The results demonstrate that our fast version allows for high-quality personalized face generation within a remarkably short training time.

\section{Additional Qualitative Comparisons}
\label{sec:additional_comparison}
\begin{figure*}
    \centering
    \renewcommand{\arraystretch}{0.3}
    \setlength{\tabcolsep}{0.5pt}

    {\footnotesize
    \begin{tabular}{c@{\hspace{0.15cm}} c c @{\hspace{0.15cm}} c c @{\hspace{0.15cm}} c c @{\hspace{0.15cm}} c c @{\hspace{0.15cm}} c c }

        Real Sample &
        \multicolumn{2}{c}{Textual Inversion} &
        \multicolumn{2}{c}{DreamBooth} &
        \multicolumn{2}{c}{NeTI} &
        \multicolumn{2}{c}{Celeb Basis} &
        \multicolumn{2}{c}{Ours} \\
        \includegraphics[width=0.08\textwidth]{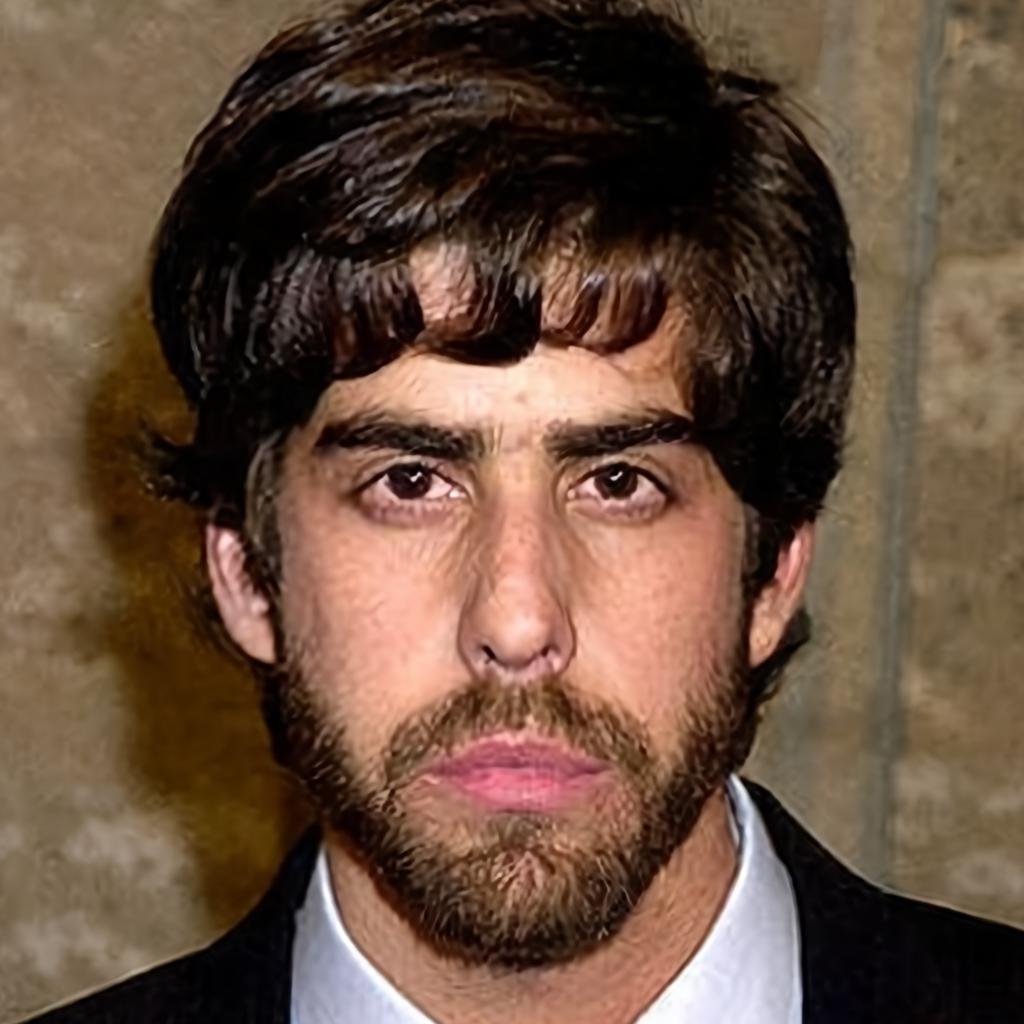} &
        \includegraphics[width=0.08\textwidth]{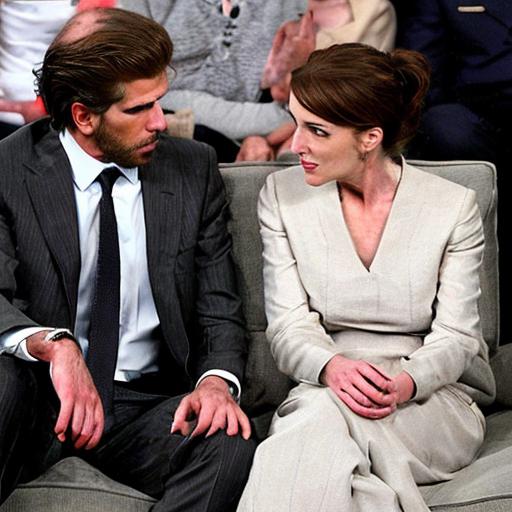} &
        \includegraphics[width=0.08\textwidth]{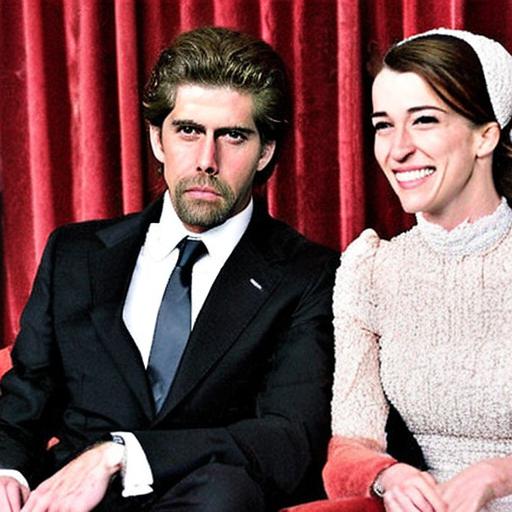} &
        \hspace{0.05cm}
        \includegraphics[width=0.08\textwidth]{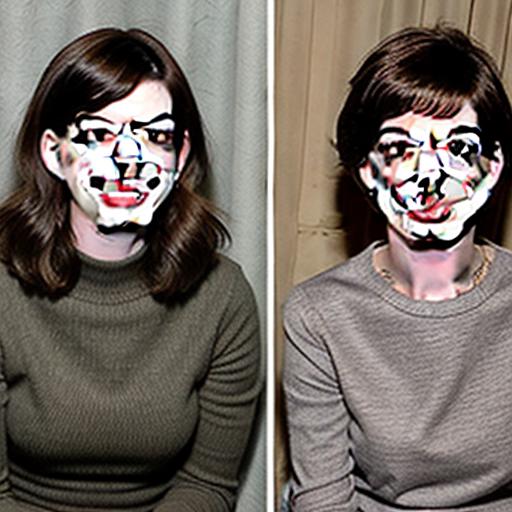} &
        \includegraphics[width=0.08\textwidth]{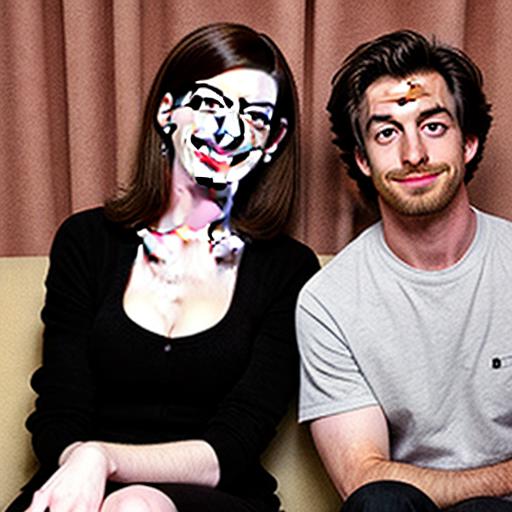} &
        \hspace{0.05cm}
        \includegraphics[width=0.08\textwidth]{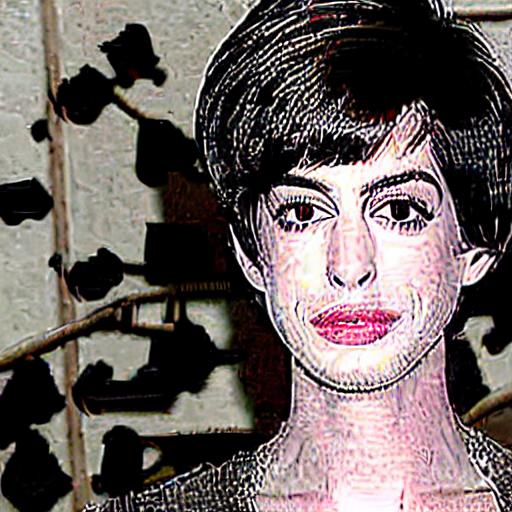} &
        \includegraphics[width=0.08\textwidth]{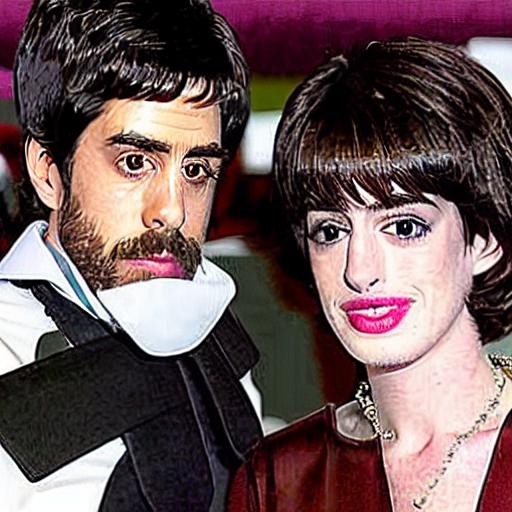} &
        \hspace{0.05cm}
        \includegraphics[width=0.08\textwidth]{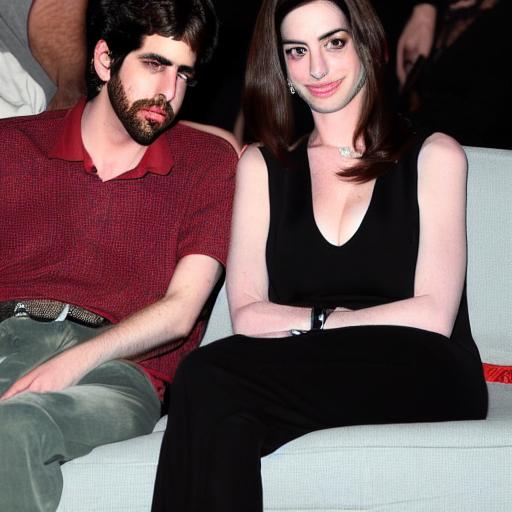} &
        \includegraphics[width=0.08\textwidth]{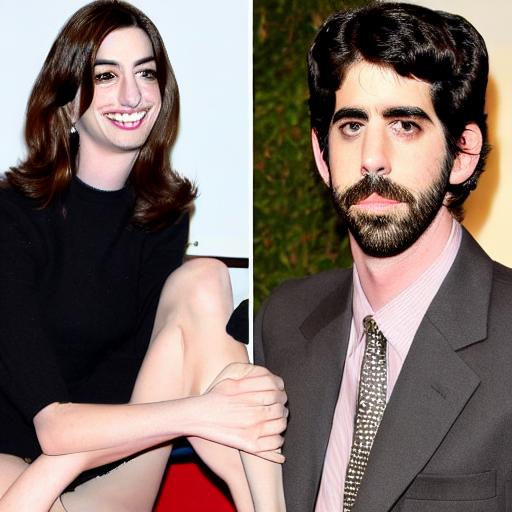} &
        \hspace{0.05cm}
        \includegraphics[width=0.08\textwidth]{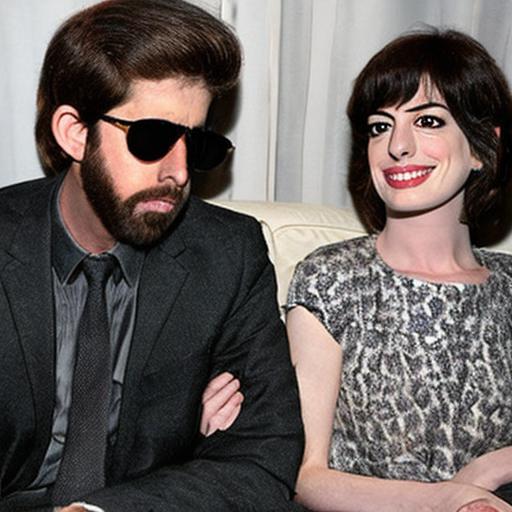} &
        \includegraphics[width=0.08\textwidth]{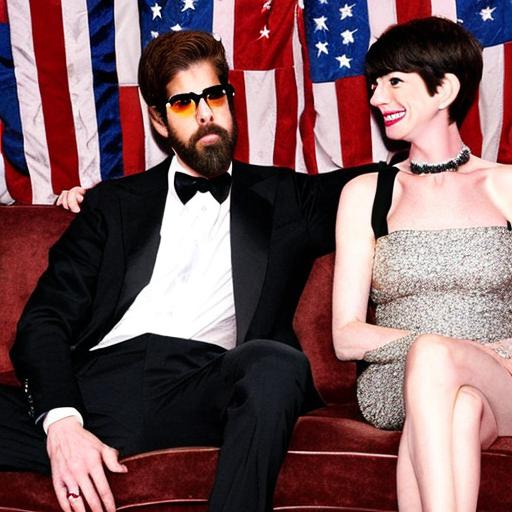} \\
        
        \raisebox{0.325in}{\begin{tabular}{c} ``$S_*$ and \\ Anne Hathaway \\ sit on a sofa''\end{tabular}} &
        \includegraphics[width=0.08\textwidth]{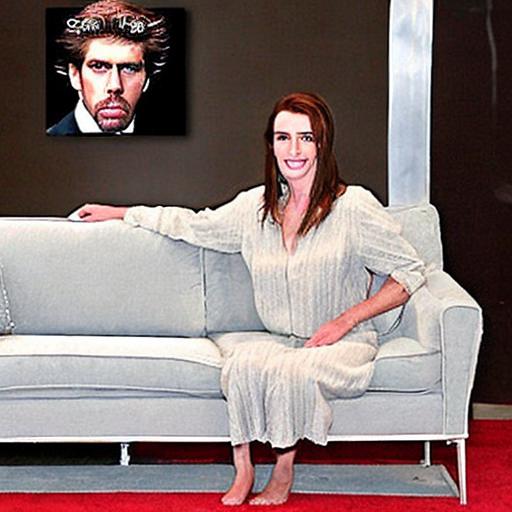} &
        \includegraphics[width=0.08\textwidth]{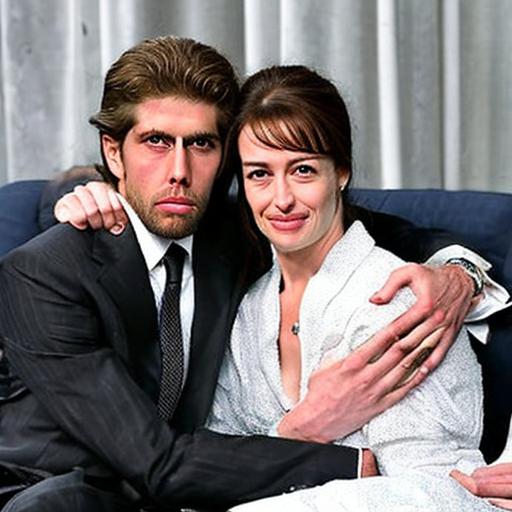} &
        \hspace{0.05cm}
        \includegraphics[width=0.08\textwidth]{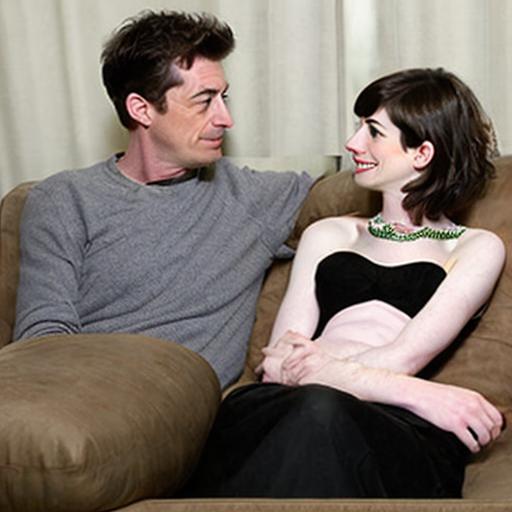} &
        \includegraphics[width=0.08\textwidth]{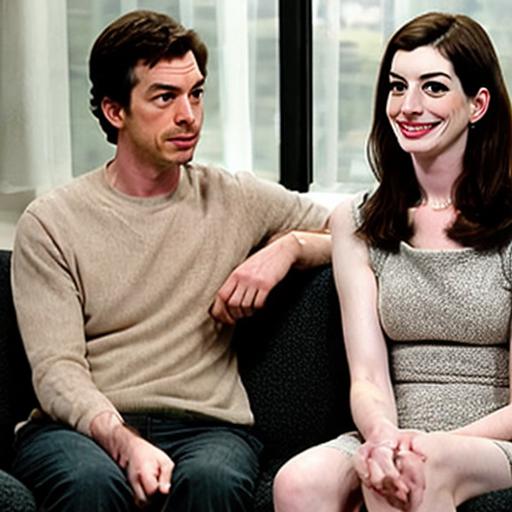} &
        \hspace{0.05cm}
        \includegraphics[width=0.08\textwidth]{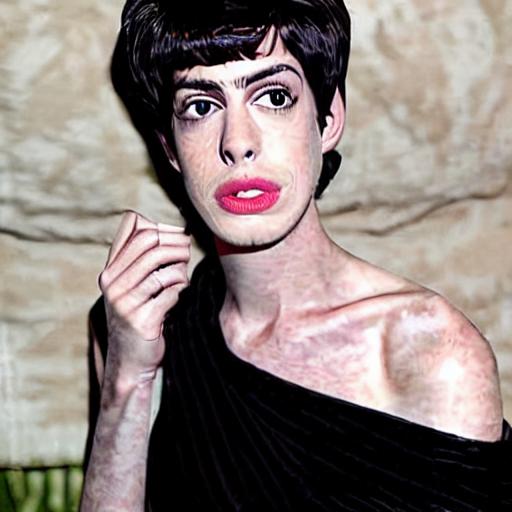} &
        \includegraphics[width=0.08\textwidth]{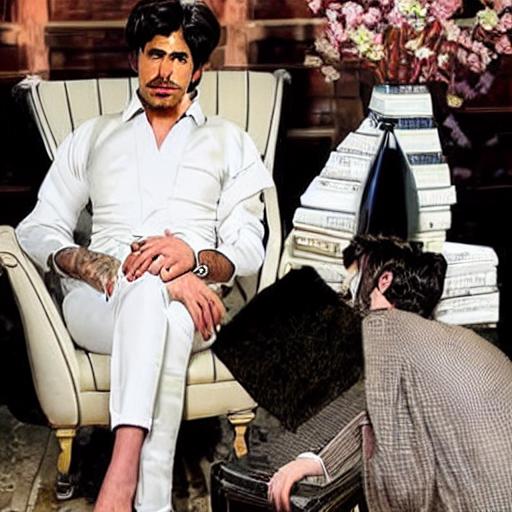} &
        \hspace{0.05cm}
        \includegraphics[width=0.08\textwidth]{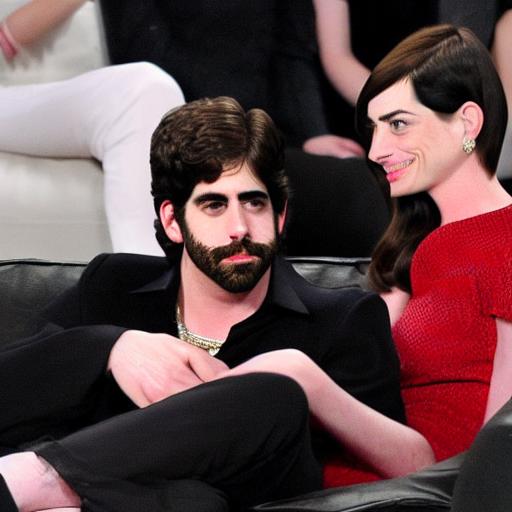} &
        \includegraphics[width=0.08\textwidth]{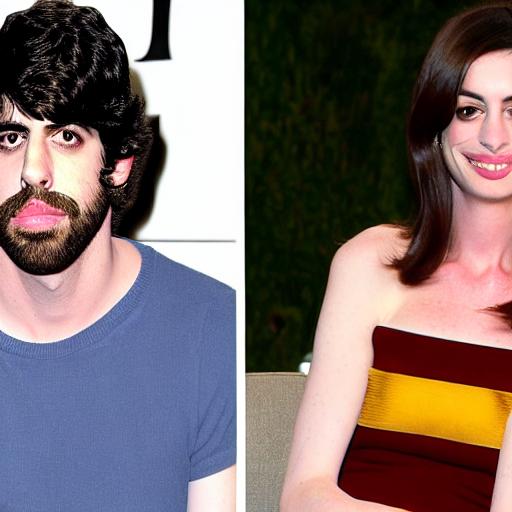} &
        \hspace{0.05cm}
        \includegraphics[width=0.08\textwidth]{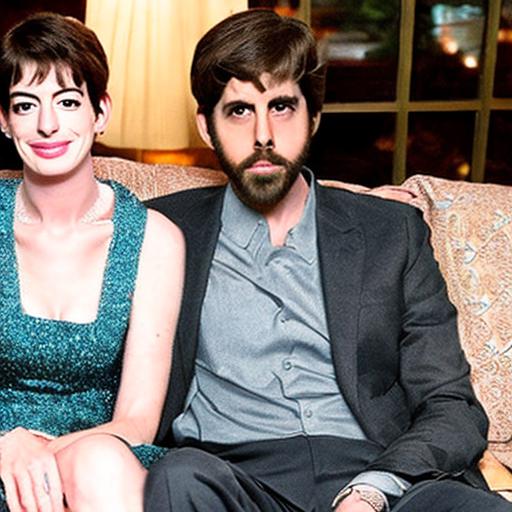} &
        \includegraphics[width=0.08\textwidth]{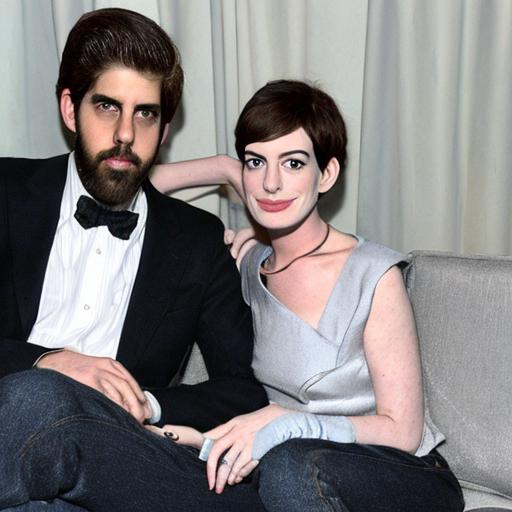} \\ \\
        
        \includegraphics[width=0.08\textwidth]{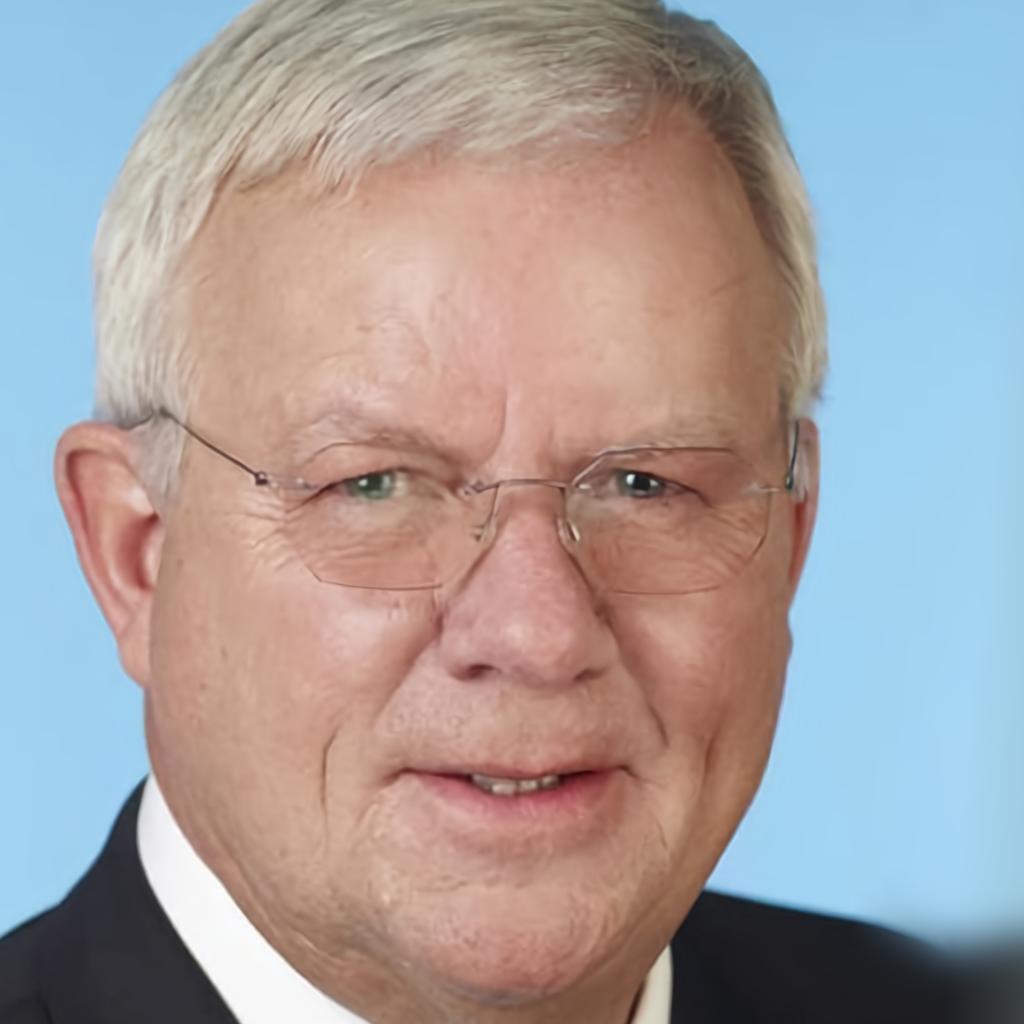} &
        \includegraphics[width=0.08\textwidth]{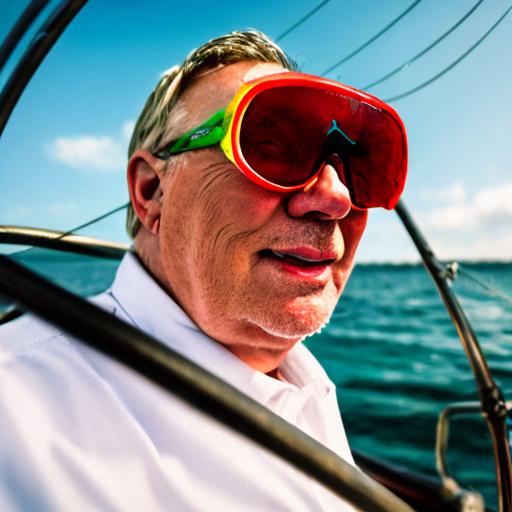} &
        \includegraphics[width=0.08\textwidth]{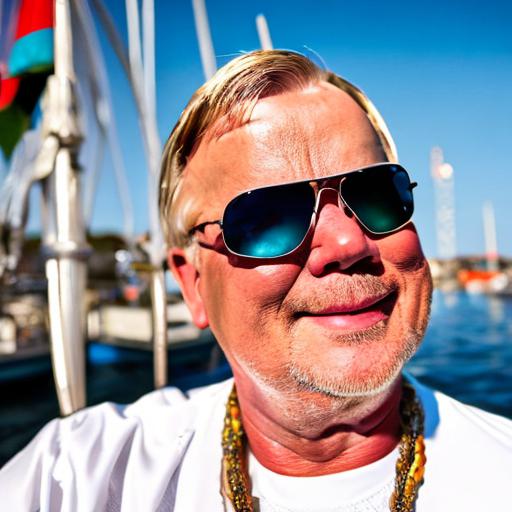} &
        \hspace{0.05cm}
        \includegraphics[width=0.08\textwidth]{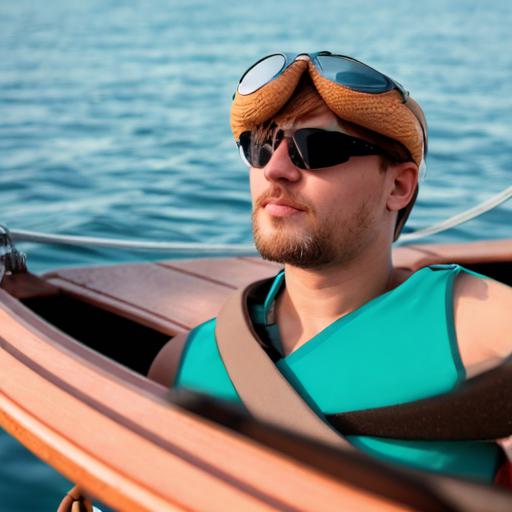} &
        \includegraphics[width=0.08\textwidth]{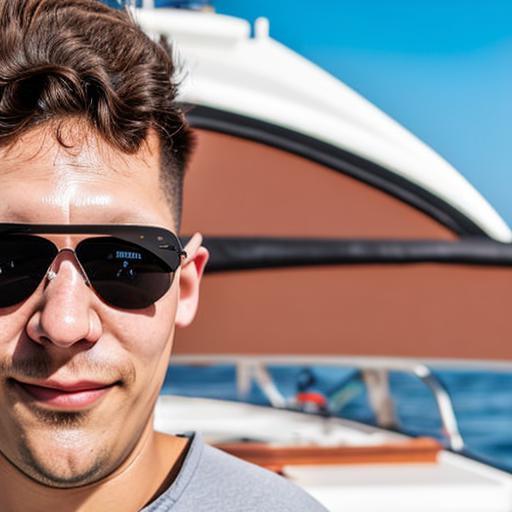} &
        \hspace{0.05cm}
        \includegraphics[width=0.08\textwidth]{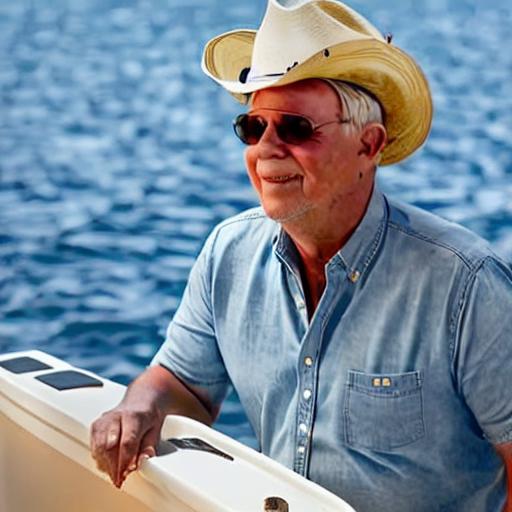} &
        \includegraphics[width=0.08\textwidth]{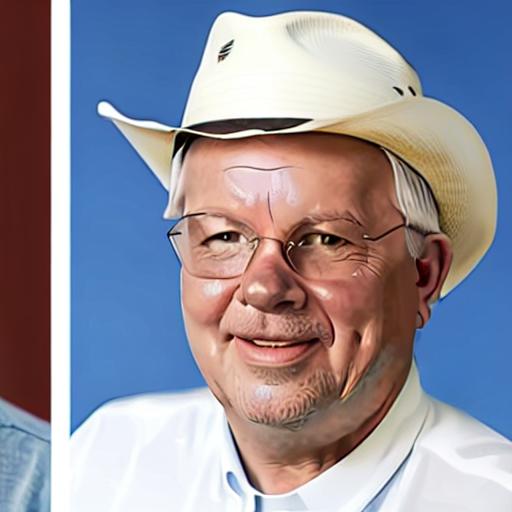} &
        \hspace{0.05cm}
        \includegraphics[width=0.08\textwidth]{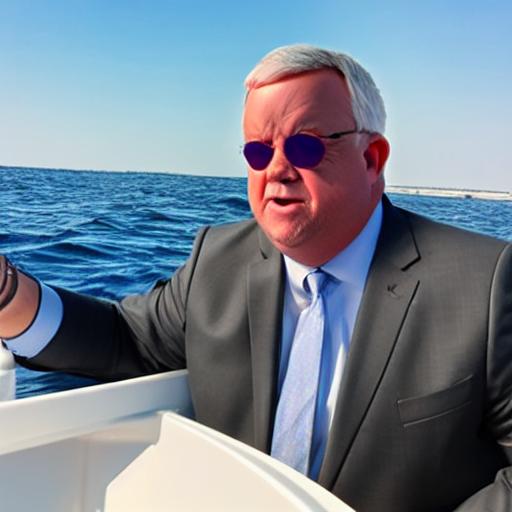} &
        \includegraphics[width=0.08\textwidth]{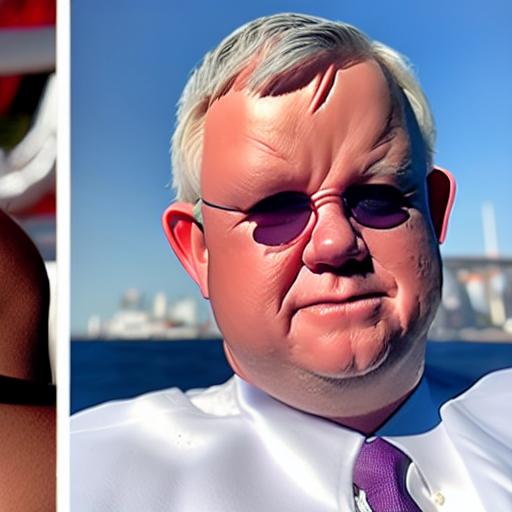} &
        \hspace{0.05cm}
        \includegraphics[width=0.08\textwidth]{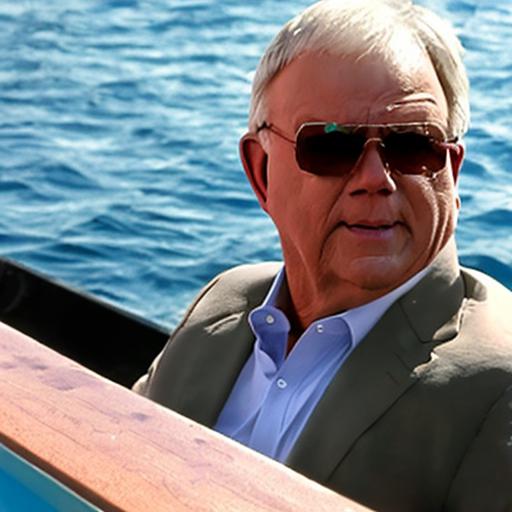} &
        \includegraphics[width=0.08\textwidth]{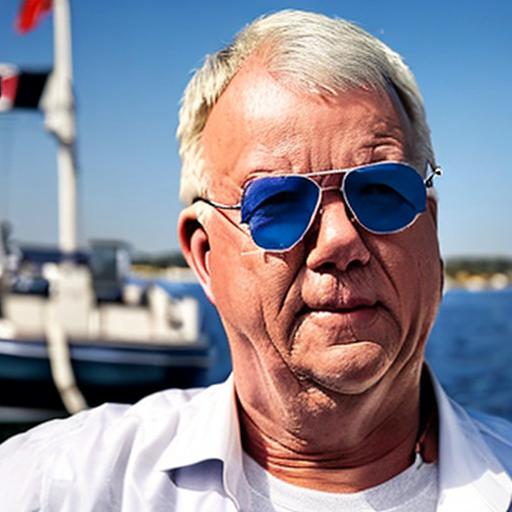} \\
        
        \raisebox{0.325in}{\begin{tabular}{c} ``$S_*$ wears \\ a sunglass
 \\ on a boat''\end{tabular}} &
        \includegraphics[width=0.08\textwidth]{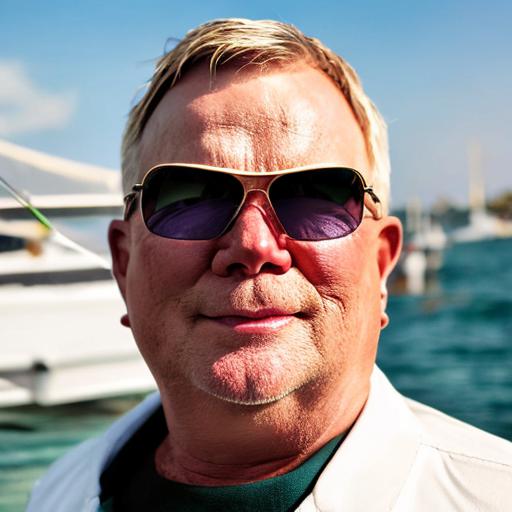} &
        \includegraphics[width=0.08\textwidth]{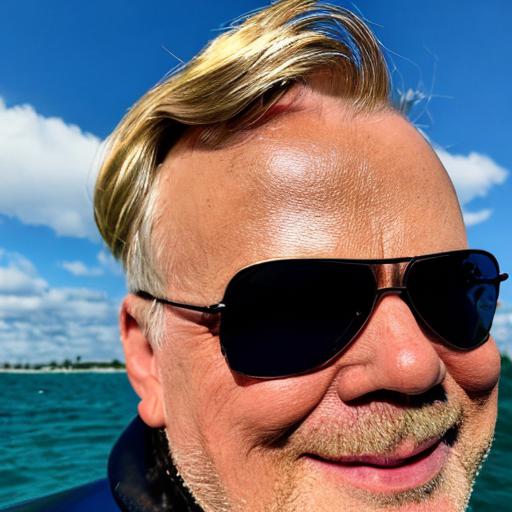} &
        \hspace{0.05cm}
        \includegraphics[width=0.08\textwidth]{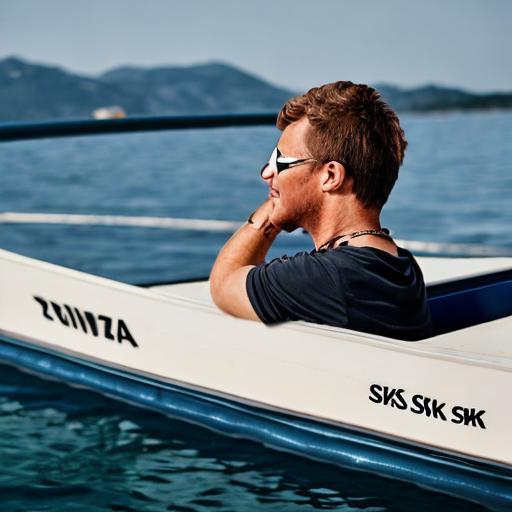} &
        \includegraphics[width=0.08\textwidth]{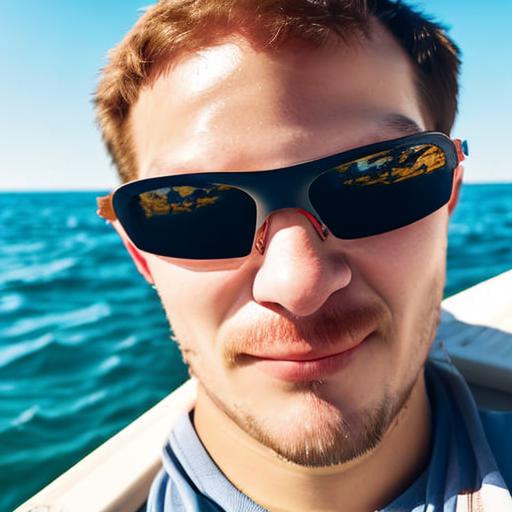} &
        \hspace{0.05cm}
        \includegraphics[width=0.08\textwidth]{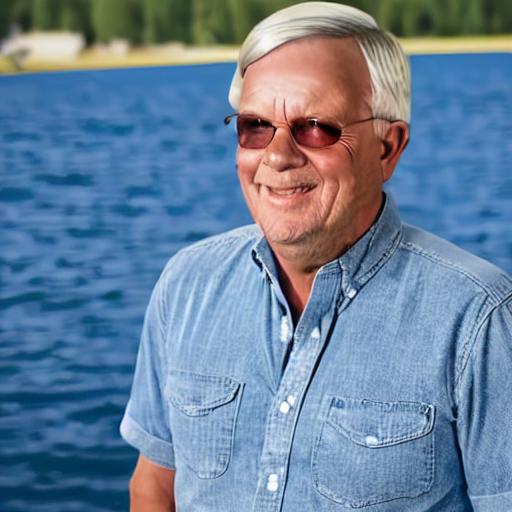} &
        \includegraphics[width=0.08\textwidth]{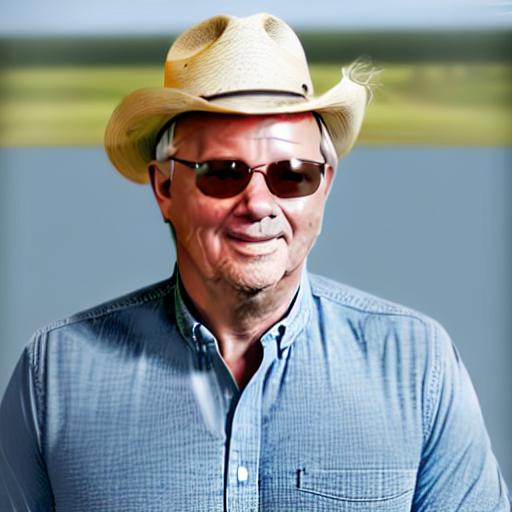} &
        \hspace{0.05cm}
        \includegraphics[width=0.08\textwidth]{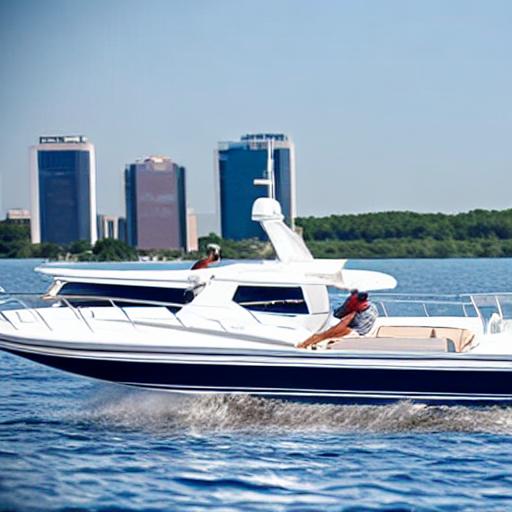} &
        \includegraphics[width=0.08\textwidth]{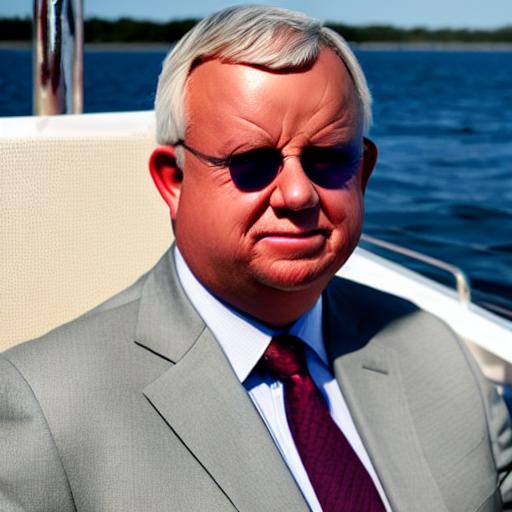} &
        \hspace{0.05cm}
        \includegraphics[width=0.08\textwidth]{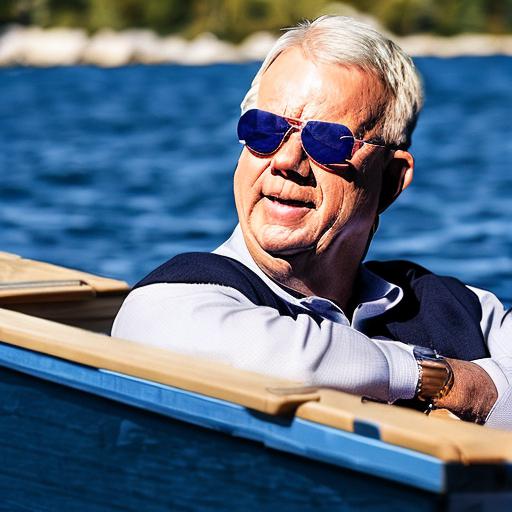} &
        \includegraphics[width=0.08\textwidth]{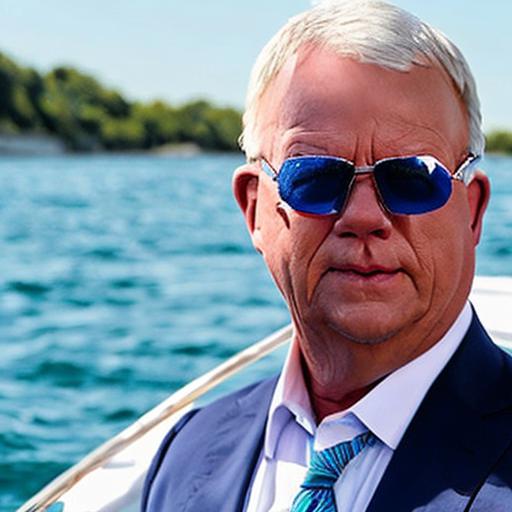} \\ \\

        \includegraphics[width=0.08\textwidth]{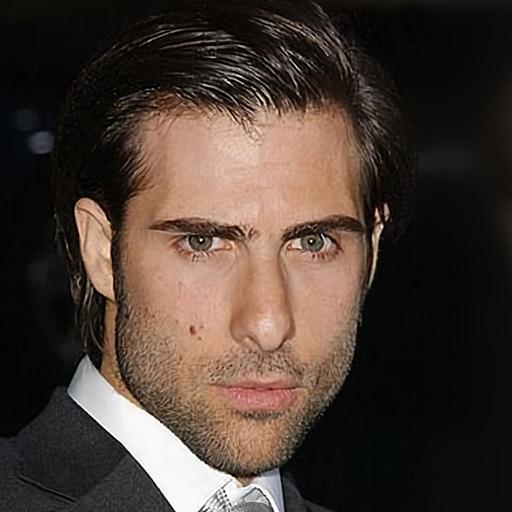} &
        \includegraphics[width=0.08\textwidth]{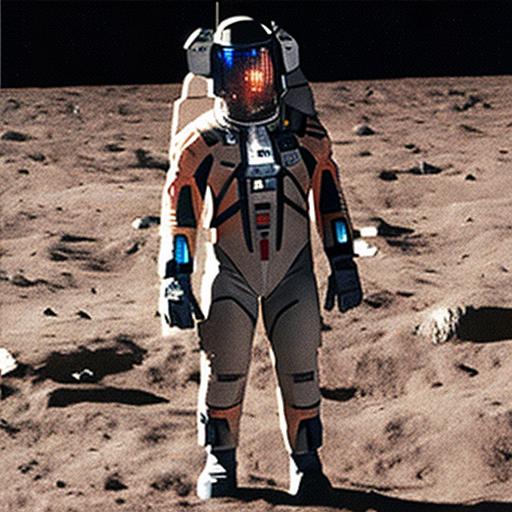} &
        \includegraphics[width=0.08\textwidth]{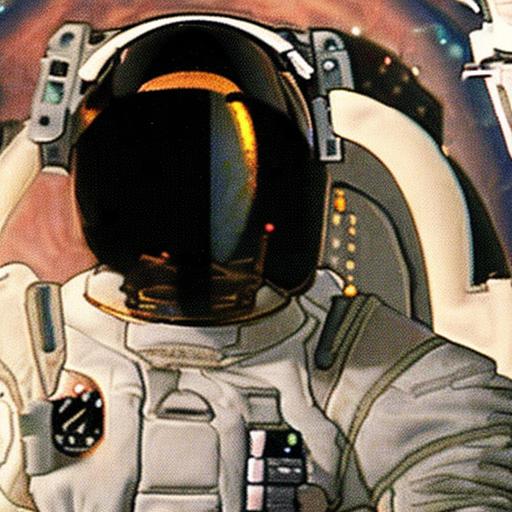} &
        \hspace{0.05cm}
        \includegraphics[width=0.08\textwidth]{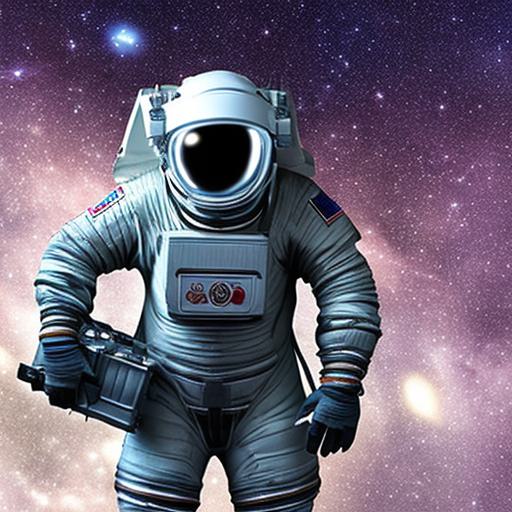} &
        \includegraphics[width=0.08\textwidth]{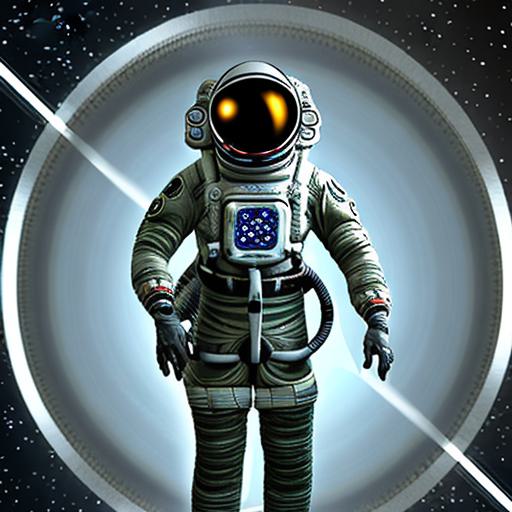} &
        \hspace{0.05cm}
        \includegraphics[width=0.08\textwidth]{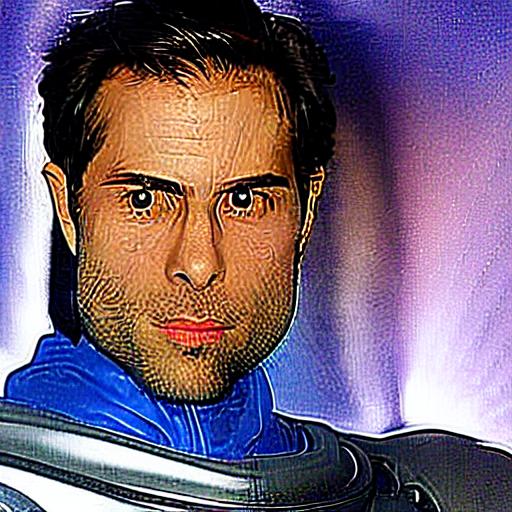} &
        \includegraphics[width=0.08\textwidth]{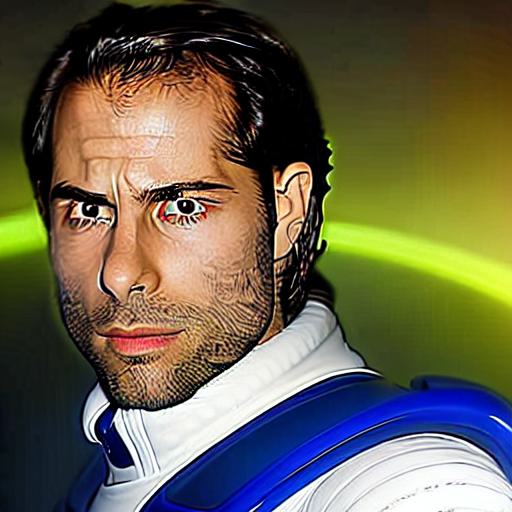} &
        \hspace{0.05cm}
        \includegraphics[width=0.08\textwidth]{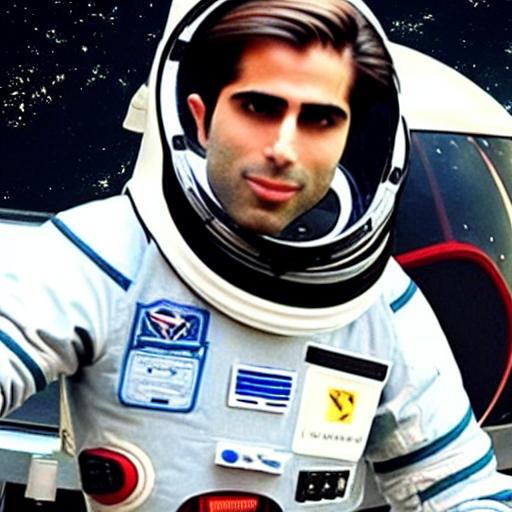} &
        \includegraphics[width=0.08\textwidth]{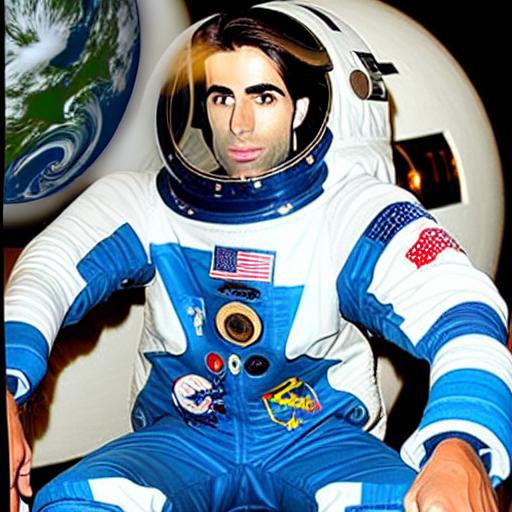} &
        \hspace{0.05cm}
        \includegraphics[width=0.08\textwidth]{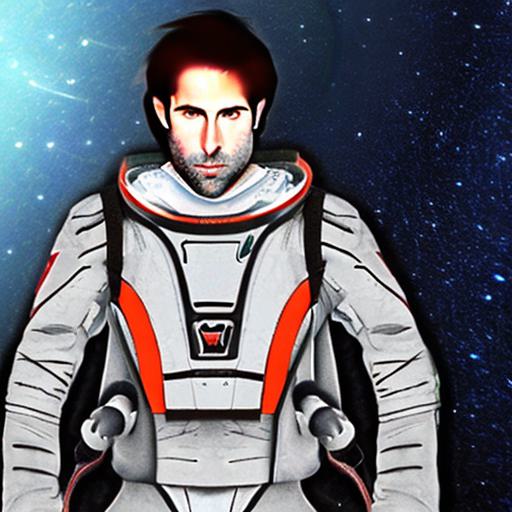} &
        \includegraphics[width=0.08\textwidth]{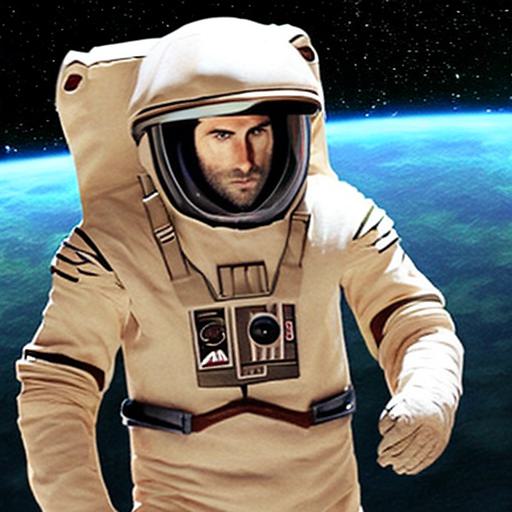} \\
        
        \raisebox{0.325in}{\begin{tabular}{c} ``$S_*$ wearing \\ \\[-0.05cm] a scifi spacesuit  \\ \\[-0.05cm]in space''\end{tabular}} &
        \includegraphics[width=0.08\textwidth]{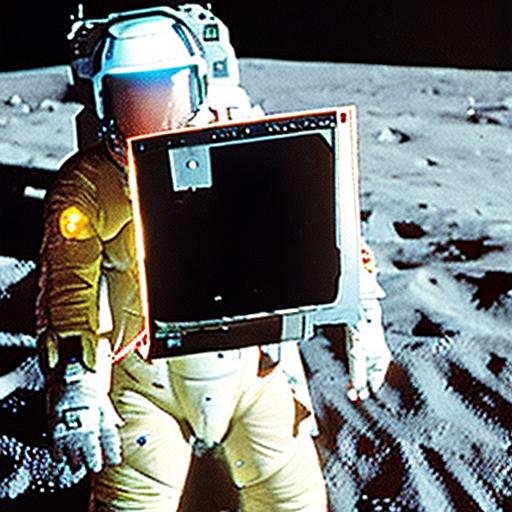} &
        \includegraphics[width=0.08\textwidth]{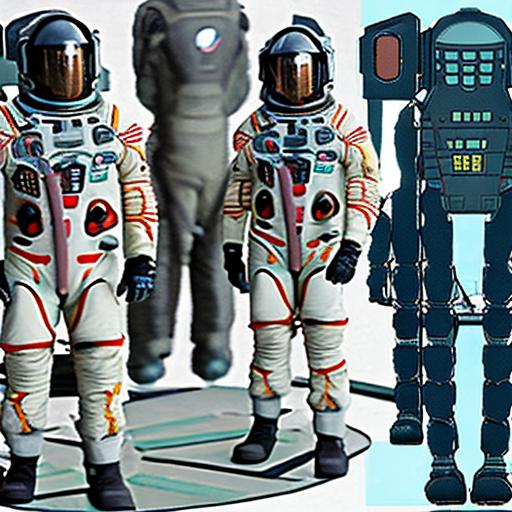} &
        \hspace{0.05cm}
        \includegraphics[width=0.08\textwidth]{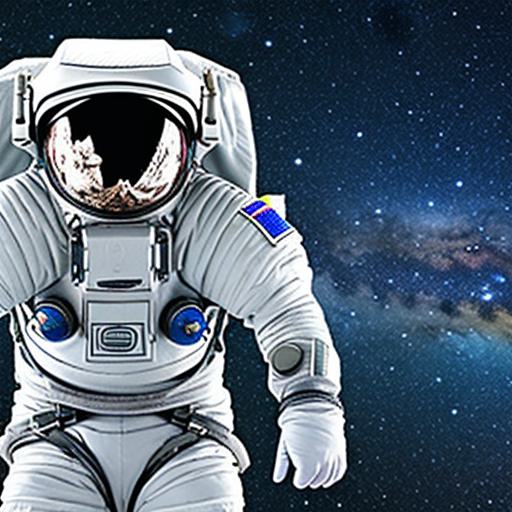} &
        \includegraphics[width=0.08\textwidth]{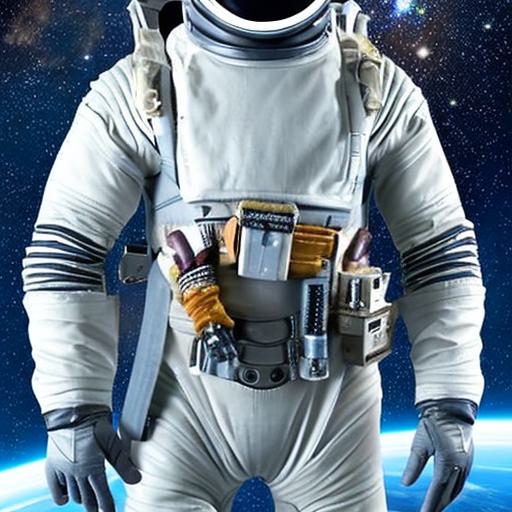} &
        \hspace{0.05cm}
        \includegraphics[width=0.08\textwidth]{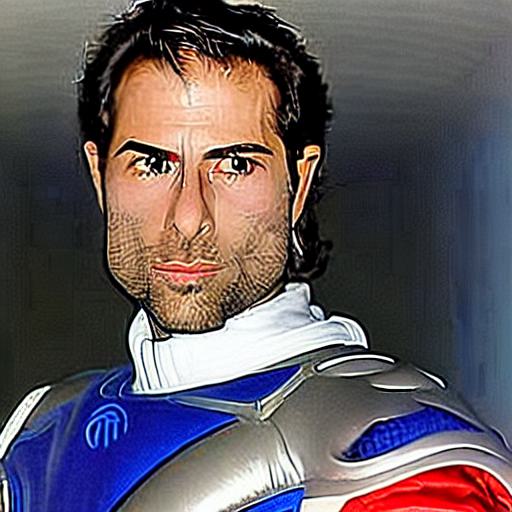} &
        \includegraphics[width=0.08\textwidth]{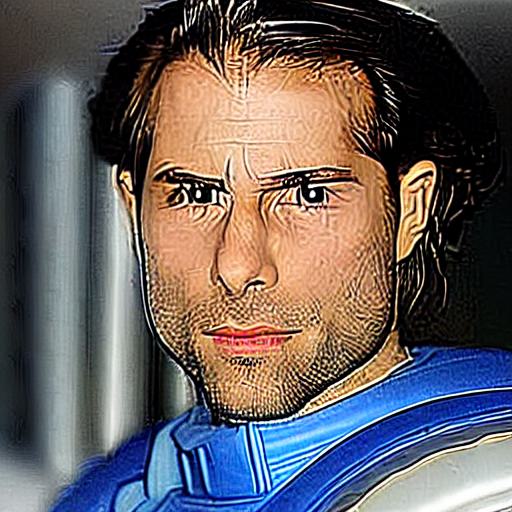} &
        \hspace{0.05cm}
        \includegraphics[width=0.08\textwidth]{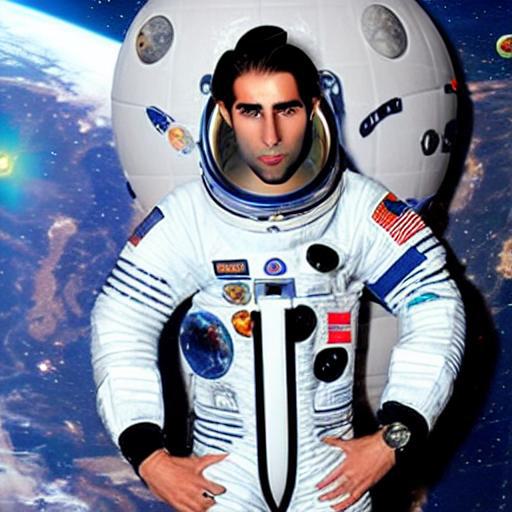} &
        \includegraphics[width=0.08\textwidth]{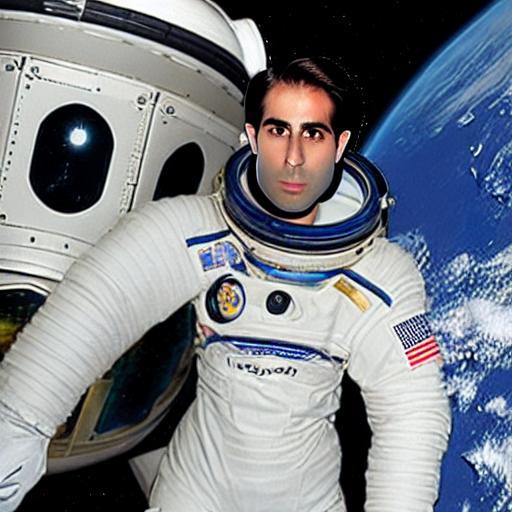} &
        \hspace{0.05cm}
        \includegraphics[width=0.08\textwidth]{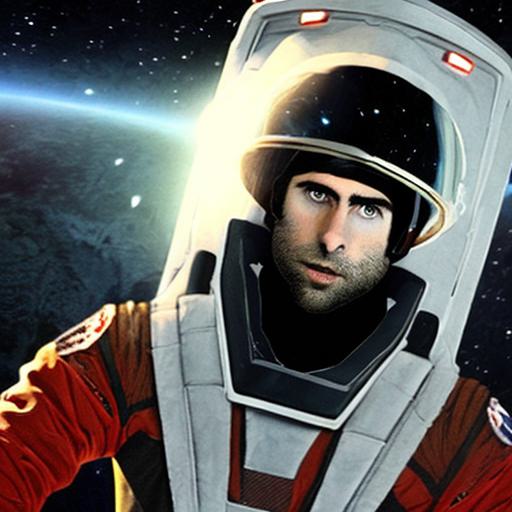} &
        \includegraphics[width=0.08\textwidth]{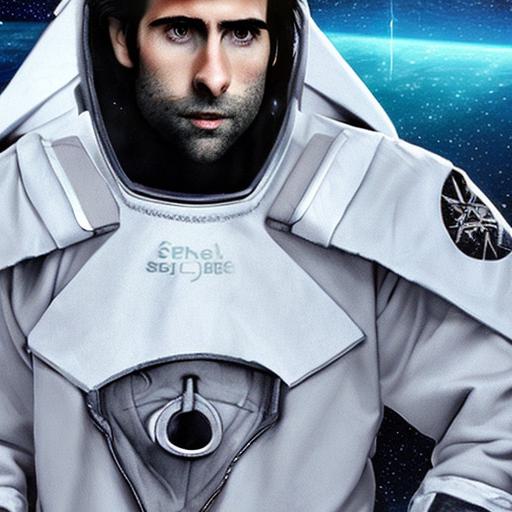} \\ \\

        \includegraphics[width=0.08\textwidth]{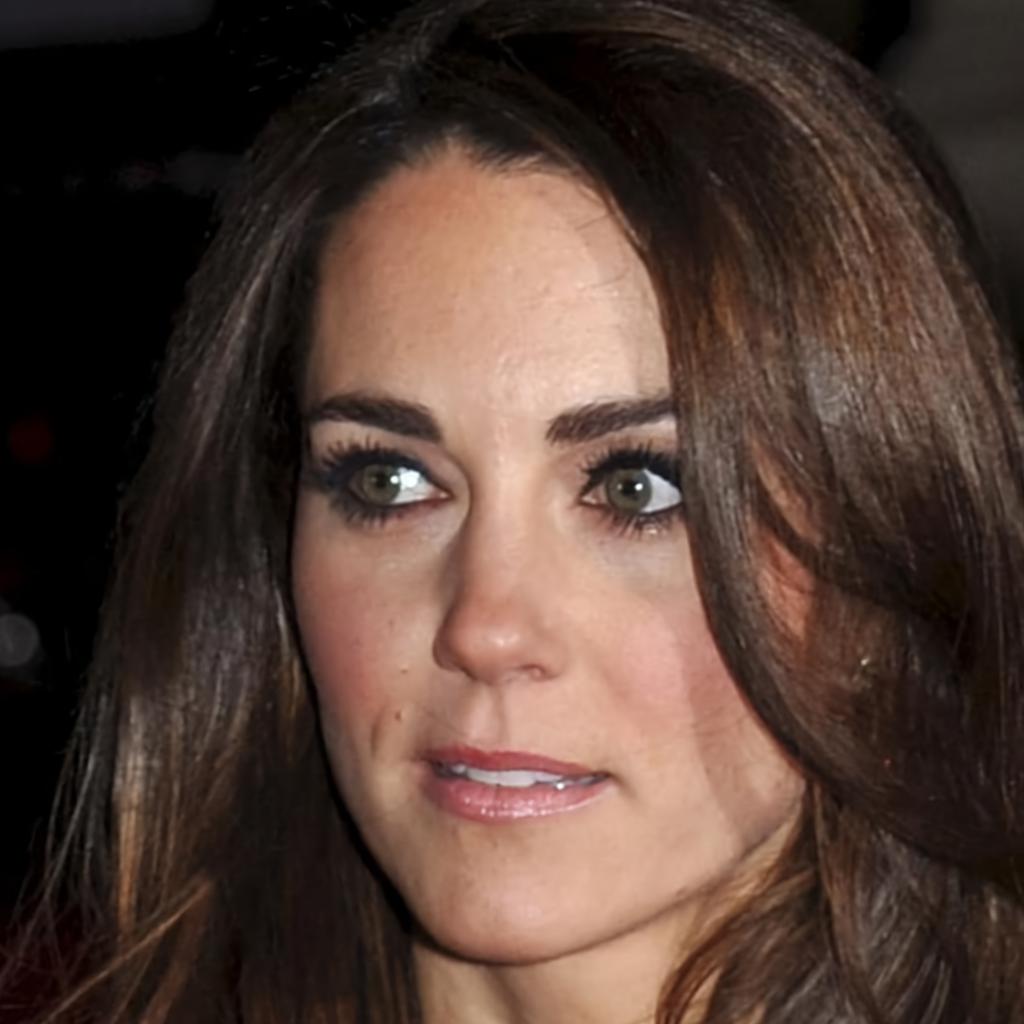} &
        \includegraphics[width=0.08\textwidth]{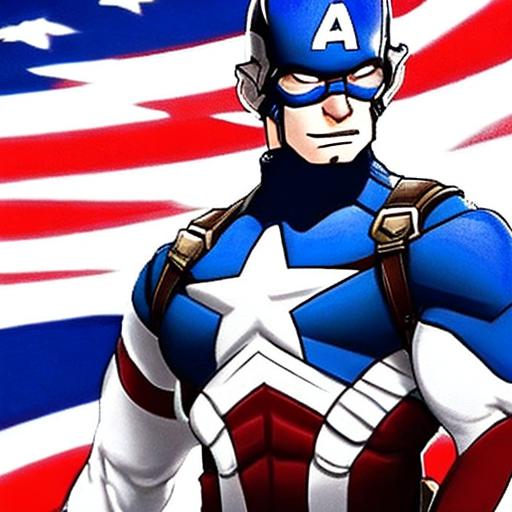} &
        \includegraphics[width=0.08\textwidth]{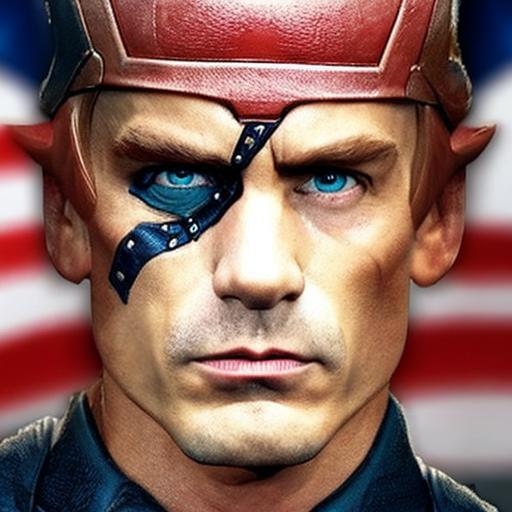} &
        \hspace{0.05cm}
        \includegraphics[width=0.08\textwidth]{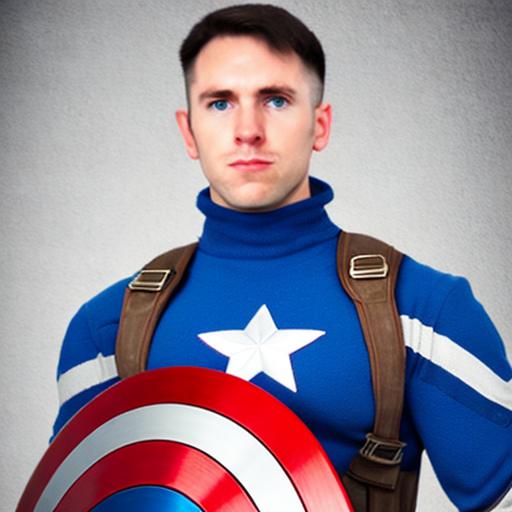} &
        \includegraphics[width=0.08\textwidth]{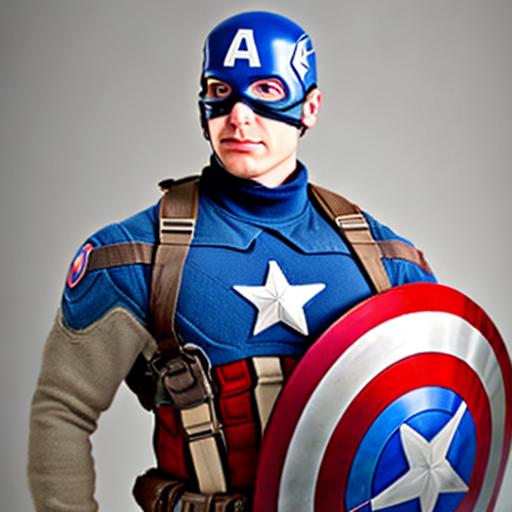} &
        \hspace{0.05cm}
        \includegraphics[width=0.08\textwidth]{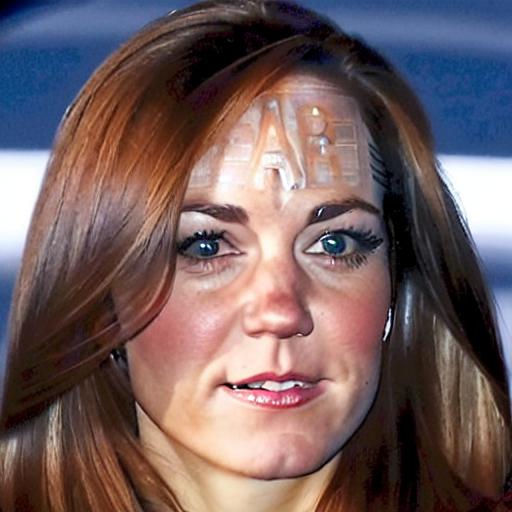} &
        \includegraphics[width=0.08\textwidth]{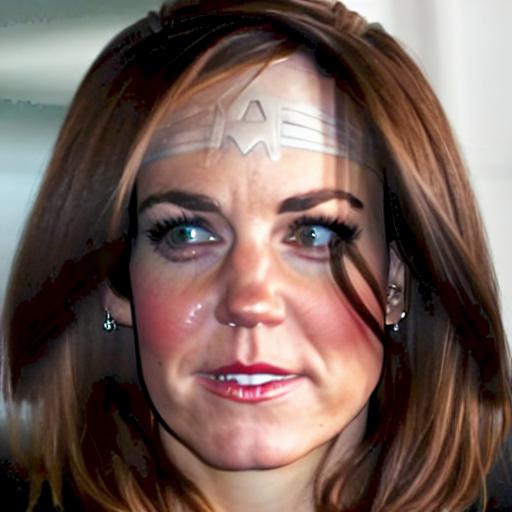} &
        \hspace{0.05cm}
        \includegraphics[width=0.08\textwidth]{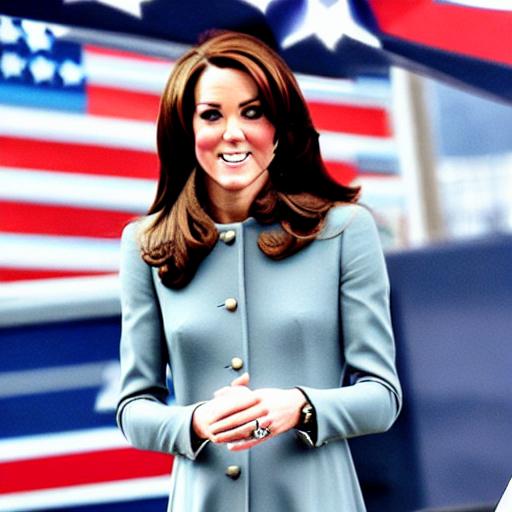} &
        \includegraphics[width=0.08\textwidth]{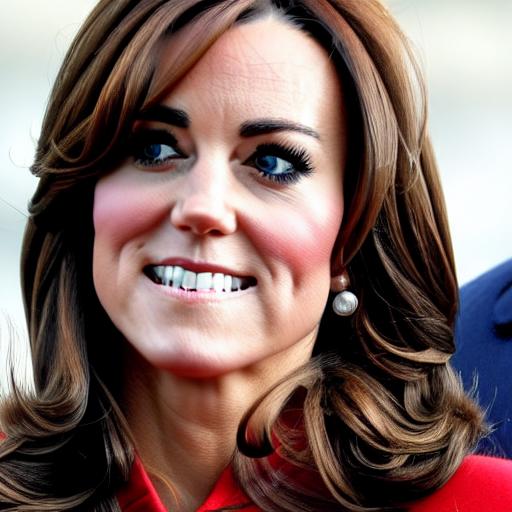} &
        \hspace{0.05cm}
        \includegraphics[width=0.08\textwidth]{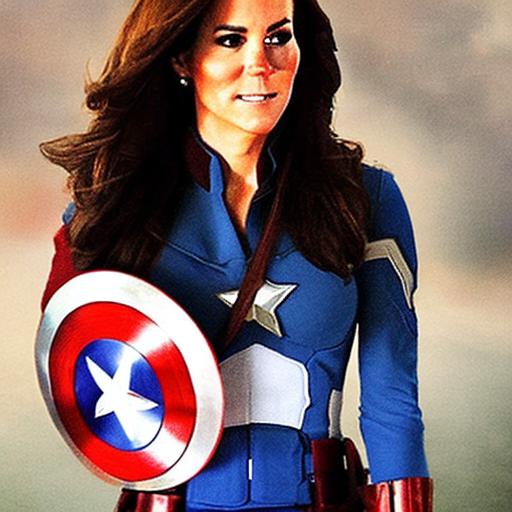} &
        \includegraphics[width=0.08\textwidth]{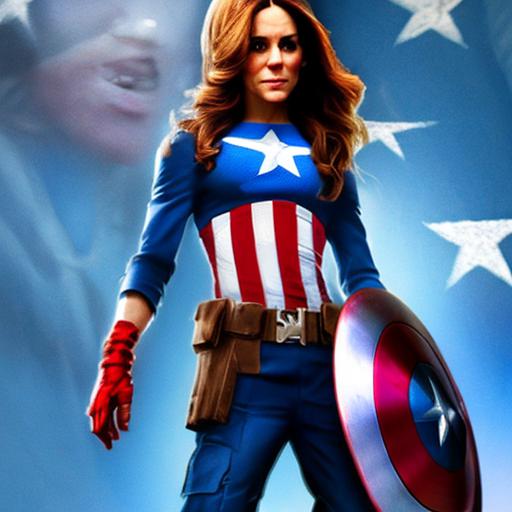} \\
        
        \raisebox{0.325in}{\begin{tabular}{c} ``$S_*$ as \\ Captain \\ America''\end{tabular}} &
        \includegraphics[width=0.08\textwidth]{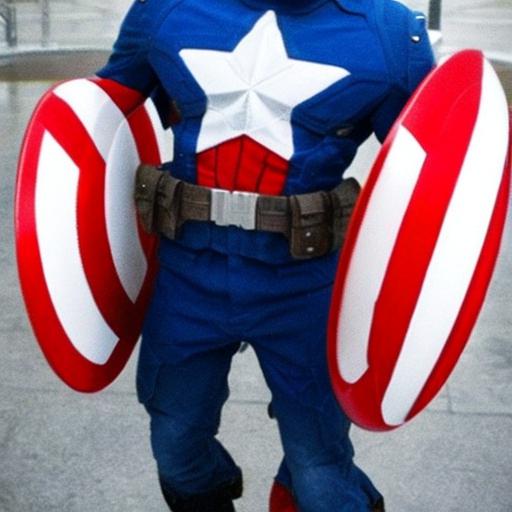} &
        \includegraphics[width=0.08\textwidth]{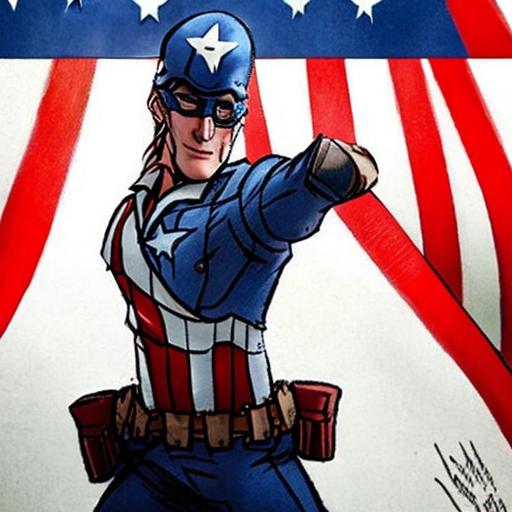} &
        \hspace{0.05cm}
        \includegraphics[width=0.08\textwidth]{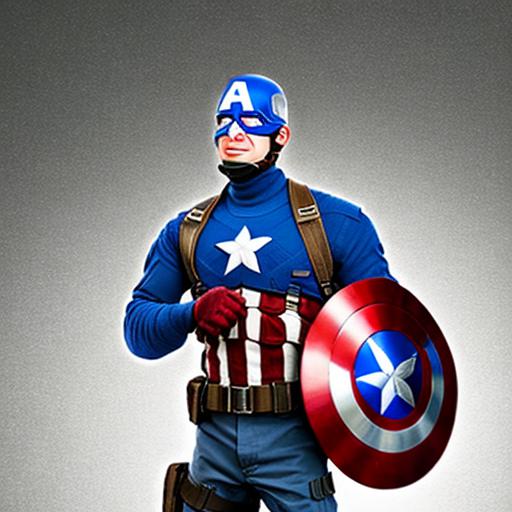} &
        \includegraphics[width=0.08\textwidth]{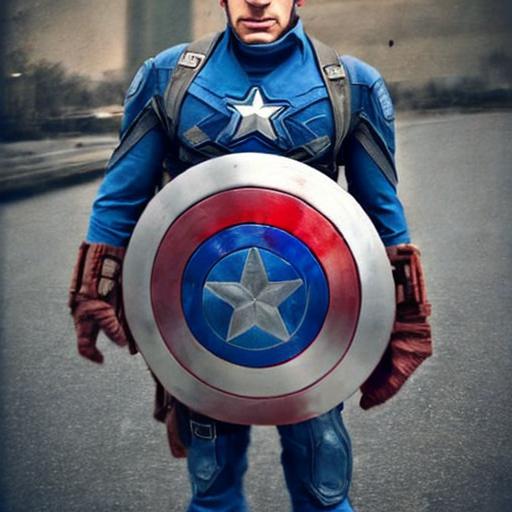} &
        \hspace{0.05cm}
        \includegraphics[width=0.08\textwidth]{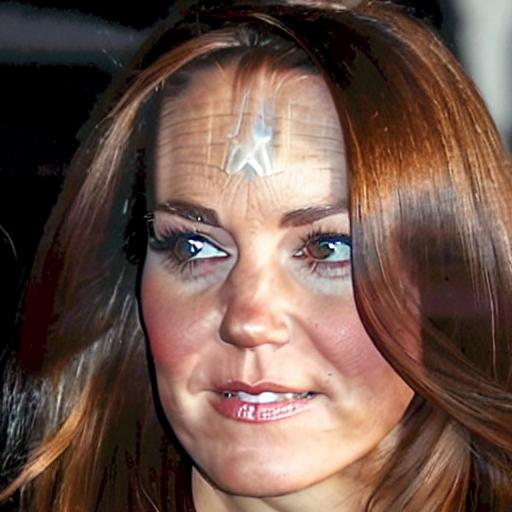} &
        \includegraphics[width=0.08\textwidth]{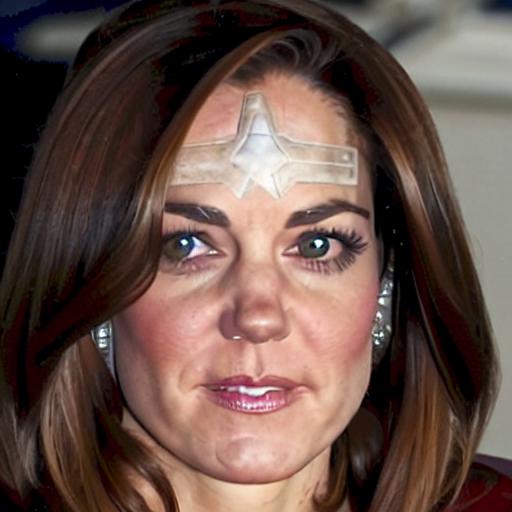} &
        \hspace{0.05cm}
        \includegraphics[width=0.08\textwidth]{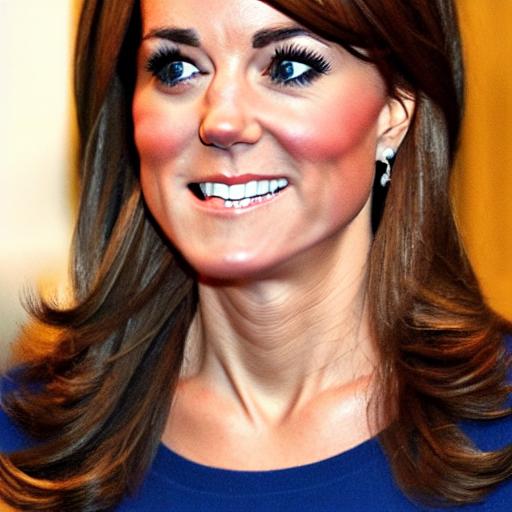} &
        \includegraphics[width=0.08\textwidth]{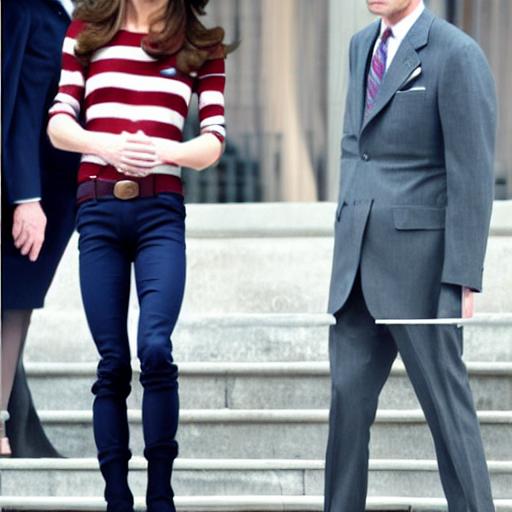} &
        \hspace{0.05cm}
        \includegraphics[width=0.08\textwidth]{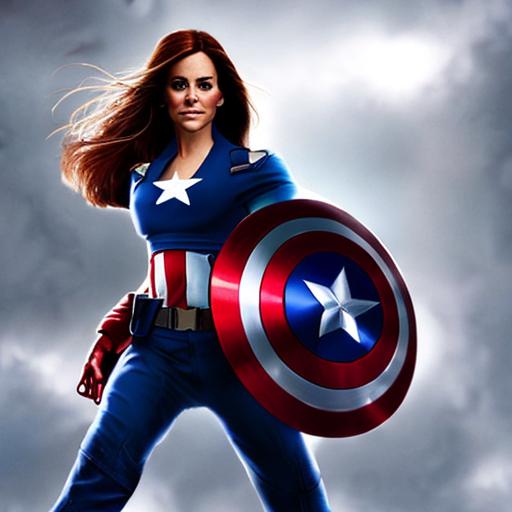} &
        \includegraphics[width=0.08\textwidth]{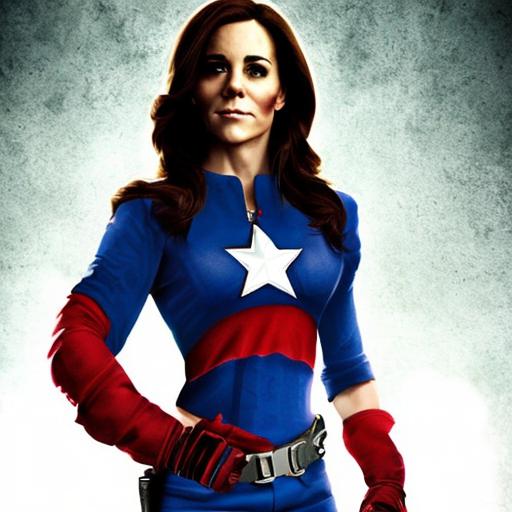} \\ \\

        \includegraphics[width=0.08\textwidth]{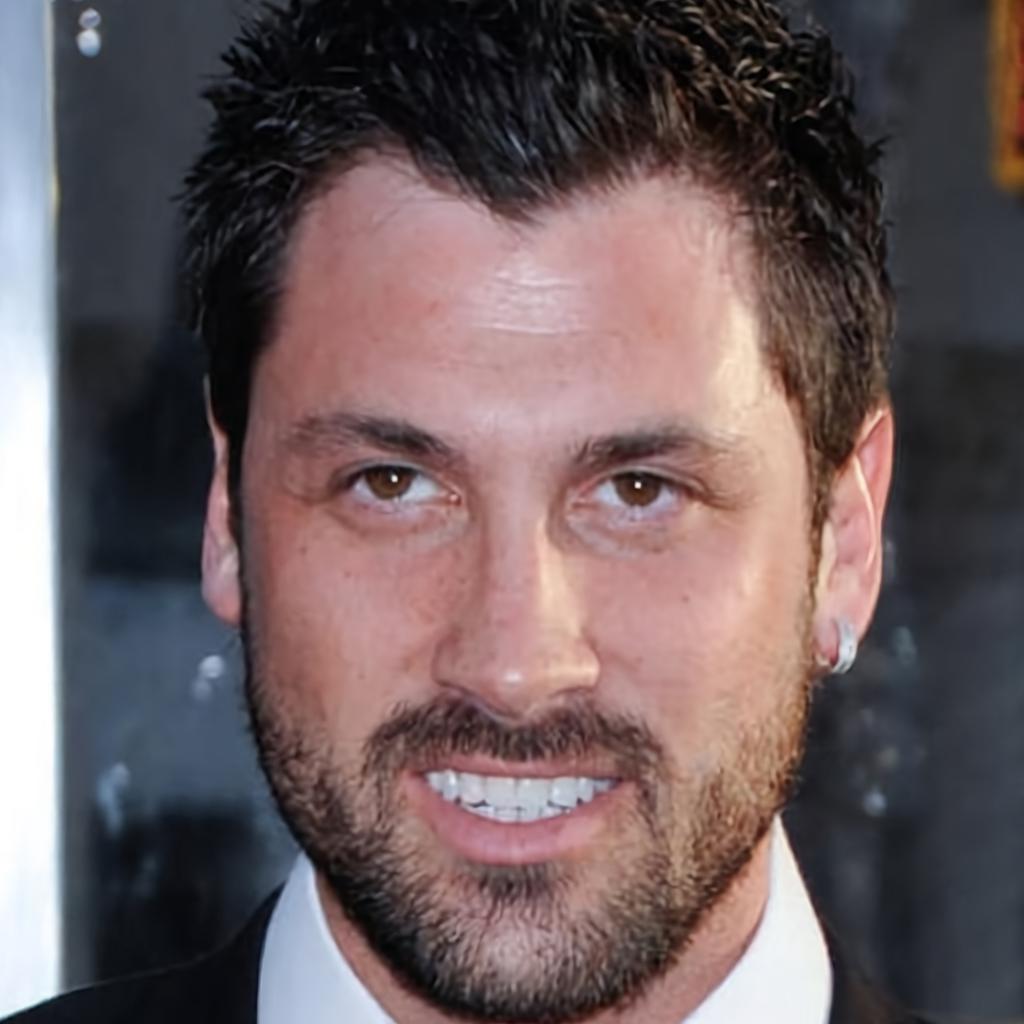} &
        \includegraphics[width=0.08\textwidth]{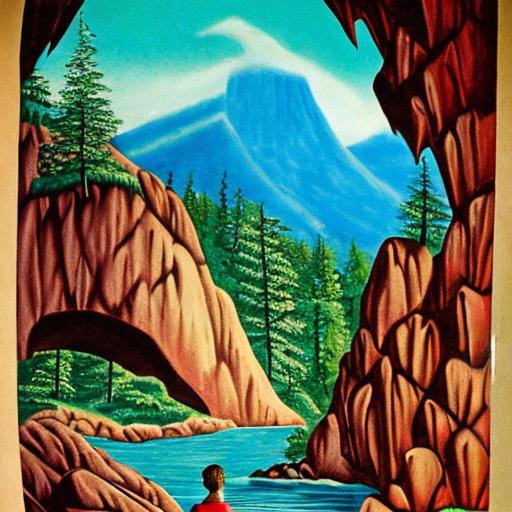} &
        \includegraphics[width=0.08\textwidth]{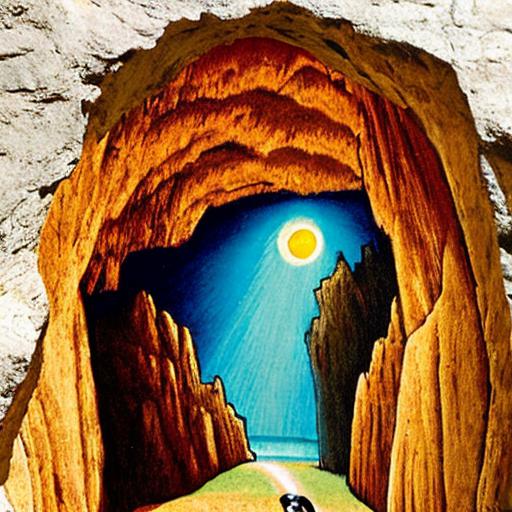} &
        \hspace{0.05cm}
        \includegraphics[width=0.08\textwidth]{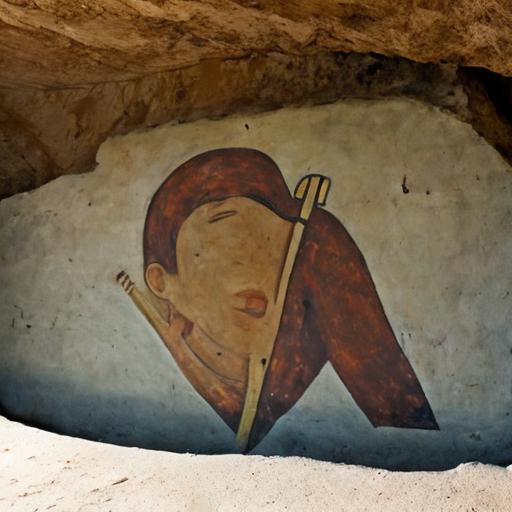} &
        \includegraphics[width=0.08\textwidth]{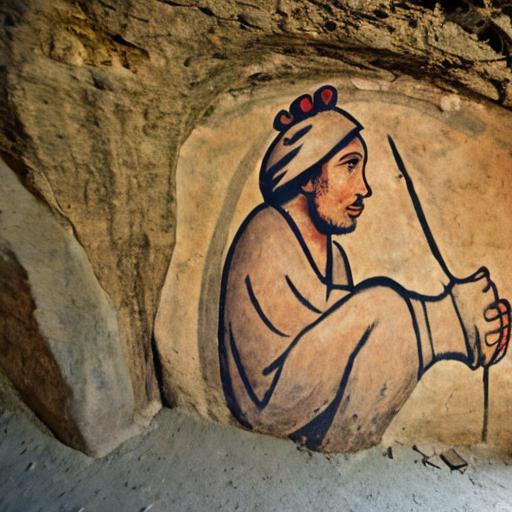} &
        \hspace{0.05cm}
        \includegraphics[width=0.08\textwidth]{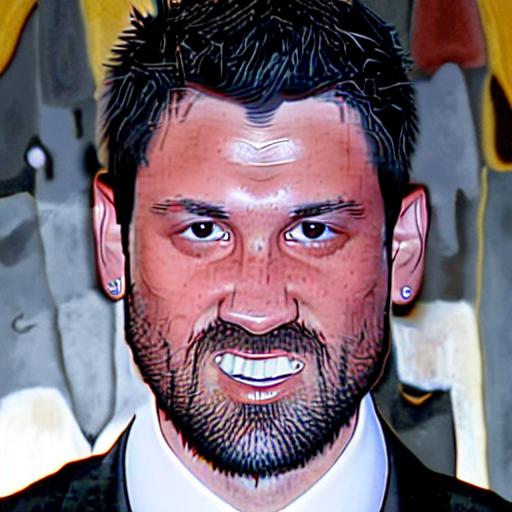} &
        \includegraphics[width=0.08\textwidth]{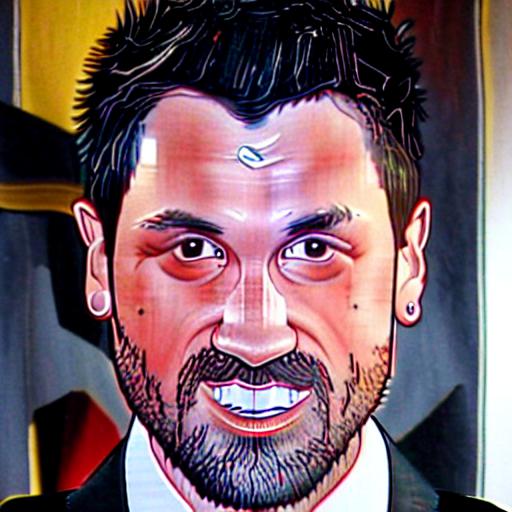} &
        \hspace{0.05cm}
        \includegraphics[width=0.08\textwidth]{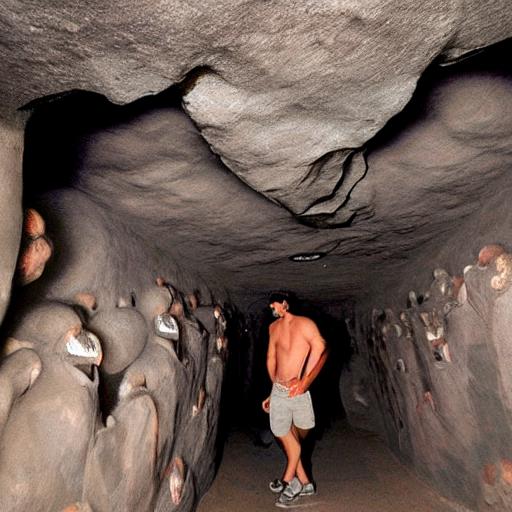} &
        \includegraphics[width=0.08\textwidth]{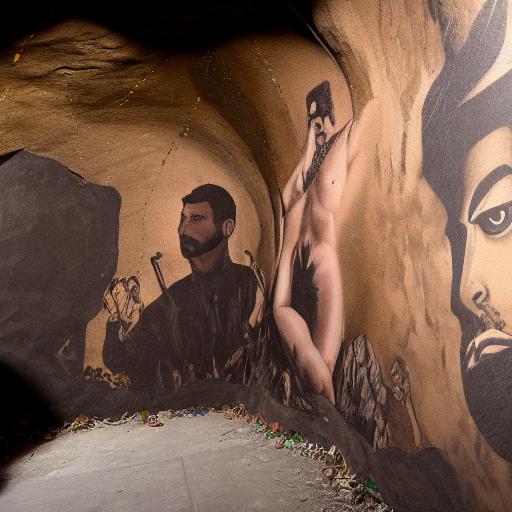} &
        \hspace{0.05cm}
        \includegraphics[width=0.08\textwidth]{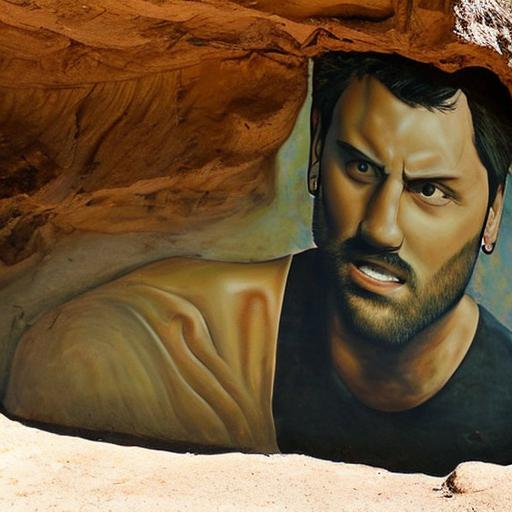} &
        \includegraphics[width=0.08\textwidth]{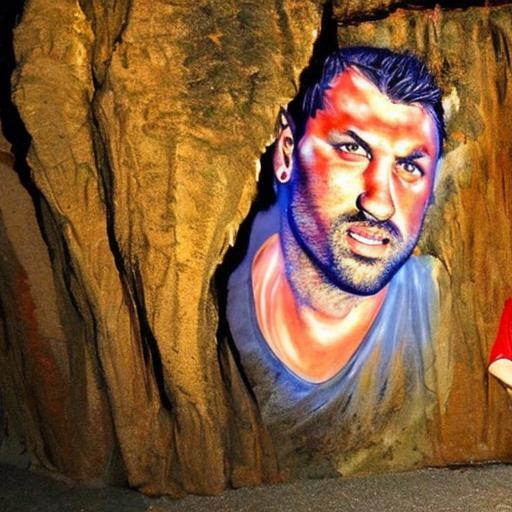} \\
        
        \raisebox{0.325in}{\begin{tabular}{c} ``cave mural \\ depicting $S_*$''\end{tabular}} &
        \includegraphics[width=0.08\textwidth]{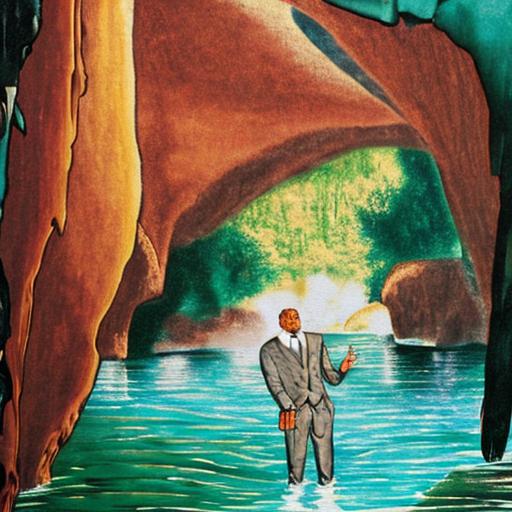} &
        \includegraphics[width=0.08\textwidth]{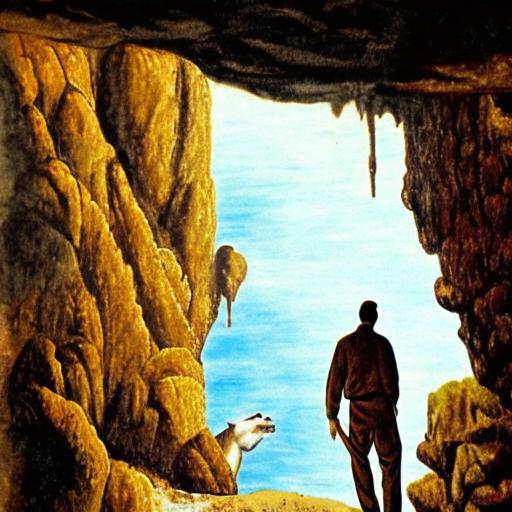} &
        \hspace{0.05cm}
        \includegraphics[width=0.08\textwidth]{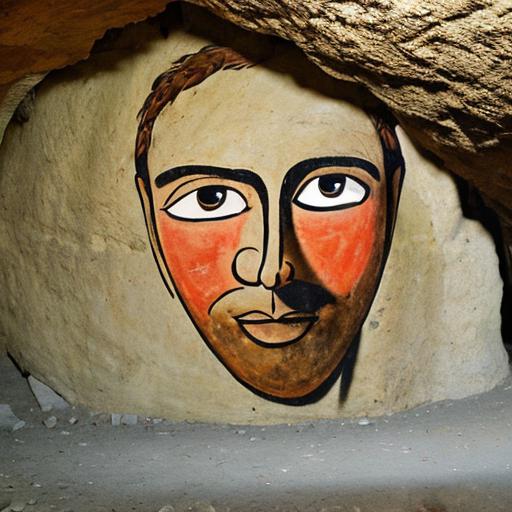} &
        \includegraphics[width=0.08\textwidth]{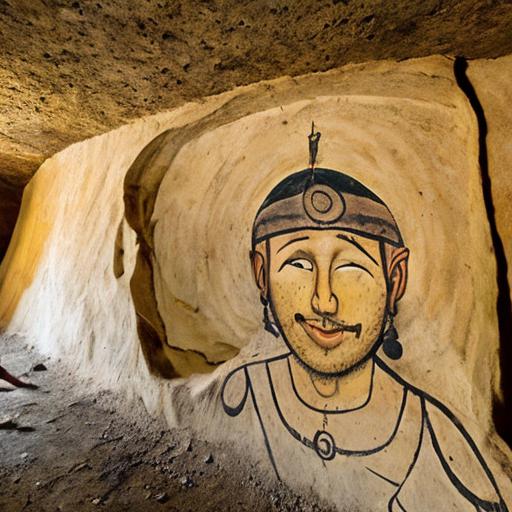} &
        \hspace{0.05cm}
        \includegraphics[width=0.08\textwidth]{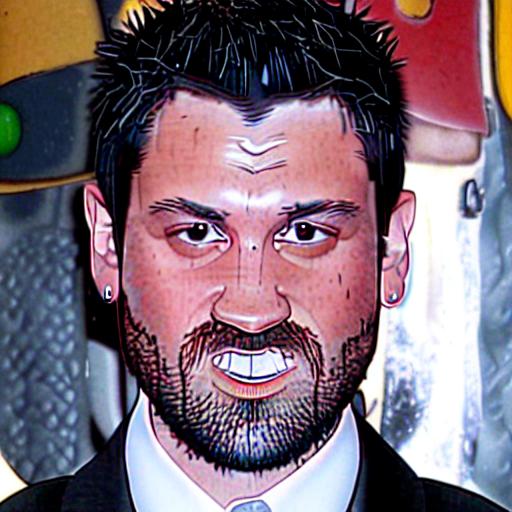} &
        \includegraphics[width=0.08\textwidth]{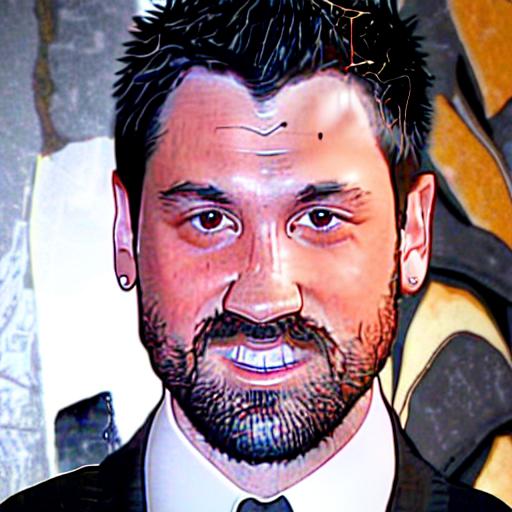} &
        \hspace{0.05cm}
        \includegraphics[width=0.08\textwidth]{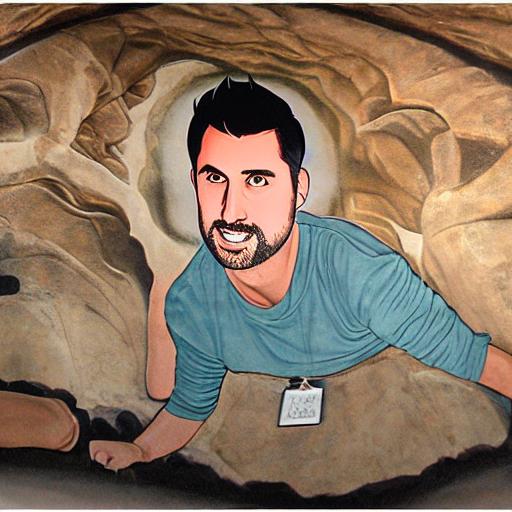} &
        \includegraphics[width=0.08\textwidth]{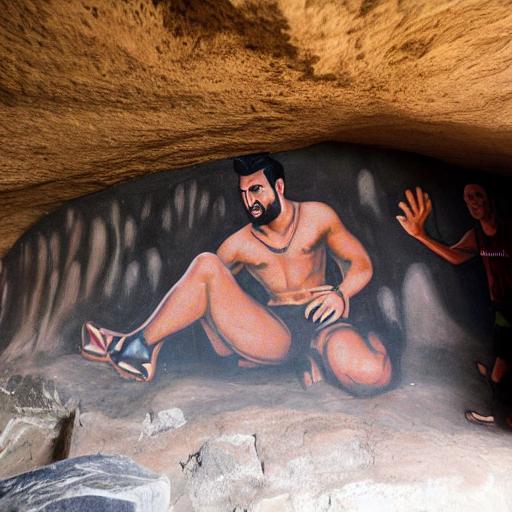} &
        \hspace{0.05cm}
        \includegraphics[width=0.08\textwidth]{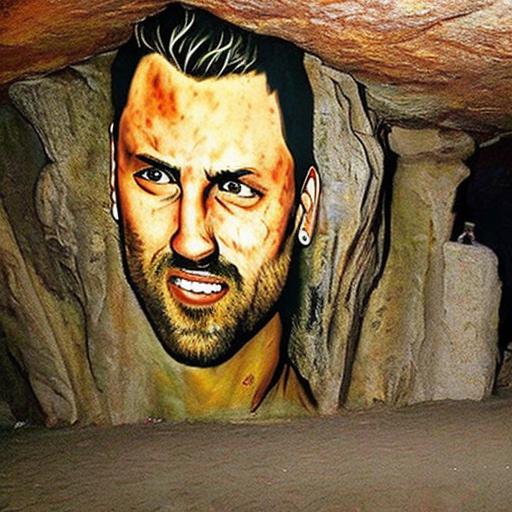} &
        \includegraphics[width=0.08\textwidth]{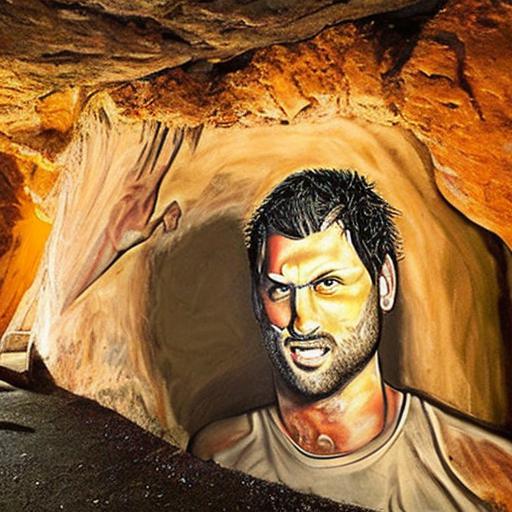} \\ \\

        \includegraphics[width=0.08\textwidth]{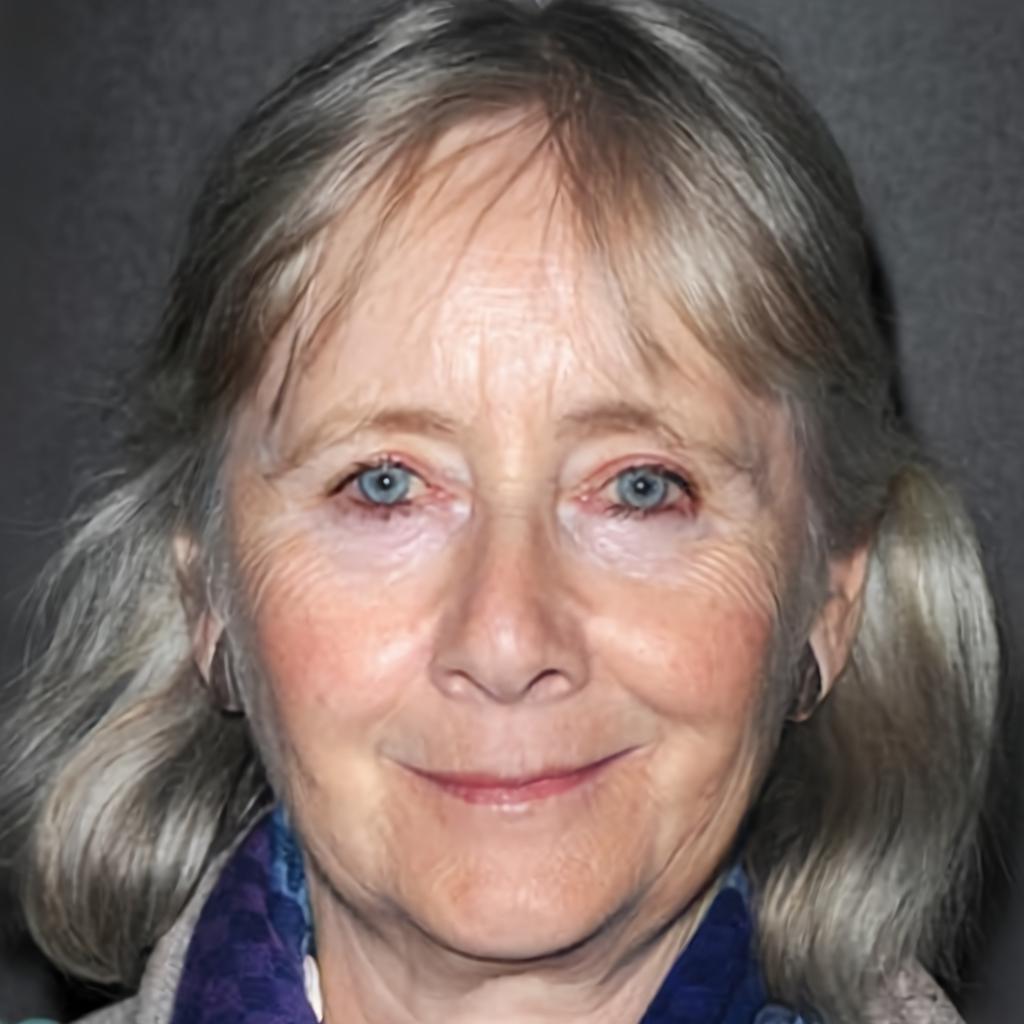} &
        \includegraphics[width=0.08\textwidth]{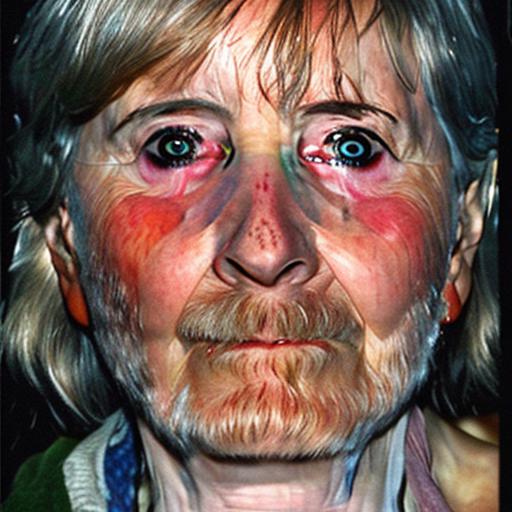} &
        \includegraphics[width=0.08\textwidth]{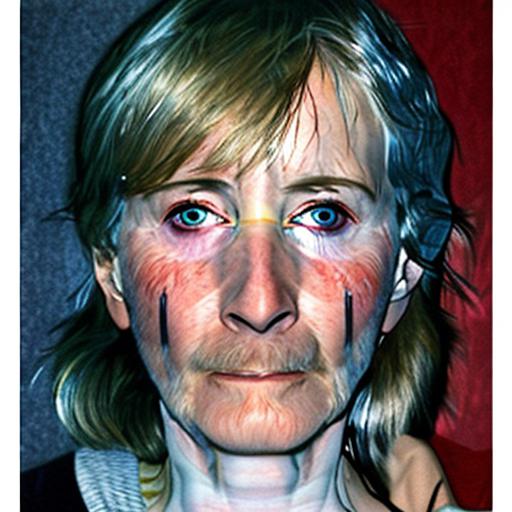} &
        \hspace{0.05cm}
        \includegraphics[width=0.08\textwidth]{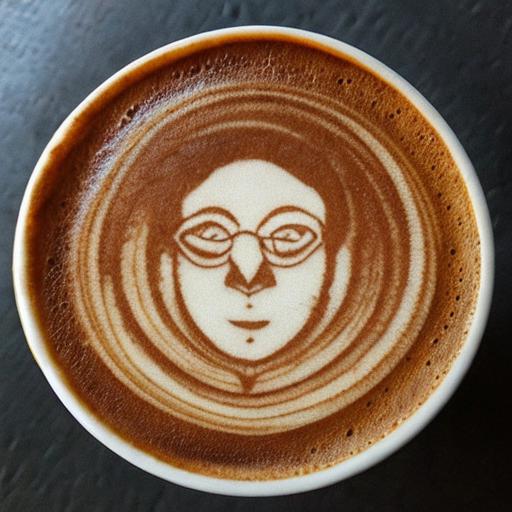} &
        \includegraphics[width=0.08\textwidth]{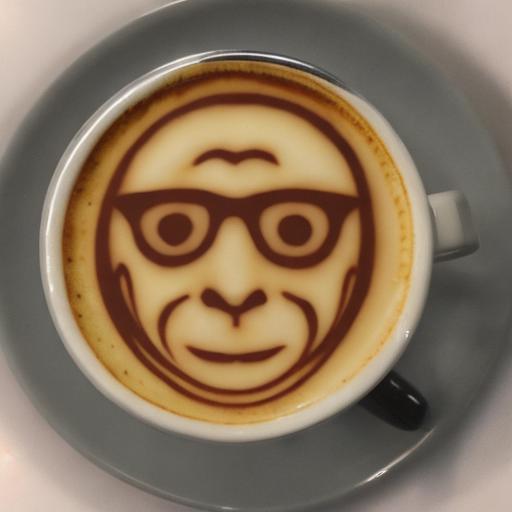} &
        \hspace{0.05cm}
        \includegraphics[width=0.08\textwidth]{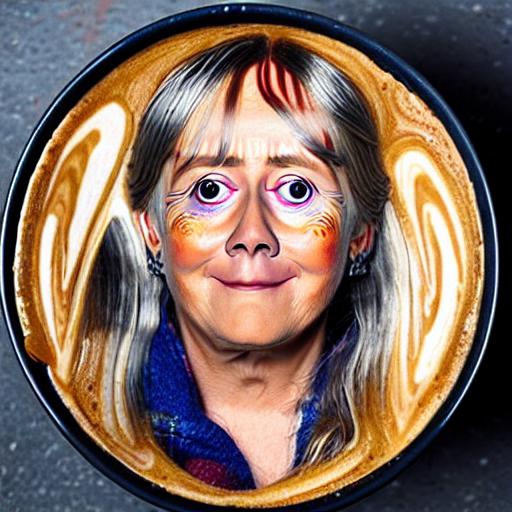} &
        \includegraphics[width=0.08\textwidth]{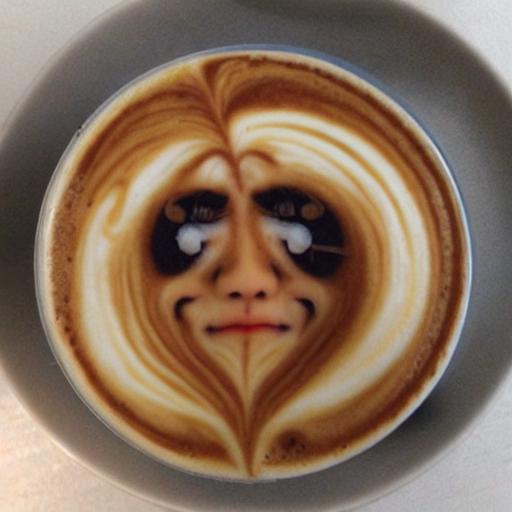} &
        \hspace{0.05cm}
        \includegraphics[width=0.08\textwidth]{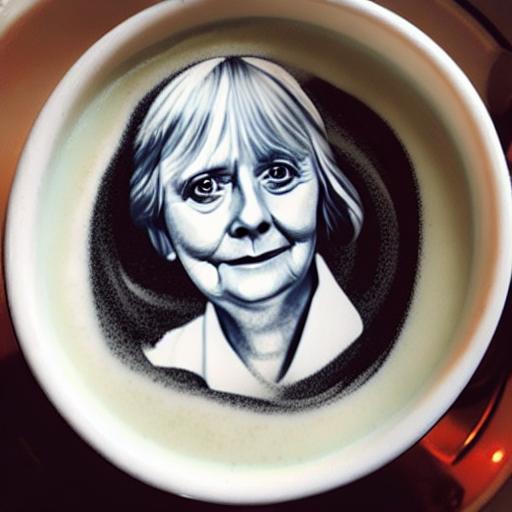} &
        \includegraphics[width=0.08\textwidth]{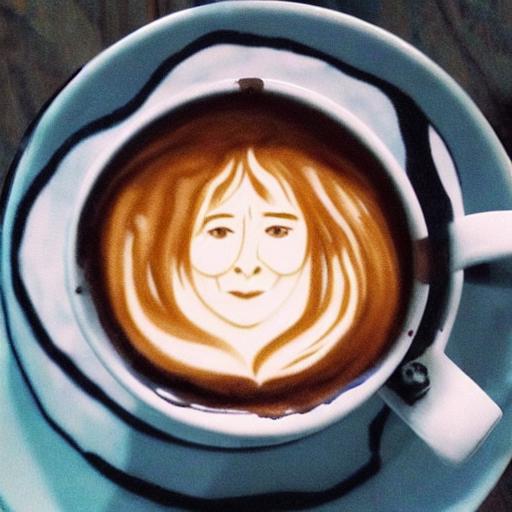} &
        \hspace{0.05cm}
        \includegraphics[width=0.08\textwidth]{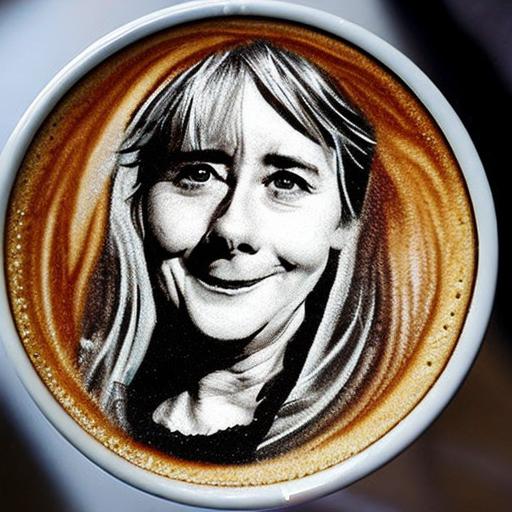} &
        \includegraphics[width=0.08\textwidth]{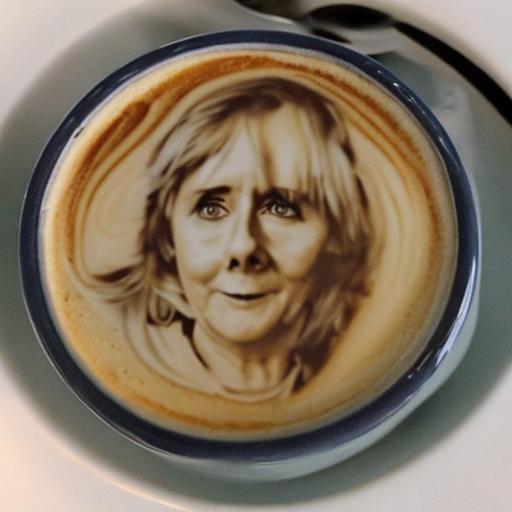} \\
        
        \raisebox{0.325in}{\begin{tabular}{c} ``$S_*$ latte art''\end{tabular}} &
        \includegraphics[width=0.08\textwidth]{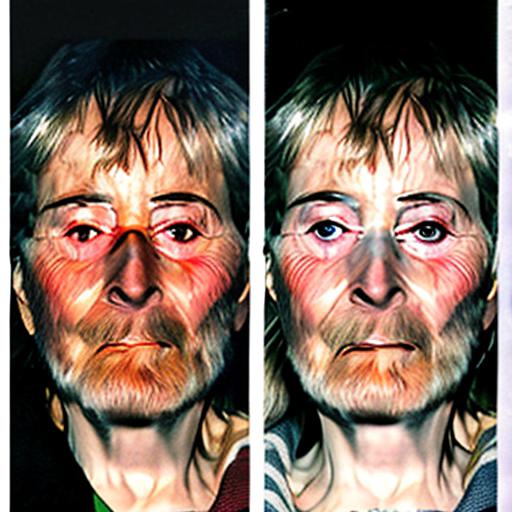} &
        \includegraphics[width=0.08\textwidth]{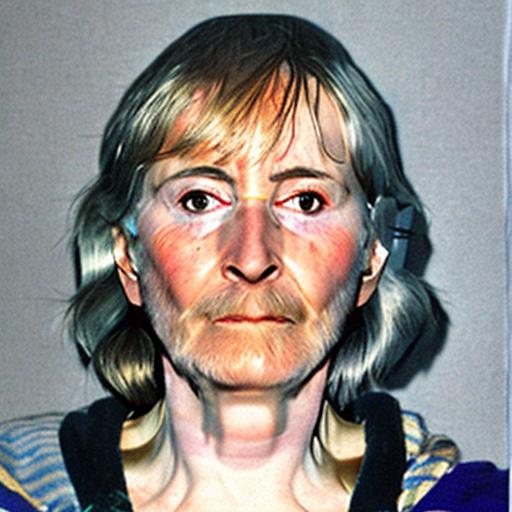} &
        \hspace{0.05cm}
        \includegraphics[width=0.08\textwidth]{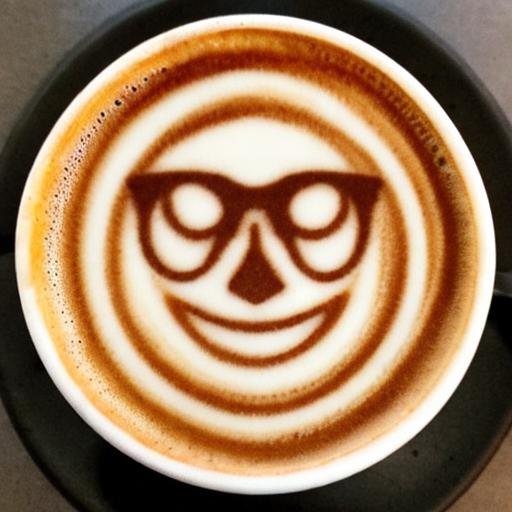} &
        \includegraphics[width=0.08\textwidth]{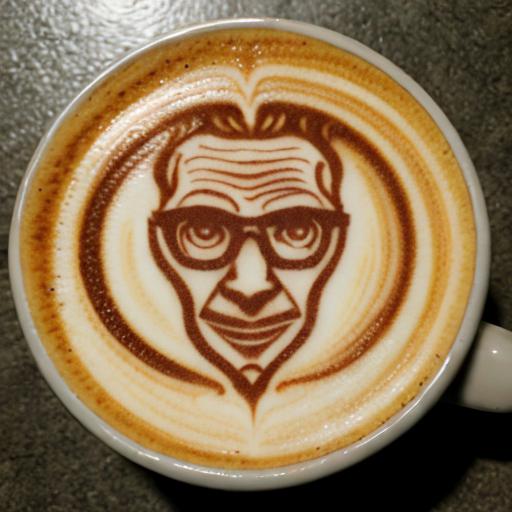} &
        \hspace{0.05cm}
        \includegraphics[width=0.08\textwidth]{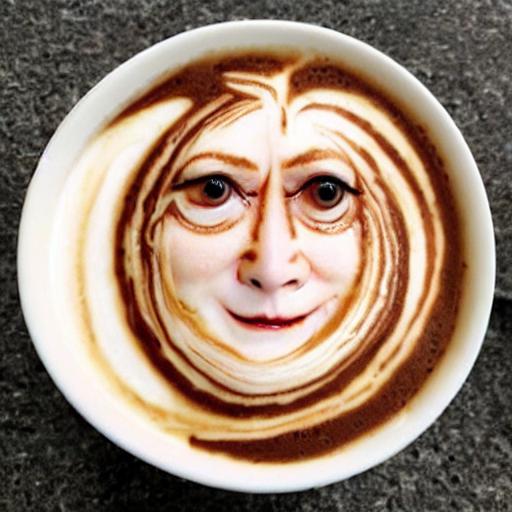} &
        \includegraphics[width=0.08\textwidth]{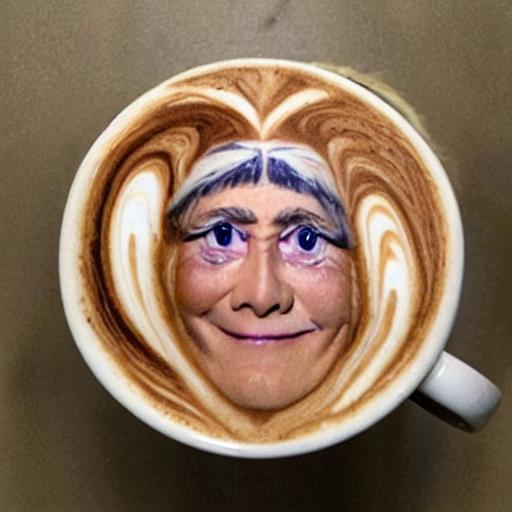} &
        \hspace{0.05cm}
        \includegraphics[width=0.08\textwidth]{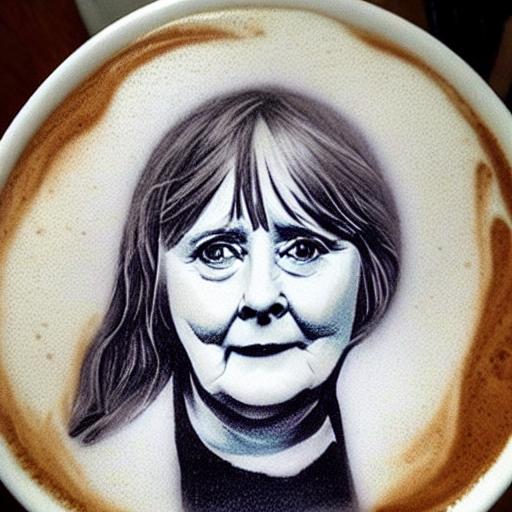} &
        \includegraphics[width=0.08\textwidth]{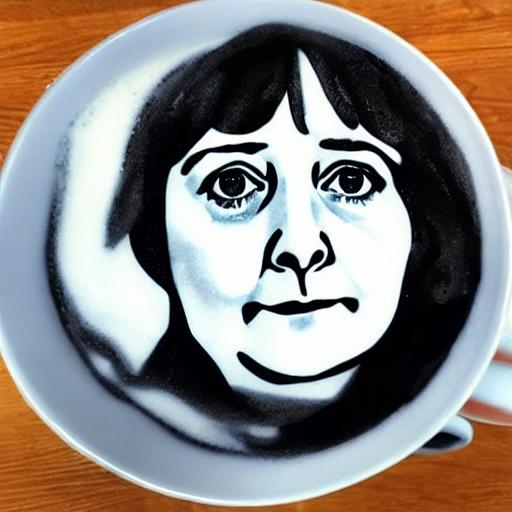} &
        \hspace{0.05cm}
        \includegraphics[width=0.08\textwidth]{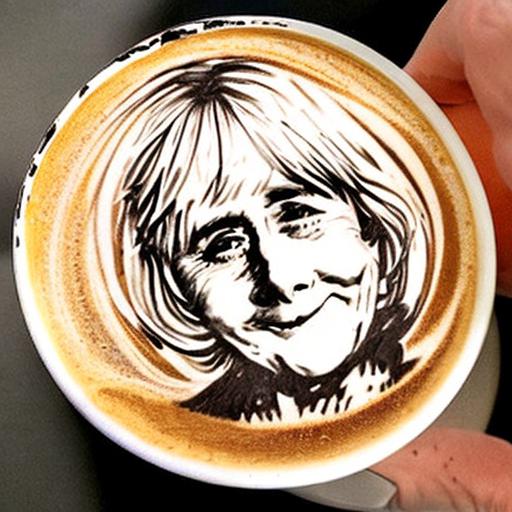} &
        \includegraphics[width=0.08\textwidth]{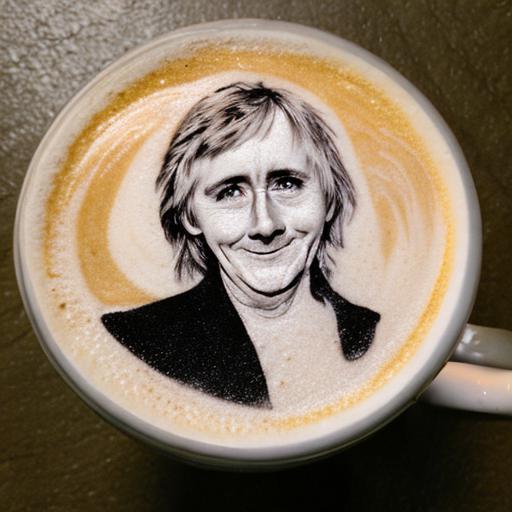} \\ \\

        \includegraphics[width=0.08\textwidth]{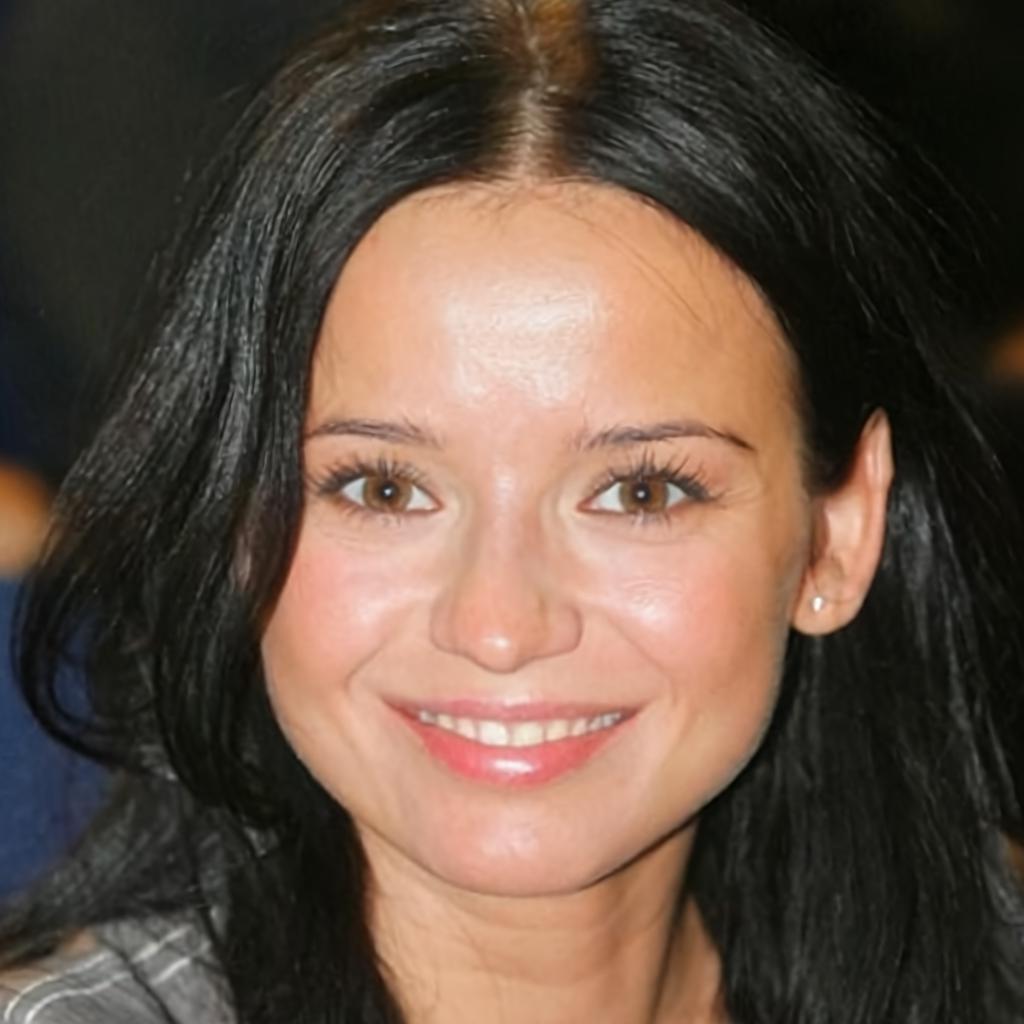} &
        \includegraphics[width=0.08\textwidth]{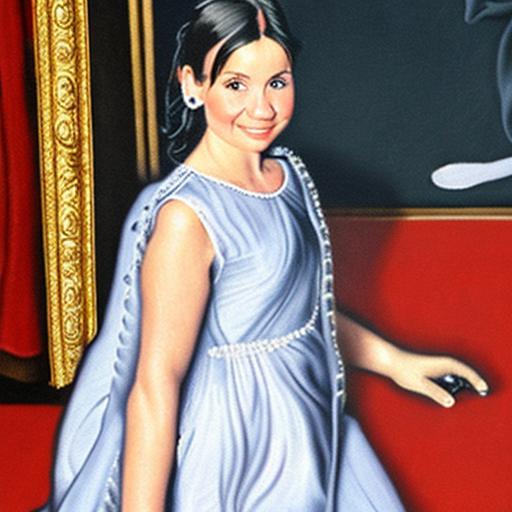} &
        \includegraphics[width=0.08\textwidth]{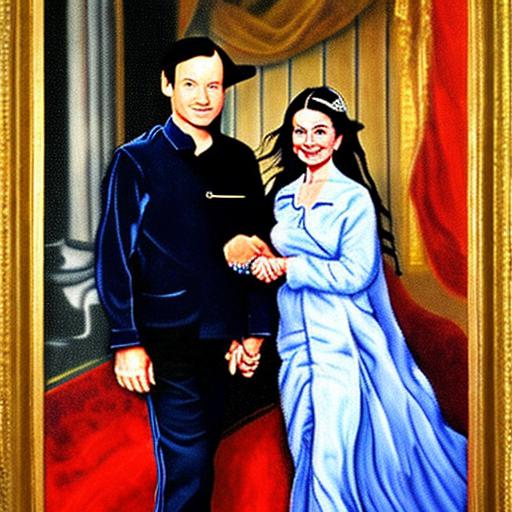} &
        \hspace{0.05cm}
        \includegraphics[width=0.08\textwidth]{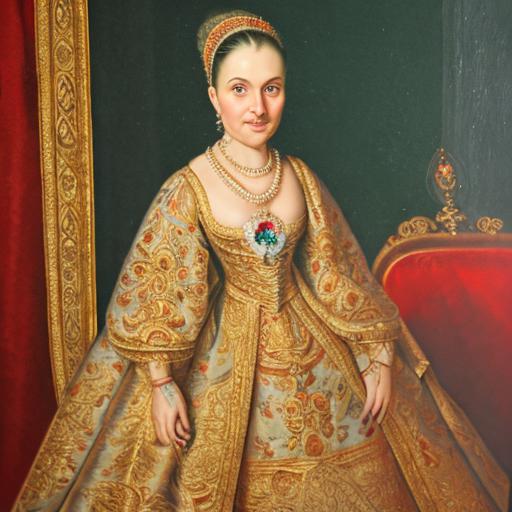} &
        \includegraphics[width=0.08\textwidth]{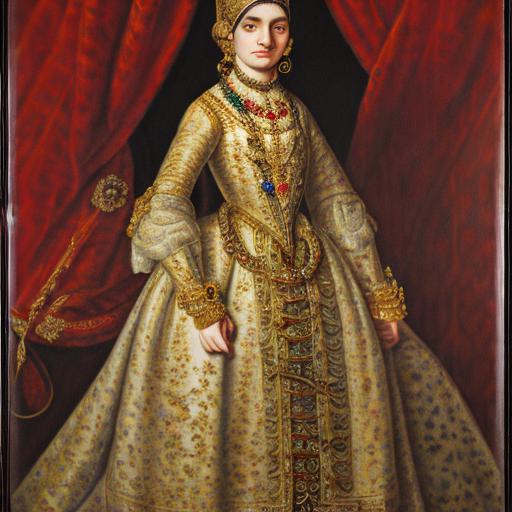} &
        \hspace{0.05cm}
        \includegraphics[width=0.08\textwidth]{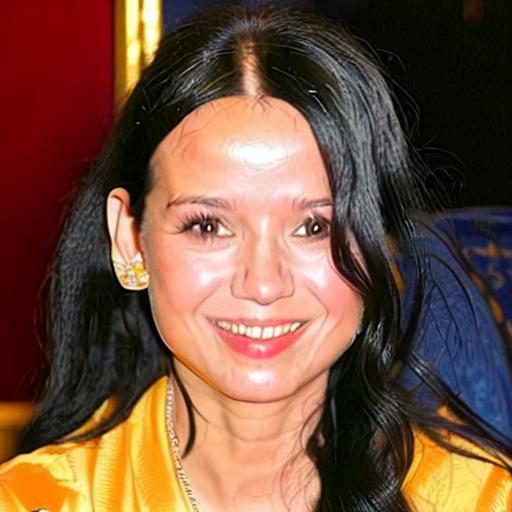} &
        \includegraphics[width=0.08\textwidth]{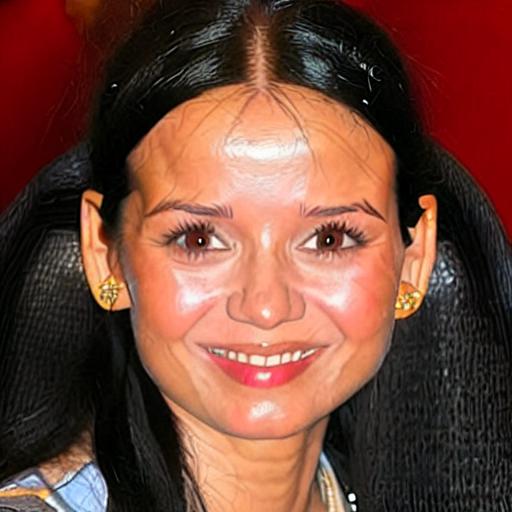} &
        \hspace{0.05cm}
        \includegraphics[width=0.08\textwidth]{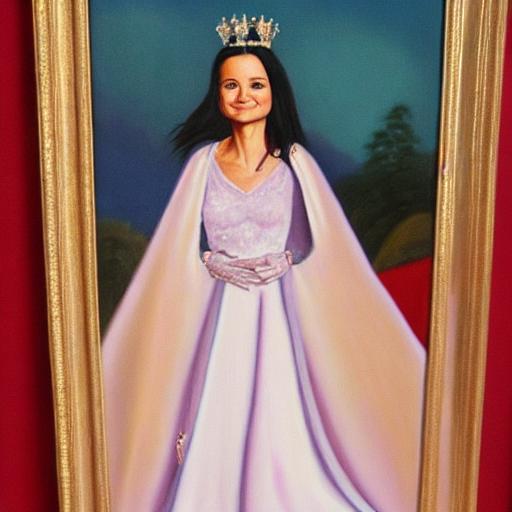} &
        \includegraphics[width=0.08\textwidth]{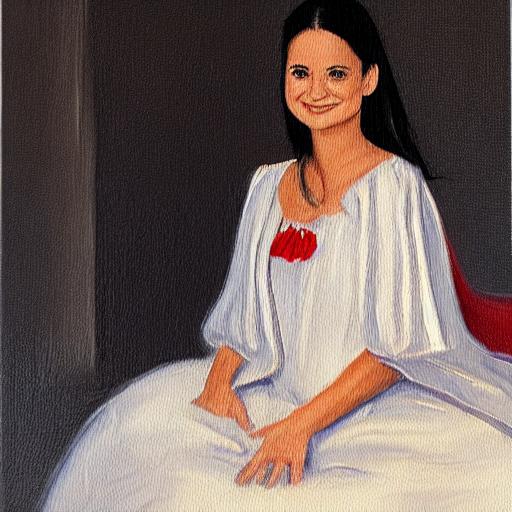} &
        \hspace{0.05cm}
        \includegraphics[width=0.08\textwidth]{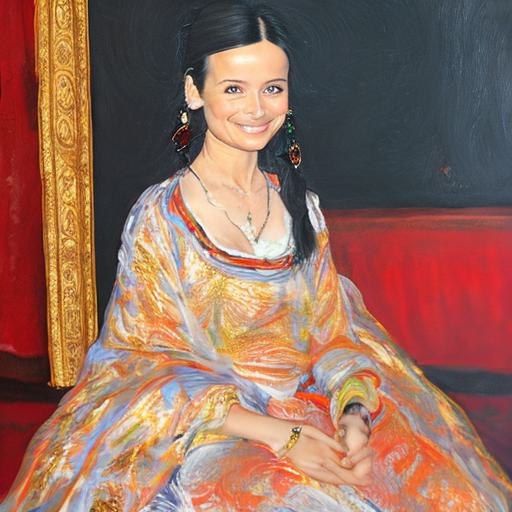} &
        \includegraphics[width=0.08\textwidth]{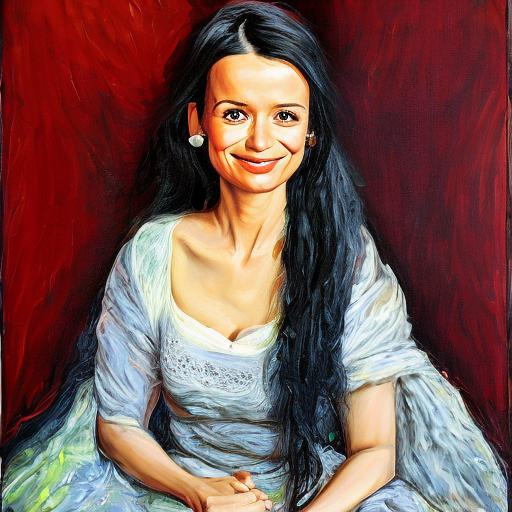} \\
        
        \raisebox{0.325in}{\begin{tabular}{c} ``A detailed oil \\ painting of  $S_*$ 
 \\ wearing a \\ royal gown \\ in a palace''\end{tabular}} &
        \includegraphics[width=0.08\textwidth]{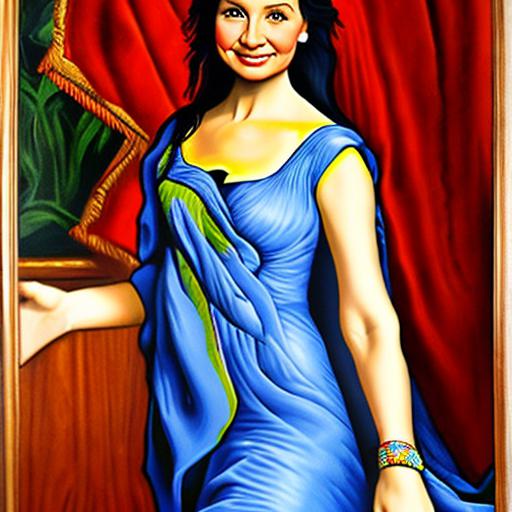} &
        \includegraphics[width=0.08\textwidth]{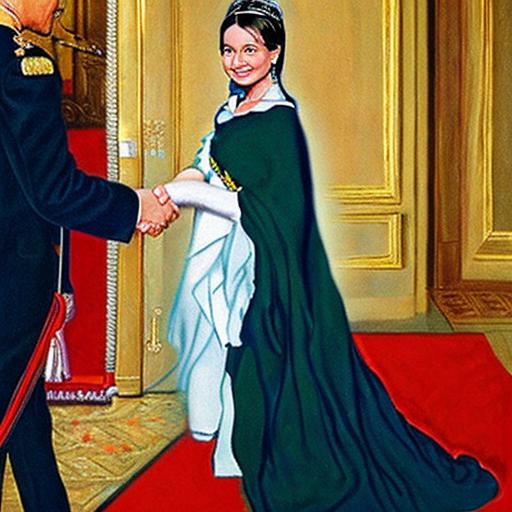} &
        \hspace{0.05cm}
        \includegraphics[width=0.08\textwidth]{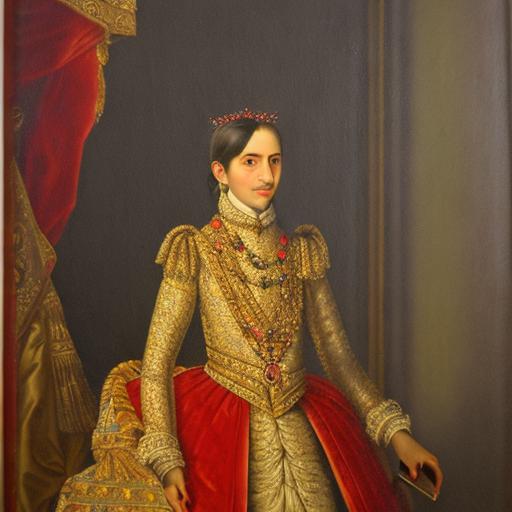} &
        \includegraphics[width=0.08\textwidth]{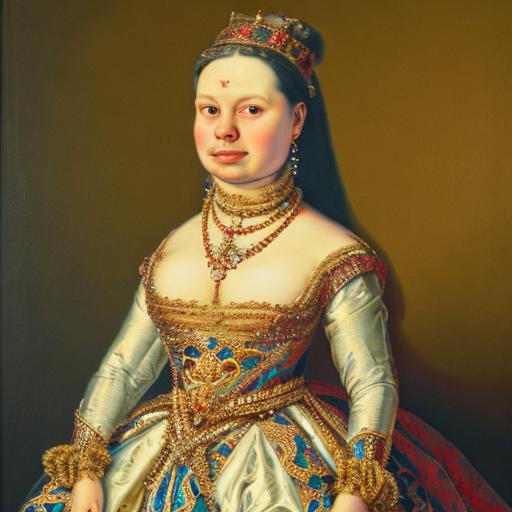} &
        \hspace{0.05cm}
        \includegraphics[width=0.08\textwidth]{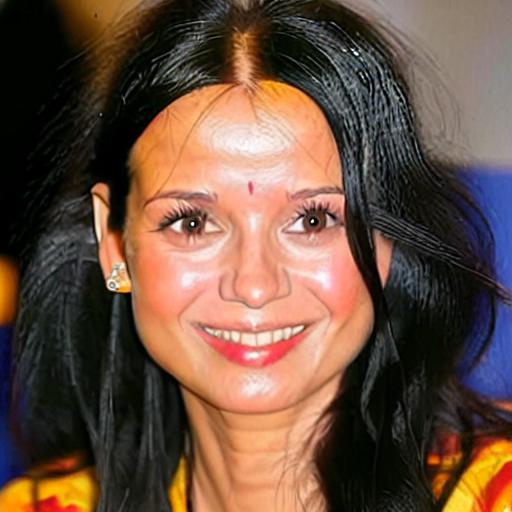} &
        \includegraphics[width=0.08\textwidth]{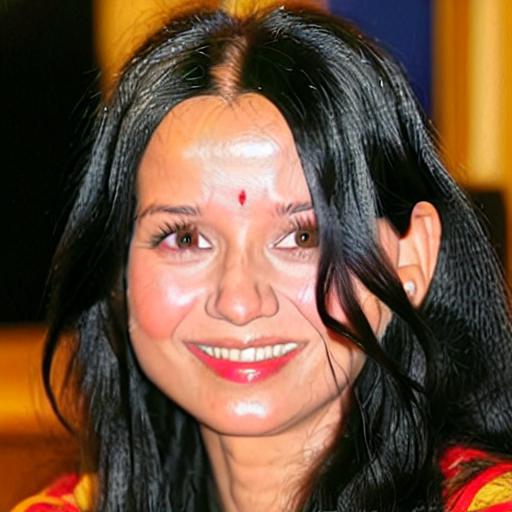} &
        \hspace{0.05cm}
        \includegraphics[width=0.08\textwidth]{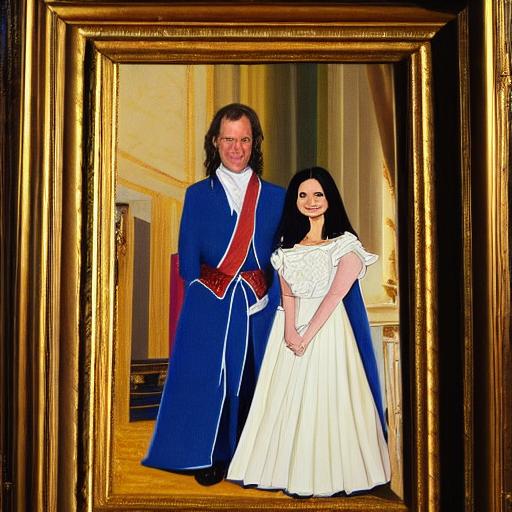} &
        \includegraphics[width=0.08\textwidth]{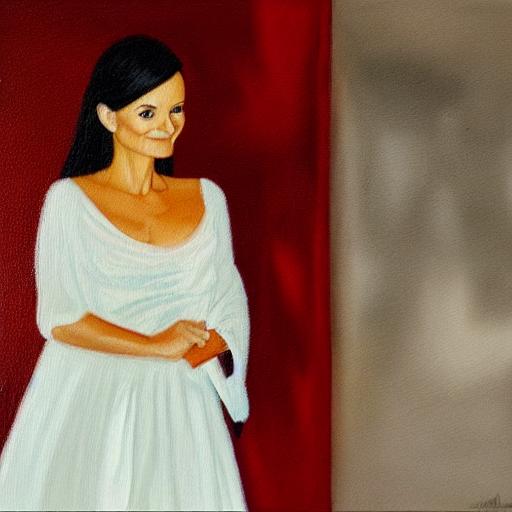} &
        \hspace{0.05cm}
        \includegraphics[width=0.08\textwidth]{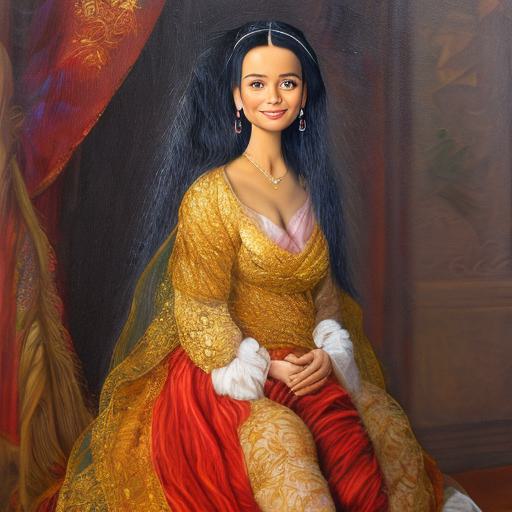} &
        \includegraphics[width=0.08\textwidth]{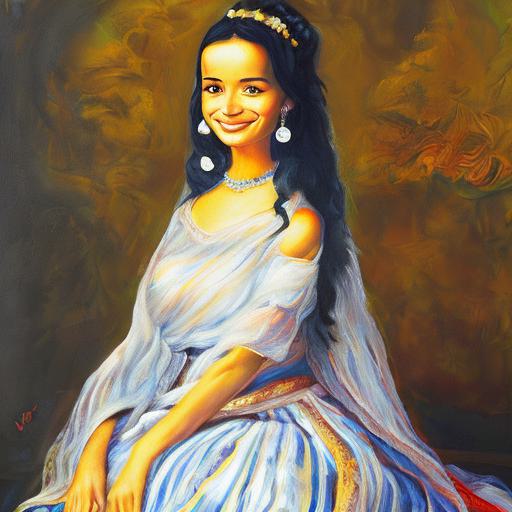} \\ \\
        
    \end{tabular}
    \\[-0.1cm]
    }
    \caption{
    Additional qualitative comparisons. Given a single input image, we present four images generated by each method using identical random seeds. Our approach demonstrates superior performance in identity preservation and editability.}
    \label{fig:appendix_qualitative_comparison}
\end{figure*}

In \cref{fig:appendix_qualitative_comparison}, we provide additional qualitative comparisons to the baseline methods on a wide range of prompts.

\section{Additional Qualitative Results}
\label{sec:additional_qualitative}
\begin{figure*}
    \centering
    \setlength{\tabcolsep}{0.1pt}
    {\footnotesize
    \begin{tabular}{c@{\hspace{0.25cm}} c@{\hspace{0.25cm}} c@{\hspace{0.25cm}} c@{\hspace{0.25cm}} c@{\hspace{0.25cm}} c@{\hspace{0.25cm}} c@{\hspace{0.25cm}}}

        \includegraphics[width=0.15\textwidth]{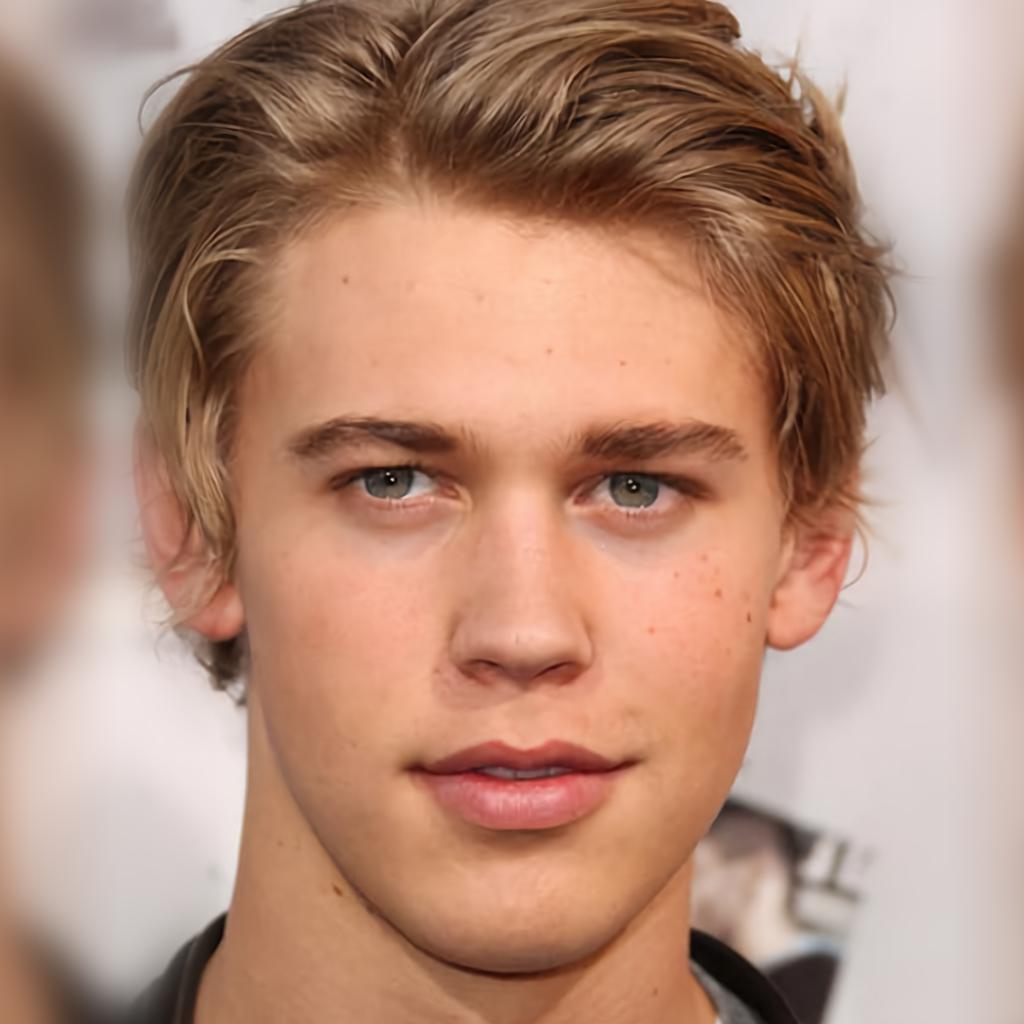} &
        \includegraphics[width=0.15\textwidth]{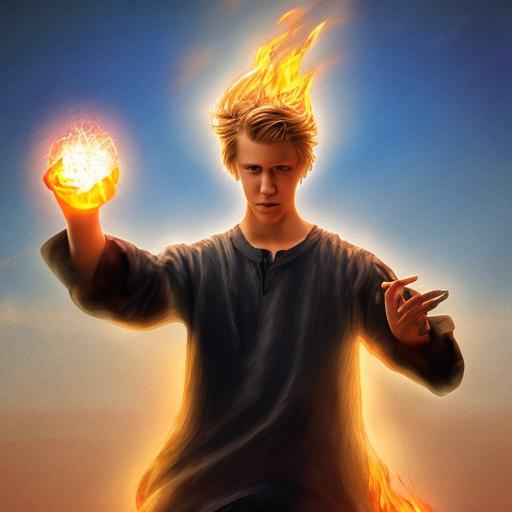} &
        \includegraphics[width=0.15\textwidth]{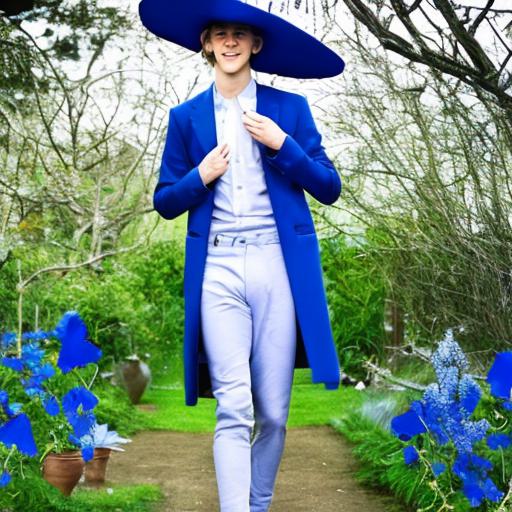} &
        \includegraphics[width=0.15\textwidth]{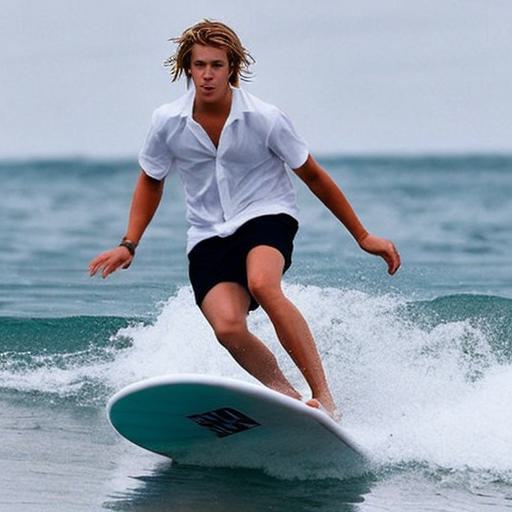} &
        \includegraphics[width=0.15\textwidth]{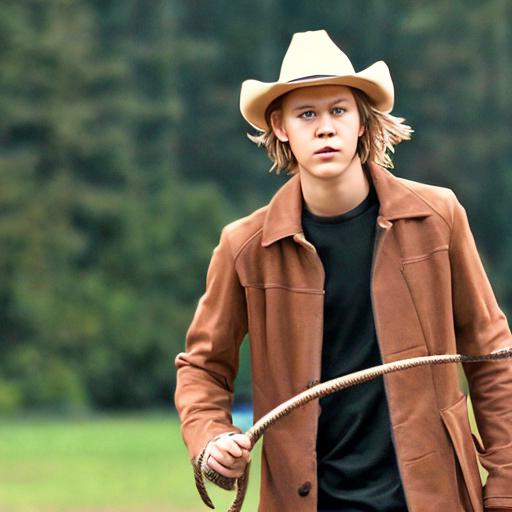} &
        \includegraphics[width=0.15\textwidth]{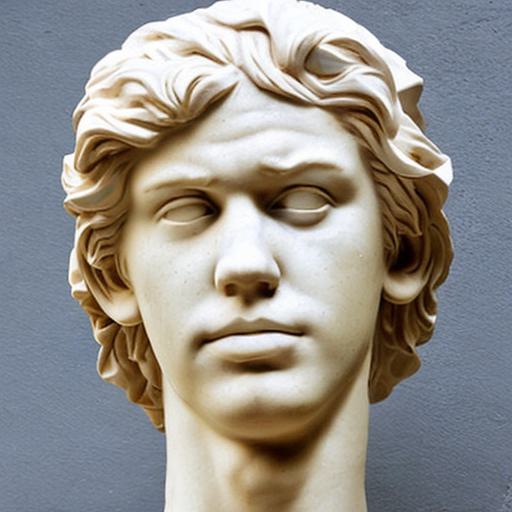} \\

        Real Sample &
        \begin{tabular}{c} ``A highly detailed \\ digital art of $S_*$ mage 
\\ casting a fire ball'' \end{tabular} &
        \begin{tabular}{c} ``$S_*$ is wearing \\ a magician hat \\ and a blue coat \\ in a garden'' \end{tabular} &
        \begin{tabular}{c} ``$S_*$ wearing a casual \\ plain white shirt 
 \\ surfing in the ocean'' \end{tabular} &
        \begin{tabular}{c} ``$S_*$ is wearing \\ a brown sports jacket \\ and a hat, holding \\ a whip in his hand'' \end{tabular} &
        \begin{tabular}{c} ``Greek sculpture \\ of $S_*$'' \end{tabular} \\ \\[-0.185cm]



        \includegraphics[width=0.15\textwidth]{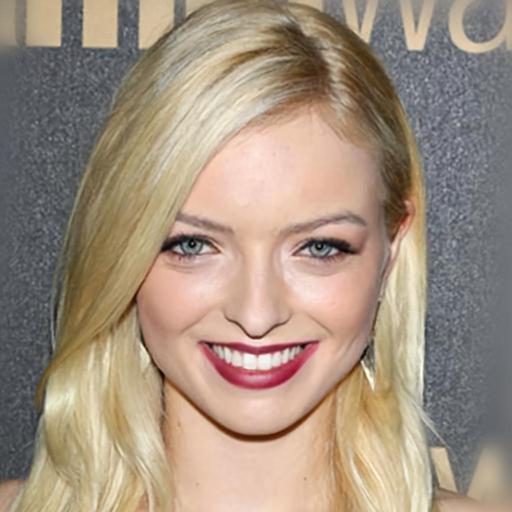} &
        \includegraphics[width=0.15\textwidth]{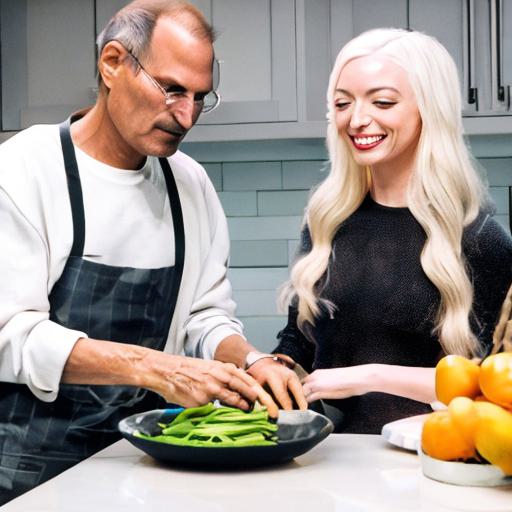} &
        \includegraphics[width=0.15\textwidth]{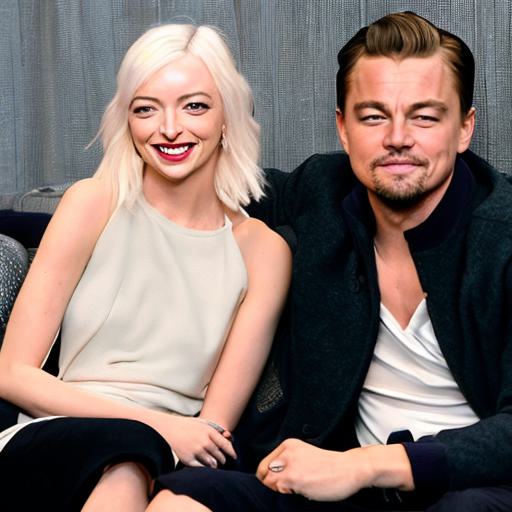} &
        \includegraphics[width=0.15\textwidth]{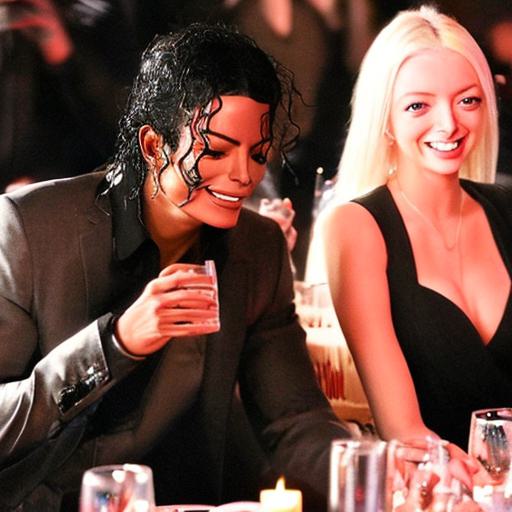} &
        \includegraphics[width=0.15\textwidth]{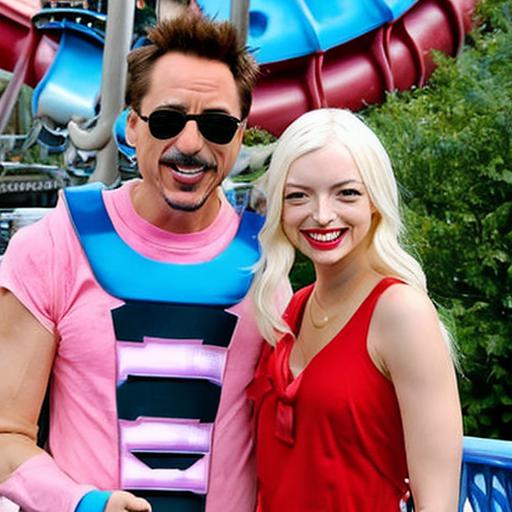} &
        \includegraphics[width=0.15\textwidth]{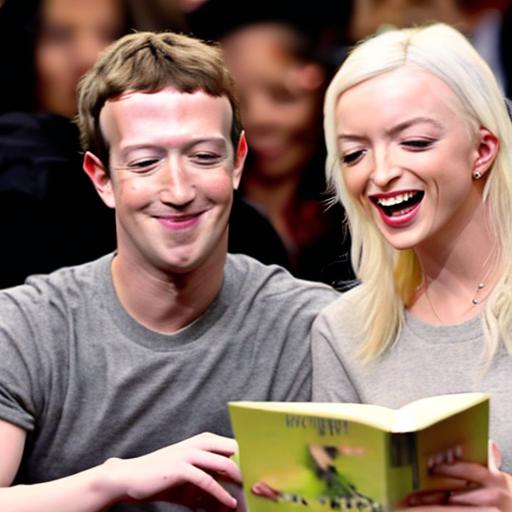} \\

        Real Sample &
        \begin{tabular}{c} ``$S_*$ and \\ Steve Jobs \\ cooking together \\ in a kitchen'' \end{tabular} &
        \begin{tabular}{c} ``$S_*$ and \\ Leonardo DiCaprio \\ sit on a sofa'' \end{tabular} &
        \begin{tabular}{c} ``$S_*$ and \\ Michael Jackson \\ enjoy a delicate \\ candlelight dinner'' \end{tabular} &
        \begin{tabular}{c} ``$S_*$ and \\ Robert Downey \\ enjoying a day \\ at an \\ amusement park'' \end{tabular} &
        \begin{tabular}{c} ``$S_*$ and \\ Mark Zuckerberg \\ are reading a \\ book together'' \end{tabular} \\ \\[-0.185cm]

        \includegraphics[width=0.15\textwidth]{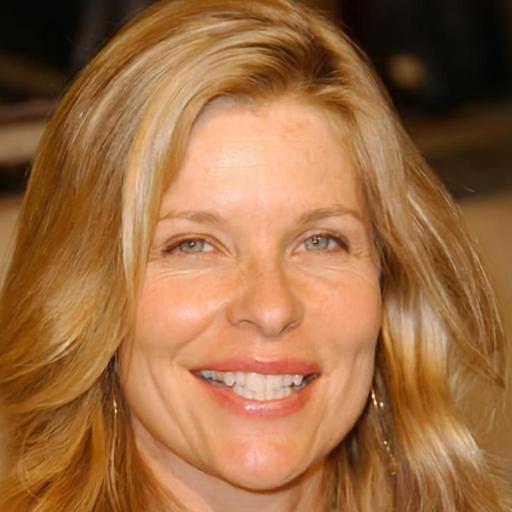} &
        \includegraphics[width=0.15\textwidth]{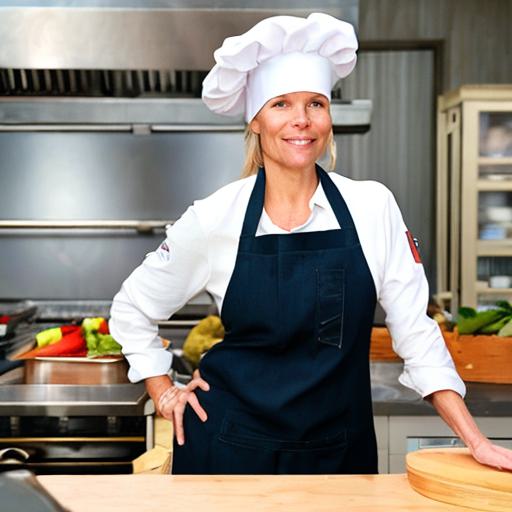} &
        \includegraphics[width=0.15\textwidth]{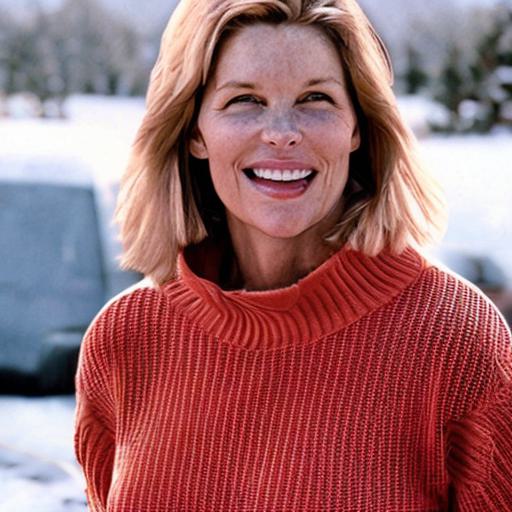} &
        \includegraphics[width=0.15\textwidth]{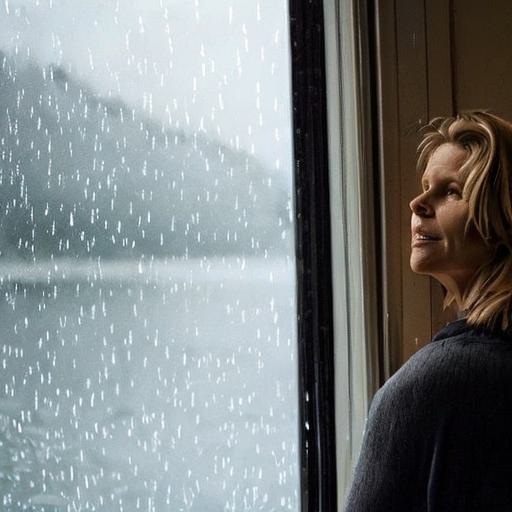} &
        \includegraphics[width=0.15\textwidth]{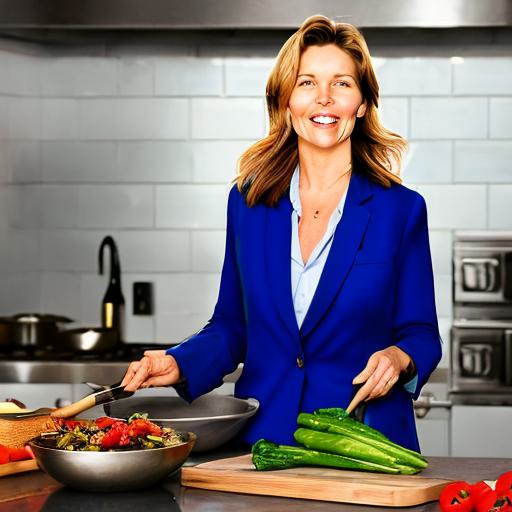} &
        \includegraphics[width=0.15\textwidth]{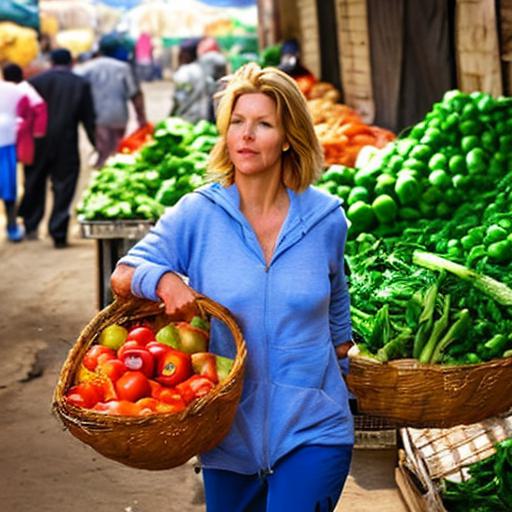} \\

        Real Sample &
        \begin{tabular}{c} ``$S_*$ wears \\ a chefs hat \\ in the kitchen'' \end{tabular} &
        \begin{tabular}{c} ``$S_*$ is wearing \\ the sweater \\ outdoors '' \end{tabular} &
        \begin{tabular}{c} ``$S_*$ is looking \\ out of a window \\ on a rainy night'' \end{tabular} &
        \begin{tabular}{c} ``$S_*$ dressed \\ in a blue suit \\ is cooking \\ a gourmet meal '' \end{tabular} &
        \begin{tabular}{c} ``$S_*$ is carrying \\ vegetables in \\ vegetable market '' \end{tabular} \\ \\[-0.185cm]

        \includegraphics[width=0.15\textwidth]{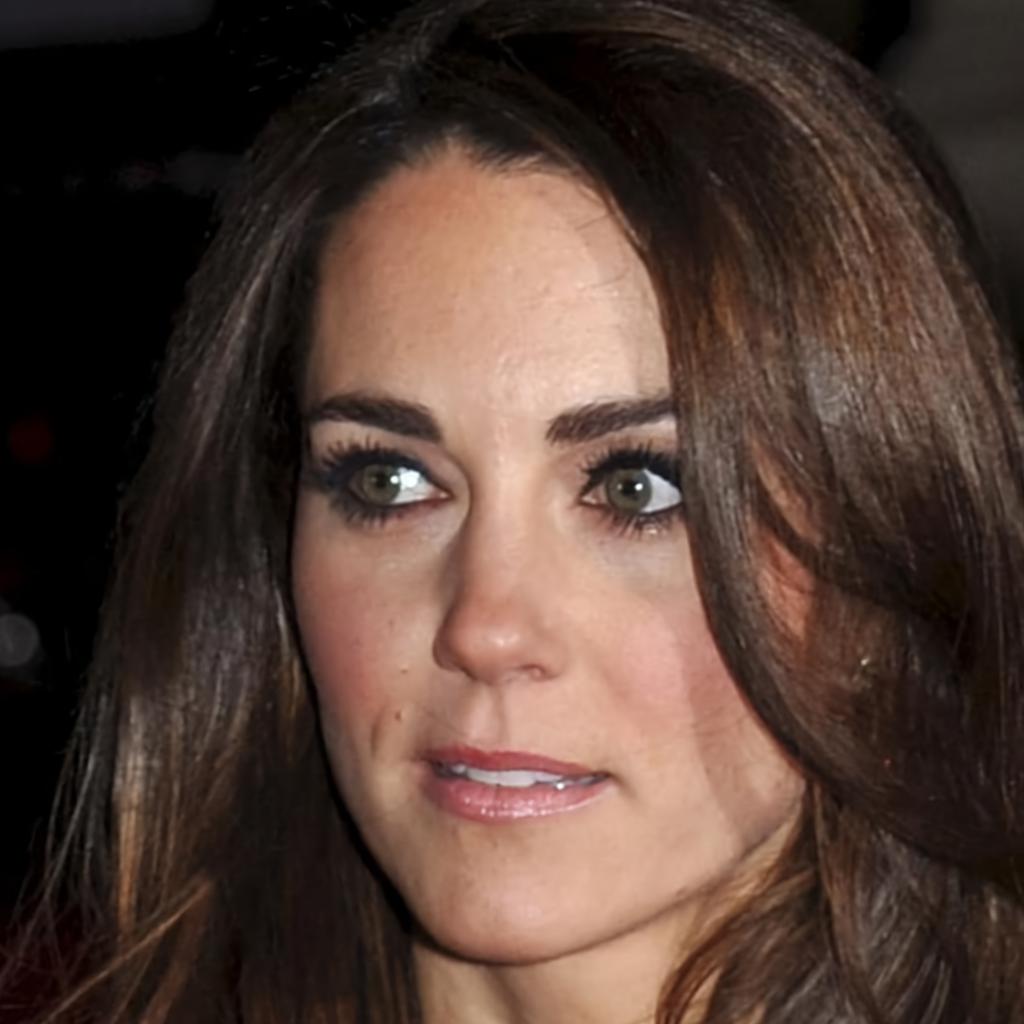} &
        \includegraphics[width=0.15\textwidth]{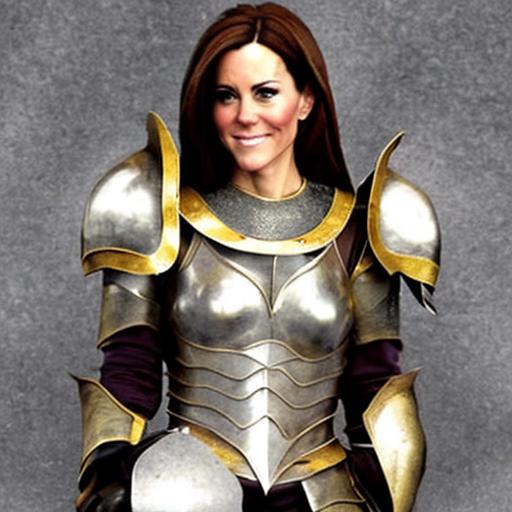} &
        \includegraphics[width=0.15\textwidth]{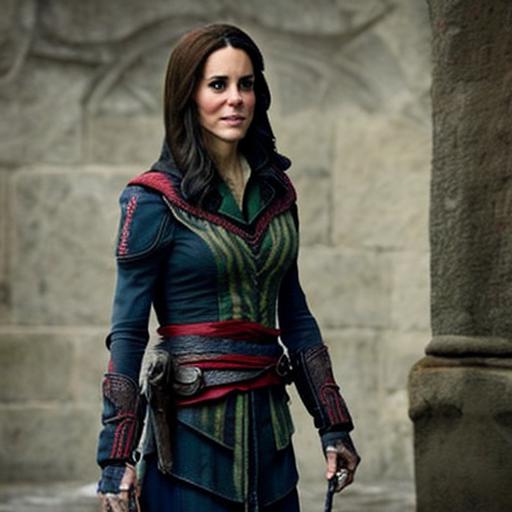} &
        \includegraphics[width=0.15\textwidth]{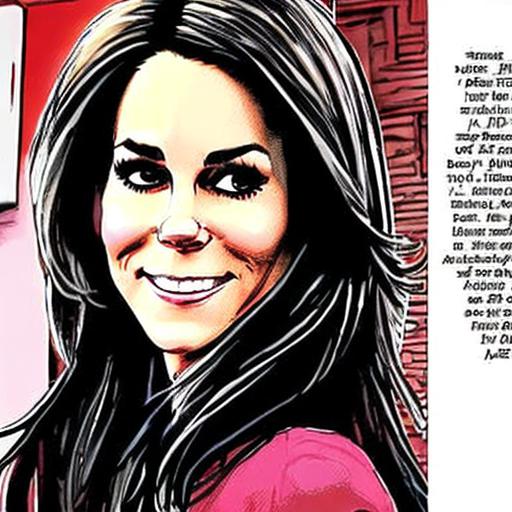} &
        \includegraphics[width=0.15\textwidth]{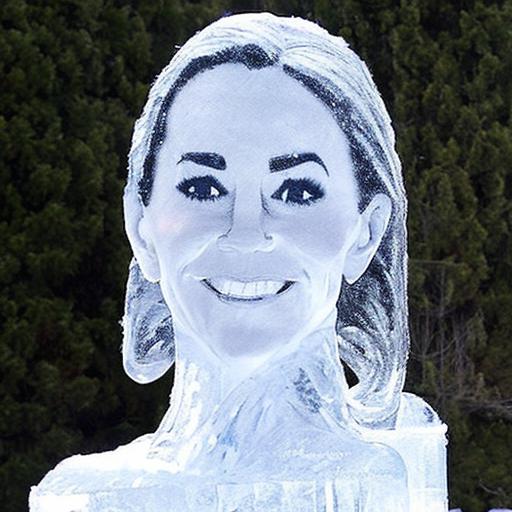} &
        \includegraphics[width=0.15\textwidth]{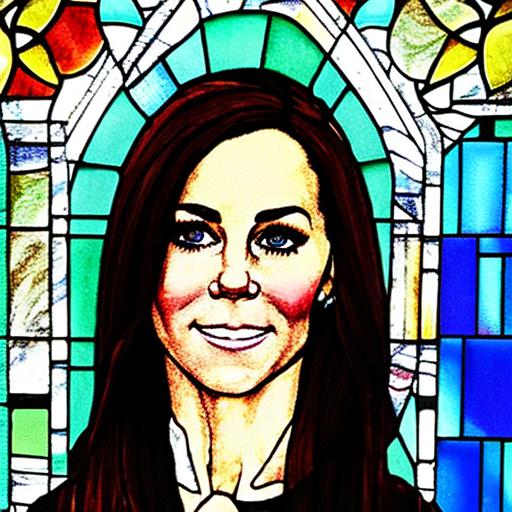} \\

        Real Sample &
        \begin{tabular}{c} ``$S_*$ as a \\ knight in \\ plate armor'' \end{tabular} &
        \begin{tabular}{c} ``$S_*$ in \\ assassins creed'' \end{tabular} &
        \begin{tabular}{c} ``$S_*$ in a \\ comic book'' \end{tabular} &
        \begin{tabular}{c} ``Ice sculpture \\ of $S_*$'' \end{tabular} &
        \begin{tabular}{c} ``$S_*$ stained \\ glass window'' \end{tabular} \\ \\[-0.185cm]

        \includegraphics[width=0.15\textwidth]{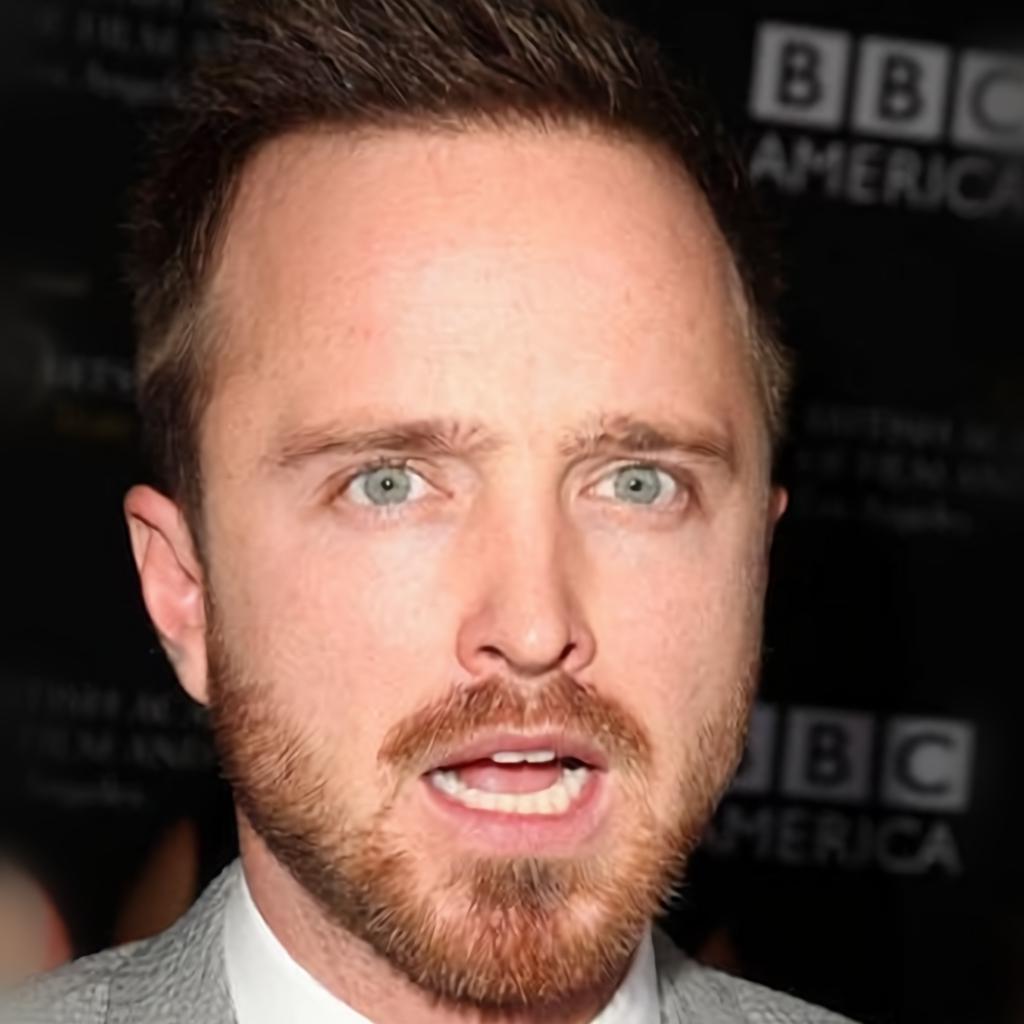} &
        \includegraphics[width=0.15\textwidth]{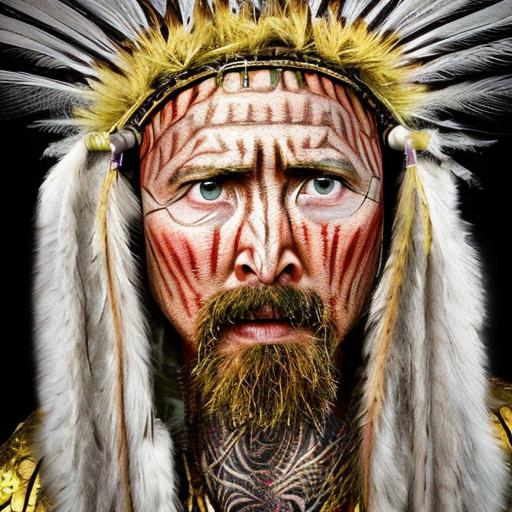} &
        \includegraphics[width=0.15\textwidth]{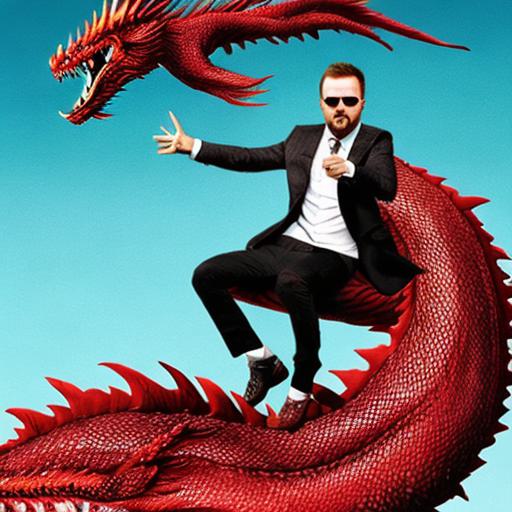} &
        \includegraphics[width=0.15\textwidth]{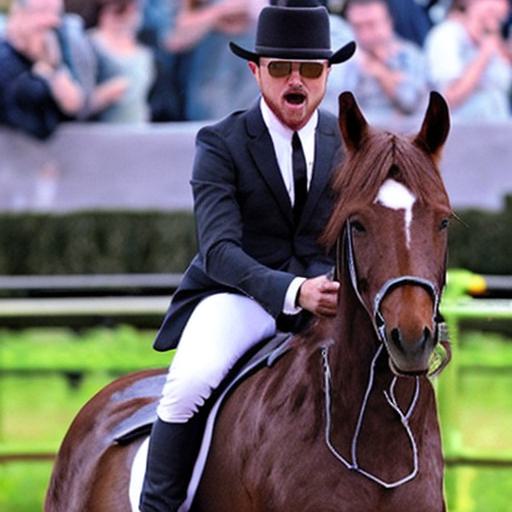} &
        \includegraphics[width=0.15\textwidth]{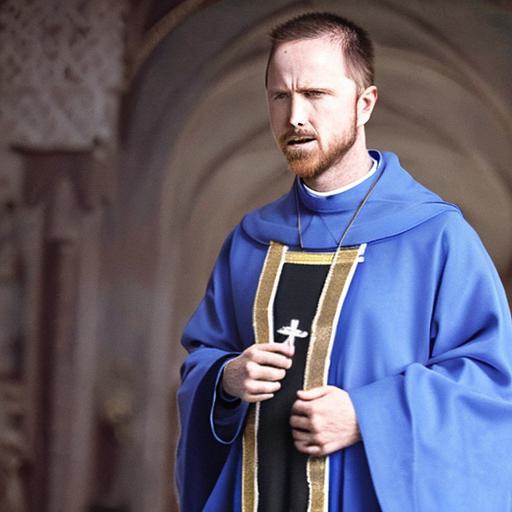} &
        \includegraphics[width=0.15\textwidth]{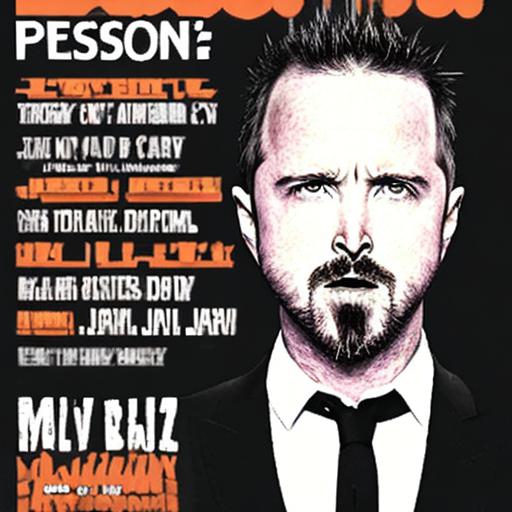} \\

        Real Sample &
        \begin{tabular}{c} ``$S_*$ portrait as \\ an asia old \\ warrior chief'' \end{tabular} &
        \begin{tabular}{c} ``$S_*$ is \\ riding a dragon'' \end{tabular} &
        \begin{tabular}{c} ``$S_*$ is \\ riding a horse'' \end{tabular} &
        \begin{tabular}{c} ``$S_*$ as a priest \\ in blue robes, \\ national geographic'' \end{tabular} &
        \begin{tabular}{c} ``A concert poster \\ of $S_*$'' \end{tabular} \\ \\[-0.185cm]



    \\[-0.4cm]        
    \end{tabular}
    }
    \caption{Additional examples of personalized text-to-image generation obtained with Cross Initialization.}
    \label{fig:appendix_ours_1}
\end{figure*}

\begin{figure*}
    \centering
    \setlength{\tabcolsep}{0.1pt}
    {\footnotesize
    \begin{tabular}{c@{\hspace{0.25cm}} c@{\hspace{0.25cm}} c@{\hspace{0.25cm}} c@{\hspace{0.25cm}} c@{\hspace{0.25cm}} c@{\hspace{0.25cm}} c@{\hspace{0.25cm}}}

        \includegraphics[width=0.15\textwidth]{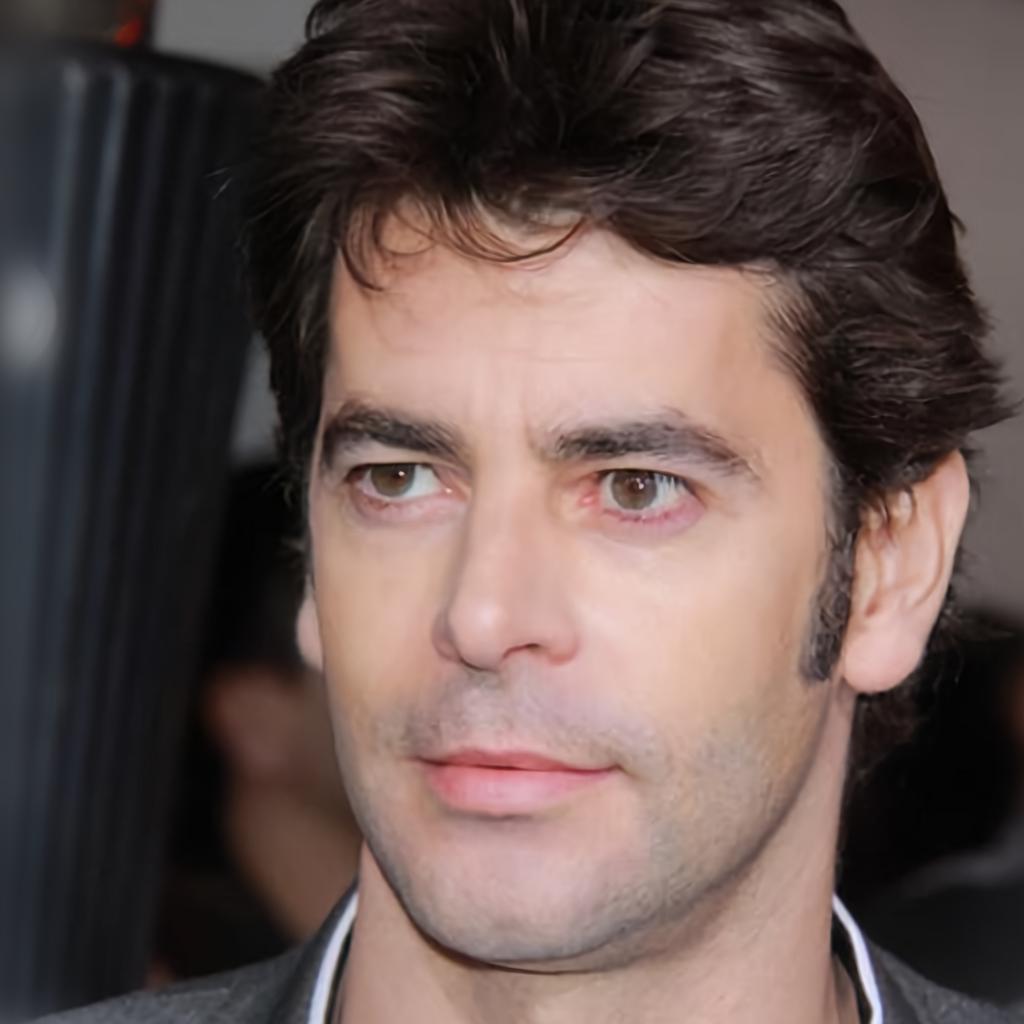} &
        \includegraphics[width=0.15\textwidth]{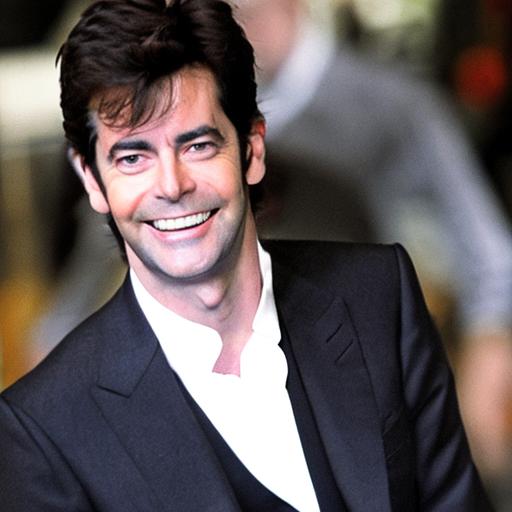} &
        \includegraphics[width=0.15\textwidth]{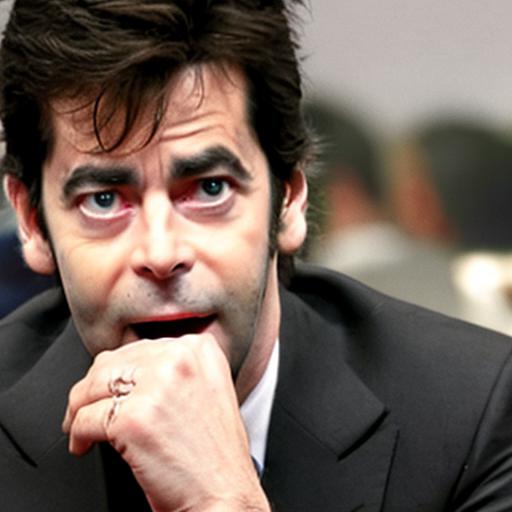} &
        \includegraphics[width=0.15\textwidth]{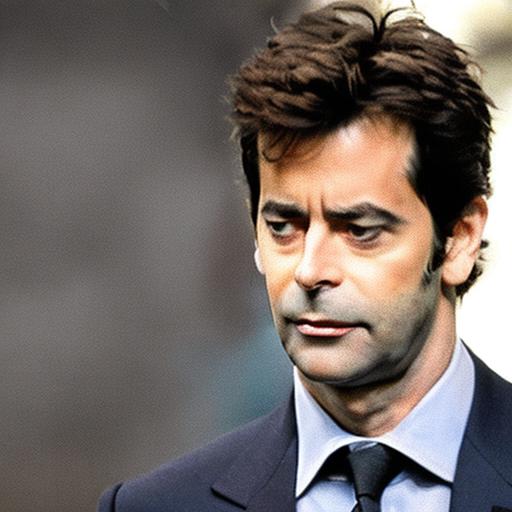} &
        \includegraphics[width=0.15\textwidth]{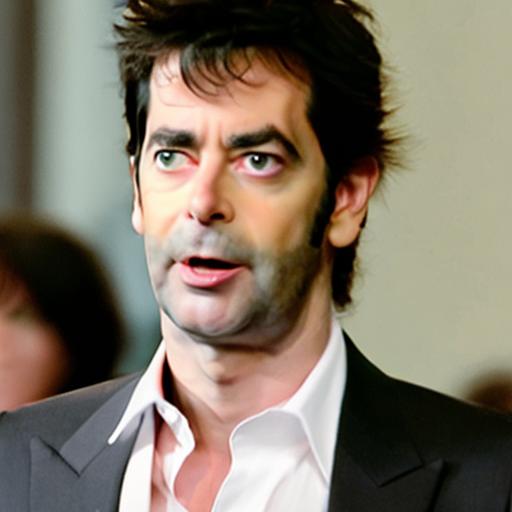} &
        \includegraphics[width=0.15\textwidth]{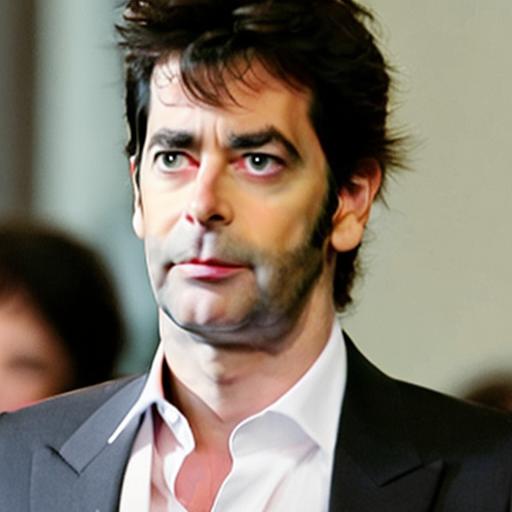} \\

        Real Sample &
        \begin{tabular}{c} ``$S_*$ with a \\ happy expression'' \end{tabular} &
        \begin{tabular}{c} ``$S_*$ with a \\ terrified expression'' \end{tabular} &
        \begin{tabular}{c} ``$S_*$ with a \\ depressed expression'' \end{tabular} &
        \begin{tabular}{c} ``$S_*$ with an \\ amazed expression'' \end{tabular} &
        \begin{tabular}{c} ``$S_*$ with a \\ confused expression'' \end{tabular} \\ \\[-0.185cm]

        \includegraphics[width=0.15\textwidth]{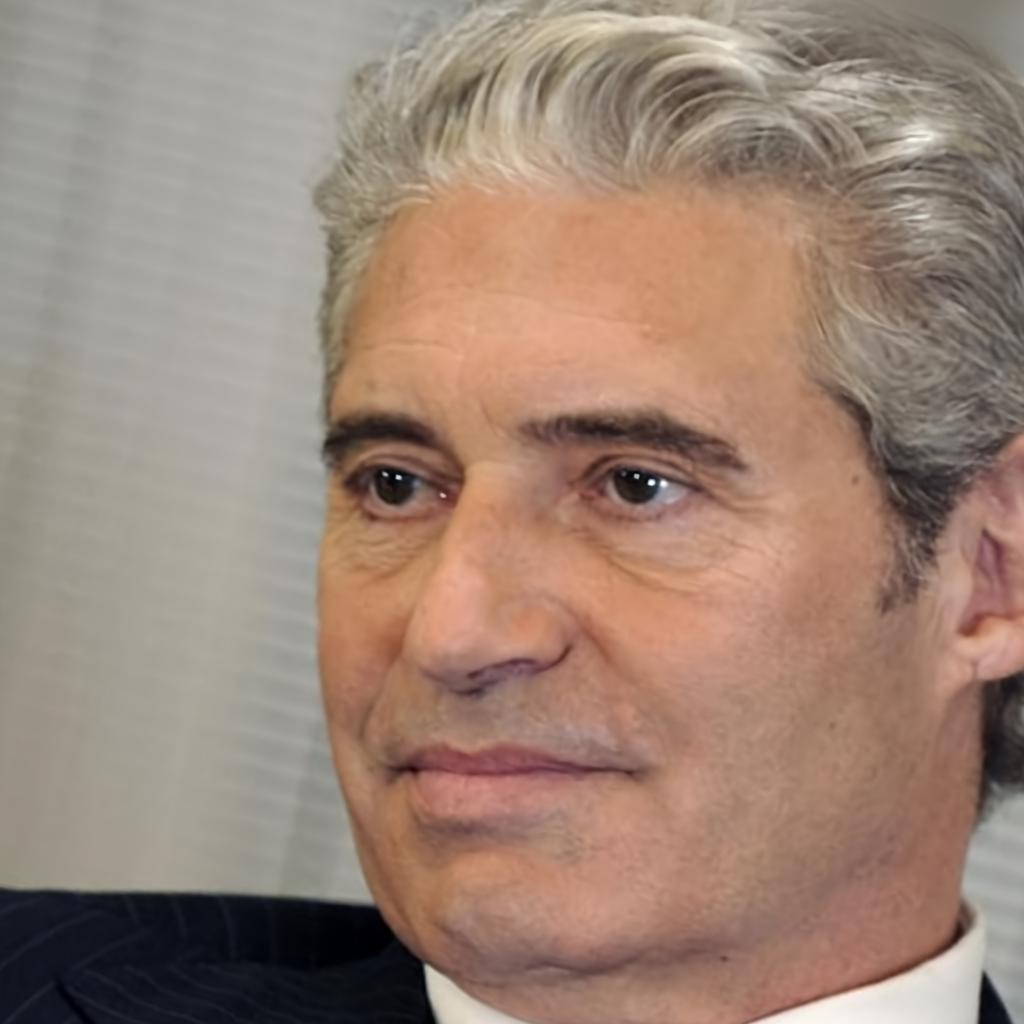} &
        \includegraphics[width=0.15\textwidth]{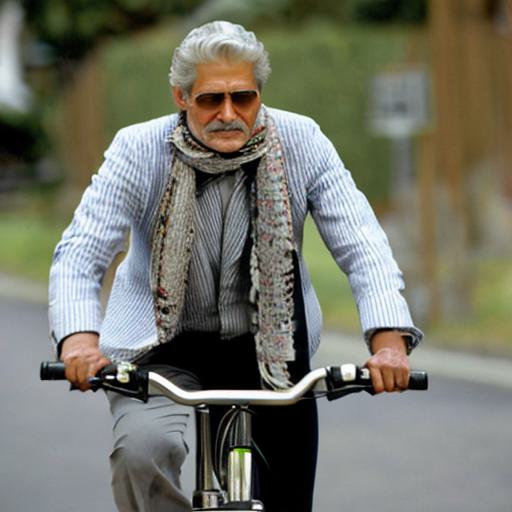} &
        \includegraphics[width=0.15\textwidth]{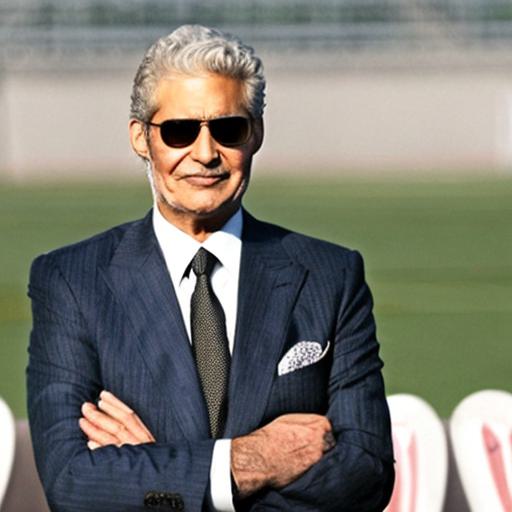} &
        \includegraphics[width=0.15\textwidth]{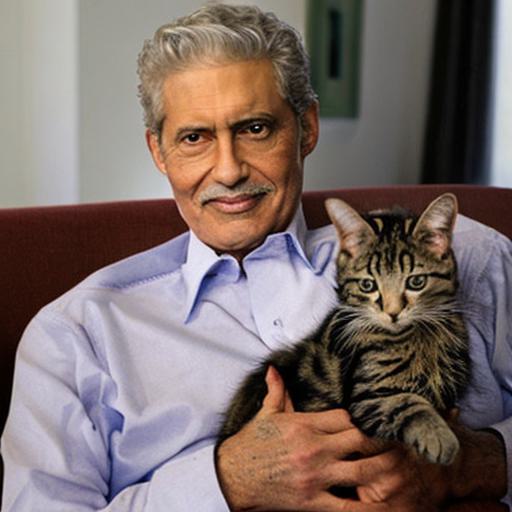} &
        \includegraphics[width=0.15\textwidth]{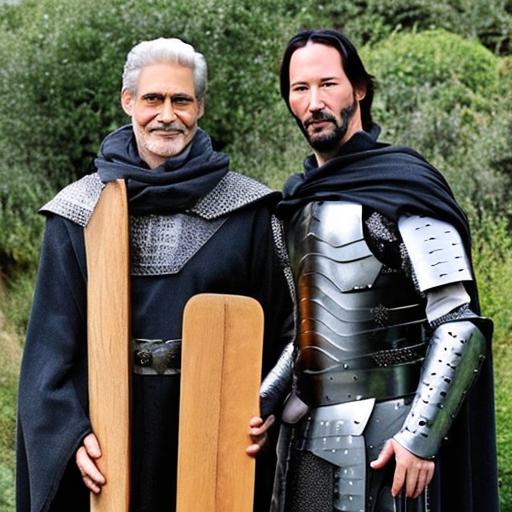} &
        \includegraphics[width=0.15\textwidth]{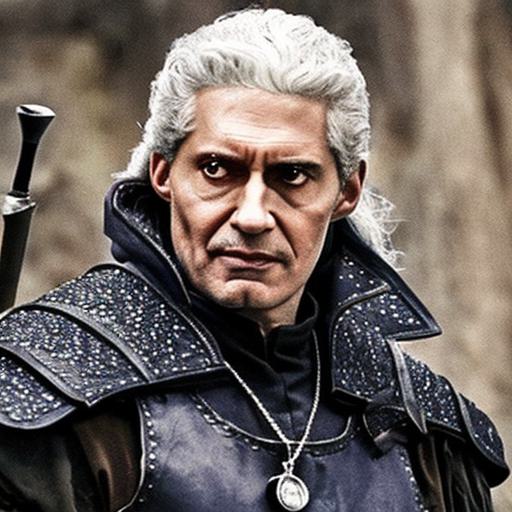} \\

        Real Sample &
        \begin{tabular}{c} ``$S_*$ is riding \\ a bicycle wearing \\ a shirt and a scarf'' \end{tabular} &
        \begin{tabular}{c} ``$S_*$ wears a suit \\ on a soccer field'' \end{tabular} &
        \begin{tabular}{c} ``$S_*$ is sitting \\ on a sofa \\ holding a cat'' \end{tabular} &
        \begin{tabular}{c} ``$S_*$ and Keanu \\ Reeves dressed as \\ knights holding a \\ wooden board'' \end{tabular} &
        \begin{tabular}{c} ``$S_*$ as \\ a Witcher'' \end{tabular} \\ \\[-0.185cm]

        \includegraphics[width=0.15\textwidth]{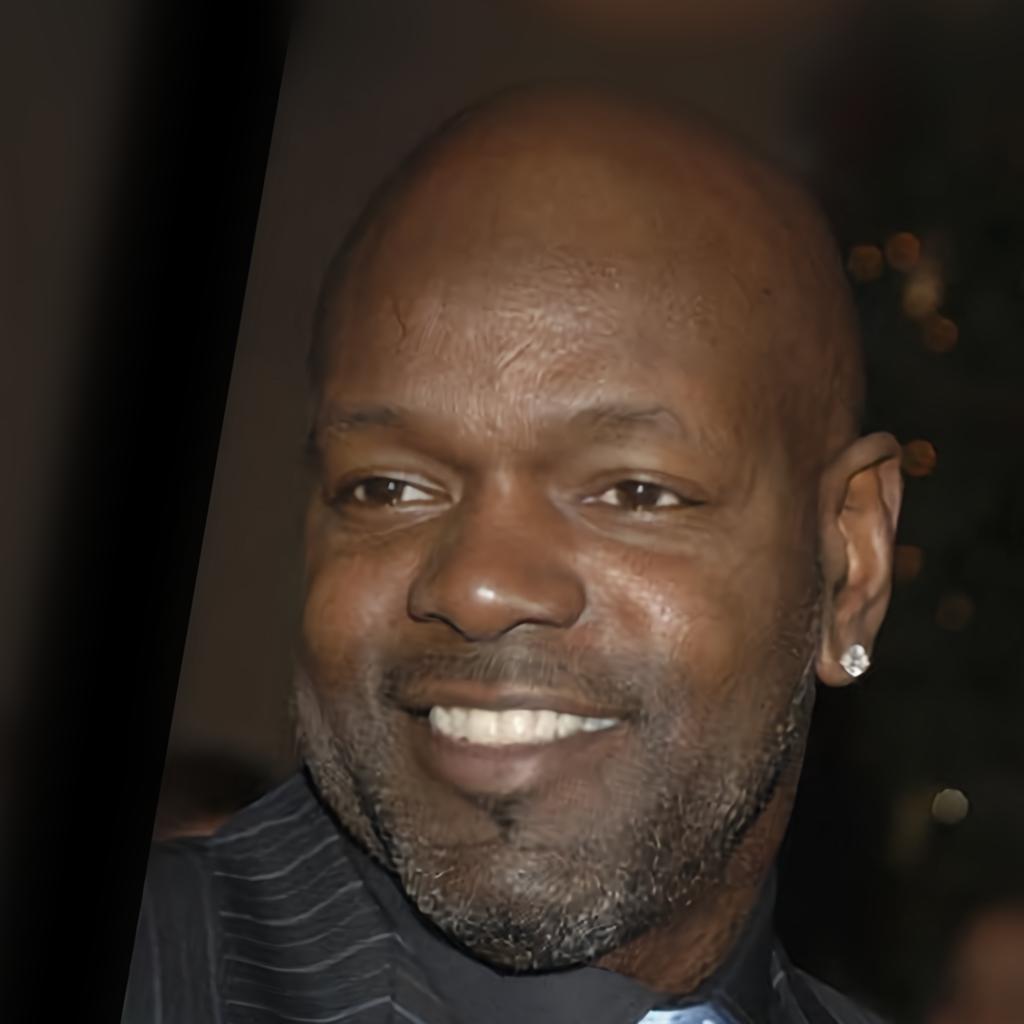} &
        \includegraphics[width=0.15\textwidth]{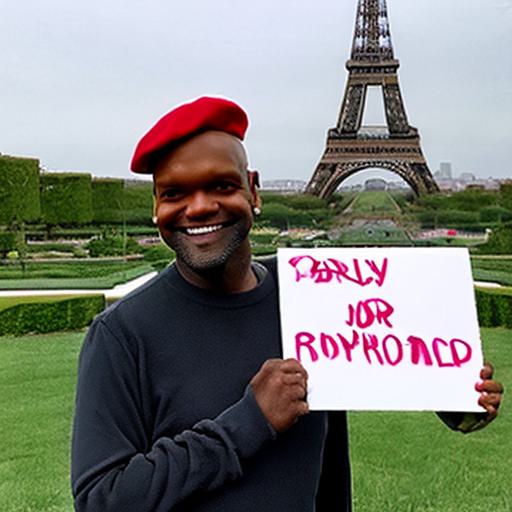} &
        \includegraphics[width=0.15\textwidth]{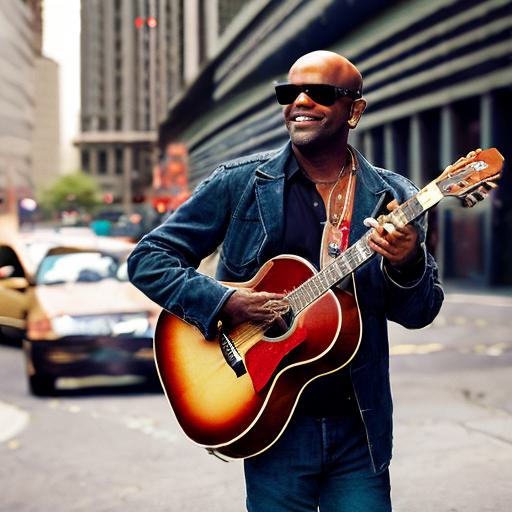} &
        \includegraphics[width=0.15\textwidth]{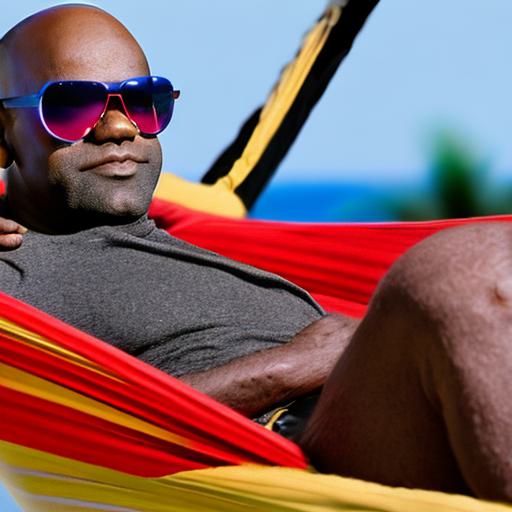} &
        \includegraphics[width=0.15\textwidth]{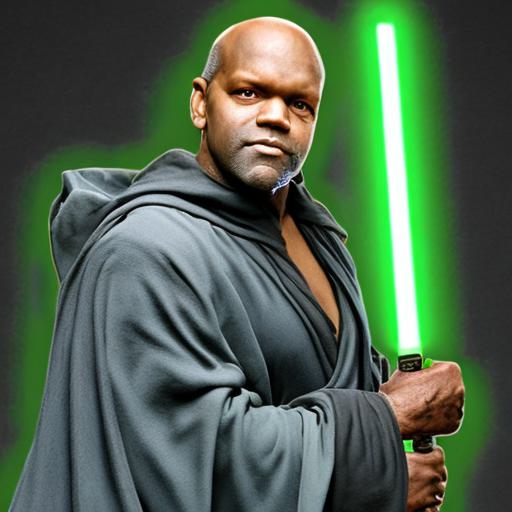} &
        \includegraphics[width=0.15\textwidth]{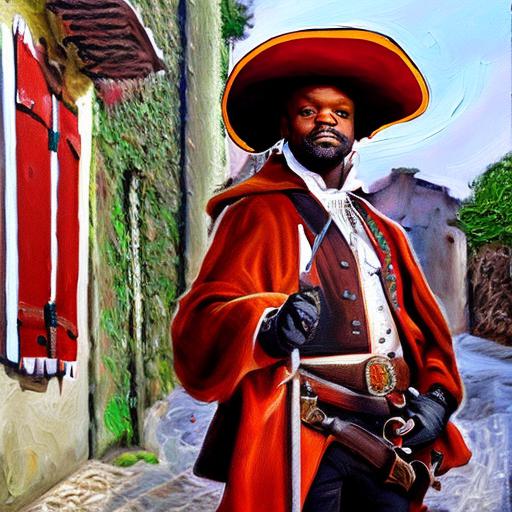} \\

        Real Sample &
        \begin{tabular}{c} ``A photo of \\ $S_*$ wearing a \\ beret holding  a \\ sign in front of \\ the Eiffel Tower'' \end{tabular} &
        \begin{tabular}{c} ``$S_*$ is playing \\ guitar in a \\ lively urban setting'' \end{tabular} &
        \begin{tabular}{c} ``$S_*$ sitting in \\ a hammock with \\ sunglasses on'' \end{tabular} &
        \begin{tabular}{c} ``$S_*$ as a Jedi'' \end{tabular} &
        \begin{tabular}{c} ``An oil painting \\ of $S_*$ dressed as \\ a musketeer in \\ an old French town'' \end{tabular} \\ \\[-0.185cm]

        \includegraphics[width=0.15\textwidth]{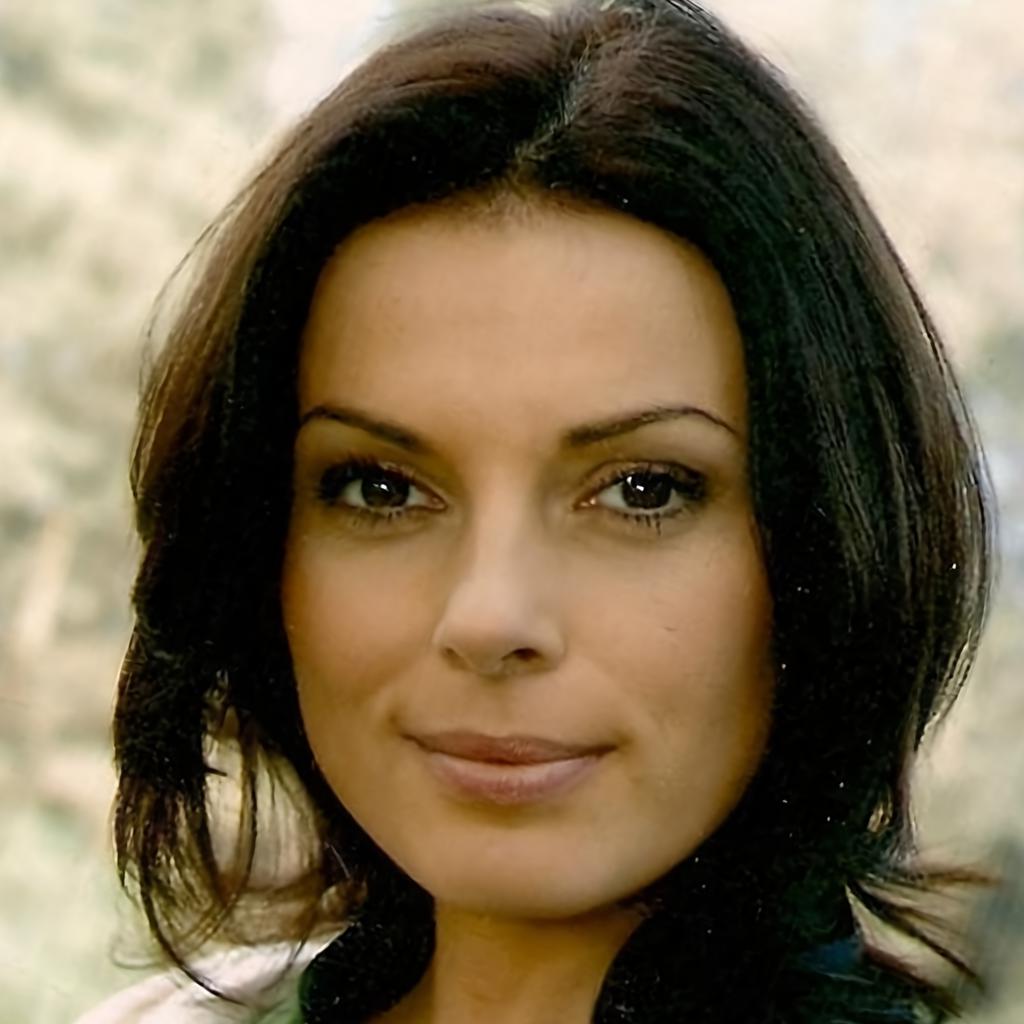} &
        \includegraphics[width=0.15\textwidth]{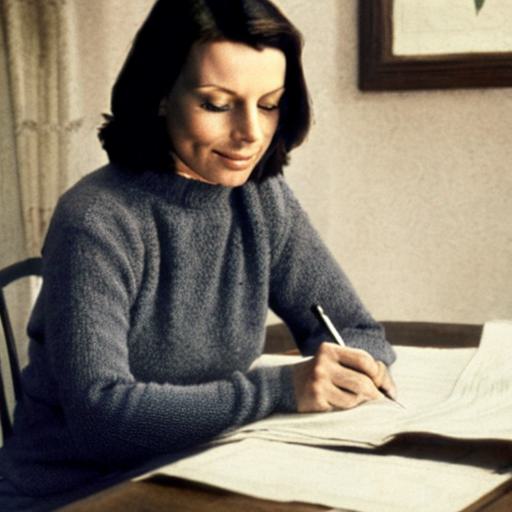} &
        \includegraphics[width=0.15\textwidth]{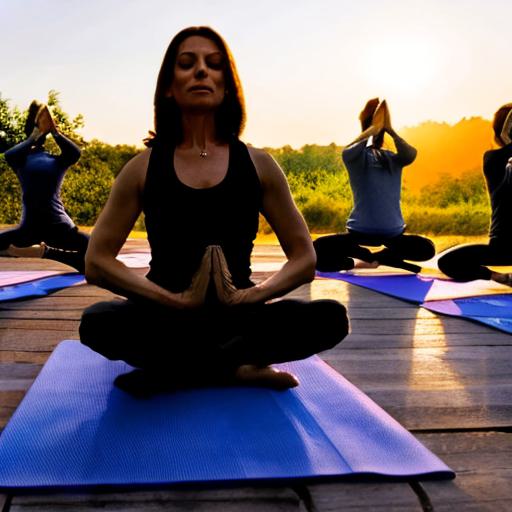} &
        \includegraphics[width=0.15\textwidth]{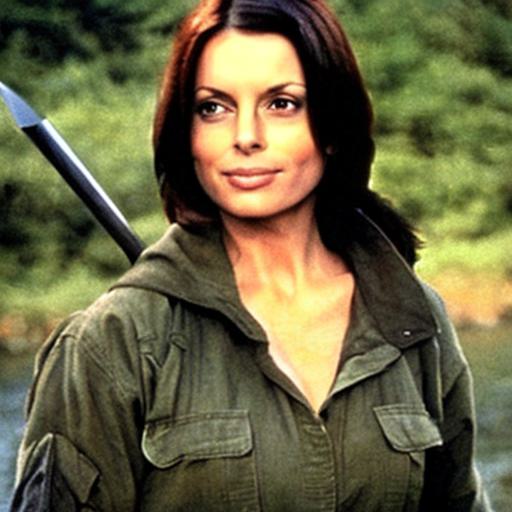} &
        \includegraphics[width=0.15\textwidth]{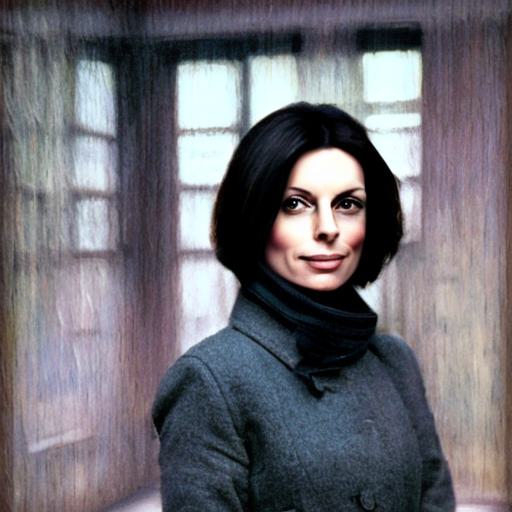} &
        \includegraphics[width=0.15\textwidth]{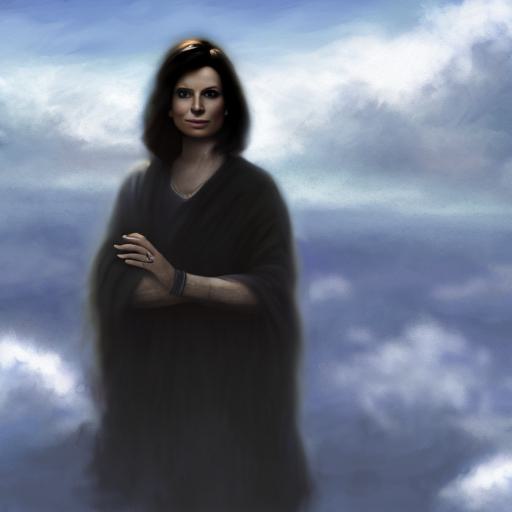} \\

        Real Sample &
        \begin{tabular}{c} ``$S_*$ in a serene \\ studio writing \\ elegant script'' \end{tabular} &
        \begin{tabular}{c} ``$S_*$ yoga instructor \\ leading a class \\ at dawn with the \\ sun in the background'' \end{tabular} &
        \begin{tabular}{c} ``$S_*$ as an \\ amazon warrior'' \end{tabular} &
        \begin{tabular}{c} ``$S_*$ in the style \\ of stefan kostic \\ and david la chapelle'' \end{tabular} &
        \begin{tabular}{c} ``A highly detailed \\ digital art of \\ $S_*$ mage \\ standing on clouds'' \end{tabular} \\ \\[-0.185cm]

        \includegraphics[width=0.15\textwidth]{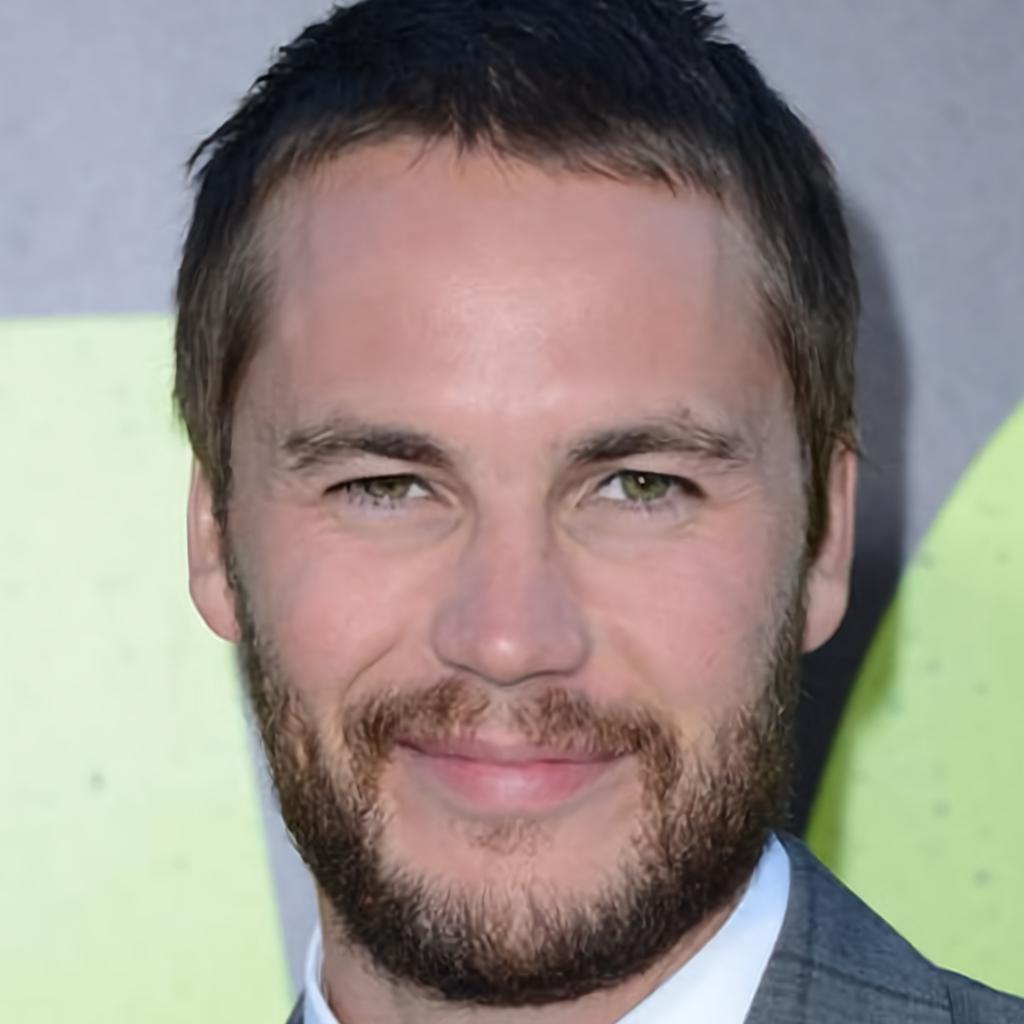} &
        \includegraphics[width=0.15\textwidth]{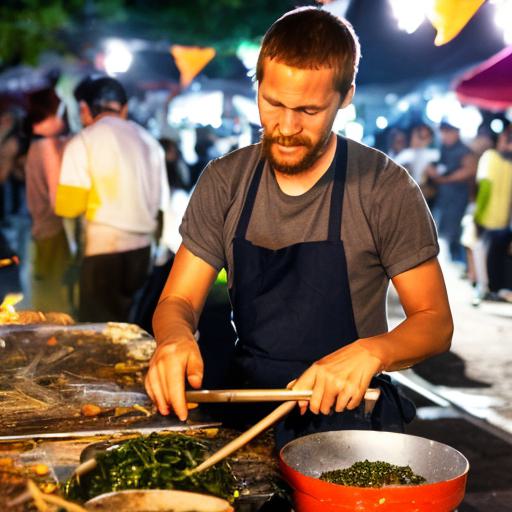} &
        \includegraphics[width=0.15\textwidth]{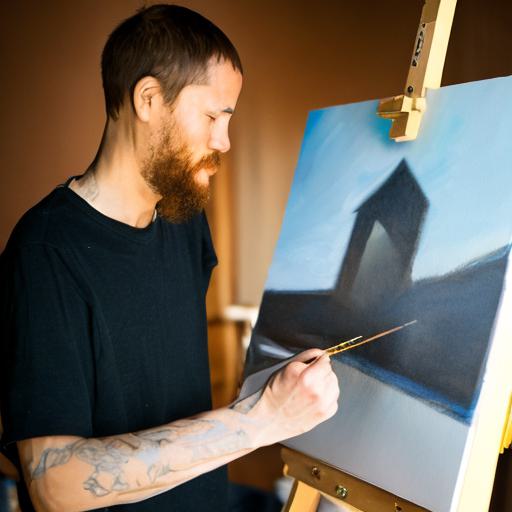} &
        \includegraphics[width=0.15\textwidth]{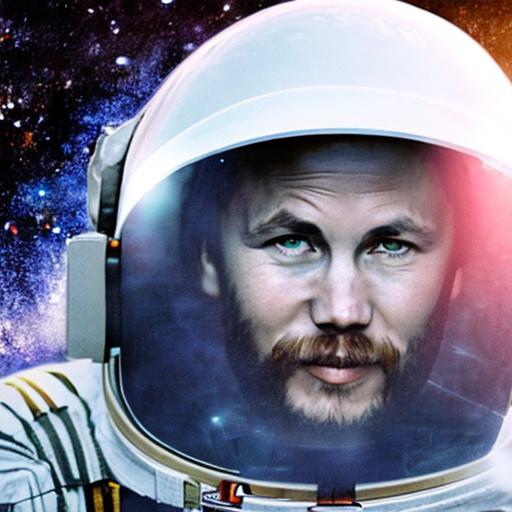} &
        \includegraphics[width=0.15\textwidth]{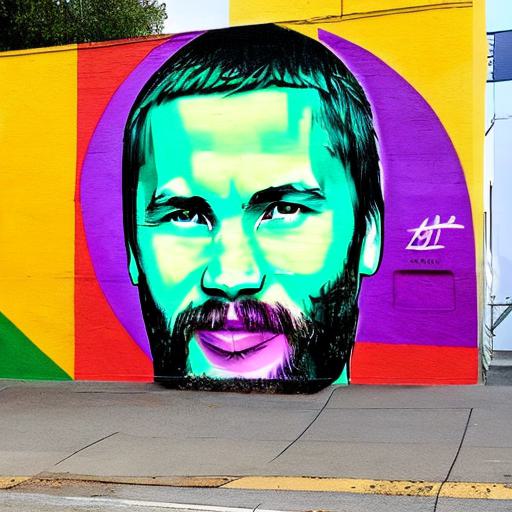} &
        \includegraphics[width=0.15\textwidth]{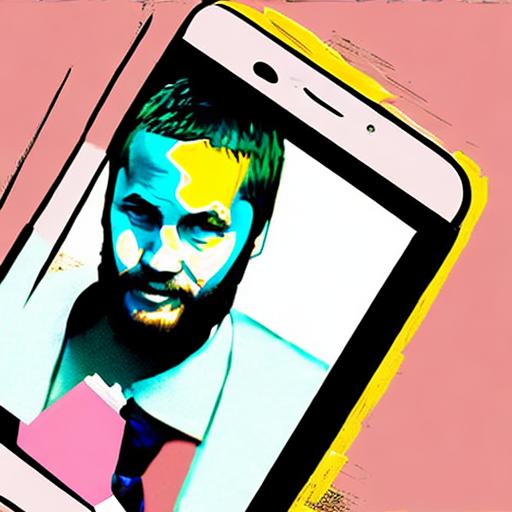} \\

        Real Sample &
        \begin{tabular}{c} ``$S_*$ cooking at \\ a night market'' \end{tabular} &
        \begin{tabular}{c} ``A dslr photo \\ of $S_*$ painting \\ in a sunlit studio'' \end{tabular} &
        \begin{tabular}{c} ``Renaissance-style \\ portrait of $S_*$ \\ astronaut in space \\ detailed starry \\ background \\ reflective helmet'' \end{tabular} &
        \begin{tabular}{c} ``A colorful mural \\ of $S_*$ on an \\ urban street wall'' \end{tabular} &
        \begin{tabular}{c} ``Pop Art painting \\ of a modern smartphone \\ with classic art pieces \\ of $S_*$ appearing \\ on the screen'' \end{tabular} \\ \\[-0.185cm]

    \\[-0.4cm]        
    \end{tabular}
    }
    \caption{Additional examples of personalized text-to-image generation obtained with Cross Initialization.}
    \label{fig:appendix_ours_2}
\end{figure*}

In \cref{fig:appendix_ours_1} and \cref{fig:appendix_ours_2}, we provide additional qualitative results obtained by our method on a diverse set of prompts.

\section{Results on Synthetic Facial Images}
\label{sec:synthetic}
\begin{figure*}
    \centering
    \setlength{\tabcolsep}{0.1pt}
    {\footnotesize
    \begin{tabular}{c@{\hspace{0.25cm}} c@{\hspace{0.25cm}} c@{\hspace{0.25cm}} c@{\hspace{0.25cm}} c@{\hspace{0.25cm}} c@{\hspace{0.25cm}} c@{\hspace{0.25cm}}}

        \includegraphics[width=0.15\textwidth]{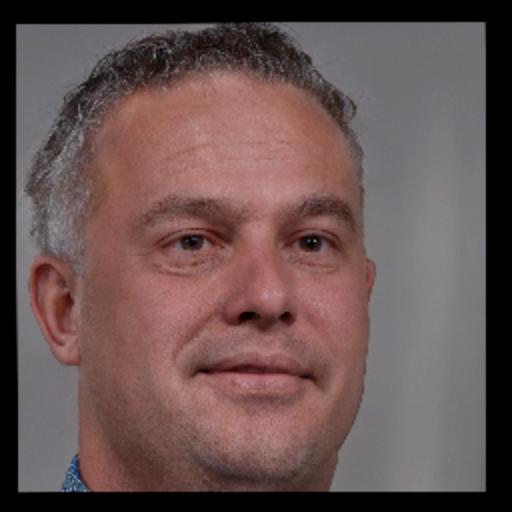} &
        \includegraphics[width=0.15\textwidth]{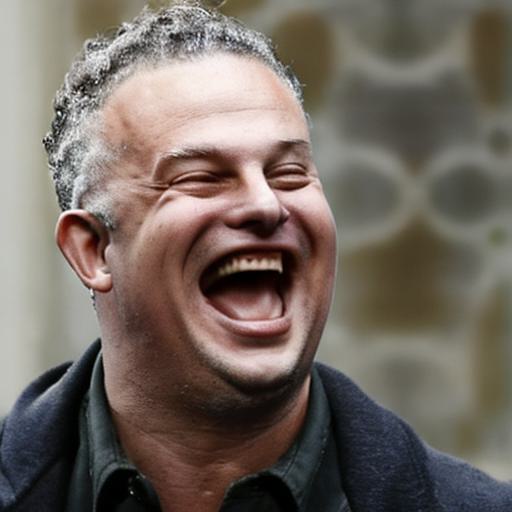} &
        \includegraphics[width=0.15\textwidth]{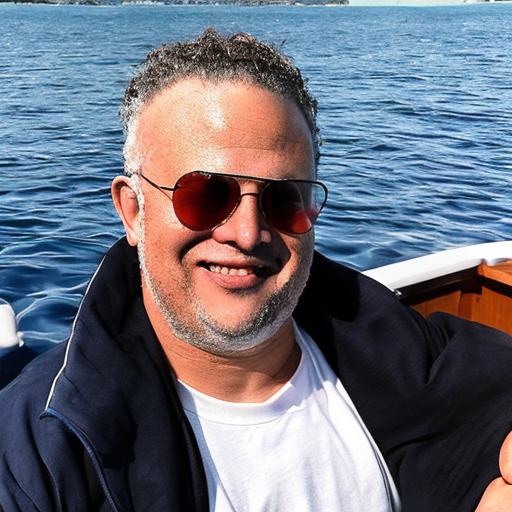} &
        \includegraphics[width=0.15\textwidth]{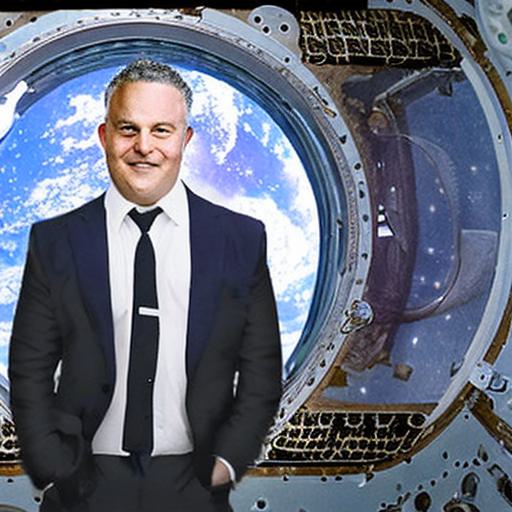} &
        \includegraphics[width=0.15\textwidth]{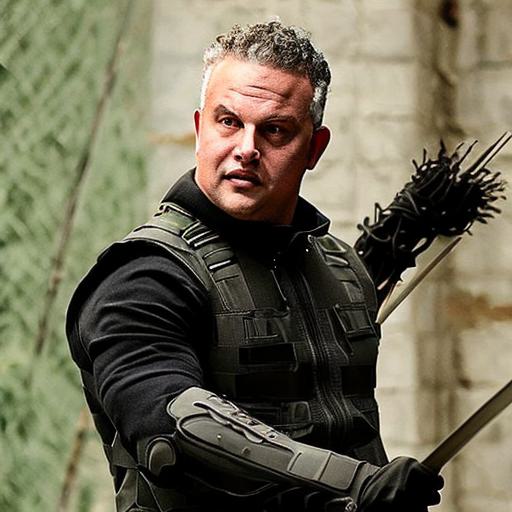} &
        \includegraphics[width=0.15\textwidth]{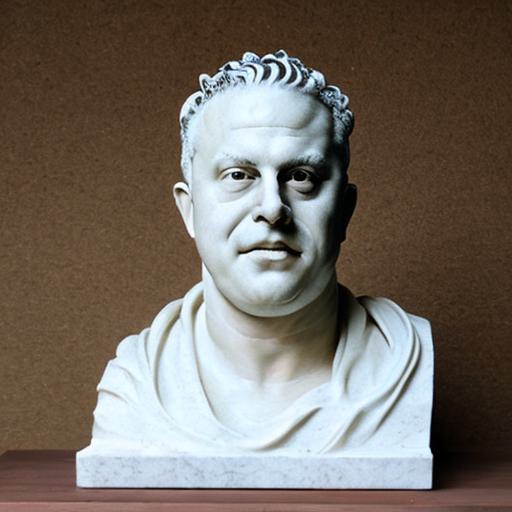} \\

        Synthetic Sample &
        \begin{tabular}{c} ``$S_*$ with an \\ ecstatic expression'' \end{tabular} &
        \begin{tabular}{c} ``$S_*$ wears a \\ sunglass on \\ a boat '' \end{tabular} &
        \begin{tabular}{c} ``$S_*$ wears a \\ suit in \\ space ship'' \end{tabular} &
        \begin{tabular}{c} ``$S_*$ as Hawkeye '' \end{tabular} &
        \begin{tabular}{c} ``Marble sculpture \\ of $S_*$'' \end{tabular} \\ \\[-0.185cm]

        \includegraphics[width=0.15\textwidth]{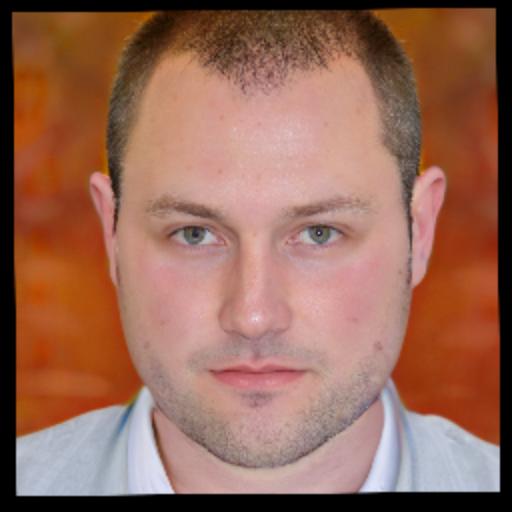} &
        \includegraphics[width=0.15\textwidth]{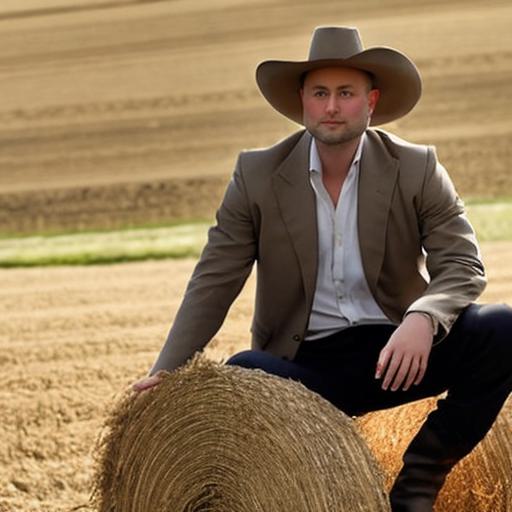} &
        \includegraphics[width=0.15\textwidth]{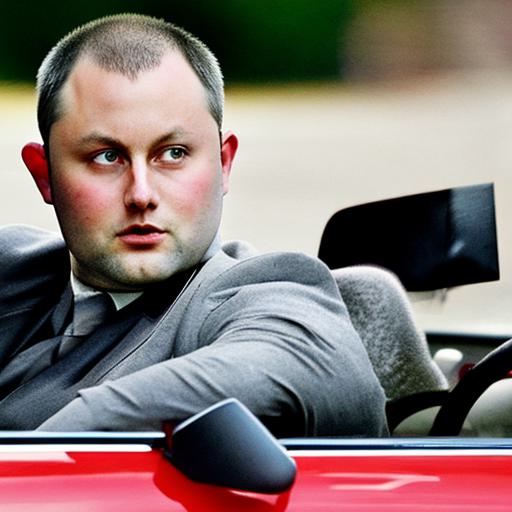} &
        \includegraphics[width=0.15\textwidth]{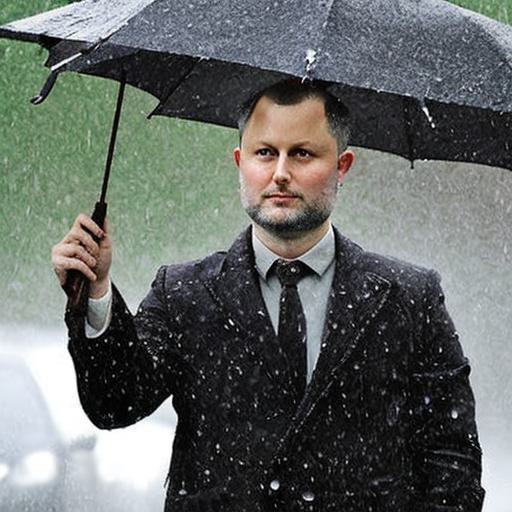} &
        \includegraphics[width=0.15\textwidth]{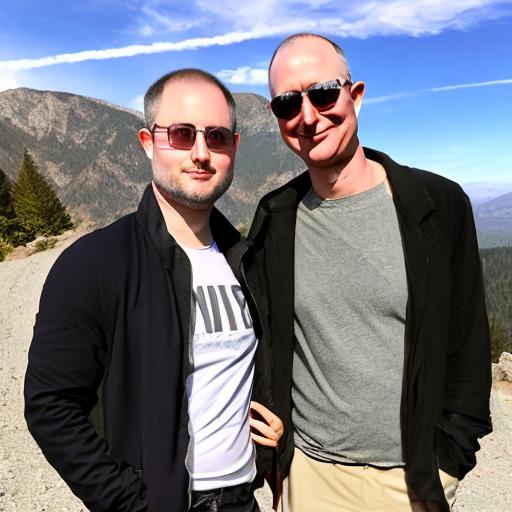} &
        \includegraphics[width=0.15\textwidth]{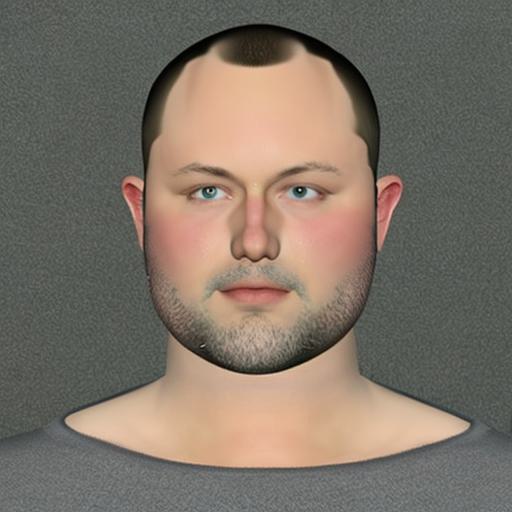} \\

        Synthetic Sample &
        \begin{tabular}{c} ``$S_*$ as a cowboy \\ sitting on hay'' \end{tabular} &
        \begin{tabular}{c} ``$S_*$ is driving \\ a car'' \end{tabular} &
        \begin{tabular}{c} ``$S_*$ stands in \\ the rain holding \\ an umbrella'' \end{tabular} &
        \begin{tabular}{c} ``$S_*$ and Jeff Bezos \\ taking a relaxing \\ hike in the mountains '' \end{tabular} &
        \begin{tabular}{c} ``3d modeling \\ of $S_*$ '' \end{tabular} \\ \\[-0.185cm]

    \\[-0.4cm]        
    \end{tabular}
    }
    \caption{Additional results on synthetic facial images generated by StyleGAN, where the input images are sourced from~\cite{yuan2023inserting}.}
    \label{fig:appendix_ours_gan}
\end{figure*}

Besides evaluating on real facial images, we also evaluate our method on synthetic facial images generated by StyleGAN. The results are shown in \cref{fig:appendix_ours_gan}. As can be seen, our method achieves high-quality personalized face generation on synthetic facial images.

\section{Additional Ablation Study Results}
\label{sec:more_ablation}
\begin{figure*}
    \centering
    \renewcommand{\arraystretch}{0.3}
    \setlength{\tabcolsep}{0.5pt}

    {\small

    \begin{tabular}{c@{\hspace{0.2cm}} c c @{\hspace{0.2cm}} c c @{\hspace{0.2cm}} c c @{\hspace{0.2cm}} c c }

        \begin{tabular}{c} Real Sample \\ \& Prompt \end{tabular} &
        \multicolumn{2}{c}{w/o CI} &
        \multicolumn{2}{c}{w/o Mean} &
        \multicolumn{2}{c}{w/o Reg} &
        \multicolumn{2}{c}{Full} \\

        \includegraphics[width=0.0975\textwidth]{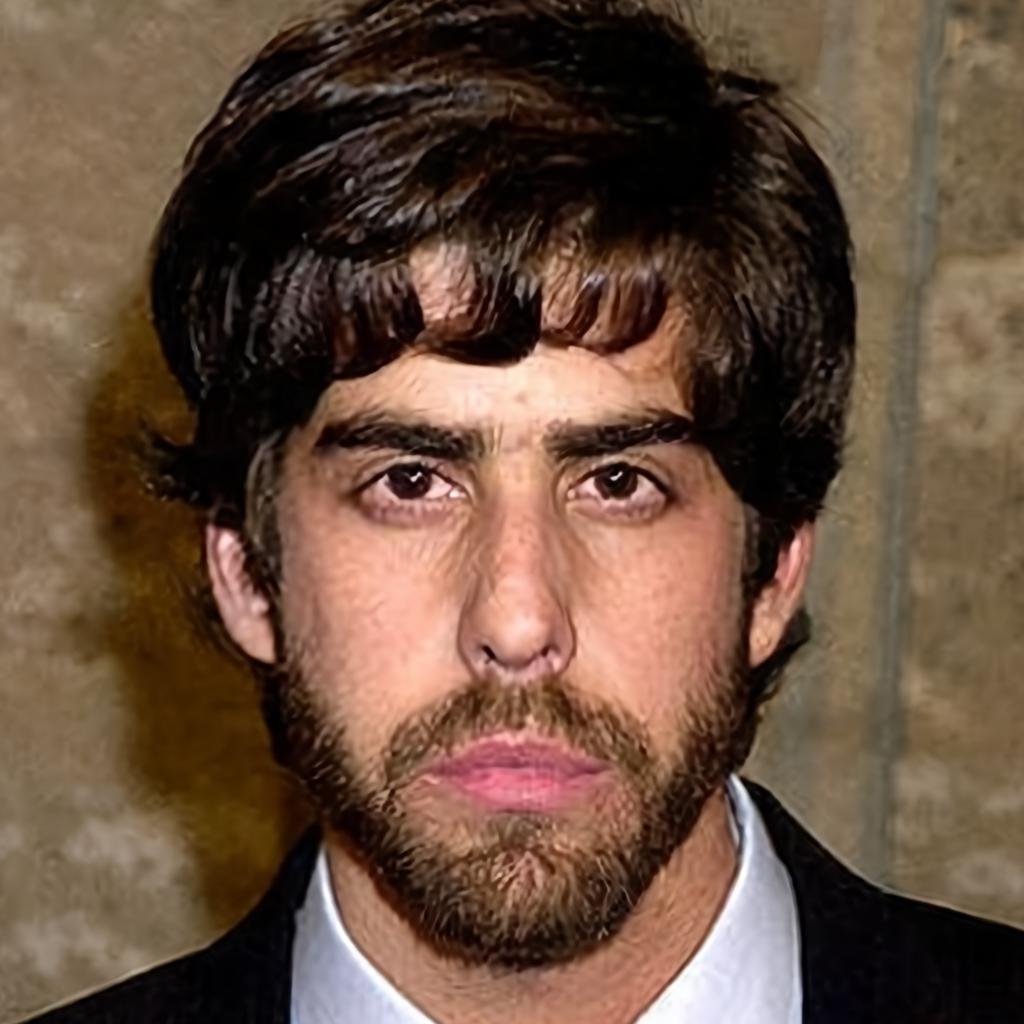} &
        \includegraphics[width=0.0975\textwidth]{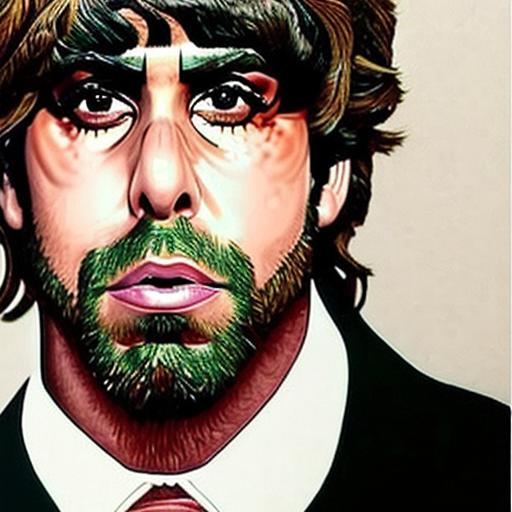} &
        \includegraphics[width=0.0975\textwidth]{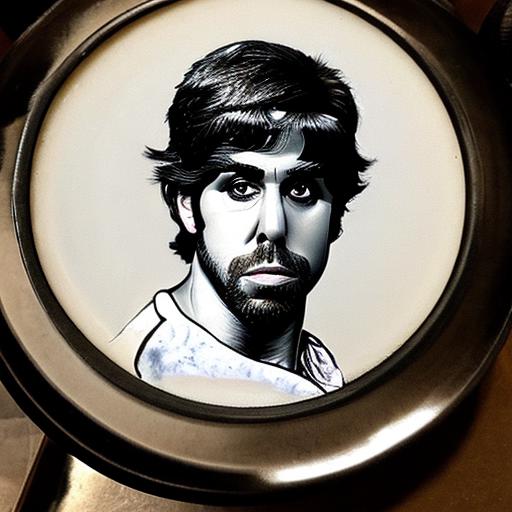} &
        \hspace{0.05cm}
        \includegraphics[width=0.0975\textwidth]{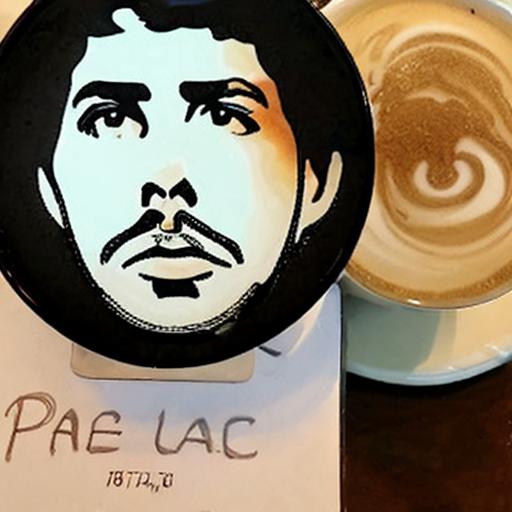} &
        \includegraphics[width=0.0975\textwidth]{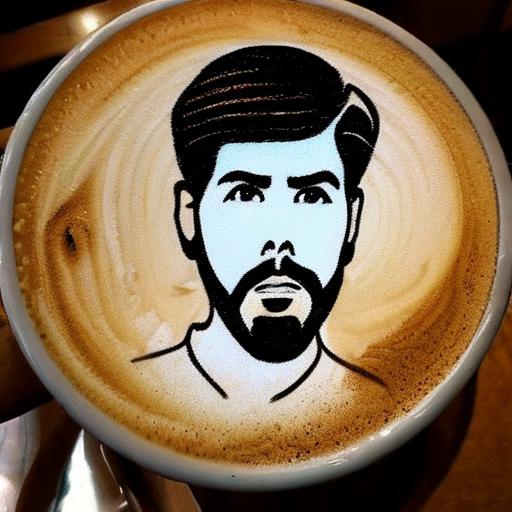} &
        \hspace{0.05cm}
        \includegraphics[width=0.0975\textwidth]{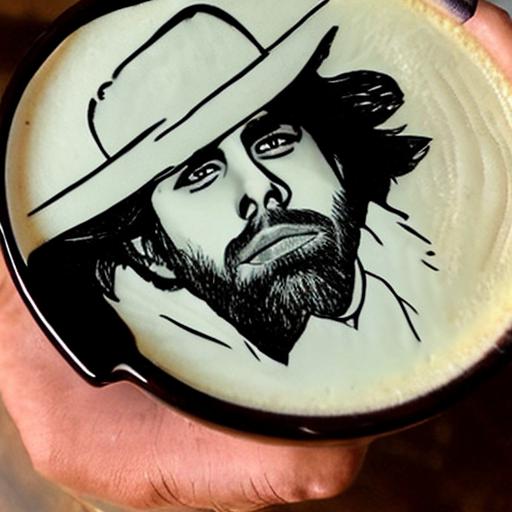} &
        \includegraphics[width=0.0975\textwidth]{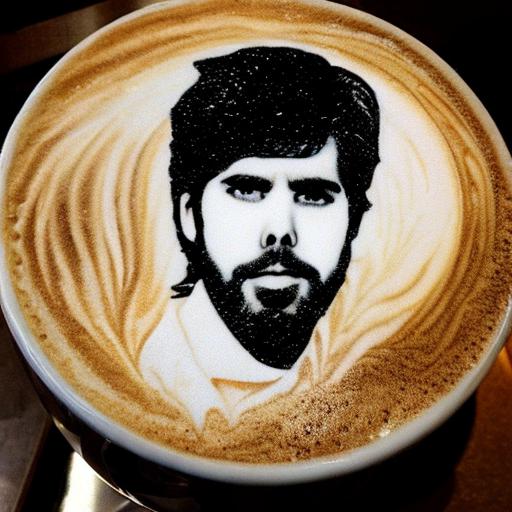} &
        \hspace{0.05cm}
        \includegraphics[width=0.0975\textwidth]{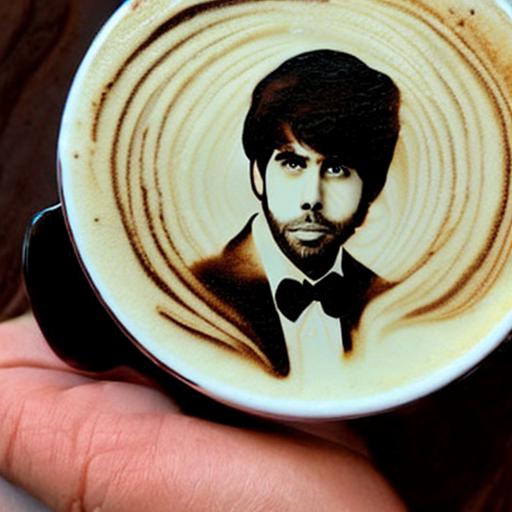} &
        \includegraphics[width=0.0975\textwidth]{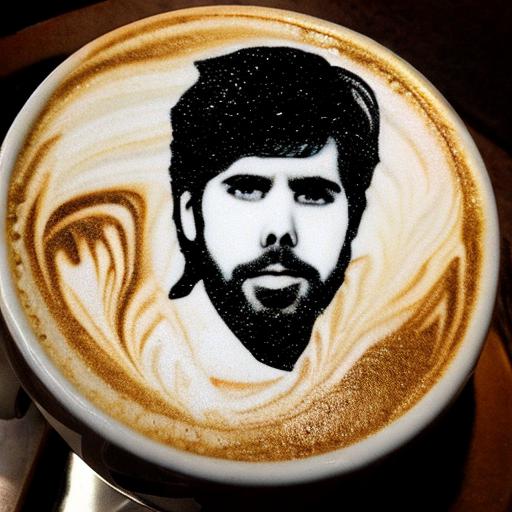} \\

        \raisebox{0.3in}{\begin{tabular}{c} ``$S_*$ latte \\ art''\end{tabular}} &
        \includegraphics[width=0.0975\textwidth]{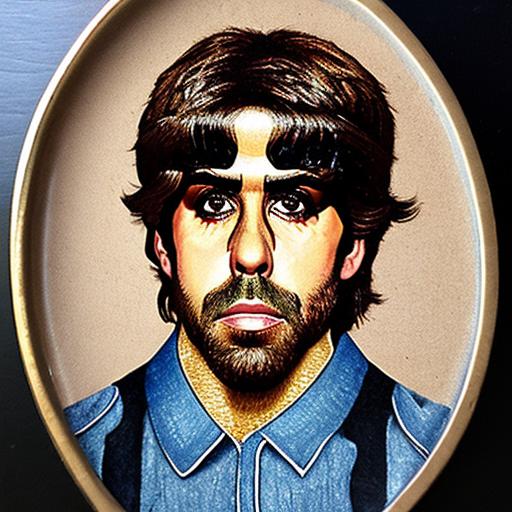} &
        \includegraphics[width=0.0975\textwidth]{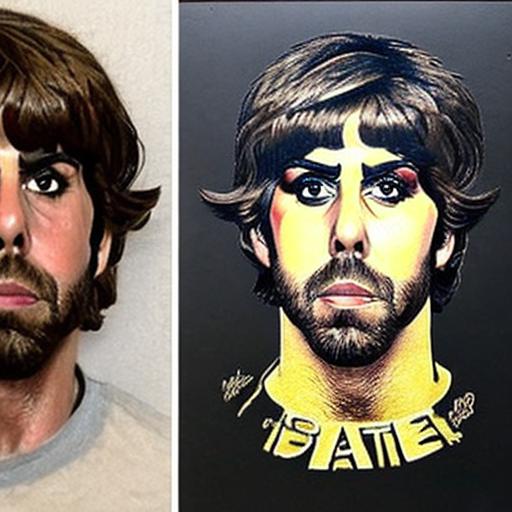} &
        \hspace{0.05cm}
        \includegraphics[width=0.0975\textwidth]{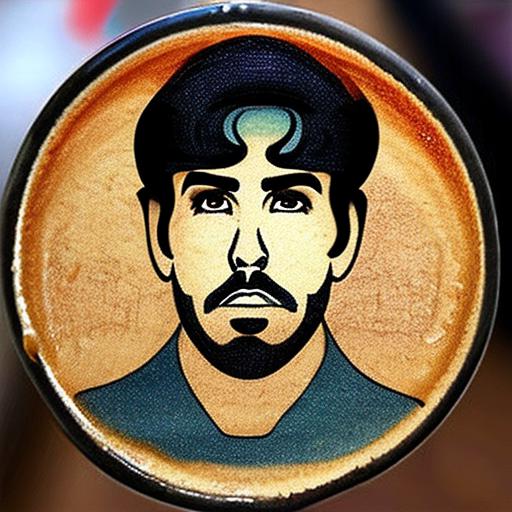} &
        \includegraphics[width=0.0975\textwidth]{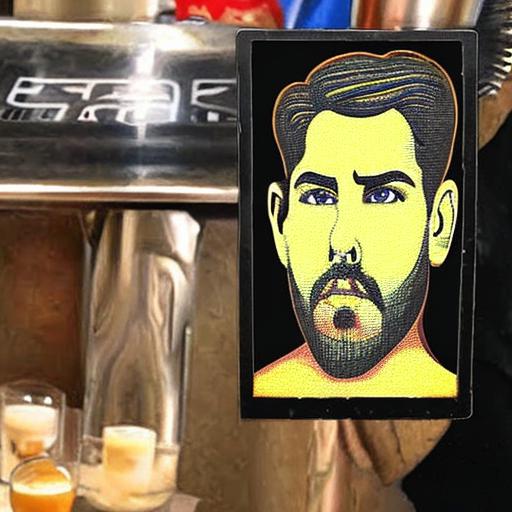} &
        \hspace{0.05cm}
        \includegraphics[width=0.0975\textwidth]{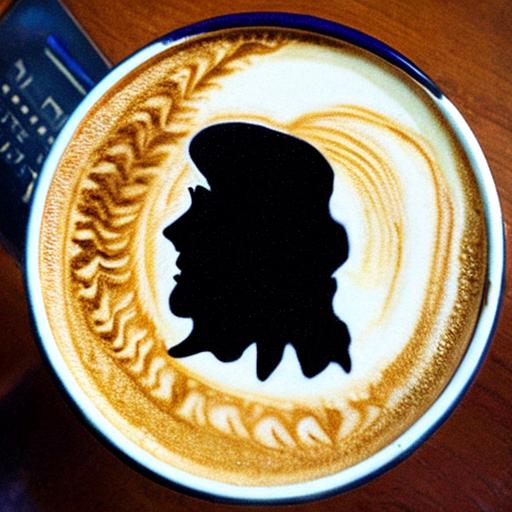} &
        \includegraphics[width=0.0975\textwidth]{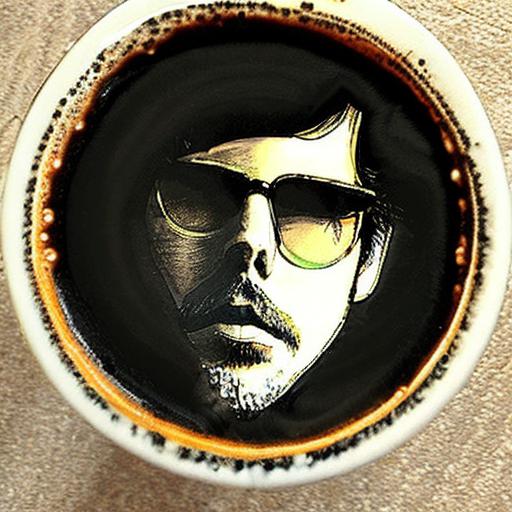} &
        \hspace{0.05cm}
        \includegraphics[width=0.0975\textwidth]{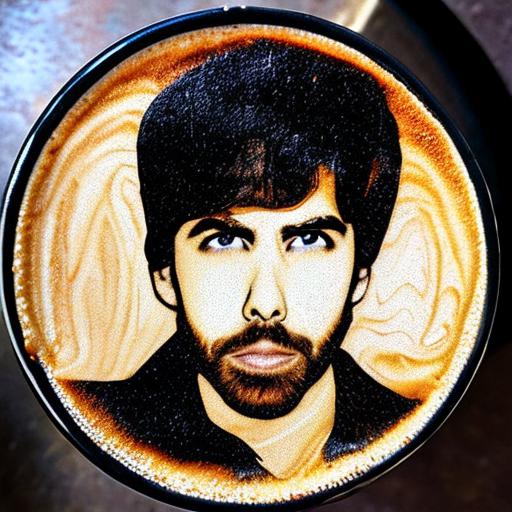} &
        \includegraphics[width=0.0975\textwidth]{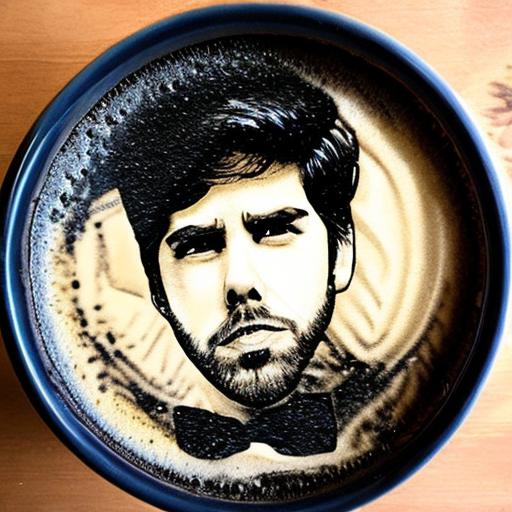} \\ \\

        \includegraphics[width=0.0975\textwidth]{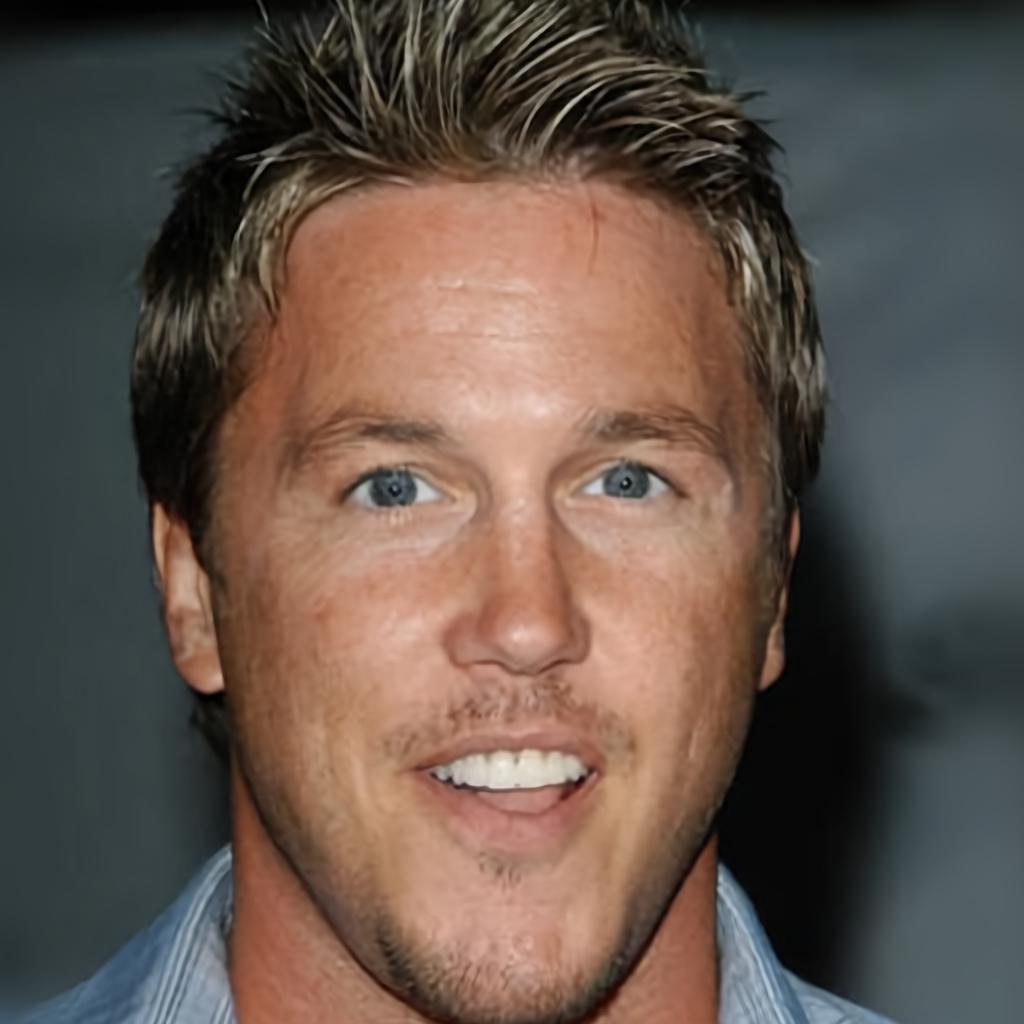} &
        \includegraphics[width=0.0975\textwidth]{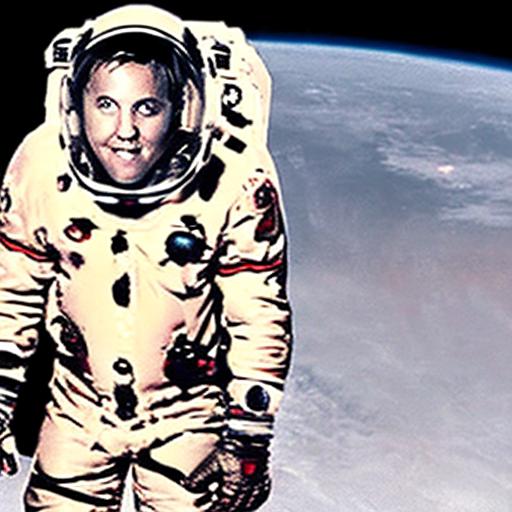} &
        \includegraphics[width=0.0975\textwidth]{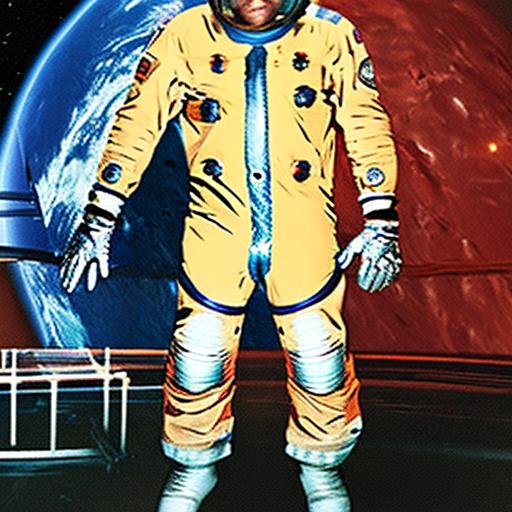} &
        \hspace{0.05cm}
        \includegraphics[width=0.0975\textwidth]{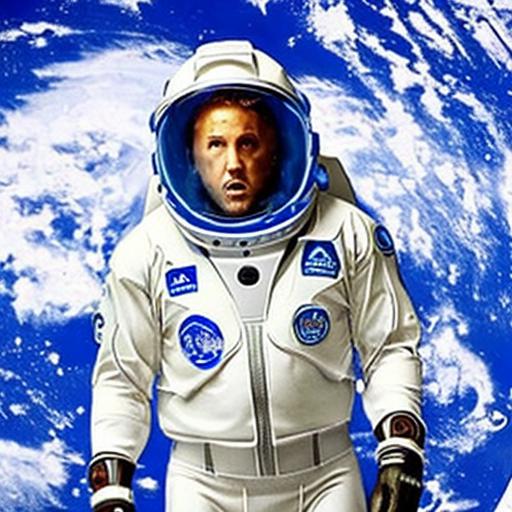} &
        \includegraphics[width=0.0975\textwidth]{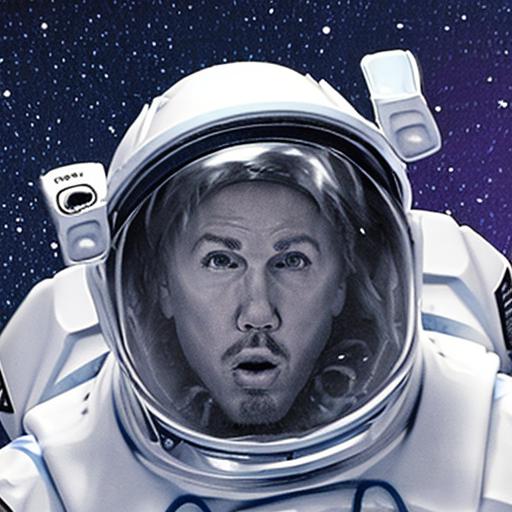} &
        \hspace{0.05cm}
        \includegraphics[width=0.0975\textwidth]{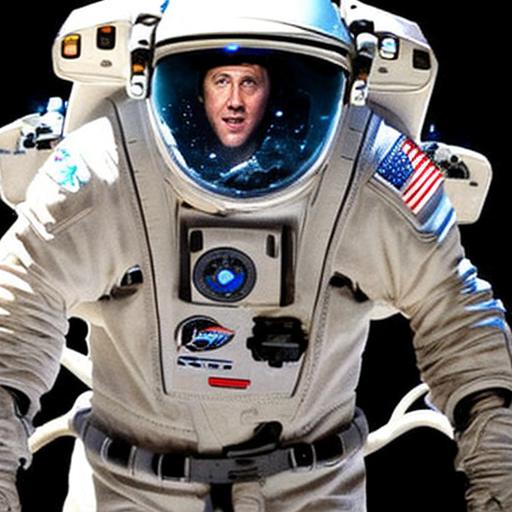} &
        \includegraphics[width=0.0975\textwidth]{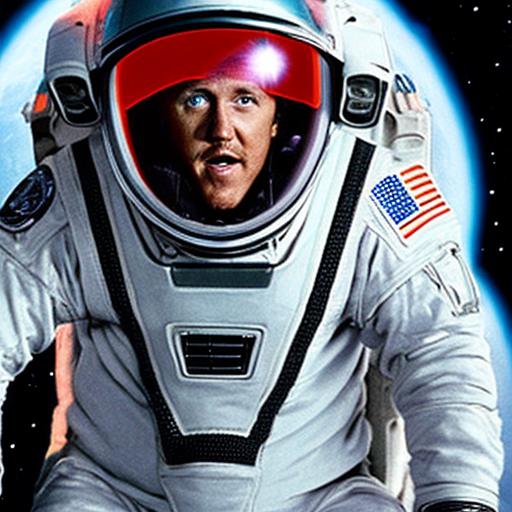} &
        \hspace{0.05cm}
        \includegraphics[width=0.0975\textwidth]{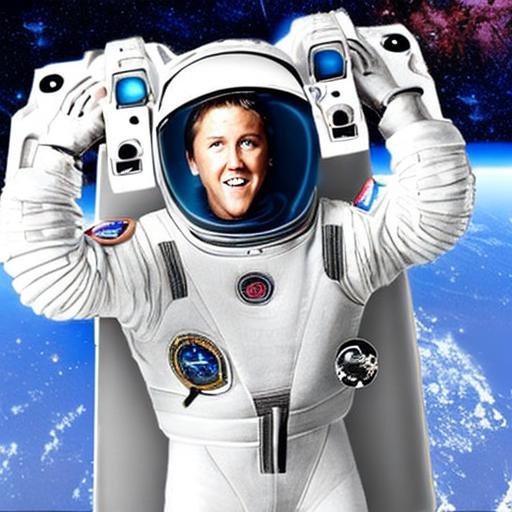} &
        \includegraphics[width=0.0975\textwidth]{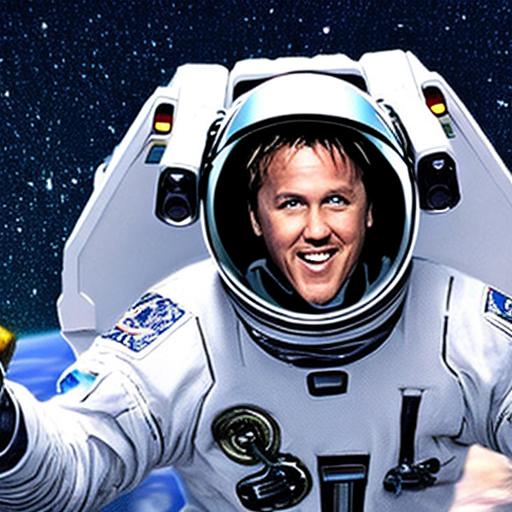} \\

        \raisebox{0.3in}{\begin{tabular}{c} ``$S_*$ wearing \\ a scifi spacesuit \\ in space''\end{tabular}} &
        \includegraphics[width=0.0975\textwidth]{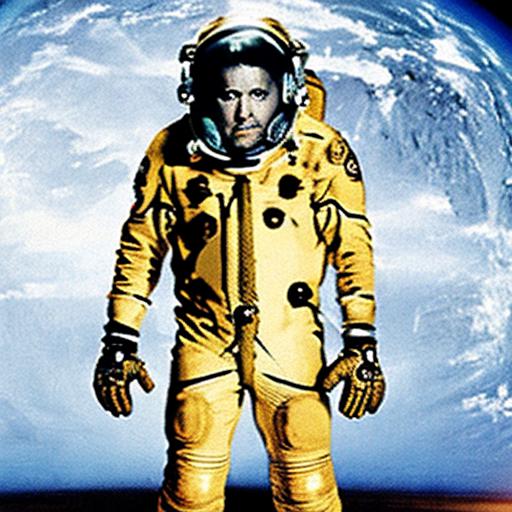} &
        \includegraphics[width=0.0975\textwidth]{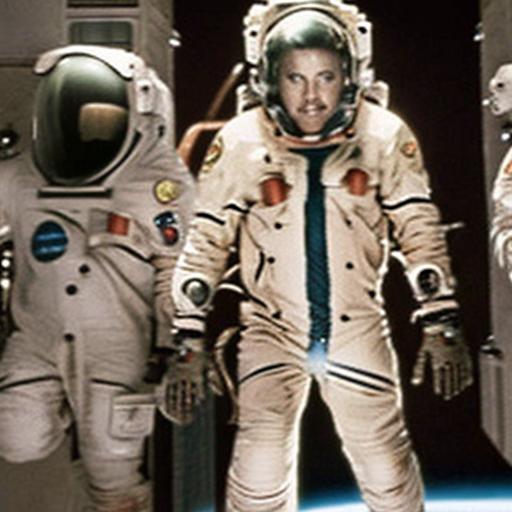} &
        \hspace{0.05cm}
        \includegraphics[width=0.0975\textwidth]{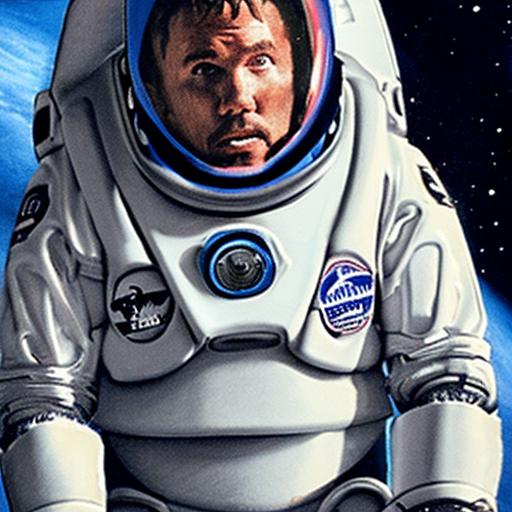} &
        \includegraphics[width=0.0975\textwidth]{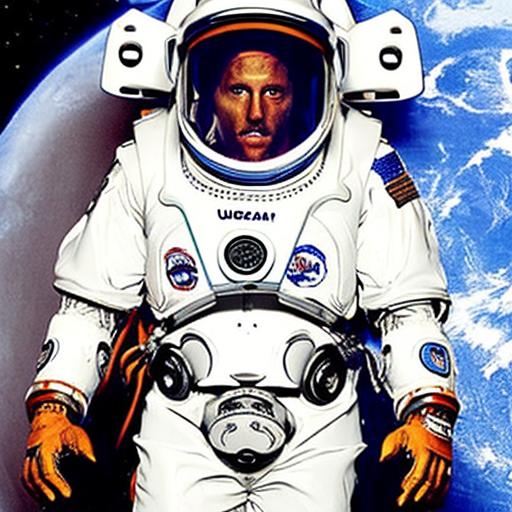} &
        \hspace{0.05cm}
        \includegraphics[width=0.0975\textwidth]{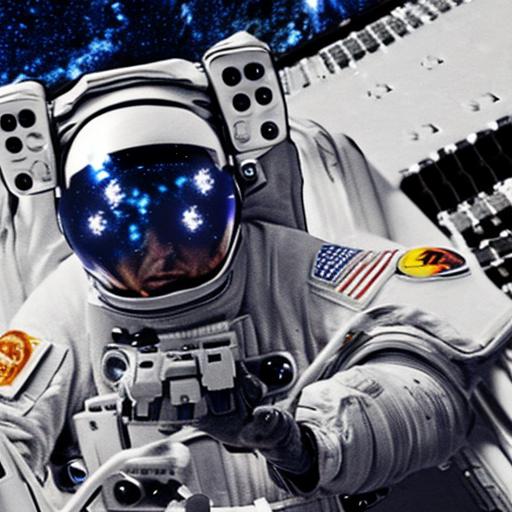} &
        \includegraphics[width=0.0975\textwidth]{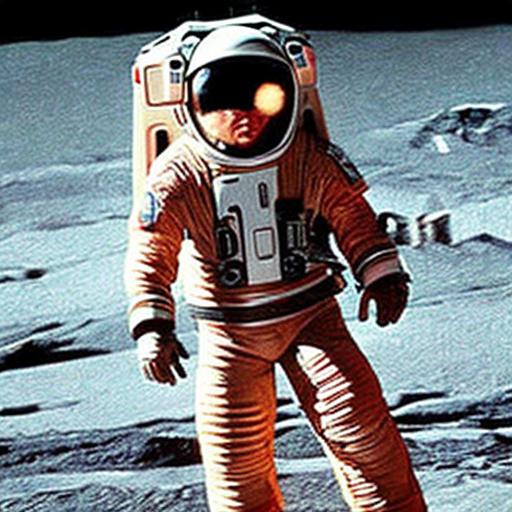} &
        \hspace{0.05cm}
        \includegraphics[width=0.0975\textwidth]{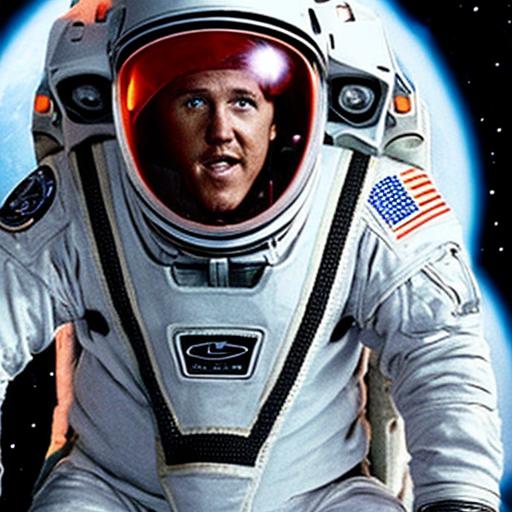} &
        \includegraphics[width=0.0975\textwidth]{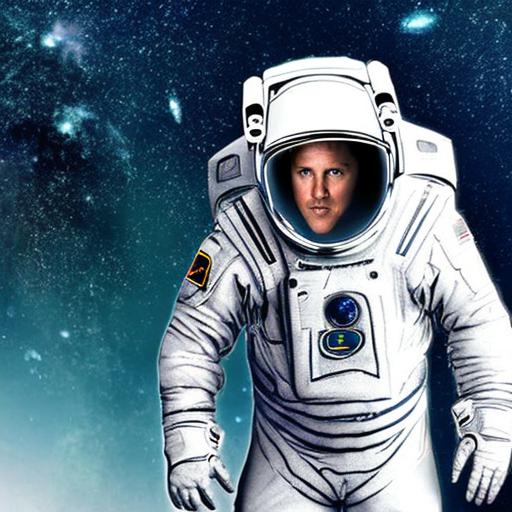} \\ \\

        \includegraphics[width=0.0975\textwidth]{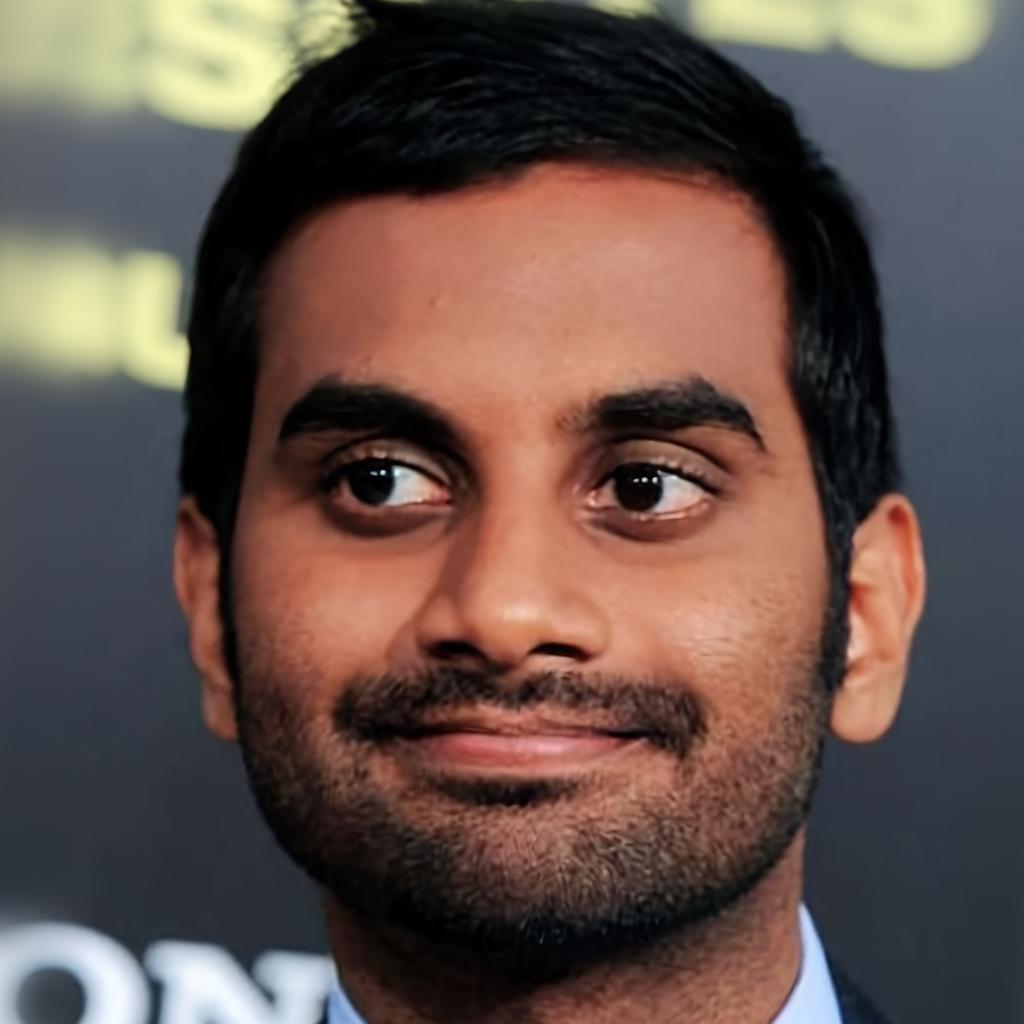} &
        \includegraphics[width=0.0975\textwidth]{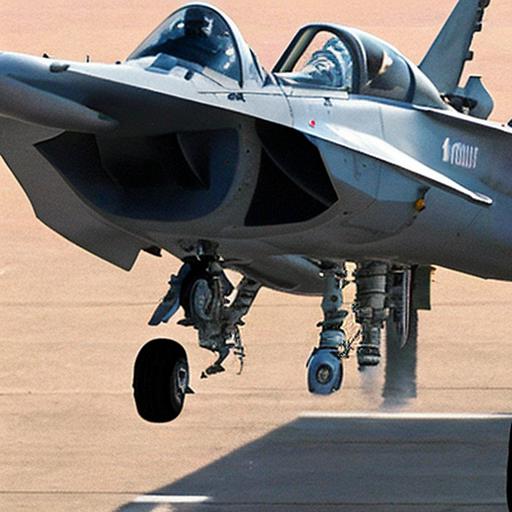} &
        \includegraphics[width=0.0975\textwidth]{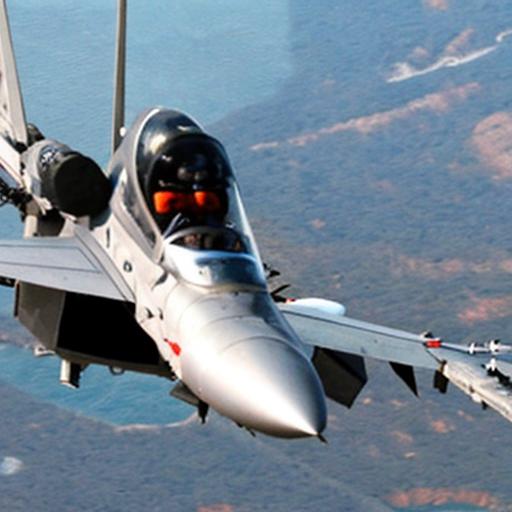} &
        \hspace{0.05cm}
        \includegraphics[width=0.0975\textwidth]{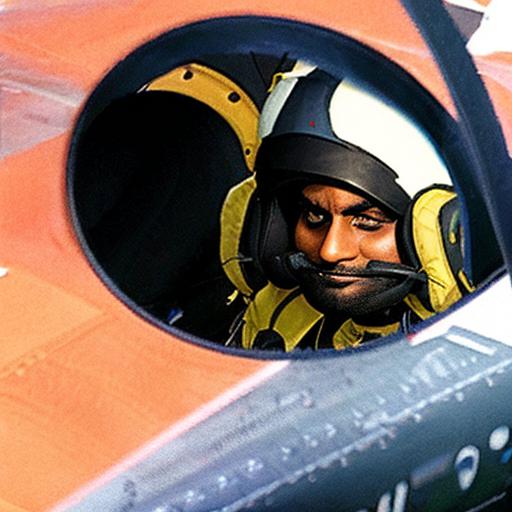} &
        \includegraphics[width=0.0975\textwidth]{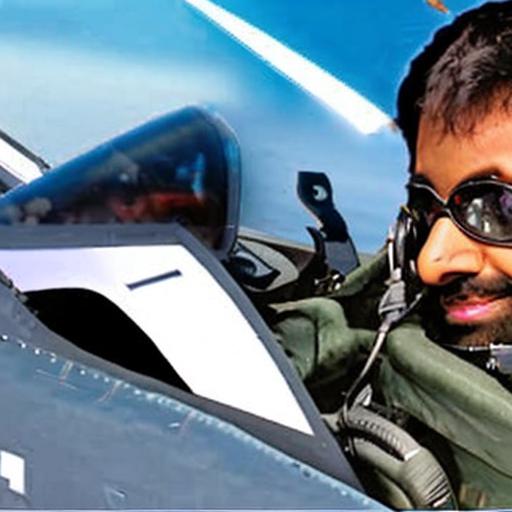} &
        \hspace{0.05cm}
        \includegraphics[width=0.0975\textwidth]{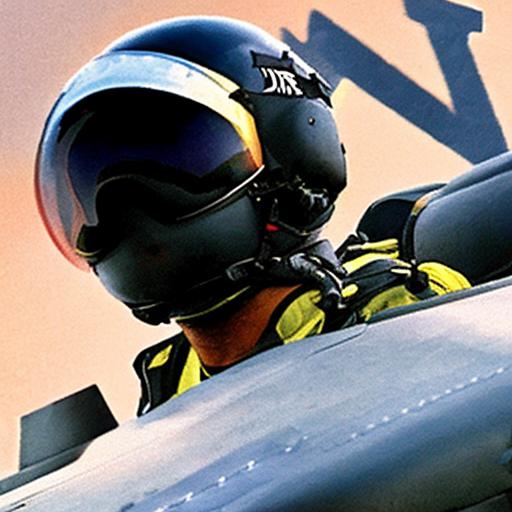} &
        \includegraphics[width=0.0975\textwidth]{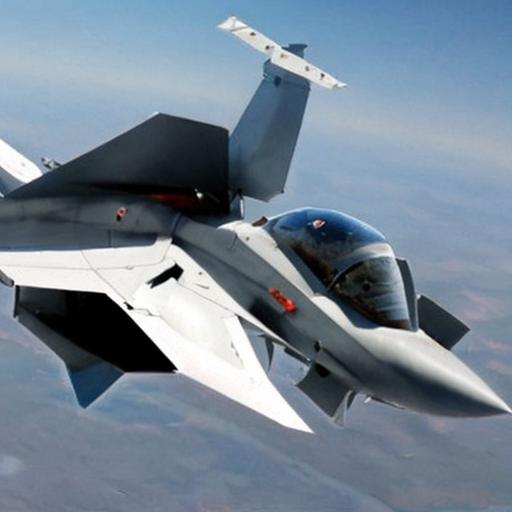} &
        \hspace{0.05cm}
        \includegraphics[width=0.0975\textwidth]{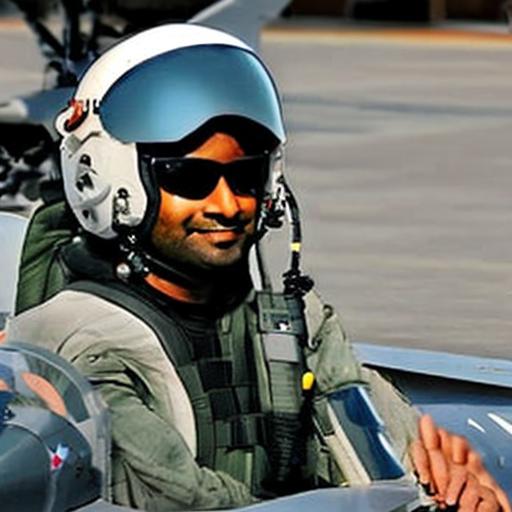} &
        \includegraphics[width=0.0975\textwidth]{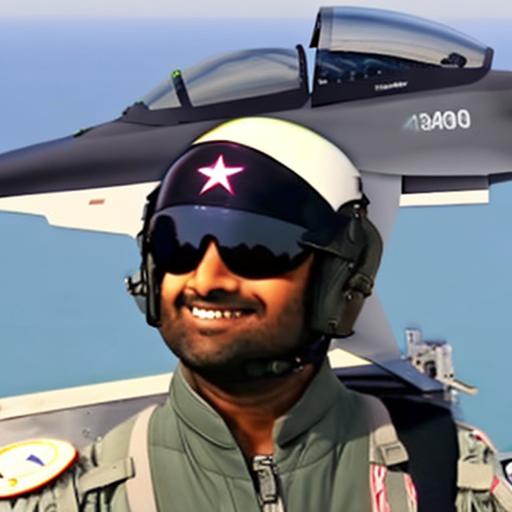} \\

        \raisebox{0.3in}{\begin{tabular}{c} ``$S_*$ piloting \\ a fighter jet''\end{tabular}} &
        \includegraphics[width=0.0975\textwidth]{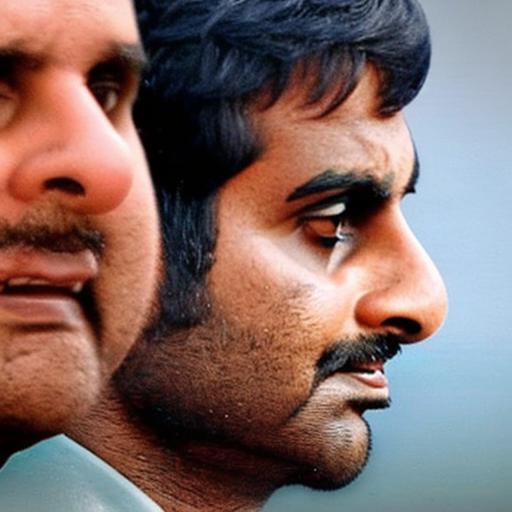} &
        \includegraphics[width=0.0975\textwidth]{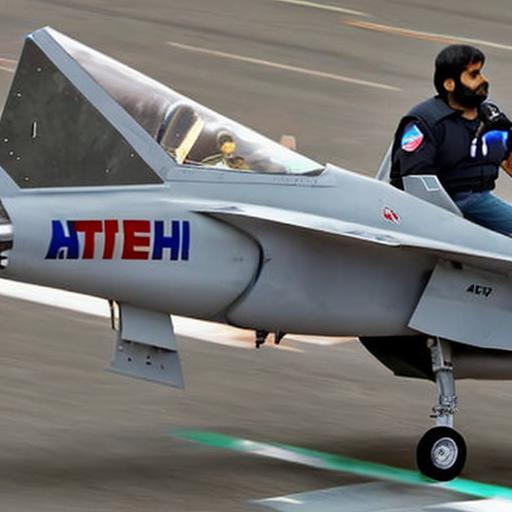} &
        \hspace{0.05cm}
        \includegraphics[width=0.0975\textwidth]{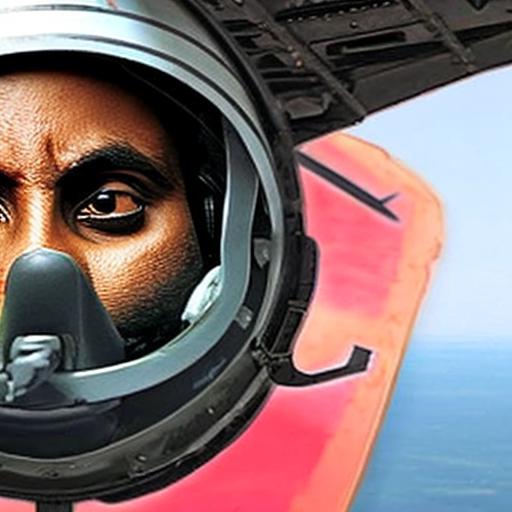} &
        \includegraphics[width=0.0975\textwidth]{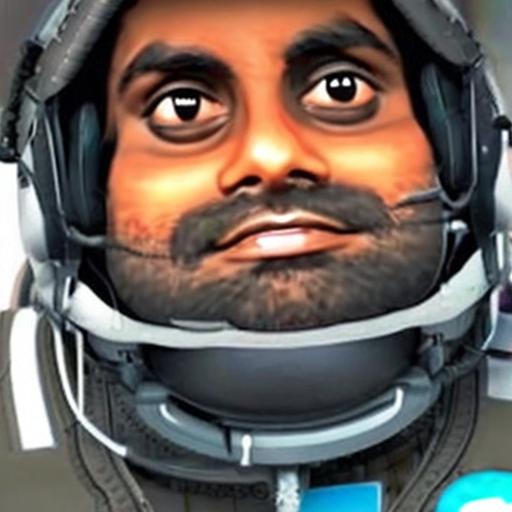} &
        \hspace{0.05cm}
        \includegraphics[width=0.0975\textwidth]{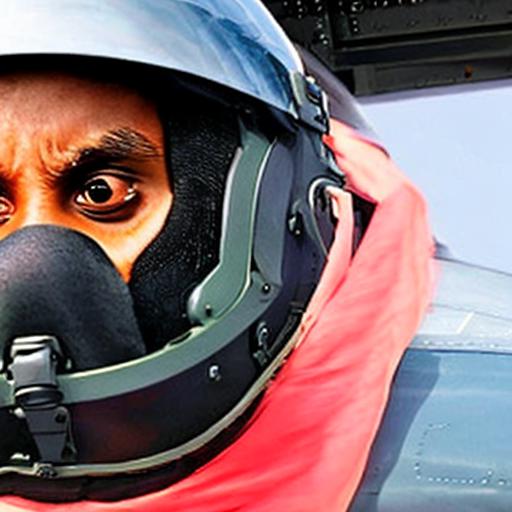} &
        \includegraphics[width=0.0975\textwidth]{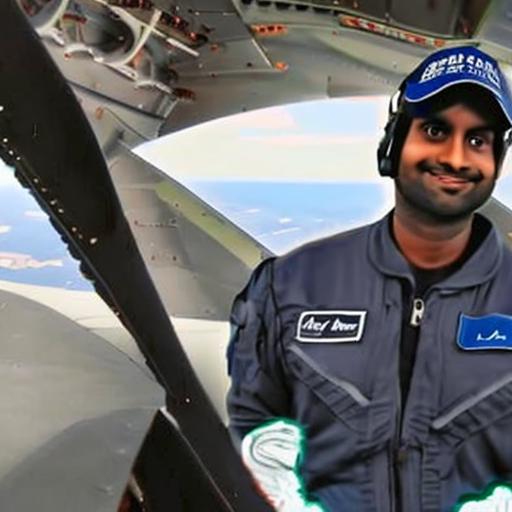} &
        \hspace{0.05cm}
        \includegraphics[width=0.0975\textwidth]{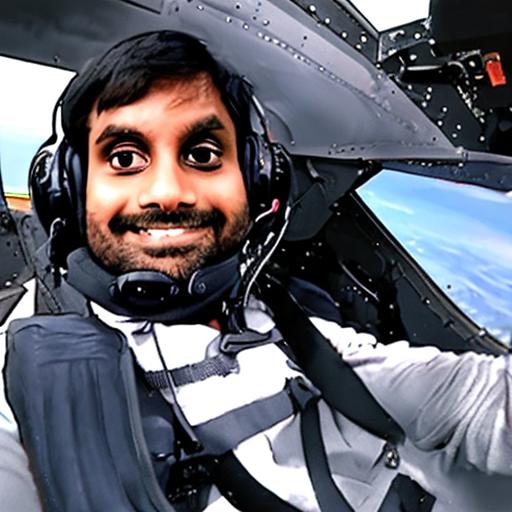} &
        \includegraphics[width=0.0975\textwidth]{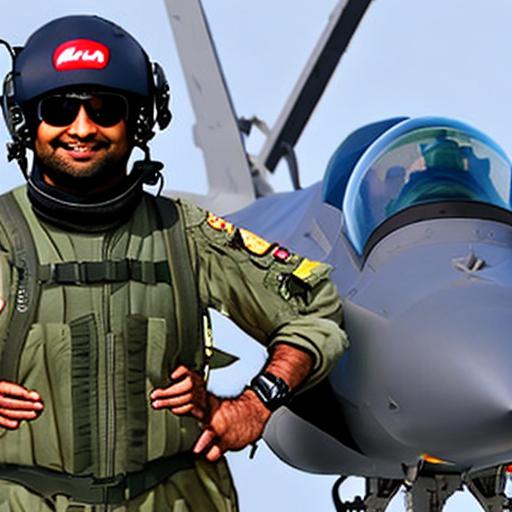} \\ \\

    \end{tabular}
    \\[-0.4cm]
    }
    \caption{Additional ablation study. We compare the models trained without Cross Initialization (w/o CI), without mean textual embedding (w/o Mean), and without regularization (w/o Reg). As can be seen, all sub-modules are essential for achieving identity-preserved and prompt-aligned personalized face generation.}
    \label{fig:appendix_ablation_study}
    \vspace{-0.3cm}
\end{figure*}

As illustrated in \cref{sec:ablation}, our ablation study involves the individual removal of the following sub-modules: 1) Cross Initialization, 2) mean textual embedding, and 3) the regularization term. Additional ablation study results for each variant are presented in \cref{fig:appendix_ablation_study}.


\end{document}